\pdfoutput=1
\documentclass[10pt,twocolumn,letterpaper]{article}

\usepackage[pagenumbers]{cvpr}      %

\usepackage{array}
\usepackage{multirow}
\usepackage{makecell}
\usepackage{pifont}
\newcommand{\model}{PhysReader}

\usepackage{adjustbox}
\usepackage{tikz}
\usetikzlibrary{arrows.meta}

\newcommand{\tauN}{eight}
\newcommand{\tauSyn}{$-0.71$}
\newcommand{\tauSweep}{$+0.50$}
\newif\ifinternalnotes \internalnotesfalse

\definecolor{cvprblue}{rgb}{0.21,0.49,0.74}
\usepackage[pagebackref,breaklinks,colorlinks,allcolors=cvprblue]{hyperref}
\hypersetup{hypertexnames=false}
\hypersetup{pdftitle={What a Drop Reveals: Sim-to-Real Material Estimation for Deformable Objects from Video}, pdfauthor={Ziyue Peng, Yanxi Hu, Junyoung Kim, Andrew Chen, Jin Yang, Chandrajit L. Bajaj, Etienne Vouga, William T. Freeman, Qixing Huang, Brandon Y. Feng}}

\title{What a Drop Reveals: Sim-to-Real Material Estimation for Deformable Objects from Video}

\author{%
Ziyue Peng\textsuperscript{1} \hspace{0.5em} Yanxi Hu\textsuperscript{1} \hspace{0.5em} Junyoung Kim\textsuperscript{1} \hspace{0.5em} Andrew Chen\textsuperscript{1} \hspace{0.5em} Jin Yang\textsuperscript{1}\\
Chandrajit L. Bajaj\textsuperscript{1} \hspace{0.5em} Etienne Vouga\textsuperscript{1} \hspace{0.5em} William T. Freeman\textsuperscript{2} \hspace{0.5em} Qixing Huang\textsuperscript{1} \hspace{0.5em} Brandon Y. Feng\textsuperscript{2}\\[0.6em]
{\normalsize \textsuperscript{1}University of Texas at Austin \quad \textsuperscript{2}Massachusetts Institute of Technology}\\[0.4em]
{\normalsize \url{https://bjxdw.github.io/PhysReader/}}%
}

\begin{document}
\twocolumn[{%
\renewcommand\twocolumn[1][]{#1}%
\maketitle

\begin{center}
  \vspace{-1.6em}
  \includegraphics[width=\textwidth]{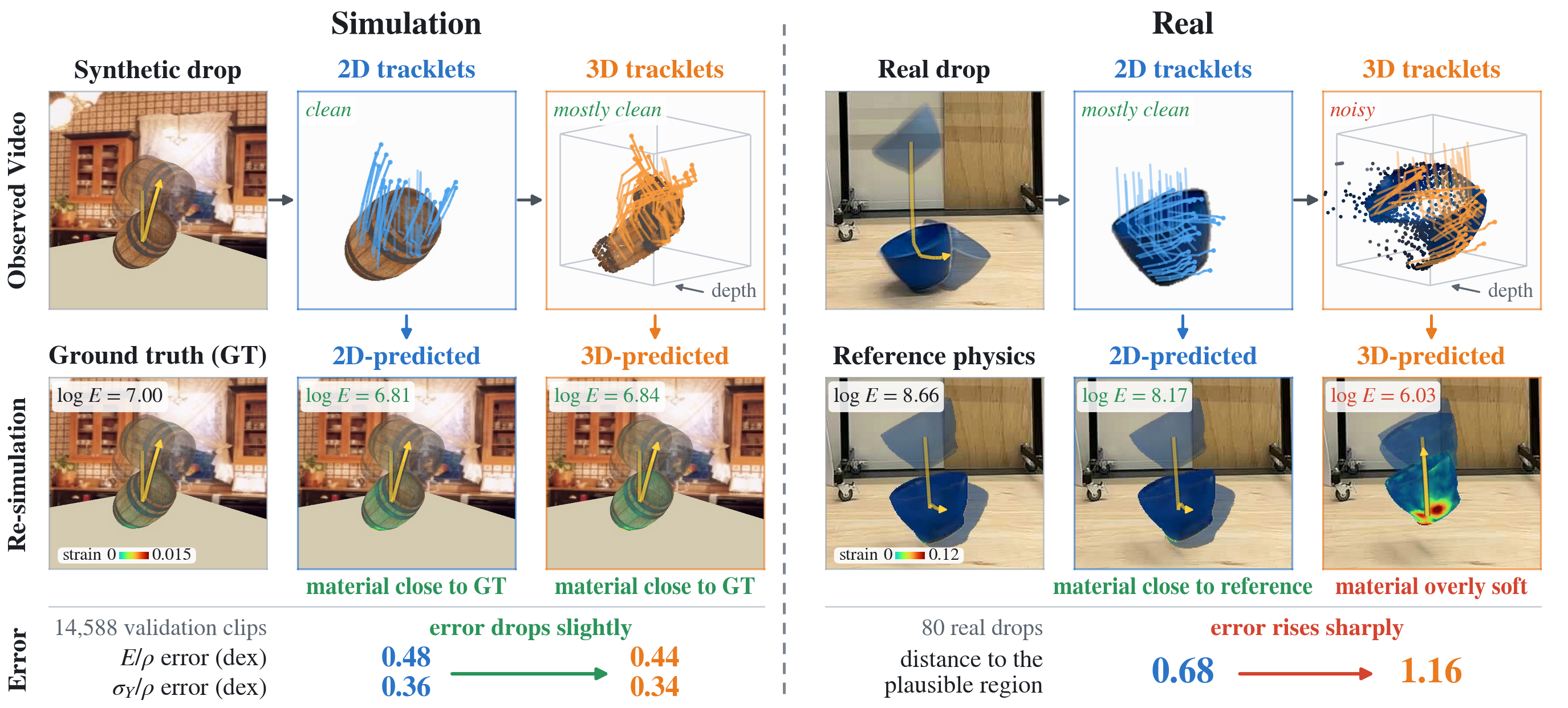}
  \vspace{-1.4em}
  \captionof{figure}{\textbf{What a drop reveals, and what transfers.}
  A predictor trained only in simulation estimates an object’s material from a video of its drop. Should it read 2D optical-flow tracklets (blue) or the same tracklets lifted to 3D with estimated depth (orange)? Lifting lowers the error slightly in simulation (left) but raises it sharply on real drops (right). Observed video: the drop at impact (earlier and later frames translucent), the 2D tracklets, and the 3D tracklets, rotated to show depth. Re-simulation: the drop re-simulated from the same initial state with the true material (simulation) or a reference material fit to the object’s drops (real), and with the materials predicted by PhysReader trained on 2D tracklets and by the same model trained on 3D tracklets. Color shows strain at impact (scales differ), and log E is $\log_{10}$ of Young’s modulus in Pa. Both predictions are close to the truth in simulation, but on the real drop the 3D one is far too soft. Error: averaged over all validation clips and real drops, in dex ($\log_{10}$ units); real drops are scored by their distance to the plausible region, the materials that reproduce the object’s drops. The 3D tracklets look noisier on the real drop, but even better depth does not close the gap (Sec. 5.3).}
  \label{fig:teaser}
\end{center}
\vspace{0.6em}

}]

\begin{abstract}
How an object deforms when it hits the ground reveals what it is made of.
From a single real video of a drop, we estimate a deformable object's density-normalized stiffness $E/\rho$, and its yield stress $\sigma_Y/\rho$ when the impact causes plastic deformation, with a feed-forward predictor trained only in simulation.
Real videos carry no material labels, so such predictors are developed and validated on synthetic data alone.
We show that synthetic validation ranks their geometric inputs in the wrong order.
Material acts through 3D deformation, and tracklets lifted to 3D with estimated depth are indeed the most accurate input in simulation.
However, on real drops they are the least accurate, and better depth does not repair them: the failure persists with simulator depth in training, LiDAR depth at test time, and exact depth on rendered replays of the real drops.
Short 2D optical-flow tracklets do the opposite: they improve nothing in simulation, yet they transfer best.
Material is therefore better read from what perception measures consistently in both domains than from what is most informative in simulation.
Our predictor, \model{}, reads the video and 2D tracklets and learns the lift to 3D in simulation by matching a teacher that sees exact 3D deformation.
To measure transfer, we introduce \model{}-Real: 16 everyday objects dropped five times each, with reference materials fit jointly to each object's drops and checked against tensile tests.
On \model{}-Real, \model{} is closer to the reference materials than every other input and baseline. The one exception is a per-scene optimizer started from \model{}'s prediction, which ends slightly closer but takes 49 minutes per video instead of 12 seconds.
\end{abstract}

\section{Introduction}
\label{sec:intro}

Dropping an object on the floor is a quick test of what it is made of: a plush toy flattens on impact, while a plastic bin barely deforms.
Reading this response from a single video as simulation-ready material parameters would let us build physical digital twins of everyday objects and re-simulate observed scenes under new conditions~\cite{xie2024physgaussian,jiang2025phystwin,Chen2025vid2sim}.
A drop is a convenient probe, since gravity and ground contact are the only external forces, but it reveals material only briefly.
Before contact, the motion depends only on the initial state, and after contact, the bounce depends on the initial pose and velocity as much as on the material.
What singles out the material is local deformation at impact (Fig.~\ref{fig:teaser}).

Per-scene inverse simulation searches for the material whose simulation best matches pixels, depth, flow, silhouettes, or a generated video~\cite{li2023pacnerf,zhong2024springgaus,zhang2024physdreamer,Liu2025physflow,jiang2025phystwin,rho2026monophysics}.
Each video needs a new search, and an overly stiff object launched slightly differently can follow nearly the observed path, so the result depends strongly on where the search starts (Table~\ref{tab:external_comparisons}).
Feed-forward predictors amortize the search~\cite{Chen2025vid2sim,lin2026analyticdynamics,yang2026matphys}.
When trained in simulation, they must transfer to real footage zero-shot, and since real videos carry no material labels, these methods are developed and validated on synthetic data alone.

Which geometric input should such a predictor read beside the video?
Material acts through 3D deformation: from the exact deformation of the whole object surface, a network recovers $E/\rho$ with a mean error of 0.145\,dex in simulation, where 1\,dex is a factor of ten (Table~\ref{tab:error_budget}).
The natural input is therefore optical-flow tracks lifted to 3D with estimated depth, and synthetic validation agrees: these lifted tracklets are the most accurate geometric input in simulation.

On real drops, the order reverses.
With the video backbone, training data, and protocol fixed, lifted tracklets are the least accurate input, worse than the same tracklets kept in 2D on 12 of 16 objects (Table~\ref{tab:representation_tradeoff}, Fig.~\ref{fig:sim_vs_real}).
Better depth does not repair them: the failure persists with simulator depth or another depth estimator in training, with LiDAR depth on the real videos, and even with exact depth on rendered replays of the real drops (Sec.~\ref{sec:3d_failure}).
2D optical-flow tracklets do the opposite. Added to the video, they leave the synthetic error unchanged, yet they give the largest gain on real drops.
Staying in 2D also costs little: in simulation, adding exact depth to the 2D tracklets lowers the error by only 0.065\,dex (Table~\ref{tab:error_budget}).
\emph{Material is better read from what perception measures consistently in both domains than from what is most informative in simulation.}
Synthetic validation misses this because it shares the training distribution, whereas renders of the target scenes, which need scans but no material labels, rank the inputs far more faithfully (Kendall $\tau$ of \tauSweep{} vs.\ \tauSyn{} with real error; Table~\ref{tab:validation_protocols}).

\begin{figure*}[t]
    \centering
    \includegraphics[width=1\linewidth]{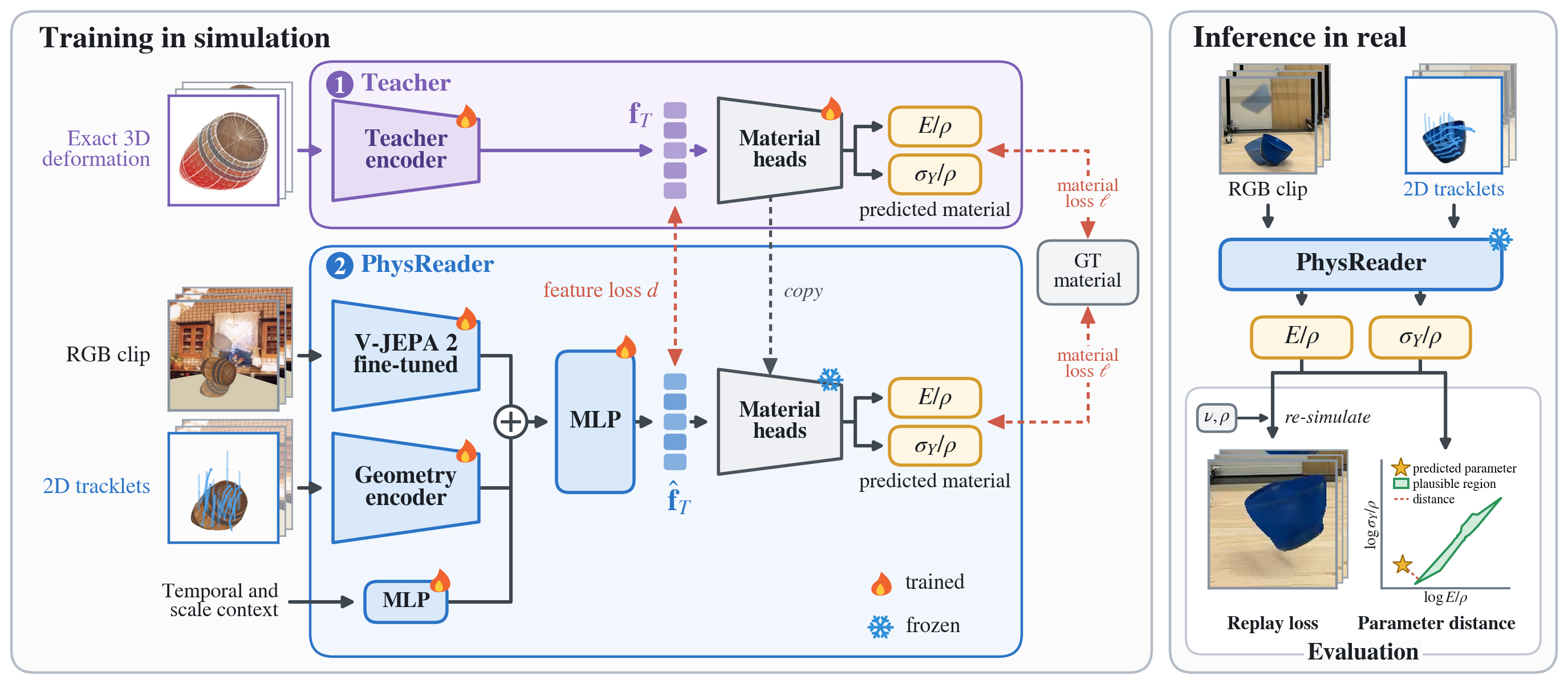}
    \vspace{-1.5em}
    \caption{\textbf{Overview of \model{}.}
    \emph{Training in simulation (left):} \ding{182}~a teacher encodes exact 3D deformation, available only in simulation, into a feature $\boldsymbol f_T$, from which material heads predict $E/\rho$ and $\sigma_Y/\rho$ (loss $\ell$ against the GT material).
    \ding{183}~\model{} sees only inputs that real video also provides: the RGB clip (V-JEPA~2 with LoRA), 2D optical-flow tracklets, and temporal and scale context.
    It maps them to $\hat{\boldsymbol f}_T$ and is trained to match $\boldsymbol f_T$ (loss $d$) and the material through a frozen copy of the teacher's heads ($\ell$; Eq.~\ref{eq:student_loss}).
    \emph{Inference in real (right):} \model{} alone predicts the material, without depth.
    For evaluation, the prediction re-simulates the drop with density $\rho$ and Poisson's ratio $\nu$, and we report the replay loss and the parameter distance to the plausible region of reference materials (Sec.~\ref{sec:benchmark}).}
    \label{fig:method}
\end{figure*}

Pose estimation offers a precedent: 3D pose can be lifted from detected 2D keypoints by networks trained without images~\cite{martinez2017simple}, and keypoint-based object-pose estimators~\cite{peng2019pvnet} can be trained on synthetic images alone~\cite{tremblay2018dope}.
Our predictor, \model{}, follows this recipe: perception supplies a 2D measurement, and the step to 3D is learned in simulation.
It reads the video with a pretrained backbone~\cite{assran2025vjepa2} and short optical-flow tracklets~\cite{wang2025waft}, needs no depth at test time, and is trained to reach the feature space of a teacher that sees exact 3D deformation (Fig.~\ref{fig:method}).

To measure transfer, we build \model{}-Real: 16 everyday deformable objects, each dropped five times in different orientations, and 40 recordings of one object under varied viewpoint, distance, background, and lighting.
Reference materials are fit jointly to the five drops of each object, with every comparably fitting material kept as a plausible region, and tensile tests place seven of eight references within or close to the measured range (Sec.~\ref{sec:benchmark}).
On \model{}-Real, \model{} is closer to the references than every other input and every feed-forward, vision-language, and agent baseline, and the materials it predicts reproduce the observed deformation best of all methods.
Of the per-scene optimizers started from its predictions, only one ends slightly closer, after 49 minutes per video instead of 12 seconds (Table~\ref{tab:external_comparisons}).

In summary, our contributions are:
\begin{itemize}
    \item A controlled study showing that synthetic validation misranks the inputs of a simulation-trained material predictor: tracklets lifted with exact, estimated, or sensor depth do not transfer, 2D optical-flow tracklets do, and renders of the target scenes rank the inputs far better.
    \item \model{}, a feed-forward predictor trained only in simulation that reads density-normalized stiffness and yield stress from video and 2D tracklets.
    \item \model{}-Real, a benchmark of repeated real drops with multi-drop reference materials checked against tensile tests.
\end{itemize}

\section{Related Work}
\label{sec:related_work}

\paragraph{Inverse physics by optimization.}
Most methods that estimate physical parameters from video simulate candidate parameters and compare the result with the observation.
Early work fit cloth parameters to calibration videos by simulated annealing~\cite{bhat2003cloth} or inferred object properties with a physics engine~\cite{wu2015galileo}.
Differentiable simulators~\cite{hu2020difftaichi,du2021diffpd} turned the search into gradient descent through simulation and rendering~\cite{jatavallabhula2021gradsim}.
Material has since been fit to multi-view videos~\cite{li2023pacnerf,zhong2024springgaus}, depth scans~\cite{arnavaz2023diffcal}, point clouds~\cite{sundaresan2022diffcloud,yang2025dpsi}, sparse RGB-D videos~\cite{jiang2025phystwin}, or a single camera~\cite{rho2026monophysics}.
Known equations can also be fit to a single video~\cite{Garcia2025Learningphysicsfromvideo}.
Generative models supply another signal.
PhysDreamer fits a generated video~\cite{zhang2024physdreamer}, Physics3D and OmniPhysGS distill video diffusion models~\cite{liu2024physics3d,lin2025omniphysgs}, and PhysFlow refines materials with optical-flow guidance from a generated video~\cite{Liu2025physflow}.
Visual observation constrains material only weakly.
Jointly fit manipulations leave many plausible parameter combinations~\cite{yang2025dpsi}, and per-scene fits of the same material can disagree~\cite{yang2026matphys}.
We use optimization only to build references for real videos, fitting one material across the five drops of each object (Sec.~\ref{sec:benchmark}).

\paragraph{Feed-forward physical inference.}
Learning amortizes this search.
From real videos, Physics~101 learns mass and density~\cite{wu2016physics101}, and Bouman~\etal regress fabric stiffness from video statistics~\cite{bouman2013fabric}.
For deformable objects, Vid2Sim predicts elastic parameters from a front-view video with a predictor trained on simulated drops, and refines them per scene~\cite{Chen2025vid2sim}.
MatPhys predicts part-level spring-mass parameters from a single view~\cite{yang2026matphys}.
Analytic Dynamics aligns video features with a representation learned from privileged simulator states~\cite{lin2026analyticdynamics}.
UniPhy fits a per-object latent of a shared neural constitutive model~\cite{Mittal2025UniPhy}.
A parallel family infers materials from static appearance: through CLIP features distilled into 3D~\cite{le2025pixie,huang2026unipixie}, single images~\cite{das2026slatphys,paul2026physvggt}, or the knowledge of language and vision-language models~\cite{Zhai2024NeRF2Physics,Shuai2025PUGS,le2026siphy,Chen2025PhysGen3D}.
Agents that write simulation code from video do not yet recover dynamics reliably~\cite{shen2026codebench}.
Appearance indicates what an object is typically made of, whereas the response of a specific object shows in how it deforms.
We read material from that deformation, and we score it against real material references rather than by rendering or replay quality alone.
Material representations and identifiability are reviewed in App.~\ref*{sec:app_related_work}.

\paragraph{Intermediate representations for sim-to-real transfer.}
A predictor trained in simulation must cope with differences in appearance, sensing, and dynamics.
Domain randomization widens the training distribution~\cite{tobin2017randomization,peng2018dynamics}, and domain adaptation aligns the two domains with unlabeled real data~\cite{shrivastava2017simgan,ganin2016dann,hoffman2018cycada,hoyer2022daformer}.
The target error of such a predictor is bounded by its source error plus a divergence between the domains~\cite{bendavid2010theory}, although invariance alone does not guarantee transfer~\cite{zhao2019invariant}.
A third strategy maps both domains to an intermediate representation with a smaller gap.
RISP predicts rendering-invariant states for system identification~\cite{ma2022risp}.
In pose estimation, 2D keypoints play this role.
Lifting networks map them to 3D pose without seeing images~\cite{martinez2017simple}.
Object-pose estimators that detect them before solving for pose are robust to occlusion~\cite{peng2019pvnet} and can be trained on synthetic images alone~\cite{tremblay2018dope}.
Vision foundation models now supply candidate intermediates off the shelf, including depth~\cite{yang2024depthanythingv2,hu2025DepthCrafter,chen2025videodepthanything}, optical flow~\cite{teed2020raft,wang2024sea_raft,wang2025waft}, point tracks~\cite{karaev2024cotracker,harley2025alltracker,xiao2025spatialtrackerv2}, and 3D and 4D reconstruction~\cite{Wang2025VGGT,Jiang2025Geo4d,luo2026fourrc}.
They are benchmarked on scene-level geometry and motion.
We ask which of them keeps the material signal of a brief impact consistent across domains.

\section{Method}
\label{Sec:methods}

We first state what a drop determines and how we train in simulation, then define the candidate intermediate representations we compare and our predictor, \model{}.

\subsection{What a Drop Determines}
\label{sec:physical_model}
We regard objects as elastoplastic deformables with Young's modulus $E$ (stiffness), yield stress $\sigma_Y$ (the threshold between elastic and plastic deformation), Poisson's ratio $\nu$, and density $\rho$. We use a Neo-Hookean elastic model with von Mises plasticity (App.~\ref*{sec:app_constitutive models}).
With deformation gradient $\boldsymbol F=\partial\boldsymbol x/\partial\boldsymbol X=\boldsymbol F_E\boldsymbol F_P$, motion obeys conservation of momentum and mass~\cite{bonet1997deformationgradient,Jiang2016MPM},
\begin{equation}
 \rho\frac{D\boldsymbol{v}}{Dt}=\nabla\cdot\boldsymbol{\sigma}+\rho\boldsymbol{g},
 \qquad \frac{D\rho}{Dt}+\rho\nabla\cdot\boldsymbol{v}=0,
 \label{eq:governing}
\end{equation}
with Cauchy stress $\boldsymbol\sigma$ and ground contact as boundary forces. We discretize these equations with the Material Point Method (MPM) and the corrections in App.~\ref*{sec:app_MPM}.
An object can be represented by one set of global material parameters~\cite{li2023pacnerf, Chen2025vid2sim, Liu2025physflow}, several parts~\cite{yang2026matphys}, or a spatial material field~\cite{zhang2024physdreamer,zhong2024springgaus,jiang2025phystwin,le2025pixie}. We use one homogeneous, isotropic, effective material per object, which a single view can constrain. The cost is a gap for strongly heterogeneous or structured objects. %

For fixed $\nu$, elastic stress is proportional to $E$, while inertia and gravity are both proportional to density $\rho$. Dividing Eq.~\eqref{eq:governing} by $\rho$ therefore shows that the motion depends on $E/\rho$ rather than on $E$ alone, as vibration-based methods also find~\cite{davis2015vibrometry,feng2022vibration}.
We therefore predict stiffness and yield stress as density-normalized ratios in log space,
\begin{equation}
 q_E=\frac{E}{\rho_0},\quad q_Y=\frac{\sigma_Y}{\rho_0},\quad z_E=\log_{10}q_E,\quad z_Y=\log_{10}q_Y,
 \label{eq:ratio_target}
\end{equation}
written $E/\rho$ and $\sigma_Y/\rho$ below. A drop does not determine density and Poisson's ratio (App.~\ref*{sec:app_identifiability}), so we take them from measurements or common values.
A drop without plastic excitation does not determine a yield threshold, so $z_Y$ is supervised and evaluated only for windows with plastic excitation.

\subsection{Training in Simulation}
\label{sec:data}
\model{}-Sim consists of simulated drops of Objaverse meshes~\cite{deitke2023objaverse1.0}, filtered for unsuitable geometry~\cite{lin2025objaverse++} and rendered from multiple cameras. Material parameters are randomized independently of semantics and appearance: $E\in[10\,\mathrm{kPa},1\,\mathrm{GPa}]$, $\sigma_Y\in[100\,\mathrm{kPa},20\,\mathrm{MPa}]$, $\nu\in[0.10,0.45]$, and $\rho\in[100,2000]\,\mathrm{kg/m^3}$.
Scales, camera distances, initial orientations, and drop heights ($0.25$--$0.75$\,m) vary. Videos have 60 frames at 30\,fps. Backgrounds from MIT Indoor67~\cite{quattoni2009indoor}, lighting, motion blur, and compression are randomized.
Training and validation are split 0.95\,:\,0.05 by object (App.~\ref*{sec:app_data_pipeline}).
In both domains, we extract optical flow with WAFT~\cite{wang2025waft}, depth with 4RC~\cite{luo2026fourrc}, and object masks with SAM~3.1~\cite{carion2026sam3}. Every geometric input the predictor sees is therefore estimated rather than simulator-derived.
We use a 16-frame window from $C-3$ to $C+12$ around the collision frame $C$. One square crop around the union of masks is expanded by 15\% and resized to $224\times224$. Depth is made metric with one robust scale per video clip, taken from rendered depth in simulation and from registered LiDAR depth on real videos (App.~\ref*{sec:app_data_pipeline}).

\subsection{Candidate Intermediate Representations}
\label{sec:representations}
The RGB branch is a V-JEPA~2 ViT-L~\cite{assran2025vjepa2} adapted with LoRA~\cite{hu2022lora}. We fix this branch, the training data, and the protocol, and vary only the geometric input.
\emph{Dense flow} and \emph{depth} are the per-pixel WAFT and 4RC maps.
\emph{2D flow tracklets} propagate points sampled in the object masks with forward flow $\boldsymbol u_t$ by bilinear sampling,
\begin{equation}
 \boldsymbol{p}_{i,t+1}=\boldsymbol{p}_{i,t}+\boldsymbol{u}_t(\boldsymbol{p}_{i,t}).
 \label{eq:short_tracks}
\end{equation}
We build four five-frame tracklets starting at clip indices $0,4,8,11$, each with up to 128 points. The $k$-nearest neighborhoods ($k\in\{8,24\}$) are fixed in the reference frame, and each point is described by its normalized reference position and displacement.
We use short tracklets because the long-range trackers we tested lose most tracks during impact (Fig.~\ref*{fig:app_tracking}).
\emph{3D tracklets} lift the same tracklets with estimated metric depth and intrinsics $\boldsymbol K$, $\boldsymbol X_{i,t}=d_t(\boldsymbol p_{i,t})\,\boldsymbol K^{-1}[\boldsymbol p_{i,t}^\top,1]^\top$, and use 3D displacements instead of 2D.
Image-plane deformation cues computed from the 2D tracklets (local strain, stretch, and non-affine residual) are a further variant (App.~\ref*{sec:feedforward}).

\subsection{\model{}: Learning the 3D Lift}
\label{sec:physreader}
\paragraph{Teacher.}
In simulation, the deformation of the whole object surface is known.
A teacher $T$ takes exact surface-deformation features, computed from simulated surface-point tracks (App.~\ref*{sec:app_oracle}). It encodes them with a graph-temporal network into a feature $\boldsymbol f_T$, from which heads $h$ predict $\boldsymbol z=(z_E,z_Y)$.
With this exact 3D deformation, it reaches 0.145\,dex on $z_E$ in simulation (Table~\ref{tab:error_budget}). %

\paragraph{Student.}
The student $S$ sees only what perception provides in both domains.
Its RGB branch encodes the unmasked clip. Its geometry branch encodes the 2D flow tracklets with three blocks of EdgeConv~\cite{wang2019dgcnn} over the fixed neighborhoods and dilated temporal convolution along each track. Masked mean and max pooling follow. A context branch encodes timing and scale: $t-t_C$, $\Delta t$, crop size over focal length, and $g\Delta t^2$. Together with the free fall, these fix the metric scale of image motion.
An MLP maps the concatenated features into the teacher's feature space, where the teacher's frozen heads predict the material (Fig.~\ref{fig:method}).
We train the student with
\begin{equation}
\mathcal L=d\bigl(S(\mathbf V,\mathcal T),\boldsymbol f_T\bigr)+\lambda\,\ell\bigl(h(S(\mathbf V,\mathcal T)),\boldsymbol z\bigr),
\label{eq:student_loss}
\end{equation}
where $\mathbf V$ is the clip, $\mathcal T$ the tracklets, and $d$ a standardized Smooth-L1 feature distance. The term $\ell$ is a regression loss on normalized targets, and its yield term is masked for windows without plastic excitation ($\lambda=1$; App.~\ref*{sec:app_teacher_student}).
At test time only the student runs, and it needs no depth.
Analytic Dynamics~\cite{lin2026analyticdynamics} also aligns a video model with a teacher on privileged simulator states, but its student sees mask videos. Ours sees RGB and 2D tracklets, the inputs that Sec.~\ref{sec:experiments} shows to transfer.

\section{\model{}-Real Benchmark}
\label{sec:benchmark}

\begin{figure*}[t]
\centering
\includegraphics[width=\textwidth]{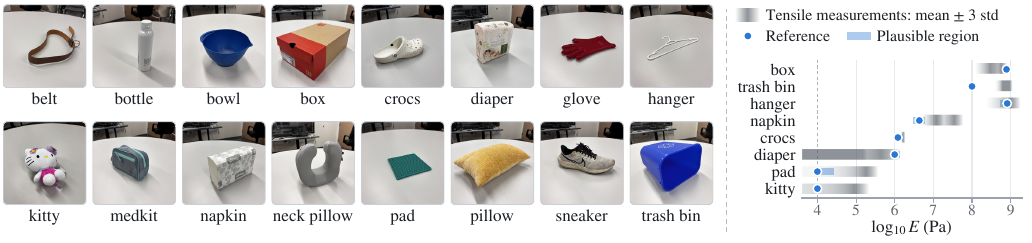}
\vspace{-1.6em}
\caption{\textbf{\model{}-Real.} Left: the 16 objects, each dropped five times in different orientations. Right: for the 8 objects we cut specimens from, the reference stiffness fit jointly to the five drops (dot) and its plausible region (blue) against the measured stiffness of tensile specimens (gray). }
\label{fig:benchmark}
\end{figure*}

\paragraph{Capture.}
\model{}-Real contains 16 everyday objects of different material types and geometries. Each is dropped five times in different orientations and recorded at $960\times720$. A supplementary set contains 40 videos of a plush toy, Kitty, under varied camera angles, backgrounds, distances, and lighting, ten per condition.
We scan the static scene with an iPhone LiDAR camera using Record3D~\cite{simonik2019record3d} and reconstruct an object mesh, which gives the simulator an explicit surface for internal filling (App.~\ref*{sec:app_real_capture}). We then fix the camera, release the object from a mechanical gripper, and clip each drop to 60 frames at 30\,fps. %

\paragraph{Reference materials.}
\label{sec:reference_protocol}
Real videos carry no material labels, so we construct object-level references by fitting one material jointly to the five drops of each object.
Drops in different orientations load the object differently, as in system identification from repeated tests of one specimen~\cite{bhat2003cloth,yang2025dpsi}. This leaves the initial states less room to compensate for a wrong material. %
Each object shares material parameters $\theta_i=(E_i,\sigma_{Y,i})$, while each drop has its own initial state $\eta_{i,d}$ (position, orientation, and linear and angular velocity). Density and Poisson's ratio are fixed per object (Tab.~\ref*{tab:fixed_reference}).
The objective combines four terms: object-center trajectory, image-plane pose, projected residual-flow thickness near the object boundary after removing the dominant affine motion, and local pair-distance changes along short tracks. In this way, local deformation, not only the trajectory, decides the material (App.~\ref*{sec:app_reference_objective}).
We alternate between two steps for up to four rounds. The first step refines each drop's initial state with the material fixed. The second searches the shared material by Bayesian optimization~\cite{jones1998ego} with the initial states fixed. We alternate because the projected deformation can be compared only after trajectory and pose are aligned.

\paragraph{Plausible material regions.}
Several coupled combinations of $E$ and $\sigma_Y$ can explain the drops comparably well. From each reference, we therefore search outward along 32 directions in the $(\log_{10}E,\log_{10}\sigma_Y)$ plane, with the initial states fixed. We keep the pairs whose multi-drop loss stays within a tolerance of the reference loss (App.~\ref*{sec:app_range}). They form a local plausible region rather than a unique label.

\paragraph{Comparison with tensile measurements.}
As in earlier checks of video-based estimates~\cite{bouman2013fabric,davis2015vibrometry}, we compare our references with tensile tests of specimens cut from eight objects. Seven fall within or close to the measured range of Young's modulus (Fig.~\ref{fig:benchmark}). %
The recorded stress--strain curves do not reliably identify yield stress, so it is not compared (App.~\ref*{sec:app_real_measure}).

\paragraph{Metrics.}
Synthetic videos are scored by the mean absolute error of $z_E$ and, for yield-excited windows, $z_Y$, in dex (App.~\ref*{sec:app_synthetic_eval}).
On real videos, the first metric is the distance from a prediction $\widehat{\mathbf z}$ to the plausible region $\mathcal A_i$ of object $i$,
\begin{equation}
d_{\mathrm{param}}^{i,d}=\min_{\mathbf z\in\mathcal A_i}\sqrt{(\widehat z_{E,i,d}-z_E)^2+m_{i,d}(\widehat z_{Y,i,d}-z_Y)^2},
\label{eq:plausible_region_distance}
\end{equation}
where $m_{i,d}$ indicates yield excitation in the reference simulation. Predictions inside the region score zero, and we average over drops.
The second metric measures whether the predicted material reproduces the observed drop. With the predicted material fixed, we refine each drop's initial state. We then report the relative increase of the total and deformation replay losses over the reference material refined the same way,
\begin{equation}
\Delta^{\mathrm{replay}}_{i,d,m}=\frac{\mathcal L^{\mathrm{pred}}_{i,d,m}-\mathcal L^{\mathrm{ref}}_{i,d,m}}{\mathcal L^{\mathrm{ref}}_{i,d,m}}\times100\%,\quad m\in\{\mathrm{total},\mathrm{deform}\}.
\label{eq:replay_loss_difference}
\end{equation}
We also report $\mathrm{Err}_{\mathrm{ref}}$, which compares trajectory, pose, and Hencky strain between simulations with the predicted and the reference material from the same initial state (App.~\ref*{sec:app_err_ref}).

\section{Experiments}
\label{sec:experiments}

All predictors are trained only on \model{}-Sim and evaluated zero-shot on \model{}-Real. Lower is better for all error metrics.

\paragraph{Implementation.}
Direct predictors and students are trained for five epochs with AdamW on 277,050 training and 14,588 validation videos.
The teacher is trained for ten epochs. Its epoch-8 checkpoint, selected on synthetic validation, is frozen.
We measure inference time per video with batch size one on one A100 GPU, including flow and masks.
Hyperparameters, appearance robustness, ablations, and the adaptation of external methods are in App.~\ref*{sec:app_architecture}, App.~\ref*{sec:app_additional_experiments}, and App.~\ref*{sec:app_baseline_adaptation}.
App.~\ref*{sec:app_revision} adds object-level intervals, per-epoch results, and other optimizer initializations, and App.~\ref*{sec:app_qualitative} shows qualitative replays.

\subsection{Does Simulation Accuracy Predict Real Accuracy?}
\label{sec:representation_tradeoff}

\begin{table}[t]
\centering
\small
\caption{\textbf{Inputs in simulation and on real drops.} Every model uses the same RGB branch, training data, and protocol, except the ``tracklets only'' rows. Syn.: $z_E$ error on synthetic validation (dex). Real: distance to the plausible region, relative increases in total and deformation replay loss, and direct simulation difference (Sec.~\ref{sec:benchmark}).}
\label{tab:representation_tradeoff}
\begin{adjustbox}{max width=\linewidth}
\begin{tabular}{@{}lccccc@{}}
\toprule
Input
& Syn.\ $\downarrow$
& \makecell{Parameter\\distance $\downarrow$}
& $\Delta_{\mathrm{total}}^{\mathrm{replay}}\downarrow$
& $\Delta_{\mathrm{deform}}^{\mathrm{replay}}\downarrow$
& $\mathrm{Err}_{\mathrm{ref}}\downarrow$ \\
\midrule
RGB & 0.478 & 0.797 & 6.17\% & 17.90\% & 0.1196 \\
RGB + dense flow & 0.472 & 0.833 & 6.82\% & 19.71\% & 0.1248 \\
RGB + depth & 0.477 & 0.789 & 7.25\% & 17.35\% & 0.1221 \\
2D tracklets only & 0.764 & 1.080 & 14.78\% & 24.74\% & 0.1501 \\
3D tracklets only (est. depth) & 0.702 & 1.337 & 25.59\% & 33.13\% & 0.1605 \\
RGB + 2D tracklets & 0.482 & 0.745 & \textbf{4.88\%} & 13.07\% & 0.1144 \\
RGB + 2D tracklets + cues & 0.487 & 0.739 & 6.18\% & 13.66\% & 0.1128 \\
RGB + 3D tracklets (est. depth) & 0.444 & 1.210 & 14.36\% & 31.23\% & 0.1504 \\
\quad + depth augmentation & 0.455 & 0.928 & 16.66\% & 21.65\% & 0.1237 \\
\midrule
Student, 3D tracklets & \textbf{0.443} & 1.164 & 17.01\% & 28.48\% & 0.1461 \\
\model{} (student, 2D tracklets) & 0.481 & \textbf{0.677} & 10.15\% & \textbf{12.32\%} & \textbf{0.1072} \\
\midrule
Random material & 1.661 & 2.071 & -- & -- & -- \\
\bottomrule
\end{tabular}
\end{adjustbox}
\end{table}

\begin{figure}[t]
\centering
\includegraphics[width=\linewidth]{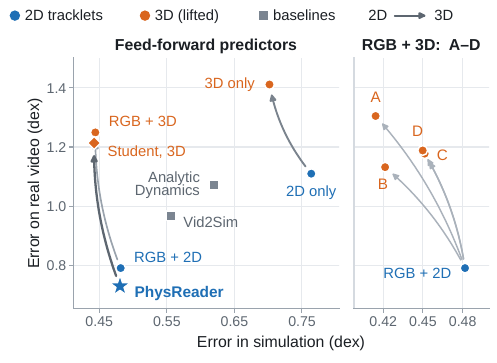}
\vspace{-1.8em}
\caption{\textbf{Simulation ranks the inputs backwards:} stiffness error on synthetic validation and on real drops. Left: arrows connect each model with 2D flow tracklets to the same model with tracklets lifted to 3D (tracklets only, RGB + tracklets, and the two students). Every arrow leads to a lower error in simulation and a higher error on real videos. The feed-forward baselines (gray) are less accurate than \model{} in both. Right: lifted tracklets, under each geometry condition of Table~\ref*{tab:3d_cross_eval}, beat 2D flow tracklets in simulation and lose to them on real videos.}
\label{fig:sim_vs_real}
\end{figure}

Table~\ref{tab:representation_tradeoff} compares geometric inputs under a fixed RGB branch, training set, and protocol.
In simulation, all inputs with RGB lie within 0.05\,dex of each other, and tracklets lifted to 3D with estimated depth give the lowest error (0.444\,dex).
On real drops, the order reverses.
Lifted tracklets are the least accurate input (parameter distance 1.210), worse even than 2D tracklets without RGB.
2D flow tracklets do not lower the synthetic error of RGB alone (0.482 vs.\ 0.478\,dex), yet they give the largest real gain (0.745 vs.\ 0.797).
Selecting inputs by synthetic validation would have chosen the input that fails and given no reason to add the one that helps.
The reversal holds on 12 of 16 objects, with an average increase of 0.465 (paired object-level bootstrap, 95\% interval $[0.073,0.843]$; App.~\ref*{sec:app_revision}), and at every training epoch (Table~\ref*{tab:app_epochs}).
Figure~\ref{fig:sim_vs_real} shows it for each pair of models that differ only in their tracklets (left), and for every combination of exact and estimated depth and tracks (right; Sec.~\ref{sec:3d_failure}).
Without RGB, lifted tracklets again win in simulation and lose on real drops (0.702 vs.\ 0.764; 1.337 vs.\ 1.080), so the reversal is not caused by the fusion with RGB.
Adding dense flow or depth to RGB changes little (0.833 and 0.789), and image-plane deformation cues add little to 2D tracklets (0.739).
2D tracklets without RGB transfer poorly (1.080), so the video backbone supplies context that the tracklets do not.
\model{}, the student given 2D tracklets, lowers the parameter distance further to 0.677, with the smallest deformation replay increase (12.32\%) and direct simulation difference (0.1072).
Its total replay increase, however, is higher than that of the direct 2D predictor (10.15\% vs.\ 4.88\%).
It is more accurate than the direct 2D predictor on 12 of 16 objects, but the paired 95\% interval of the difference, $[-0.136,0.009]$, includes zero.
All predictors respond to material: in a held-out simulated sweep in which only $E$ changes, every RGB-based predictor orders stiffness correctly for 90--95\% of pairs, against 71.45\% for 2D tracklets alone (Table~\ref*{tab:material_sensitivity}).

\paragraph{Which validation predicts transfer?}
Table~\ref{tab:validation_protocols} asks how well each validation set orders the \tauN{} predictors that share the RGB branch.
Synthetic validation, drawn from the training distribution, is anti-correlated with real accuracy (Kendall $\tau$ = \tauSyn) and selects a lifted model.
Renders of the real scenes, the real drops re-simulated with $E$ swept from $10^4$ to $10^9$\,Pa (App.~\ref*{sec:material_appearance}), need scanned scenes but no material labels.
They correlate with real accuracy (\tauSweep) and select \model{}, though they do not separate the closely spaced RGB, flow, and 2D inputs.
Replays with the reference materials order the inputs as real video does ($\tau$ = 0.929), and four real objects with references select the best input for the remaining objects in 65.8\% of subsets.

\begin{table}[t]
\centering
\small
\caption{\textbf{Which validation predicts transfer?} Kendall's $\tau$ between each validation error and the real parameter distance over the \tauN{} predictors of Table~\ref{tab:representation_tradeoff} that share the RGB branch (without the depth-augmentation variant), and the predictor each would select (its real parameter distance). Intervals and the $k$-object protocol are in App.~\ref*{sec:app_revision}.}
\label{tab:validation_protocols}
\begin{adjustbox}{max width=\linewidth}
\begin{tabular}{@{}llcl@{}}
\toprule
Validation data & Requires & $\tau\uparrow$ & Selects (real) \\
\midrule
Synthetic validation & simulation only & \tauSyn & Student, 3D tracklets (1.164) \\
Real scenes, $E$ swept & scanned objects, scenes & \tauSweep & \model{} (0.677) \\
Replays of real drops & + reference materials & 0.929 & \model{} (0.677) \\
4 real objects & + real videos, references & 0.568 & \model{} (0.677; 65.8\% of subsets) \\
\bottomrule
\end{tabular}
\end{adjustbox}
\end{table}

\subsection{Where Does the Error Come From?}
\label{sec:error_budget}

\begin{table}[t]
\centering
\small
\caption{\textbf{Error budget} for $\log_{10}(E/\rho)$ (MAE, dex). Each row is one step from the row above. Synthetic rows use the 10,202 validation videos on which every row is defined; real rows use the 80 \model{}-Real drops, scored against the reference materials. Exact geometry comes from the simulator; flow tracks from WAFT~\cite{wang2025waft}; estimated depth from 4RC~\cite{luo2026fourrc}.}
\label{tab:error_budget}
\begin{adjustbox}{max width=\linewidth}
\begin{tabular}{@{}llcc@{}}
\toprule
Input & Domain & Error & Change \\
\midrule
Exact 3D deformation, full surface (teacher) & Syn & 0.145 & \\
Exact 3D deformation, visible surface (teacher) & Syn & 0.217 & +0.072 \\
RGB + exact 3D tracks of the visible surface & Syn & 0.406 & +0.189 \\
\quad flow tracks instead of exact tracks & Syn & 0.414 & +0.008 \\
\quad 2D flow tracks, no depth & Syn & 0.478 & +0.065 \\
\quad same 2D model on real videos & Real & 0.792 & +0.313 \\
\midrule
RGB + flow tracks lifted with estimated depth & Syn & 0.444 & $-$0.034 vs.\ 2D \\
\quad same model on real videos & Real & 1.188 & +0.744 \\
\bottomrule

\end{tabular}
\end{adjustbox}
\end{table}

Table~\ref{tab:error_budget} changes one source of error at a time.
With exact geometry, replacing simulator correspondences by optical-flow tracks costs 0.008\,dex, and dropping depth costs 0.065\,dex.
2D flow tracklets therefore keep most of what exact depth adds for this predictor.
Moving to real video costs the 2D model 0.313\,dex, whereas the model on lifted tracklets, 0.034\,dex better in simulation, loses 0.744\,dex.
The largest synthetic step (+0.189\,dex) separates the teacher, which reads exact deformation of the visible surface, from a predictor given exact 3D tracks of the same surface.
The teacher--student training of \model{} targets this gap.
It leaves the synthetic error unchanged (0.481 vs.\ 0.482\,dex) but improves every real-video measure in Table~\ref{tab:representation_tradeoff} except total replay.
Linear probes show that the student's features retain strain of the visible surface ($R^2=0.625$; teacher 0.726) better than strain near the ground (0.506; App.~\ref*{sec:app_teacher_student}).

\subsection{Why Does Lifting with Estimated Depth Fail?}
\label{sec:3d_failure}

\begin{table}[t]
\centering
\small
\caption{\textbf{Same drops, three domains:} $z_E$ error (dex) of epoch-5 predictors on synthetic validation, on rendered replays of the real drops, and on real videos. The second column names the training depth and its alignment (one scale per clip or frame). ``Rendered'' and ``LiDAR'' replace the estimated depth at evaluation. Dashes: not applicable.}
\label{tab:same_drops}
\begin{adjustbox}{max width=\linewidth}
\begin{tabular}{@{}llccccc@{}}
\toprule
Input
& \makecell[l]{Depth /\\alignment}
& Syn.
& Re-sim.
& \makecell{Re-sim.,\\rendered}
& Real
& \makecell{Real,\\LiDAR} \\
\midrule
RGB + 2D tracklets
& -- & 0.482 & 0.453 & -- & 0.792 & -- \\
\midrule
RGB + 3D tracklets
& 4RC / clip scale
& 0.444 & 0.808 & 0.809 & 1.249 & 1.220 \\
& DA3 / clip scale
& 0.448 & 0.762 & 0.770 & 1.302 & 1.269 \\
& DA3 / frame scale
& 0.426 & 0.990 & 0.928 & 1.267 & 1.387 \\
& simulator (A) & 0.414 & 0.820 & 1.310 & 1.305 & 1.289 \\
\midrule
\model{}
& -- & 0.481 & 0.435 & -- & 0.731 & -- \\
\bottomrule
\end{tabular}
\end{adjustbox}
\end{table}

To separate the representation from its estimation, we train the RGB + 3D tracklets predictor with each combination of simulator or 4RC depth and simulator or WAFT tracks (A--D; App.~\ref*{sec:app_3d_cross_eval} and Fig.~\ref{fig:sim_vs_real}, right).
A model trained on exact geometry loses 0.12\,dex when WAFT replaces simulator tracks but 0.41\,dex when 4RC replaces simulator depth.
Training on 4RC depth removes this cost in simulation: these models are within 0.04\,dex of those trained on simulator depth and insensitive to which depth they receive.
Yet all four lifted models remain 0.34--0.51\,dex worse than 2D tracklets on real drops (1.132--1.305 vs.\ 0.792\,dex).

Better depth does not repair the transfer (Table~\ref{tab:same_drops}).
4RC is the most accurate of the 17 depth estimators we tested on our data (App.~\ref*{sec:app_depth_comparison}).
Depth Anything~3~\cite{lin2026depthanything3} with per-frame scale lowers the synthetic error to 0.426\,dex but raises the real error.
Replacing estimated depth with registered LiDAR depth on the real videos changes the error little (1.249 to 1.220 for the lifted model, 1.305 to 1.289 for A).
Training with depth augmentation (App.~\ref*{sec:app_revision}) reduces the real parameter distance from 1.210 to 0.928 but does not close the gap to 2D tracklets (0.745).
The gap already appears on rendered replays of the real drops, which share their objects, scenes, and initial states: lifted tracklets score 0.808 against 0.453\,dex for 2D tracklets, and 0.809 with the renderer's depth.
Model A, trained on simulator depth and tracks and given them on the replays, still scores 1.118.
This suggests that lifting ties the predictor to 3D statistics of the training scenes, which change with the scene and the sensor.
2D optical flow behaves alike across these shifts, and the free fall fixes the metric scale of image motion.
In domain-adaptation terms, the real error of a predictor is bounded by its synthetic error plus a divergence between the domains as seen through its input~\cite{bendavid2010theory}.
Lifting with depth lowers the first term and raises the second.

\paragraph{Consistency.}
Lifted tracklets are also the least consistent input (App.~\ref*{sec:recording_stability}).
Across 40 recordings of one object under changed viewpoint, distance, background, and lighting, their predictions spread by 1.018\,dex, against 0.241 for \model{} and 0.309 for RGB.
Their predictions from real drops and from the rendered replays differ by 0.704\,dex, against 0.530 for \model{}, which also varies least across the five drops of an object (0.319\,dex).
Analytic Dynamics, which reads object masks, is the most stable across the recordings (0.195\,dex) but varies most across drops (0.613).

\subsection{Comparison with Prior Methods}
\label{sec:external_comparisons}

\begin{table}[t]
\centering
\small
\caption{\textbf{Comparison with prior methods} on real videos (Sec.~\ref{sec:benchmark}), on all 80 drops.
PhysDreamer, PhysFlow, and MonoPhysics use Neo-Hookean elasticity and von Mises plasticity.
PhysTwin retains its mass--spring model.
For the per-scene optimizers, the initialization is given in parentheses.}
\label{tab:external_comparisons}
\begin{adjustbox}{max width=\linewidth}
\begin{tabular}{@{}lccccc@{}}
\toprule
Method
& \makecell{Param.\\dist.\ $\downarrow$}
& $\Delta_{\mathrm{deform}}^{\mathrm{replay}}\downarrow$
& $\Delta_{\mathrm{total}}^{\mathrm{replay}}\downarrow$
& $\mathrm{Err}_{\mathrm{ref}}\downarrow$
& \makecell{Time\\(min) $\downarrow$} \\
\midrule
PhysDreamer (\model{} init)
& 1.203 & 40.20\% & 21.00\%
& 0.1469 & 148.95 \\
PhysFlow (\model{} init)
& \textbf{0.668} & 17.19\% & \textbf{5.50\%}
& 0.1149 & 49.36 \\
PhysFlow (VLM-image init)
& 1.266 & 30.31\% & 21.52\%
& 0.1556 & 34.45 \\
MonoPhysics (\model{} init)
& 0.698 & 13.86\% & 8.83\%
& 0.1100 & 17.44 \\
MonoPhysics (Original init)
& 1.933 & 96.75\% & 26.07\%
& 0.2578 & 6.66 \\
PhysTwin
& -- & 74.41\% & 35.68\%
& 0.3264 & 12.47 \\
Vid2Sim predictor
& 0.960 & 26.78\% & 11.82\% & 0.1317 & \textbf{0.004} \\
Analytic Dynamics
& 1.084 & 18.53\% & 17.12\% & 0.1463 & 0.074 \\
\midrule
VLM (GPT-6 Astra, image input)
& 1.267 & 34.95\% & 30.54\% & 0.1683 & 0.196 \\
VLM (GPT-6 Astra, video input)
& 1.117 & 26.98\% & 29.18\% & 0.1686 & 0.199 \\
Agent (GPT-6 Astra High)
& 1.131 & 36.59\% & 9.52\% & 0.1181 & 55.19 \\
\midrule
\model{}
& 0.677 & \textbf{12.32\%} & 10.15\%
& \textbf{0.1072} & 0.200 \\
\bottomrule
\end{tabular}
\end{adjustbox}
\end{table}

We compare with the per-scene optimizers PhysDreamer~\cite{zhang2024physdreamer}, PhysFlow~\cite{Liu2025physflow}, MonoPhysics~\cite{rho2026monophysics}, and PhysTwin~\cite{jiang2025phystwin}, and with the feed-forward predictors of Vid2Sim~\cite{Chen2025vid2sim} and Analytic Dynamics~\cite{lin2026analyticdynamics}.
We also compare with a vision-language model, GPT-6 Astra, given the first frame or the same 16 frames as our predictors (App.~\ref*{sec:app_vlm_baselines}), and with a Codex agent that estimates material parameters from the full video, given the object mesh and our MPM simulator (App.~\ref*{sec:app_agent_baseline}).

PhysDreamer and PhysFlow run in our simulator on our meshes, MonoPhysics keeps its Gaussian representation and is evaluated through replay on our meshes, and all three use our fixed density and Poisson's ratio.
PhysTwin keeps its mass--spring model and is compared through replay only, and PhysDreamer's material field through its volume-weighted average (App.~\ref*{sec:app_baseline_adaptation}).

\model{} has the lowest deformation replay increase and the lowest direct simulation difference of all methods (Table~\ref{tab:external_comparisons}), and it takes 12\,s per video.
The other feed-forward predictors, the vision-language model, and the agent have a higher parameter distance than even our RGB-only predictor (0.797; Table~\ref{tab:representation_tradeoff}).
The per-scene optimizers mostly end close to where they start.
From our predictions (0.677), PhysFlow reaches 0.668 and a lower total replay increase after about 49 minutes per video, and MonoPhysics reaches 0.698 in 17 minutes.
From a VLM-image prediction, as in its original paper, PhysFlow moves from 1.267 to 1.266, and from its original initialization, MonoPhysics moves from 1.972 to 1.933.
Only PhysDreamer moves far: from our predictions, it increases the parameter distance from 0.677 to 1.203.

\section{Conclusion}
\label{sec:conclusion}

How an object deforms on impact reveals its material.
A predictor trained only in simulation can read it from real video if its input is measured alike in both domains.
Synthetic validation does not find that input.
Tracklets lifted with estimated depth are the most accurate input in simulation and the least accurate on real drops, for every training condition, depth estimator, and alignment we tried.
2D optical-flow tracklets lose little in simulation and transfer.
Renders of the target scenes rank the inputs far better.
We therefore suggest that simulation-trained estimators of physical properties validate their inputs on such renders rather than on the training distribution.
As with 2D keypoints in pose estimation, perception supplies a measurement that is consistent across domains.
The step to 3D is left to a network trained in simulation.
\model{}, built on this choice, is closest to the reference materials among all inputs and baselines in their original form, in 12\,s per video.

\paragraph{Limitations.}
\model{}-Real covers 16 objects and 80 drops, recorded with one fixed camera and a gripper release; more objects and in-the-wild capture are left to future work.
It is costly to extend because every object needs a scan, repeated drops, and a multi-drop fit.
We therefore report object-level intervals (App.~\ref*{sec:app_revision}).
Tensile tests cover eight of the objects, since they destroy the specimen, and they check stiffness only.
Yield stress is determined only when the impact causes plastic deformation, so it is less constrained than stiffness.
The references are effective parameters of one homogeneous material under our simulator.
They are not intrinsic constants of each object, and they can depart from tensile measurements of cut specimens for structured objects.

\paragraph{Acknowledgments.}
This project was supported by NSF grants 2047677, 2413161, 2504906, and 2515626; gifts from Adobe, Google, and the Chen \& Meng Family Foundation; and computing support on the Lonestar6 GPU cluster through the Center for Generative AI (CGAI) and the Texas Advanced Computing Center (TACC) at UT Austin.
{
    \small
    \bibliographystyle{ieeenat_fullname}
    \bibliography{main}

\begin{thebibliography}{90}
\providecommand{\natexlab}[1]{#1}
\providecommand{\url}[1]{\texttt{#1}}
\expandafter\ifx\csname urlstyle\endcsname\relax
  \providecommand{\doi}[1]{doi: #1}\else
  \providecommand{\doi}{doi: \begingroup \urlstyle{rm}\Url}\fi

\bibitem[Arnavaz et~al.(2023)Arnavaz, Nielsen, Kry, Macklin, and
  Erleben]{arnavaz2023diffcal}
K. Arnavaz, M.~Kragballe Nielsen, P.~G. Kry, M. Macklin, and K. Erleben.
\newblock {Differentiable Depth for Real2Sim Calibration of Soft Body
  Simulations}.
\newblock \emph{Computer Graphics Forum}, 42\penalty0 (1):\penalty0 277--289,
  2023.

\bibitem[Assran et~al.(2025)Assran, Bardes, Fan, Garrido, Howes, Komeili,
  Muckley, Rizvi, Roberts, Sinha, Zholus, Arnaud, Gejji, Martin, Robert~Hogan,
  Dugas, Bojanowski, Khalidov, Labatut, Massa, Szafraniec, Krishnakumar, Li,
  Ma, Chandar, Meier, LeCun, Rabbat, and Ballas]{assran2025vjepa2}
Mido Assran, Adrien Bardes, David Fan, Quentin Garrido, Russell Howes, Mojtaba
  Komeili, Matthew Muckley, Ammar Rizvi, Claire Roberts, Koustuv Sinha, Artem
  Zholus, Sergio Arnaud, Abha Gejji, Ada Martin, Francois Robert~Hogan, Daniel
  Dugas, Piotr Bojanowski, Vasil Khalidov, Patrick Labatut, Francisco Massa,
  Marc Szafraniec, Kapil Krishnakumar, Yong Li, Xiaodong Ma, Sarath Chandar,
  Franziska Meier, Yann LeCun, Michael Rabbat, and Nicolas Ballas.
\newblock {V-JEPA 2: Self-Supervised Video Models Enable Understanding,
  Prediction and Planning}.
\newblock \emph{arXiv preprint arXiv:2506.09985}, 2025.

\bibitem[{ASTM International}(2025)]{astmE132}
{ASTM International}.
\newblock {{ASTM E132-17(2025)e1}: Standard Test Method for Poisson's Ratio at
  Room Temperature}, 2025.

\bibitem[Baraff(1991)]{baraff1991coulomb_friction}
David Baraff.
\newblock {Coping with friction for non-penetrating rigid body simulation}.
\newblock \emph{ACM SIGGRAPH computer graphics}, 25\penalty0 (4):\penalty0
  31--41, 1991.

\bibitem[Ben-David et~al.(2010)Ben-David, Blitzer, Crammer, Kulesza, Pereira,
  and Vaughan]{bendavid2010theory}
Shai Ben-David, John Blitzer, Koby Crammer, Alex Kulesza, Fernando Pereira, and
  Jennifer~Wortman Vaughan.
\newblock {A Theory of Learning from Different Domains}.
\newblock \emph{Machine Learning}, 79\penalty0 (1--2):\penalty0 151--175, 2010.

\bibitem[Bhat et~al.(2003)Bhat, Twigg, Hodgins, Khosla, Popovic, and
  Seitz]{bhat2003cloth}
Kiran~S. Bhat, Christopher~D. Twigg, Jessica~K. Hodgins, Pradeep~K. Khosla,
  Zoran Popovic, and Steven~M. Seitz.
\newblock {Estimating Cloth Simulation Parameters from Video}.
\newblock In \emph{ACM SIGGRAPH/Eurographics Symposium on Computer Animation},
  2003.

\bibitem[Bonet and Wood(1997)]{bonet1997deformationgradient}
Javier Bonet and Richard~D Wood.
\newblock \emph{{Nonlinear continuum mechanics for finite element analysis}}.
\newblock Cambridge university press, 1997.

\bibitem[Bouman et~al.(2013)Bouman, Xiao, Battaglia, and
  Freeman]{bouman2013fabric}
Katherine~L. Bouman, Bei Xiao, Peter Battaglia, and William~T. Freeman.
\newblock {Estimating the Material Properties of Fabric from Video}.
\newblock In \emph{Proceedings of the IEEE International Conference on Computer
  Vision}, 2013.

\bibitem[Butcher(2016)]{butcher2016ode_RK2_midpoint}
John~Charles Butcher.
\newblock \emph{{Numerical methods for ordinary differential equations}}.
\newblock John Wiley \& Sons, 2016.

\bibitem[Carion et~al.(2026)Carion, Gustafson, Hu, Debnath, Hu, Suris
  Coll-Vinent, Ryali, Alwala, Khedr, Huang, Lei, Ma, Guo, Kalla, Marks, Greer,
  Wang, Sun, R{\"a}dle, Afouras, Mavroudi, Xu, Wu, Zhou, Momeni, Hazra, Ding,
  Vaze, Porcher, Li, Li, Kamath, Cheng, Dollar, Ravi, Saenko, Zhang, and
  Feichtenhofer]{carion2026sam3}
Nicolas Carion, Laura Gustafson, Yuan-Ting Hu, Shoubhik Debnath, Ronghang Hu,
  Didac Suris Coll-Vinent, Chaitanya Ryali, Kalyan~Vasudev Alwala, Haitham
  Khedr, Andrew Huang, Jie Lei, Tengyu Ma, Baishan Guo, Arpit Kalla, Markus
  Marks, Joseph Greer, Meng Wang, Peize Sun, Roman R{\"a}dle, Triantafyllos
  Afouras, Effrosyni Mavroudi, Katherine Xu, Tsung-Han Wu, Yu Zhou, Liliane
  Momeni, Rishi Hazra, Shuangrui Ding, Sagar Vaze, Francois Porcher, Feng Li,
  Siyuan Li, Aishwarya Kamath, Ho~Kei Cheng, Piotr Dollar, Nikhila Ravi, Kate
  Saenko, Pengchuan Zhang, and Christoph Feichtenhofer.
\newblock {{SAM} 3: Segment Anything with Concepts}.
\newblock In \emph{International Conference on Learning Representations
  (ICLR)}, 2026.

\bibitem[Chang et~al.(2026)Chang, Ye, Wu, Chen, Zhang, Luo, Li, Zhi, and
  Han]{chang2025reconviagen}
Jiahao Chang, Chongjie Ye, Yushuang Wu, Yuantao Chen, Yidan Zhang, Zhongjin
  Luo, Chenghong Li, Yihao Zhi, and Xiaoguang Han.
\newblock {ReconViaGen: Towards Accurate Multi-view 3D Object Reconstruction
  via Generation}.
\newblock In \emph{International Conference on Learning Representations
  (ICLR)}, 2026.

\bibitem[Chen et~al.(2025{\natexlab{a}})Chen, Jiang, Liu, Gupta, Li, Zhao, and
  Wang]{Chen2025PhysGen3D}
Boyuan Chen, Hanxiao Jiang, Shaowei Liu, Saurabh Gupta, Yunzhu Li, Hao Zhao,
  and Shenlong Wang.
\newblock {PhysGen3D: Crafting a Miniature Interactive World from a Single
  Image}.
\newblock In \emph{Proceedings of the Computer Vision and Pattern Recognition
  Conference (CVPR)}, pages 6178--6189, 2025{\natexlab{a}}.

\bibitem[Chen et~al.(2025{\natexlab{b}})Chen, Dou, Wang, Huang, Chen, Feng, Gu,
  and Liu]{Chen2025vid2sim}
Chuhao Chen, Zhiyang Dou, Chen Wang, Yiming Huang, Anjun Chen, Qiao Feng,
  Jiatao Gu, and Lingjie Liu.
\newblock {Vid2Sim: Generalizable, Video-based Reconstruction of Appearance,
  Geometry and Physics for Mesh-free Simulation}.
\newblock In \emph{Proceedings of the IEEE/CVF Conference on Computer Vision
  and Pattern Recognition (CVPR)}, pages 26545--26555, 2025{\natexlab{b}}.

\bibitem[Chen et~al.(2025{\natexlab{c}})Chen, Guo, Zhu, Zhang, Huang, Feng, and
  Kang]{chen2025videodepthanything}
Sili Chen, Hengkai Guo, Shengnan Zhu, Feihu Zhang, Zilong Huang, Jiashi Feng,
  and Bingyi Kang.
\newblock {Video Depth Anything: Consistent Depth Estimation for Super-Long
  Videos}.
\newblock In \emph{Proceedings of the IEEE/CVF Conference on Computer Vision
  and Pattern Recognition (CVPR)}, pages 22831--22840, 2025{\natexlab{c}}.

\bibitem[Cheng and Cheng(2005)]{cheng2005dirichlet}
Alexander H-D Cheng and Daisy~T Cheng.
\newblock {Heritage and early history of the boundary element method}.
\newblock \emph{Engineering analysis with boundary elements}, 29\penalty0
  (3):\penalty0 268--302, 2005.

\bibitem[Courant et~al.(1928)Courant, Friedrichs, and
  Lewy]{courant1928CFL_condition}
Richard Courant, Kurt Friedrichs, and Hans Lewy.
\newblock {{\"U}ber die partiellen Differenzengleichungen der mathematischen
  Physik}.
\newblock \emph{Mathematische annalen}, 100\penalty0 (1):\penalty0 32--74,
  1928.

\bibitem[Courant et~al.(1967)Courant, Friedrichs, and
  Lewy]{courant1967CFL_condition}
Richard Courant, Kurt Friedrichs, and Hans Lewy.
\newblock {On the partial difference equations of mathematical physics}.
\newblock \emph{IBM journal of Research and Development}, 11\penalty0
  (2):\penalty0 215--234, 1967.

\bibitem[Das and Manocha(2026)]{das2026slatphys}
Rocktim~Jyoti Das and Dinesh Manocha.
\newblock {SLAT-Phys: Fast Material Property Field Prediction from Structured
  3D Latents}.
\newblock \emph{arXiv preprint arXiv:2603.23973}, 2026.

\bibitem[Davis et~al.(2015)Davis, Bouman, Chen, Rubinstein, Durand, and
  Freeman]{davis2015vibrometry}
Abe Davis, Katherine~L. Bouman, Justin~G. Chen, Michael Rubinstein, Fr{\'e}do
  Durand, and William~T. Freeman.
\newblock {Visual Vibrometry: Estimating Material Properties from Small Motions
  in Video}.
\newblock In \emph{Proceedings of the IEEE Conference on Computer Vision and
  Pattern Recognition (CVPR)}, pages 5335--5343, 2015.

\bibitem[Deitke et~al.(2023)Deitke, Schwenk, Salvador, Weihs, Michel,
  VanderBilt, Schmidt, Ehsani, Kembhavi, and Farhadi]{deitke2023objaverse1.0}
Matt Deitke, Dustin Schwenk, Jordi Salvador, Luca Weihs, Oscar Michel, Eli
  VanderBilt, Ludwig Schmidt, Kiana Ehsani, Aniruddha Kembhavi, and Ali
  Farhadi.
\newblock {Objaverse: A Universe of Annotated 3D Objects}.
\newblock In \emph{Proceedings of the IEEE/CVF Conference on Computer Vision
  and Pattern Recognition}, pages 13142--13153, 2023.

\bibitem[Deng et~al.(2026)Deng, Song, Zhang, Wang, Wang, and
  Yue]{deng2026physreal}
Yinan Deng, Jianqiao Song, Yisi Zhang, Yuhan Wang, Jiahui Wang, and Yufeng Yue.
\newblock {PhysReal: Learning Real-World Deformable Object Physics via Hybrid
  Constitutive Modeling}, 2026.
\newblock arXiv preprint.

\bibitem[Drucker(1954)]{drucker1954coulomb_friction}
Daniel~Charles Drucker.
\newblock {Coulomb friction, plasticity, and limit loads}.
\newblock \emph{Journal of Applied Mechanics}, 21\penalty0 (1):\penalty0
  71--74, 1954.

\bibitem[Du et~al.(2022)Du, Wu, Ma, Wah, Spielberg, Rus, and
  Matusik]{du2021diffpd}
Tao Du, Kui Wu, Pingchuan Ma, Sebastien Wah, Andrew Spielberg, Daniela Rus, and
  Wojciech Matusik.
\newblock {DiffPD: Differentiable Projective Dynamics}.
\newblock \emph{ACM Transactions on Graphics}, 41\penalty0 (2):\penalty0
  13:1--13:21, 2022.

\bibitem[Feng et~al.(2022)Feng, Ogren, Daraio, and Bouman]{feng2022vibration}
Berthy~T. Feng, Alexander~C. Ogren, Chiara Daraio, and Katherine~L. Bouman.
\newblock {Visual Vibration Tomography: Estimating Interior Material Properties
  From Monocular Video}.
\newblock In \emph{Proceedings of the IEEE/CVF Conference on Computer Vision
  and Pattern Recognition (CVPR)}, pages 16231--16240, 2022.

\bibitem[Ganin et~al.(2016)Ganin, Ustinova, Ajakan, Germain, Larochelle,
  Laviolette, Marchand, and Lempitsky]{ganin2016dann}
Yaroslav Ganin, Evgeniya Ustinova, Hana Ajakan, Pascal Germain, Hugo
  Larochelle, Fran{\c c}ois Laviolette, Mario Marchand, and Victor Lempitsky.
\newblock {Domain-Adversarial Training of Neural Networks}.
\newblock \emph{Journal of Machine Learning Research}, 17\penalty0
  (59):\penalty0 1--35, 2016.

\bibitem[Garcia et~al.(2025)Garcia, Warchocki, van Gemert, Brinks, and
  T{\"o}men]{Garcia2025Learningphysicsfromvideo}
Alejandro~Casta{\~n}eda Garcia, Jan Warchocki, Jan van Gemert, Daan Brinks, and
  Nergis T{\"o}men.
\newblock {Learning Physics From Video: Unsupervised Physical Parameter
  Estimation for Continuous Dynamical Systems}.
\newblock In \emph{Proceedings of the IEEE/CVF Conference on Computer Vision
  and Pattern Recognition (CVPR)}, pages 27924--27933, 2025.

\bibitem[Harley et~al.(2025)Harley, You, Sun, Zheng, Raghuraman, Gu, Liang,
  Chu, Dave, Tokmakov, You, Ambrus, Fragkiadaki, and
  Guibas]{harley2025alltracker}
Adam~W. Harley, Yang You, Xinglong Sun, Yang Zheng, Nikhil Raghuraman, Yunqi
  Gu, Sheldon Liang, Wen-Hsuan Chu, Achal Dave, Pavel Tokmakov, Suya You, Rares
  Ambrus, Katerina Fragkiadaki, and Leonidas~J. Guibas.
\newblock {AllTracker: Efficient Dense Point Tracking at High Resolution}.
\newblock In \emph{Proceedings of the IEEE/CVF International Conference on
  Computer Vision (ICCV)}, pages 5253--5262, 2025.

\bibitem[Hoffman et~al.(2018)Hoffman, Tzeng, Park, Zhu, Isola, Saenko, Efros,
  and Darrell]{hoffman2018cycada}
Judy Hoffman, Eric Tzeng, Taesung Park, Jun-Yan Zhu, Phillip Isola, Kate
  Saenko, Alexei~A. Efros, and Trevor Darrell.
\newblock {CyCADA: Cycle-Consistent Adversarial Domain Adaptation}.
\newblock In \emph{International Conference on Machine Learning}, 2018.

\bibitem[Hoyer et~al.(2022)Hoyer, Dai, and Gool]{hoyer2022daformer}
Lukas Hoyer, Dengxin Dai, and Luc~Van Gool.
\newblock {DAFormer: Improving Network Architectures and Training Strategies
  for Domain-Adaptive Semantic Segmentation}.
\newblock In \emph{Proceedings of the IEEE/CVF Conference on Computer Vision
  and Pattern Recognition}, 2022.

\bibitem[Hu et~al.(2022)Hu, Shen, Wallis, Allen-Zhu, Li, Wang, Wang, and
  Chen]{hu2022lora}
Edward~J. Hu, Yelong Shen, Phillip Wallis, Zeyuan Allen-Zhu, Yuanzhi Li, Shean
  Wang, Lu Wang, and Weizhu Chen.
\newblock {LoRA}: Low-rank adaptation of large language models.
\newblock In \emph{International Conference on Learning Representations
  (ICLR)}, 2022.

\bibitem[Hu et~al.(2025)Hu, Gao, Li, Zhao, Cun, Zhang, Quan, and
  Shan]{hu2025DepthCrafter}
Wenbo Hu, Xiangjun Gao, Xiaoyu Li, Sijie Zhao, Xiaodong Cun, Yong Zhang, Long
  Quan, and Ying Shan.
\newblock {DepthCrafter: Generating Consistent Long Depth Sequences for
  Open-world Videos}.
\newblock In \emph{Proceedings of the IEEE/CVF Conference on Computer Vision
  and Pattern Recognition (CVPR)}, pages 2005--2015, 2025.

\bibitem[Hu et~al.(2020)Hu, Anderson, Li, Sun, Carr, Ragan-Kelley, and
  Durand]{hu2020difftaichi}
Yuanming Hu, Luke Anderson, Tzu-Mao Li, Qi Sun, Nathan Carr, Jonathan
  Ragan-Kelley, and Fr{\'e}do Durand.
\newblock {DiffTaichi: Differentiable Programming for Physical Simulation}.
\newblock In \emph{International Conference on Learning Representations}, 2020.

\bibitem[Huang et~al.(2026)Huang, Huynh, Le, Wang, Chen, Lucas, Eaton, and
  Liu]{huang2026unipixie}
Qilin Huang, Quynh~Anh Huynh, Long Le, Chen Wang, Chuhao Chen, Ryan Lucas, Eric
  Eaton, and Lingjie Liu.
\newblock {UniPixie: Unified and Probabilistic 3D Physics Learning via Flow
  Matching}.
\newblock In \emph{Proceedings of the IEEE/CVF Conference on Computer Vision
  and Pattern Recognition}, pages 19907--19916, 2026.

\bibitem[Jatavallabhula et~al.(2021)Jatavallabhula, Macklin, Golemo, Voleti,
  Petrini, Weiss, Considine, Parent-Levesque, Xie, Erleben, Paull, Shkurti,
  Nowrouzezahrai, and Fidler]{jatavallabhula2021gradsim}
Krishna~Murthy Jatavallabhula, Miles Macklin, Florian Golemo, Vikram Voleti,
  Linda Petrini, Martin Weiss, Breandan Considine, Jerome Parent-Levesque,
  Kevin Xie, Kenny Erleben, Liam Paull, Florian Shkurti, Derek Nowrouzezahrai,
  and Sanja Fidler.
\newblock {gradSim: Differentiable Simulation for System Identification and
  Visuomotor Control}.
\newblock In \emph{International Conference on Learning Representations}, 2021.

\bibitem[Jiang et~al.(2016)Jiang, Schroeder, Teran, Stomakhin, and
  Selle]{Jiang2016MPM}
Chenfanfu Jiang, Craig Schroeder, Joseph Teran, Alexey Stomakhin, and Andrew
  Selle.
\newblock {The Material Point Method for Simulating Continuum Materials}.
\newblock In \emph{ACM SIGGRAPH 2016 Courses}, New York, NY, USA, 2016.
  Association for Computing Machinery.

\bibitem[Jiang et~al.(2025{\natexlab{a}})Jiang, Hsu, Zhang, Yu, Wang, and
  Li]{jiang2025phystwin}
Hanxiao Jiang, Hao-Yu Hsu, Kaifeng Zhang, Hsin-Ni Yu, Shenlong Wang, and Yunzhu
  Li.
\newblock {PhysTwin: Physics-Informed Reconstruction and Simulation of
  Deformable Objects from Videos}.
\newblock In \emph{Proceedings of the IEEE/CVF International Conference on
  Computer Vision (ICCV)}, pages 7219--7230, 2025{\natexlab{a}}.

\bibitem[Jiang et~al.(2025{\natexlab{b}})Jiang, Zheng, Laina, Larlus, and
  Vedaldi]{Jiang2025Geo4d}
Zeren Jiang, Chuanxia Zheng, Iro Laina, Diane Larlus, and Andrea Vedaldi.
\newblock {Geo4D: Leveraging Video Generators for Geometric 4D Scene
  Reconstruction}.
\newblock In \emph{Proceedings of the IEEE/CVF International Conference on
  Computer Vision (ICCV)}, pages 20658--20671, 2025{\natexlab{b}}.

\bibitem[Jones et~al.(1998)Jones, Schonlau, and Welch]{jones1998ego}
Donald~R. Jones, Matthias Schonlau, and William~J. Welch.
\newblock {Efficient Global Optimization of Expensive Black-Box Functions}.
\newblock \emph{Journal of Global Optimization}, 13:\penalty0 455--492, 1998.

\bibitem[Karaev et~al.(2024)Karaev, Rocco, Graham, Neverova, Vedaldi, and
  Rupprecht]{karaev2024cotracker}
Nikita Karaev, Ignacio Rocco, Benjamin Graham, Natalia Neverova, Andrea
  Vedaldi, and Christian Rupprecht.
\newblock {CoTracker: It is Better to Track Together}.
\newblock In \emph{European conference on computer vision}, pages 18--35.
  Springer, 2024.

\bibitem[Kutta(1901)]{kutta1901RK2}
Wilhelm Kutta.
\newblock \emph{{Beitrag zur n{\"a}herungsweisen Integration totaler
  Differentialgleichungen}}.
\newblock Teubner, 1901.

\bibitem[Le et~al.(2026)Le, Kwon, Ismayilzada, Zhang, and Cui]{le2026siphy}
Hoang Le, Joonwoo Kwon, Elkhan Ismayilzada, Yufei Zhang, and Zijun Cui.
\newblock {SiPhy: Single-Image Physical Property Reasoning}.
\newblock In \emph{European Conference on Computer Vision (ECCV)}, pages 1--19,
  2026.

\bibitem[Le et~al.(2025)Le, Lucas, Wang, Chen, Jayaraman, Eaton, and
  Liu]{le2025pixie}
Long Le, Ryan Lucas, Chen Wang, Chuhao Chen, Dinesh Jayaraman, Eric Eaton, and
  Lingjie Liu.
\newblock {Pixie: Fast and Generalizable Supervised Learning of 3D Physics from
  Pixels}.
\newblock \emph{arXiv preprint arXiv:2508.17437}, 2025.

\bibitem[Li et~al.(2025)Li, Song, Zhou, and Yang]{Li2025FreeGave}
Jinxi Li, Ziyang Song, Siyuan Zhou, and Bo Yang.
\newblock {FreeGave: 3D Physics Learning from Dynamic Videos by Gaussian
  Velocity}.
\newblock In \emph{Proceedings of the Computer Vision and Pattern Recognition
  Conference (CVPR)}, pages 12433--12443, 2025.

\bibitem[Li et~al.(2023)Li, Qiao, Chen, Jatavallabhula, Lin, Jiang, and
  Gan]{li2023pacnerf}
Xuan Li, Yi-Ling Qiao, Peter~Yichen Chen, Krishna~Murthy Jatavallabhula, Ming
  Lin, Chenfanfu Jiang, and Chuang Gan.
\newblock {PAC-NeRF: Physics Augmented Continuum Neural Radiance Fields for
  Geometry-Agnostic System Identification}.
\newblock In \emph{International Conference on Learning Representations
  (ICLR)}, 2023.

\bibitem[Lin et~al.(2025{\natexlab{a}})Lin, Liu, Lin, Bright, Tang, He, Liu,
  Zhu, and Le]{lin2025objaverse++}
Chendi Lin, Heshan Liu, Qunshu Lin, Zachary Bright, Shitao Tang, Yihui He,
  Minghao Liu, Ling Zhu, and Cindy Le.
\newblock {Objaverse++: Curated 3D Object Dataset with Quality Annotations}.
\newblock In \emph{Proceedings of the IEEE/CVF International Conference on
  Computer Vision Workshops (ICCVW)}, pages 6872--6881, 2025{\natexlab{a}}.

\bibitem[Lin et~al.(2026{\natexlab{a}})Lin, Chen, Liew, Chen, Li, Zhao, Peng,
  Guo, Zhou, Shi, Feng, and Kang]{lin2026depthanything3}
Haotong Lin, Sili Chen, Jun~Hao Liew, Donny~Y. Chen, Zhenyu Li, Yang Zhao, Sida
  Peng, Hengkai Guo, Xiaowei Zhou, Guang Shi, Jiashi Feng, and Bingyi Kang.
\newblock {Depth Anything 3: Recovering the Visual Space from Any Views}.
\newblock In \emph{International Conference on Learning Representations
  (ICLR)}, 2026{\natexlab{a}}.

\bibitem[Lin et~al.(2026{\natexlab{b}})Lin, Zhang, and
  Sun]{lin2026analyticdynamics}
Jailing Lin, Jikuan Zhang, and Jianhua Sun.
\newblock {Analytic Dynamics: Learning Physics-Grounded Representation for Fast
  Intrinsic Dynamics Inference from Monocular Videos}.
\newblock \emph{arXiv preprint arXiv:2608.31025}, 2026{\natexlab{b}}.

\bibitem[Lin et~al.(2025{\natexlab{b}})Lin, Lin, Xu, and Mu]{lin2025omniphysgs}
Yuchen Lin, Chenguo Lin, Jianjin Xu, and Yadong Mu.
\newblock {OmniPhysGS: 3D Constitutive Gaussians for General Physics-Based
  Dynamics Generation}.
\newblock In \emph{International Conference on Learning Representations
  (ICLR)}, 2025{\natexlab{b}}.

\bibitem[Liu et~al.(2024)Liu, Wang, Yao, Zhang, Zhou, and
  Duan]{liu2024physics3d}
Fangfu Liu, Hanyang Wang, Shunyu Yao, Shengjun Zhang, Jie Zhou, and Yueqi Duan.
\newblock {Physics3D: Learning Physical Properties of 3D Gaussians via Video
  Diffusion}.
\newblock \emph{arXiv preprint arXiv:2406.04338}, 2024.

\bibitem[Liu et~al.(2025)Liu, Ye, Luximon, Wan, and Zhang]{Liu2025physflow}
Zhuoman Liu, Weicai Ye, Yan Luximon, Pengfei Wan, and Di Zhang.
\newblock {Unleashing the Potential of Multi-modal Foundation Models and Video
  Diffusion for 4D Dynamic Physical Scene Simulation}.
\newblock In \emph{Proceedings of the IEEE/CVF Conference on Computer Vision
  and Pattern Recognition (CVPR)}, pages 11016--11025, 2025.

\bibitem[Luo et~al.(2026)Luo, Zhou, Lan, Pan, and Loy]{luo2026fourrc}
Yihang Luo, Shangchen Zhou, Yushi Lan, Xingang Pan, and Chen~Change Loy.
\newblock {4RC: 4D Reconstruction via Conditional Querying Anytime and
  Anywhere}.
\newblock In \emph{Proceedings of the 43rd International Conference on Machine
  Learning}, pages 83198--83214. PMLR, 2026.

\bibitem[Ma et~al.(2022)Ma, Du, Tenenbaum, Matusik, and Gan]{ma2022risp}
Pingchuan Ma, Tao Du, Joshua~B. Tenenbaum, Wojciech Matusik, and Chuang Gan.
\newblock {RISP: Rendering-Invariant State Predictor with Differentiable
  Simulation and Rendering for Cross-Domain Parameter Estimation}.
\newblock In \emph{International Conference on Learning Representations}, 2022.

\bibitem[Ma et~al.(2023)Ma, Chen, Deng, Tenenbaum, Du, Gan, and
  Matusik]{ma2023nclaw}
Pingchuan Ma, Peter~Yichen Chen, Bolei Deng, Joshua~B. Tenenbaum, Tao Du,
  Chuang Gan, and Wojciech Matusik.
\newblock {Learning Neural Constitutive Laws from Motion Observations for
  Generalizable PDE Dynamics}.
\newblock In \emph{International Conference on Machine Learning}, 2023.

\bibitem[Martinez et~al.(2017)Martinez, Hossain, Romero, and
  Little]{martinez2017simple}
Julieta Martinez, Rayat Hossain, Javier Romero, and James~J. Little.
\newblock {A Simple yet Effective Baseline for 3D Human Pose Estimation}.
\newblock In \emph{Proceedings of the IEEE International Conference on Computer
  Vision (ICCV)}, 2017.

\bibitem[{Microsoft}(n.d.)]{trellis2repo}
{Microsoft}.
\newblock {{TRELLIS.2}: Native and Compact Structured Latents for 3D
  Generation}, n.d.
\newblock Official implementation; accessed October 2026.

\bibitem[Mittal et~al.(2025)Mittal, Zhuang, Lee, and
  Tulsiani]{Mittal2025UniPhy}
Himangi Mittal, Peiye Zhuang, Hsin-Ying Lee, and Shubham Tulsiani.
\newblock {UniPhy: Learning a Unified Constitutive Model for Inverse Physics
  Simulation}.
\newblock In \emph{Proceedings of the IEEE/CVF Conference on Computer Vision
  and Pattern Recognition (CVPR)}, pages 16208--16218, 2025.

\bibitem[Mostaghel and Davis(1997)]{mostaghel1997coulomb_friction}
Naser Mostaghel and Todd Davis.
\newblock {Representations of Coulomb friction for dynamic analysis}.
\newblock \emph{Earthquake engineering \& structural dynamics}, 26\penalty0
  (5):\penalty0 541--548, 1997.

\bibitem[Paul et~al.(2026)Paul, Wu, Liu, and Bai]{paul2026physvggt}
Sneha Paul, Guile Wu, Bingbing Liu, and Dongfeng Bai.
\newblock {PhysVGGT: Feed-Forward Dense Physical Property Estimation from A
  Single Image}.
\newblock \emph{arXiv preprint arXiv:2609.18920}, 2026.

\bibitem[Peng et~al.(2019)Peng, Liu, Huang, Zhou, and Bao]{peng2019pvnet}
Sida Peng, Yuan Liu, Qixing Huang, Xiaowei Zhou, and Hujun Bao.
\newblock {PVNet: Pixel-wise Voting Network for 6DoF Pose Estimation}.
\newblock In \emph{Proceedings of the IEEE/CVF Conference on Computer Vision
  and Pattern Recognition (CVPR)}, 2019.

\bibitem[Peng et~al.(2018)Peng, Andrychowicz, Zaremba, and
  Abbeel]{peng2018dynamics}
Xue~Bin Peng, Marcin Andrychowicz, Wojciech Zaremba, and Pieter Abbeel.
\newblock {Sim-to-Real Transfer of Robotic Control with Dynamics
  Randomization}.
\newblock In \emph{IEEE International Conference on Robotics and Automation},
  2018.

\bibitem[Pfaff et~al.(2021)Pfaff, Fortunato, Sanchez-Gonzalez, and
  Battaglia]{pfaff2021meshgraphnets}
Tobias Pfaff, Meire Fortunato, Alvaro Sanchez-Gonzalez, and Peter~W. Battaglia.
\newblock {Learning Mesh-Based Simulation with Graph Networks}.
\newblock In \emph{International Conference on Learning Representations}, 2021.

\bibitem[Quattoni and Torralba(2009)]{quattoni2009indoor}
Ariadna Quattoni and Antonio Torralba.
\newblock {Recognizing Indoor Scenes}.
\newblock In \emph{Proceedings of the IEEE Conference on Computer Vision and
  Pattern Recognition (CVPR)}, pages 413--420, 2009.

\bibitem[Rho et~al.(2026)Rho, Choi, Thornton, Dey, and
  Sengupta]{rho2026monophysics}
Daniel Rho, Jun~Myeong Choi, Matthew Thornton, Biswadip Dey, and Roni Sengupta.
\newblock {MonoPhysics: Estimating Geometry, Appearance, and Physical
  Parameters from Monocular Videos}.
\newblock In \emph{Advances in Neural Information Processing Systems}, 2026.

\bibitem[Sanchez-Gonzalez et~al.(2020)Sanchez-Gonzalez, Godwin, Pfaff, Ying,
  Leskovec, and Battaglia]{sanchez2020gns}
Alvaro Sanchez-Gonzalez, Jonathan Godwin, Tobias Pfaff, Rex Ying, Jure
  Leskovec, and Peter~W. Battaglia.
\newblock {Learning to Simulate Complex Physics with Graph Networks}.
\newblock In \emph{International Conference on Machine Learning}, 2020.

\bibitem[Shen et~al.(2026)Shen, Kova{\v c}i{\v c}, Kulits, Wang, Li, Tenenbaum,
  Yuille, Chen, and Wu]{shen2026codebench}
Ruihong Shen, {\v Z}iga Kova{\v c}i{\v c}, Peter Kulits, Xingrui Wang, Zizhang
  Li, Joshua~B. Tenenbaum, Alan Yuille, Jieneng Chen, and Jiajun Wu.
\newblock {4DCodeBench: Benchmarking Agents on Inverse Graphics of Dynamic
  Scenes}.
\newblock \emph{arXiv preprint arXiv:2610.03715}, 2026.

\bibitem[Shrivastava et~al.(2017)Shrivastava, Pfister, Tuzel, Susskind, Wang,
  and Webb]{shrivastava2017simgan}
Ashish Shrivastava, Tomas Pfister, Oncel Tuzel, Josh Susskind, Wenda Wang, and
  Russ Webb.
\newblock {Learning from Simulated and Unsupervised Images through Adversarial
  Training}.
\newblock In \emph{Proceedings of the IEEE Conference on Computer Vision and
  Pattern Recognition}, 2017.

\bibitem[Shuai et~al.(2025)Shuai, Yu, Chen, Jiang, Song, Wang, Zheng, Ma, Yang,
  Wang, Ding, and Zhao]{Shuai2025PUGS}
Yinghao Shuai, Ran Yu, Yuantao Chen, Zijian Jiang, Xiaowei Song, Nan Wang, Jv
  Zheng, Jianzhu Ma, Meng Yang, Zhicheng Wang, Wenbo Ding, and Hao Zhao.
\newblock {PUGS: Zero-shot Physical Understanding with Gaussian Splatting}.
\newblock In \emph{IEEE International Conference on Robotics and Automation
  (ICRA)}, 2025.

\bibitem[{\v{S}}imon{\'\i}k(2019)]{simonik2019record3d}
Marek {\v{S}}imon{\'\i}k.
\newblock {Record3D}.
\newblock iOS application, 2019.

\bibitem[Stomakhin et~al.(2013)Stomakhin, Schroeder, Chai, Teran, and
  Selle]{SIG2013SnowMPM}
Alexey Stomakhin, Craig Schroeder, Lawrence Chai, Joseph Teran, and Andrew
  Selle.
\newblock {A material point method for snow simulation}.
\newblock \emph{ACM Trans. Graph.}, 32\penalty0 (4), 2013.

\bibitem[Sundaresan et~al.(2022)Sundaresan, Antonova, and
  Bohg]{sundaresan2022diffcloud}
Priya Sundaresan, Rika Antonova, and Jeannette Bohg.
\newblock {DiffCloud: Real-to-Sim from Point Clouds with Differentiable
  Simulation and Rendering of Deformable Objects}.
\newblock In \emph{IEEE/RSJ International Conference on Intelligent Robots and
  Systems (IROS)}, pages 10828--10835, 2022.

\bibitem[Teed and Deng(2020)]{teed2020raft}
Zachary Teed and Jia Deng.
\newblock {RAFT: Recurrent All-Pairs Field Transforms for Optical Flow}.
\newblock In \emph{European Conference on Computer Vision (ECCV)}, pages
  402--419, 2020.

\bibitem[Tobin et~al.(2017)Tobin, Fong, Ray, Schneider, Zaremba, and
  Abbeel]{tobin2017randomization}
Josh Tobin, Rachel Fong, Alex Ray, Jonas Schneider, Wojciech Zaremba, and
  Pieter Abbeel.
\newblock {Domain Randomization for Transferring Deep Neural Networks from
  Simulation to the Real World}.
\newblock In \emph{IEEE/RSJ International Conference on Intelligent Robots and
  Systems}, 2017.

\bibitem[Tremblay et~al.(2018)Tremblay, To, Sundaralingam, Xiang, Fox, and
  Birchfield]{tremblay2018dope}
Jonathan Tremblay, Thang To, Balakumar Sundaralingam, Yu Xiang, Dieter Fox, and
  Stan Birchfield.
\newblock {Deep Object Pose Estimation for Semantic Robotic Grasping of
  Household Objects}.
\newblock In \emph{Proceedings of the Conference on Robot Learning (CoRL)},
  pages 306--316, 2018.

\bibitem[Wang et~al.(2025)Wang, Chen, Karaev, Vedaldi, Rupprecht, and
  Novotny]{Wang2025VGGT}
Jianyuan Wang, Minghao Chen, Nikita Karaev, Andrea Vedaldi, Christian
  Rupprecht, and David Novotny.
\newblock {VGGT: Visual Geometry Grounded Transformer}.
\newblock In \emph{Proceedings of the Computer Vision and Pattern Recognition
  Conference (CVPR)}, pages 5294--5306, 2025.

\bibitem[Wang et~al.(2026)Wang, He, Sun, Gu, Mo, Cao, Zhang, and
  Xu]{wang2026mosa}
Jiaxu Wang, Junhao He, Jingkai Sun, Yi Gu, Yunyang Mo, Jiahang Cao, Qiang
  Zhang, and Renjing Xu.
\newblock {MoSA: Motion-constrained Stress Adaptation for Mitigating
  Real-to-Sim Gap in Continuum Dynamics via Learning Residual Anisotropy}.
\newblock In \emph{Proceedings of the International Conference on Machine
  Learning (ICML)}, pages 126644--126668. PMLR, 2026.

\bibitem[Wang and Deng(2026)]{wang2025waft}
Yihan Wang and Jia Deng.
\newblock {WAFT: Warping-Alone Field Transforms for Optical Flow}.
\newblock In \emph{International Conference on Learning Representations
  (ICLR)}, 2026.

\bibitem[Wang et~al.(2019)Wang, Sun, Liu, Sarma, Bronstein, and
  Solomon]{wang2019dgcnn}
Yue Wang, Yongbin Sun, Ziwei Liu, Sanjay~E. Sarma, Michael~M. Bronstein, and
  Justin~M. Solomon.
\newblock {Dynamic Graph CNN for Learning on Point Clouds}.
\newblock \emph{ACM Transactions on Graphics}, 38\penalty0 (5), 2019.

\bibitem[Wang et~al.(2024)Wang, Lipson, and Deng]{wang2024sea_raft}
Yihan Wang, Lahav Lipson, and Jia Deng.
\newblock {SEA-RAFT: Simple, Efficient, Accurate RAFT for Optical Flow}.
\newblock In \emph{European Conference on Computer Vision (ECCV)}, pages
  36--54, 2024.

\bibitem[Wu et~al.(2015)Wu, Yildirim, Lim, Freeman, and
  Tenenbaum]{wu2015galileo}
Jiajun Wu, Ilker Yildirim, Joseph~J. Lim, William~T. Freeman, and Joshua~B.
  Tenenbaum.
\newblock {Galileo: Perceiving Physical Object Properties by Integrating a
  Physics Engine with Deep Learning}.
\newblock In \emph{Advances in Neural Information Processing Systems}, 2015.

\bibitem[Wu et~al.(2016)Wu, Lim, Zhang, Tenenbaum, and
  Freeman]{wu2016physics101}
Jiajun Wu, Joseph~J. Lim, Hongyi Zhang, Joshua~B. Tenenbaum, and William~T.
  Freeman.
\newblock {Physics 101: Learning Physical Object Properties from Unlabeled
  Videos}.
\newblock In \emph{Proceedings of the British Machine Vision Conference
  (BMVC)}, pages 39.1--39.12. BMVA Press, 2016.

\bibitem[Xiao et~al.(2025)Xiao, Wang, Xue, Karaev, Makarov, Kang, Zhu, Bao,
  Shen, and Zhou]{xiao2025spatialtrackerv2}
Yuxi Xiao, Jianyuan Wang, Nan Xue, Nikita Karaev, Yuri Makarov, Bingyi Kang,
  Xing Zhu, Hujun Bao, Yujun Shen, and Xiaowei Zhou.
\newblock {SpatialTrackerV2: Advancing 3D Point Tracking with Explicit Camera
  Motion}.
\newblock In \emph{Proceedings of the IEEE/CVF International Conference on
  Computer Vision}, pages 6726--6737, 2025.

\bibitem[Xie et~al.(2024)Xie, Zong, Qiu, Li, Feng, Yang, and
  Jiang]{xie2024physgaussian}
Tianyi Xie, Zeshun Zong, Yuxing Qiu, Xuan Li, Yutao Feng, Yin Yang, and
  Chenfanfu Jiang.
\newblock {PhysGaussian: Physics-Integrated 3D Gaussians for Generative
  Dynamics}.
\newblock In \emph{Proceedings of the IEEE/CVF Conference on Computer Vision
  and Pattern Recognition (CVPR)}, pages 4389--4398, 2024.

\bibitem[Yang et~al.(2024)Yang, Kang, Huang, Zhao, Xu, Feng, and
  Zhao]{yang2024depthanythingv2}
Lihe Yang, Bingyi Kang, Zilong Huang, Zhen Zhao, Xiaogang Xu, Jiashi Feng, and
  Hengshuang Zhao.
\newblock {Depth Anything V2}.
\newblock In \emph{Advances in Neural Information Processing Systems}, pages
  21875--21911, 2024.

\bibitem[Yang et~al.(2025)Yang, Ji, and Lai]{yang2025dpsi}
Xintong Yang, Ze Ji, and Yu-Kun Lai.
\newblock {Differentiable physics-based system identification for robotic
  manipulation of elastoplastic materials}.
\newblock \emph{The International Journal of Robotics Research}, 44\penalty0
  (13):\penalty0 2126--2155, 2025.

\bibitem[Yang et~al.(2026)Yang, Wang, Liu, and Iwamoto]{yang2026matphys}
Yang Yang, Yiyan Wang, Zheming Liu, and Naoya Iwamoto.
\newblock {MatPhys: Learning Material-Aware Physics Parameters for Deformable
  Object Simulation from Videos}.
\newblock \emph{arXiv preprint arXiv:2605.19386}, 2026.

\bibitem[Zhai et~al.(2024)Zhai, Shen, Chen, Wang, Wang, Wang, Guan, and
  Wang]{Zhai2024NeRF2Physics}
Albert~J. Zhai, Yuan Shen, Emily~Y. Chen, Gloria~X. Wang, Xinlei Wang, Sheng
  Wang, Kaiyu Guan, and Shenlong Wang.
\newblock {Physical Property Understanding from Language-Embedded Feature
  Fields}.
\newblock In \emph{Proceedings of the IEEE/CVF Conference on Computer Vision
  and Pattern Recognition (CVPR)}, pages 28296--28305, 2024.

\bibitem[Zhang et~al.(2024)Zhang, Yu, Wu, Feng, Zheng, Snavely, Wu, and
  Freeman]{zhang2024physdreamer}
Tianyuan Zhang, Hong-Xing Yu, Rundi Wu, Brandon~Y. Feng, Changxi Zheng, Noah
  Snavely, Jiajun Wu, and William~T. Freeman.
\newblock {PhysDreamer: Physics-Based Interaction with 3D Objects via Video
  Generation}.
\newblock In \emph{European Conference on Computer Vision (ECCV)}, pages
  388--406, 2024.

\bibitem[Zhao et~al.(2019)Zhao, Combes, Zhang, and Gordon]{zhao2019invariant}
Han Zhao, Remi Tachet~Des Combes, Kun Zhang, and Geoffrey~J. Gordon.
\newblock {On Learning Invariant Representations for Domain Adaptation}.
\newblock In \emph{International Conference on Machine Learning}, 2019.

\bibitem[Zhao et~al.(2025)Zhao, Chen, Liu, Li, Herrmann, Hur, Li, Yang, Raj,
  and Xu]{zhao2025masiv}
Yizhou Zhao, Haoyu Chen, Chunjiang Liu, Zhenyang Li, Charles Herrmann, Junhwa
  Hur, Yinxiao Li, Ming-Hsuan Yang, Bhiksha Raj, and Min Xu.
\newblock {Toward Material-Agnostic System Identification from Videos}.
\newblock In \emph{Proceedings of the IEEE/CVF International Conference on
  Computer Vision (ICCV)}, pages 5944--5956, 2025.

\bibitem[Zhong et~al.(2024)Zhong, Yu, Wu, and Li]{zhong2024springgaus}
Licheng Zhong, Hong-Xing Yu, Jiajun Wu, and Yunzhu Li.
\newblock {Reconstruction and Simulation of Elastic Objects with Spring-Mass 3D
  Gaussians}.
\newblock In \emph{European Conference on Computer Vision (ECCV)}, pages
  407--423, 2024.

\end{thebibliography}
}

\clearpage
\twocolumn[\centering\Large\textbf{\thetitle}\\\vspace{0.5em}Appendix\\\vspace{1.0em}]
\appendix
\setcounter{figure}{0}
\setcounter{table}{0}
\setcounter{equation}{0}
\renewcommand{\thefigure}{S\arabic{figure}}
\renewcommand{\thetable}{S\arabic{table}}
\renewcommand{\theequation}{S\arabic{equation}}
\setcounter{dbltopnumber}{4}\setcounter{topnumber}{3}\setcounter{totalnumber}{6}
\renewcommand{\dbltopfraction}{0.97}\renewcommand{\topfraction}{0.9}\renewcommand{\textfraction}{0.03}
\renewcommand{\floatpagefraction}{0.7}\renewcommand{\dblfloatpagefraction}{0.7}
\makeatletter\newcommand{\suppresetcolumn}{\global\@colroom\@colht\global\vsize\@colroom}\makeatother
\section{Extended Related Work}
\label{sec:app_related_work}
\paragraph{Material representation and identifiability.}
Methods that estimate material from video differ in what they estimate and simulate.
Material can be one homogeneous parameter set per object~\cite{li2023pacnerf,Chen2025vid2sim}, one set per semantic part~\cite{yang2026matphys}, or a spatial field over particles, springs, or voxels~\cite{zhang2024physdreamer,zhong2024springgaus,jiang2025phystwin,le2025pixie}.
It drives a material point method~\cite{Jiang2016MPM,li2023pacnerf,xie2024physgaussian,zhang2024physdreamer}, a spring-mass system~\cite{zhong2024springgaus,jiang2025phystwin}, a finite-element model~\cite{arnavaz2023diffcal}, or a reduced model based on linear blend skinning~\cite{Chen2025vid2sim}, and UniPixie predicts parameters for several of these solvers~\cite{huang2026unipixie}.
Neural constitutive models relax the analytic material law~\cite{ma2023nclaw,Mittal2025UniPhy,zhao2025masiv} or learn residual corrections to it~\cite{wang2026mosa,deng2026physreal}, learned simulators replace the solver~\cite{sanchez2020gns,pfaff2021meshgraphnets}, and FreeGave bypasses material parameters by learning per-Gaussian physics codes that drive a velocity field~\cite{Li2025FreeGave}.
Richer representations fit observed motion more closely but are harder to infer from one view and harder to compare across methods.
We use one homogeneous elastoplastic material per object in MPM: it cannot capture internal variation, but it yields a compact effective quantity that can be compared with references and measurements.
What a video can determine is also limited by physics.
Vibration-based methods recover the elasticity-to-density ratio of rods~\cite{davis2015vibrometry} and stiffness and density fields only up to a common scale~\cite{feng2022vibration}; in a drop, gravity and inertia both scale with density, so we likewise predict $E/\rho$ (Sec.~\ref{sec:physical_model}).
Finally, simulator-fitted parameters need not equal laboratory values: Arnavaz~\etal found that manufacturer-specified silicone parameters did not reproduce the observed deformation in their simulator~\cite{arnavaz2023diffcal}.
Video-based estimates have been validated against measured or known properties of fabrics and rods~\cite{bouman2013fabric,davis2015vibrometry}; we compare impact-based reference materials of everyday objects with tensile tests.

\section{Constitutive Models}
\label{sec:app_constitutive models}
\textbf{Elasticity: Neo-Hookean model.} The strain energy density is 
\[\Psi(\boldsymbol{F}) = \frac{\mu}{2} (\operatorname{tr}(\boldsymbol{F}^{T} \boldsymbol{F}) - 3) - \mu \operatorname{ln}J + \frac{\lambda}{2} (\operatorname{ln}J)^2\]
where \(J = \det \boldsymbol{F}\), and \(\lambda = \frac{E\nu}{(1+\nu)(1-2\nu)}\) and \(\mu = \frac{E}{2(1 + \nu)}\) are Lam\'{e} parameters describing the resistance to volumetric deformation and shear deformations. 

The first Piola--Kirchhoff stress is
\[\boldsymbol{P} = \frac{\partial \Psi (\boldsymbol{F})}{\partial \boldsymbol{F}} = \mu (\boldsymbol{F} - \boldsymbol{F}^{-T}) + \lambda \operatorname{ln} J \, \boldsymbol{F}^{-T}.\]
The simulator uses the Kirchhoff stress \(\boldsymbol{\tau}\), which is related to the Cauchy stress \(\boldsymbol{\sigma}\) and to \(\boldsymbol{P}\) by
\[\boldsymbol{\tau} = J \boldsymbol{\sigma} = \boldsymbol{P} \boldsymbol{F}^T = \mu (\boldsymbol{F} \boldsymbol{F}^T - \boldsymbol{I}) + \lambda \operatorname{ln} J \, \boldsymbol{I}.\]
With plasticity, \(\boldsymbol{F}\) and \(J\) in these expressions are the elastic part \(\boldsymbol{F}_E\) defined below and \(J_E = \det \boldsymbol{F}_E\).

\textbf{Plasticity: von Mises model.} The return mapping \(\mathcal{Z}\) splits the deformation into an elastic and a plastic part, \(\boldsymbol{F}_E = \mathcal{Z}(\boldsymbol{F}) = \boldsymbol{F F}_P^{-1}\), and gives the plastic dissipation \(W_P\).

We first compute the singular value decomposition \(\boldsymbol{F = U \Sigma V}^T\) and define the Hencky strain
\[\boldsymbol{\varepsilon} = \ln \boldsymbol{\Sigma}\]

This strain is split into the volumetric \(\bar{\varepsilon}\) and deviatoric parts \(\boldsymbol{\varepsilon'}\) \[\bar{\varepsilon} = \frac{1}{3} \operatorname{tr}(\boldsymbol{\varepsilon})   \quad \quad \quad \boldsymbol{\varepsilon'} = \boldsymbol{\varepsilon} - \bar{\varepsilon} \boldsymbol{I}\]

The trial Kirchhoff stress and its deviatoric part are
\[\begin{gathered}
\boldsymbol{\tau_{trial}} = 2 \mu \boldsymbol{\varepsilon} + \lambda \operatorname{tr}(\boldsymbol{\varepsilon}) \boldsymbol{I} \\ 
\boldsymbol{s}_{trial} = \operatorname{dev} \boldsymbol{\tau}_{trial} = \boldsymbol{\tau}_{trial} - \frac{1}{3} \operatorname{tr}(\boldsymbol{\tau}_{trial}) \boldsymbol{I}
\end{gathered}\]

Define the equivalent trial stress
\[\tau_{eq}^{trial} = \sqrt{\frac{3}{2} \boldsymbol{s}_{trial}:\boldsymbol{s}_{trial}}\]

If $\tau_{eq}^{trial} > \tau_Y = J \sigma_Y$,
yielding occurs.
The volumetric strain \(\bar{\varepsilon}\) remains unchanged, while the deviatoric strain \(\boldsymbol{\varepsilon'}\) is scaled back to the yield surface via the plastic multiplier
\[\Delta \gamma = \frac{ \tau_{eq}^{trial} - \tau_Y }{2 \mu}\]

We then update
\[\boldsymbol{\varepsilon}_{E} = \boldsymbol{\varepsilon} - \frac{\Delta\gamma}{\lVert \boldsymbol{\varepsilon'} \rVert} \boldsymbol{\varepsilon'} \quad \quad \quad 
\boldsymbol{\Sigma}_E = \operatorname{exp}(\boldsymbol{\varepsilon}_{E})\]
and construct the elastic part of the deformation gradient
\[\boldsymbol{F_E} = \boldsymbol{U \Sigma_E V}^T.\]
The associated plastic energy dissipation is 
\[\Delta W_p = \sqrt{\frac{2}{3}} \tau_Y \Delta \gamma.\]

\section{Physically Accurate MPM Simulator}
\label{sec:app_MPM}
MPM represents matter with Lagrangian particles that store the complete history of strain and stress, while using a background Eulerian grid to evaluate spatial derivatives required for force calculations \cite{Jiang2016MPM, SIG2013SnowMPM}. 
Each time step proceeds in a particle-to-grid transfer (P2G), grid-processing, and grid-to-particle transfer (G2P) loop.
Below, we detail the standard formulation and our specific enhancements for physical accuracy.

\paragraph{1. Particles-to-Grid Transfer (P2G)}  Mass and momentum are mapped from particles \(p\) to grids \(i\):
\[\begin{gathered}
m_n^i = \sum_p \omega_{ip}^n m_p \\ 
m_n^i \boldsymbol{v}_i^n = \sum_p \omega_{ip}^n m_p (\boldsymbol{v}_p^n + \boldsymbol{C}_p^n (\boldsymbol{x}_i - \boldsymbol{x}_p^n) ) 
\end{gathered}\]
Here \(\omega_{ip}^n\) is the \(b\)-spline kernel centered at particle \(p\) and evaluated at node \(i\) at time \(t_n\). 
Each particle \(p\) carries volume \(V_p\), mass \(m_p = \rho V_p\), position \(\boldsymbol{x}_p\), velocity \(\boldsymbol{v}_p\), affine velocity \(\boldsymbol{C}_p^n\), and deformation gradient \(\boldsymbol{F}^n\).

\paragraph{2. Grid Velocity Update with Force} Grid velocities are updated by gravity and internal stress:
\[ \boldsymbol{v}_i^{n+1} = \boldsymbol{v}_i^{n} + \boldsymbol{g} \Delta t - \frac{\Delta t}{m_i} \sum_p \boldsymbol{\tau}_p^n V_p \nabla \omega_{ip}^n \]

\paragraph{2a. Improvement: Collision Handling}
Instead of using the Dirichlet collider (boundary conditions \(\boldsymbol{v} = \boldsymbol{0}\)) \cite{cheng2005dirichlet} that simply sets the grid velocity at the boundary to zero (which will lose extra energy and momentum), we employ a Coulomb friction model \cite{baraff1991coulomb_friction, drucker1954coulomb_friction, mostaghel1997coulomb_friction} that is aligned with realistic physics. 

We decompose the incident grid velocity into the normal and tangential components \(\boldsymbol{v} = \boldsymbol{v}_{n} + \boldsymbol{v}_{t}\) and set the rebounded normal speed to \(\boldsymbol{v'}_n = - e \boldsymbol{v}_n\) with restitution \(e\). 
The normal impulse is then
\[\boldsymbol{J}_n = m (\boldsymbol{v'}_n - \boldsymbol{v}_n) = -m(1 + e)\boldsymbol{v}_n.\]

Coulomb's law limits the tangential impulse, with friction coefficient \(\mu_f\):
\[ \lVert \boldsymbol{J}_t \rVert \leq \mu_f \lVert \boldsymbol{J}_n \rVert. \]

When the contact point is sliding, the tangential impulse reaches its Coulomb limit, and the tangential velocity is reduced:
\[\begin{gathered}
  \lVert \boldsymbol{J}_t \rVert = \mu_f \lVert \boldsymbol{J}_n \rVert \\ 
\boldsymbol{v'}_t = \frac{\lVert m\boldsymbol{v}_t \rVert - \lVert \boldsymbol{J}_t \rVert}{ \lVert m\boldsymbol{v}_t \rVert} \boldsymbol{v}_t = (1 - \frac{\mu_f(1+e) \lVert \boldsymbol{v}_n \rVert}{\lVert \boldsymbol{v}_t \rVert})\boldsymbol{v}_t
\end{gathered}\]

If the contact point sticks (static friction), the tangential impulse remains below the Coulomb threshold, and the tangential velocity after impact vanishes
\[  \lVert \boldsymbol{J}_t \rVert < \mu_f \lVert \boldsymbol{J}_n \rVert  \quad \quad \quad \boldsymbol{v'}_t = \boldsymbol{0}\]

The final velocity is 
\[\boldsymbol{v}_i^{n+1} = \boldsymbol{v'}_{n} + \boldsymbol{v'}_{t}\]

\paragraph{3. Grids-to-Particles Transfer (G2P)} Transfer grid velocities back to particles and update particle states.

In the original APIC-MPM, particle velocity, position, and affine velocity are updated by
\[
\boldsymbol{v}_p^{n+1} = \sum_i \omega_{ip}^n \boldsymbol{v}_i^{n+1} \quad \quad \quad 
\boldsymbol{x}_p^{n+1} = \boldsymbol{x}_p^n + \boldsymbol{v}_p^{n+1} \Delta t 
\]
\[
\boldsymbol{C}_p^{n+1} = \frac{12}{(b+1) \Delta x^2} \sum_i \omega_{ip}^n \boldsymbol{v}_i^{n+1} (\boldsymbol{x}_i^n - \boldsymbol{x}_p^n)^T
\]
here \(b\) is the B-spline degree, and \(\Delta x\) is the grid spacing.
In the continuous setting, the deformation gradient \(\boldsymbol{F}\) satisfies
\[\dot{\boldsymbol{F}} = \boldsymbol{L} \boldsymbol{F} \quad \quad \quad
\boldsymbol{L} = \nabla \boldsymbol{v}.\]

Discretely, the spatial velocity gradient at particle \(p\) at time step \(t^n\) is approximated by
\[ \boldsymbol{L}_p^{n+1} = \nabla \boldsymbol{v}_p^{n+1} = \sum_i (\nabla \omega_{ip}^n)^T \boldsymbol{v}_i^{n+1}.\]

After that, the deformation gradient update is approximated with a first-order Taylor expansion
\[\boldsymbol{F}_{trial}^{n+1} = (\boldsymbol{I} + \boldsymbol{L}_p^{n+1} \Delta t) \boldsymbol{F}_E^n.\]

We subsequently project \(\boldsymbol{F}_{trial}^{n+1}\) to the elastic manifold, compute Kirchhoff stress \(\boldsymbol{\tau}^{n+1}\) and continue to the next P2G2P loop.
\[ \boldsymbol{F}_E^{n+1} = \mathcal{Z}(\boldsymbol{F}_{trial}^{n+1}) \quad \quad \quad
\boldsymbol{\tau}^{n+1} = \boldsymbol{\tau}(\boldsymbol{F}_E^{n+1}).\]

However, the G2P process in the original MPM is neither physically nor numerically accurate. 
Thus, we implement some modifications as follows.
 
\paragraph{3a. Improvement: Computation of velocity and velocity gradient} When the particles are near boundaries, the grid nodes outside the boundaries have their velocities clamped to \(\boldsymbol{0}\). 
Including these zeros biases the estimated velocity gradients and decreases particle velocities.
We address this issue by fitting a local affine velocity field via moving least squares (MLS) with Tikhonov regularization and boundary pre-conditioning.

\subparagraph{MLS with Tikhonov regularization.} We approximate the velocity near the particle \(p\) by a first-order model
\[\boldsymbol{v}(\boldsymbol{x}) = \boldsymbol{a}_p + \boldsymbol{B}_p (\boldsymbol{x} - \boldsymbol{x}_p)\]
To avoid boundary pollution from outside grids, we only include grid nodes inside the domain (including boundary nodes) in the neighbor set \(S_p\). 
We solve the Tikhonov-regularized weighted least squares
\[\min_{\boldsymbol{a}_p, \boldsymbol{B}_p} \sum_{i \in S_p} \omega_{ip} \lVert \boldsymbol{v}_i - \boldsymbol{a}_p - \boldsymbol{B}_p (\boldsymbol{x}_i - \boldsymbol{x}_p) \rVert^2 + \lambda \lVert \boldsymbol{B}_p \rVert_F^2\]
Let \(\boldsymbol{r}_i = \boldsymbol{x}_i - \boldsymbol{x}_p\), we define the weighted average velocity and offset:
\[\bar{\boldsymbol{v}} = \frac{\sum_i \omega_{ip} \boldsymbol{v}_i}{\sum_i \omega_{ip}} \quad \quad \quad 
\bar{\boldsymbol{r}} = \frac{\sum_i \omega_{ip} \boldsymbol{r}_i}{\sum_i \omega_{ip}} \quad \quad \quad \]
and the centered moment matrices:
\[\begin{gathered}
\boldsymbol{M}_{rr} = \sum_i \omega_{ip} (\boldsymbol{r}_i - \bar{\boldsymbol{r}}) (\boldsymbol{r}_i - \bar{\boldsymbol{r}}) ^ T \\ 
\boldsymbol{M}_{vr} = \sum_i \omega_{ip} (\boldsymbol{v}_i - \bar{\boldsymbol{v}}) (\boldsymbol{r}_i - \bar{\boldsymbol{r}}) ^ T. 
\end{gathered}\]
The closed-form solution is
\[\boldsymbol{B}_p = \boldsymbol{M}_{vr} (\boldsymbol{M}_{rr} + \lambda \boldsymbol{I})^{-1}
\quad \quad \quad
\boldsymbol{a}_p = \bar{\boldsymbol{v}} - \boldsymbol{B}_p\bar{\boldsymbol{r}}\]
By definition, we set
\[\boldsymbol{C}_p^{n+1} = \boldsymbol{L}_p^{n+1} = \boldsymbol{B}_p  \quad \quad \quad
\boldsymbol{v}_p^{n+1} = \boldsymbol{a}_p\]

\subparagraph{Boundary Preconditioning} 
Near solid boundaries, the neighborhood becomes strongly anisotropic: grids outside the domain are excluded, so \(\boldsymbol{M}_{rr}\) is ill-conditioned with small eigenvalues along the boundary surface normal.
Let \(\tilde{\boldsymbol{r}}_i=\boldsymbol{r}_i-\bar{\boldsymbol{r}}\).
We precondition by whitening the offsets with a diagonal scaling
\[
\boldsymbol{P}=\operatorname{diag}(s_x,s_y,s_z)\quad \quad \quad
s_d=\frac{1}{\sqrt{(\boldsymbol{M}_{rr})_{dd}+\varepsilon}},
\]
so that the transformed offsets \(\tilde{\boldsymbol{r}}'_i=\boldsymbol{P}\ \tilde{\boldsymbol{r}}_i\) have comparable variances across axes. 
We then solve the Tikhonov regularized MLS in the transformed coordinates. 
We use \(\lambda'=\eta\,\operatorname{tr}(\boldsymbol{M}_{rr}')/3\) with \(\eta\in[10^{-4},10^{-2}]\). 
\[
\boldsymbol{M}_{rr}^{'} = \sum_i \omega_{ip} \tilde{\boldsymbol{r}}_i^{'} \tilde{\boldsymbol{r}}_i^{'T}
\quad \quad \quad 
\boldsymbol{M}_{vr}^{'} = \sum_i \omega_{ip} (\boldsymbol{v}_i - \bar{\boldsymbol{v}}) \tilde{\boldsymbol{r}}_i^{'T} 
\]
\[
\boldsymbol{B}_p^{'} = \boldsymbol{M}_{vr}^{'} (\boldsymbol{M}_{rr}^{'} + \lambda^{'} \boldsymbol{I})^{-1} 
\]
Since \((\boldsymbol{v}_i-\bar{\boldsymbol{v}})\approx \boldsymbol{B}_p\,\tilde{\boldsymbol{r}}_i=\boldsymbol{B}_p'(\boldsymbol{P}\tilde{\boldsymbol{r}}_i)\), we map back to the original coordinates by
\[\boldsymbol{B}_p = \boldsymbol{B}_p^{'} \, \boldsymbol{P}\]

\paragraph{3b. Improvement: Local Deformation Gradient Update} The original deformation gradient update is first-order accurate (global error \(O(\Delta t)\)) because it extrapolates along the tangent at the end of the step. 
Thus, the energy drift accumulates, and the Hamiltonian sometimes increases over time, which is not correct. 
To reduce it, we adopt the second-order mid-point (RK-2) scheme \cite{butcher2016ode_RK2_midpoint, kutta1901RK2}. 

First, compute the mid-point velocity gradient
\[ \boldsymbol{L}_p^{n+\frac{1}{2}} =  \nabla \boldsymbol{v}_p^{n+\frac{1}{2}} = \nabla [\frac{1}{2}(\boldsymbol{v}_p^n + \boldsymbol{v}_p^{n+1})] = \frac{1}{2} (\boldsymbol{L}_p^{n} + \boldsymbol{L}_p^{n+1})  \]

Then update \(\boldsymbol{F}\) in two sub-steps
\[\begin{gathered}
\boldsymbol{F}_{trial}^{n+\frac{1}{2}} = (\boldsymbol{I} + \frac{1}{2} \boldsymbol{L} \Delta t) \boldsymbol{F}_E^n \\ 
\boldsymbol{F}_{trial}^{n+1} = \boldsymbol{F}_E^n + \boldsymbol{L} \Delta t  \boldsymbol{F}_{trial}^{n+\frac{1}{2}} 
\end{gathered}\]
Combining the two steps yields
\[\boldsymbol{F}_{trial}^{n+1} = (\boldsymbol{I} + \boldsymbol{L} \Delta t + \frac{1}{2} \boldsymbol{L}^2 \Delta t^2)\boldsymbol{F}_E^n\]
The scheme is second-order accurate (global error \(O(\Delta t^2)\)), which greatly reduces energy drift in stiff materials.

\paragraph{4. Energy Monitoring} We compute the system's kinetic energy \(T\) , gravitational potential energy \(U_g\) , strain energy \(U_e\), and plastic dissipation \(\Delta W_p\) by
\[
T = \sum_p \frac{1}{2} m_p \lVert \boldsymbol{v}_p \rVert ^2 \quad \quad \quad
U_g = -\sum_p m_p \boldsymbol{g} \cdot \boldsymbol{x}_p 
\]
\[
U_e = \sum_p  V_p\Psi_p \quad \quad \quad
\Delta W_p = \sum_p V_p\sqrt{\tfrac{2}{3}}\,\tau_Y \Delta \gamma_p \]
so that the Hamiltonian \(H\) is \[H = T + U_g + U_e\]
We use these energies only to check the simulator. While we developed it, the Hamiltonian sometimes increased during a drop, which is not physical, and this led us to the problems that the improvements of this section fix. The predictors, the reference fit, and the evaluation do not use the energies.

\paragraph{5. Improvement: Adaptive Sub-Stepping of Time} 
Explicit time integration schemes must satisfy the Courant-Friedrichs-Lewy (CFL) condition \cite{courant1928CFL_condition,courant1967CFL_condition}
\[\Delta t \leq \frac{\alpha \Delta x}{c_{max} + \lVert \boldsymbol{v}_{max} \rVert}\]
where \(\Delta x\) is the grid spacing, \(c_{max} \approx \sqrt{\frac{E}{\rho}}\) is the maximum elastic wave speed, \(\lVert \boldsymbol{v}_{max} \rVert\) is the maximum grid velocity, and \(\alpha \leq 1\) is a safety factor.
In our simulations, we set
\[\Delta t = 0.4\frac{\Delta x}{c_{max}} = 0.4\Delta x \sqrt{\frac{\rho}{E}}\]
because \(\lVert\boldsymbol{v}_{max}\rVert \ll c_{max}\) for all tested cases.

\section{Scaling of Physical Parameters}
\label{sec:app_param_scaling}

The reconstructed scenes used for the reference fit and the replays need an absolute metric scale, which the RGB-D scan provides; \model{} itself reads no depth. To show why the scale matters, we analyze how a global rescaling of the reconstructed geometry affects the physical parameters used in simulation.
Specifically, suppose we isotropically scale the input point cloud by a factor \(s\). To keep the discrete equations dimensionally consistent, every quantity that carries the dimension of length must be rescaled by the appropriate power of \(s\).  Denoting dimensional exponents in the form \([\mathrm{L}^{\alpha}\mathrm{M}^{\beta}\mathrm{T}^{\gamma}]\), and keeping the time step \(\Delta t\) unchanged, we obtain
\[
\renewcommand{\arraystretch}{1.2}
\begin{array}{lll}
\text{Position }(\boldsymbol{x}) & [\mathrm L] & \boldsymbol{x}' = s\,\boldsymbol{x},\\
\text{Acceleration }(\boldsymbol g) & [\mathrm L\,\mathrm T^{-2}]
  & \boldsymbol g' = s\,\boldsymbol g,\\
\text{Young's modulus }(E) & [\mathrm L^{-1}\mathrm M\,\mathrm T^{-2}]
  & E' = s^{-1}E,\\
\text{Yield stress }(\sigma_Y) & [\mathrm L^{-1}\mathrm M\,\mathrm T^{-2}]
  & \sigma_Y' = s^{-1}\sigma_Y,\\
\text{Poisson's ratio }(\nu) & [1]
  & \nu' = \nu,\\
\text{Density }(\rho) & [\mathrm L^{-3}\mathrm M]
  & \rho' = s^{-3}\rho.
\end{array}
\]
Energy has dimension \([\mathrm L^{2}\mathrm M\,\mathrm T^{-2}]\); in the scaled space it therefore scales with \(s^{2}\). When we monitor the energies of a scaled scene (App.~\ref{sec:app_MPM}), we multiply them by \(s^{-2}\) to map them back to the original physical scale.

\section{Parameter Identifiability and Units}
\label{sec:app_identifiability}
Write the homogeneous constitutive stress, at fixed Poisson's ratio, as
\begin{equation}
\boldsymbol\sigma=E\,\boldsymbol\Phi(\boldsymbol F,\boldsymbol F_p,
\nu,\sigma_Y/E).
\end{equation}
Dividing momentum conservation by density gives a stress-acceleration contribution controlled by $E/\rho$, while the relative plastic threshold is $\sigma_Y/E=(\sigma_Y/\rho)/(E/\rho)$. A joint rescaling by $a>0$ of $E,\sigma_Y,\rho$ therefore leaves the continuum acceleration unchanged under these assumptions. In a drop, every external force scales with the mass: gravity, and the force with which the ground stops the object. No external force acts independently of the density, so the drop does not reveal the density itself. Even the teacher, which reads the exact 3D deformation, does not learn the density (Fig.~\ref{fig:app_four_heads}, App.~\ref{sec:app_controls}). This is distinct from geometrically scaling the scene in App.~\ref{sec:app_param_scaling}. Spatially varying density, unscaled external loads, constitutive terms with additional dimensional constants, and numerical/contact regularization can break the simple invariance.

Poisson's ratio still affects the simulation. It shows in how much the object expands sideways and changes its volume when it is compressed at impact. In exact 3D deformation this can be seen: the teacher, which reads it, predicts part of $\nu$ (Fig.~\ref{fig:app_four_heads}, top row). In a video, the change is so small that it is hard to see in pixels (App.~\ref{sec:app_nu}), and the student, which reads the video, does not learn $\nu$ (Fig.~\ref{fig:app_four_heads}, bottom row). We therefore fix $\nu$ per object (Table~\ref{tab:fixed_reference}). Likewise, a drop without plastic deformation shows only that the yield stress was not reached; it does not give its value. These limits are why we predict the two ratios and mask the yield term for drops without plastic deformation. They do not mean that the ratios determine every aspect of the motion.

All evaluation ratios use $E,\sigma_Y$ in Pa and $\rho$ in $\mathrm{kg/m^3}$, so $q_E$ and $q_Y$ have units $\mathrm{m^2/s^2}$. The numerical logarithms refer to these SI values. Let $z_E=\log_{10}q_E$ and $z_Y=\log_{10}q_Y$. The prediction heads use
\begin{equation}
\widetilde z_E=(z_E-3.5)/2.5,\qquad
\widetilde z_Y=(z_Y-3.1505)/1.1505.
\end{equation}
Decoding applies exactly the inverse transforms. To convert a predicted ratio into a modulus in Pa, we multiply it by the object's fixed density. This does not predict density. The error of $\log_{10}E$ then equals the error of $\log_{10}(E/\rho)$ only if the prediction and the reference use the same density.

\section{Data and Observation Processing Details}
\label{sec:app_data_pipeline}
\paragraph{Split.}
\model{}-Sim has 291,769 simulated videos. We remove 131 whose estimated depth cannot be calibrated to metric scale (see below), for every model, including those that do not read depth. The remaining 291,638 videos are split by source mesh into 277,050 training and 14,588 validation videos (0.95\,:\,0.05); no mesh appears on both sides. Each simulated trajectory is rendered from up to three cameras, and all views of a trajectory stay in the same split. The replays of the real drops (Table~\ref{tab:same_drops}) and the stiffness sweep (App.~\ref{sec:material_appearance}) are simulated in the reconstructed real scenes and belong to neither split.

\paragraph{Background and camera effects.}
Each video is composited on one photograph from the MIT Indoor67 training set~\cite{quattoni2009indoor}, chosen by its sample seed. Premultiplied foreground color and alpha are warped together with stored flow, accumulated across 11 exposure samples, and composited on the photograph. Color is approximately linearized with exponent 2.2 before accumulation. Exposure is sampled in $[0.5,1.0]$ frame intervals. H.264 uses yuv420p, CRF 23--28, and the source frame rate.

Flow, depth, and masks are estimated from these composited, blurred, and compressed videos, not from the clean renders. SAM~3.1 is prompted with one box per frame. In simulation the box comes from the simulator's silhouette of the object, enlarged to cover the exposure interval and by a margin of at most 4 pixels or 5\%; on real videos we annotate the boxes by hand on the RGB frames. Masks select and validate the tracked points and define the crop. RGB, flow, and depth keep their background.

\paragraph{Temporal window and crop.}
On the real videos, we chose the collision frame $C$ by hand, once per video. It is the same for every model. The window has 16 RGB and depth frames, and a 17th RGB frame for the 16th forward flow. The object masks take up only a small portion of the original $960\times720$ frames, with different distributions in the synthetic (2.7\%) and real videos (1.2\%), so the same cropping rule is applied in both domains. A fixed union-of-masks square crop with 15\% margin is applied to every frame and modality; individual frames are not recentered. The window is resized to $224\times224$. Flow vectors are rescaled into the same crop coordinates. Flow and depth are estimated on the full frames and cropped afterwards.

\paragraph{Depth alignment.}
We need to recover the metric scale of relative depth; App.~\ref{sec:app_param_scaling} explains why scale matters for simulation and for the predicted ratios. For predicted camera-$z$ depth $\widetilde d$, reference depth $d_{\rm ref}$, and valid paired calibration pixels $\mathcal P$, we use one positive scale for the whole clip, a robust scalar alignment:
\begin{equation}
s=\operatorname{median}_{p\in\mathcal P}
 \frac{d_{\rm ref}(p)}{\widetilde d(p)},\qquad d(p)=s\widetilde d(p).
 \label{eq:depth_scale}
\end{equation}
Synthetic references use rendered metric depth; re-simulation references use the corresponding renderer depth. On real clips, $d_{\rm ref}$ is registered LiDAR depth and support is static background outside the union of object masks dilated by a $21\times21$ kernel. The scale is estimated on the first frame of the window that has at least 20 valid pairs (in simulation, among its first three frames) and is then kept for the whole clip. Simulated videos without such a frame are the 131 removed above. This is scale calibration, not a per-frame scale-and-shift fit: it avoids fitting a separate scale to the deforming foreground at every time step. No depth shift is added.

4RC is run on the 16 frames of the window. Invalid depth is set to zero and marked as invalid; valid background depth is kept.

Depth is normalized with the mean and standard deviation of valid foreground log-depth in the training set:
\begin{equation}
z_d=\frac{\ln d-0.3373}{0.2170}.
\end{equation}
The same constants are applied to the full valid crop in training and evaluation.

\paragraph{Flow and RGB encoding.}
RGB is divided by 255 and normalized using the pretrained V-JEPA processor statistics. Raw WAFT displacement is converted to crop-relative displacement $f$. Its signed compression is
\begin{equation}
z_f=\operatorname{asinh}(f/0.0762027679)/\operatorname{asinh}(1).
\end{equation}
The dense-flow branch does not remove the dominant affine component; the non-affine residual cue (App.~\ref{sec:app_cue_details}) removes it locally. Invalid pixels and cue observations are explicitly marked.

\begin{figure*}[t]
\centering
\includegraphics[width=\textwidth]{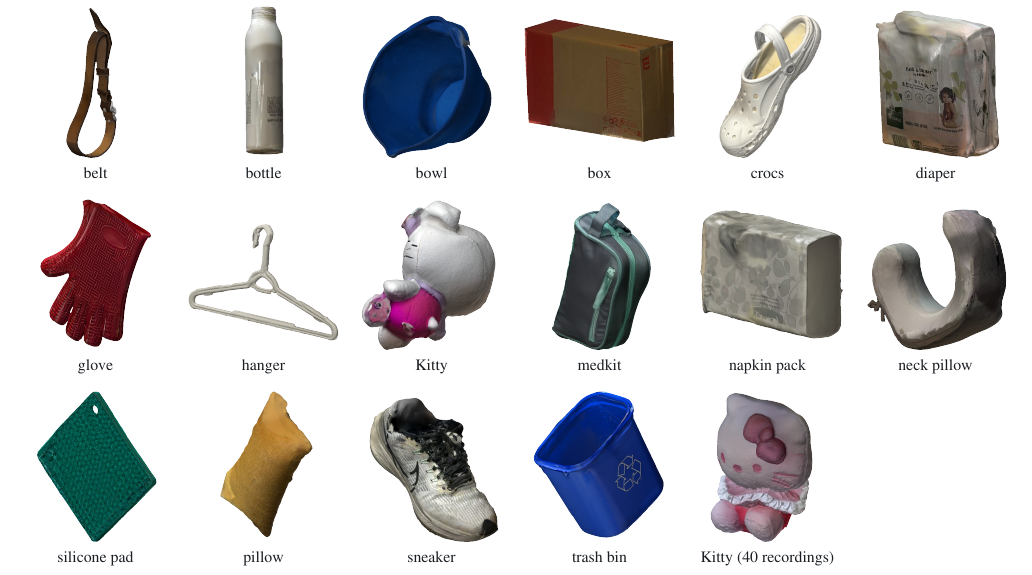}
\caption{\textbf{Reconstructed object meshes} used in our simulations (App.~\ref{sec:app_real_capture}): the 16 objects of \model{}-Real and, last, the Kitty of the 40 additional recordings (App.~\ref{sec:app_stability_metrics}). Each mesh is shown from the side that it is best seen from, with its vertex colors; the meshes are not drawn to a common scale.}
\label{fig:app_meshes}
\end{figure*}

\paragraph{Real capture and scene reconstruction.}
\label{sec:app_real_capture}
We place each object in a gripper, scan the static scene with Record3D, then open the gripper to start the drop and record it with the camera fixed.
The RGB-D sensor gives the absolute scale of the scene, on which the inferred parameters depend (App.~\ref{sec:app_param_scaling}).
We reconstruct the object mesh from the scan with the pipeline of ReconViaGen~\cite{chang2025reconviagen}: VGGT~\cite{Wang2025VGGT} gives a point cloud from multi-view images of the object, which is passed to the second and third stages of TRELLIS.2~\cite{trellis2repo}.
This gives one candidate mesh per scan, five per object; we pick the best one by hand, trim outliers, voxelize it, and fill holes from coarse to fine before converting it back to a mesh. Fig.~\ref{fig:app_meshes} shows the meshes of the 16 objects and of the Kitty of the 40 additional recordings (App.~\ref{sec:app_stability_metrics}).
From the same scan, we also reconstruct the whole static scene as 3D Gaussians, with the gsplat implementation of Gaussian splatting in Nerfstudio. Among the baselines, MonoPhysics needs the object as 3D Gaussians; we initialize them from the particles of our object mesh (App.~\ref{sec:app_baseline_adaptation}). In these Gaussians, we segment the wooden platform and the object by hand. We align the scene so that the platform lies on the ground plane of the MPM simulation, at its center, with the $x$--$z$ plane as the floor and the $y$ axis vertical, so that gravity is $\boldsymbol g=[0,-9.8,0]^\top$. The side of the platform on which the object lies sets the direction of the platform normal, that is, which way is up. The mesh is then placed at the segmented object, and its pose is refined for each drop against the video (Sec.~\ref{sec:reference_protocol}).

\section{Video and Cue Architecture}
\label{sec:app_architecture}
\label{sec:appendix_mlp_heads}

\subsection{Deformation cue definitions}
\label{sec:app_cue_details}

\paragraph{Short correspondences.}
We use short correspondences derived from optical flow rather than long correspondences from a point tracker, because the long-range trackers that we tested are not stable across our drops. On the 80 real drops, we tested eleven long-range trackers and tracker modes (App.~\ref{sec:app_more_trackers}). Each of them covers less than half of the object at the end of the window, $C+12$, in some drops: in 10 drops for the best one (CoWTracker), and in 16 to 67 drops for the others. Our short tracklets restart at window frames 0, 4, 8, and 11 and cover more than half of the object at $C+12$ in every drop. Fig.~\ref{fig:app_tracking} looks at one tracker in detail, CoTracker3, a successor of CoTracker~\cite{karaev2024cotracker}, in its online mode (App.~\ref{sec:app_qualitative}). We start its points before the collision. About ten frames after the collision, it reports most of these points as no longer visible, because the object rotates and blurs. Tracks are propagated with Eq.~\eqref{eq:short_tracks}.

\paragraph{2D tracklets.}
The models on 2D tracklets (RGB + 2D tracklets and \model{}) read the 2D tracklets of the main paper directly, without derived cues. For each point $i$ and each frame $t$ of a tracklet, it receives the position $\boldsymbol p_{i,r}$ of the point in the reference frame $r$, in which the tracklet starts, and the displacement from that position, both divided by the crop size of 224 pixels:
\begin{equation}
 \boldsymbol{g}^{\rm raw}_{i,t}
 =\left[\frac{\boldsymbol{p}_{i,r}}{224},
         \frac{\boldsymbol{p}_{i,t}-\boldsymbol{p}_{i,r}}{224}\right].
 \label{eq:raw_trajectory}
\end{equation}
This input contains the local reference geometry and the motion of every point, which is what is needed to compute deformation. It is therefore more than the motion of the object's centroid. The control uses the same graph and geometry encoder as the cue models.

\paragraph{Local affine fit and 2D strain.}
For a point $i$ and one of its neighbors $j$, define the offset between the two points in the reference frame $r$ and in the current frame $t$:
\begin{equation}
 \boldsymbol{a}_{ij}=\boldsymbol{p}_{j,r}-\boldsymbol{p}_{i,r},
 \qquad \boldsymbol{b}_{ij,t}=\boldsymbol{p}_{j,t}-\boldsymbol{p}_{i,t}.
\end{equation}
We robustly fit a local affine map from the reference offsets to the current offsets. The fit runs over the $k$ nearest neighbors $\mathcal N_k(i)$ of point $i$, with the weights $w_{ij,t}$ described below:
\begin{equation}
 \boldsymbol{A}_{i,t}^{(k)}
 =\operatorname*{arg\,min}_{\boldsymbol A}
 \sum_{j\in\mathcal N_k(i)}w_{ij,t}
     \|\boldsymbol{b}_{ij,t}-\boldsymbol A\boldsymbol{a}_{ij}\|_2^2.
 \label{eq:app_local_affine}
\end{equation}
We fit the map four times: first with Gaussian weights, then after each of three reweighting steps. The initial weight $w_{ij,t}$ is zero for a neighbor that is not valid in frame $t$. For a valid neighbor, it is a Gaussian function of the distance between the neighbor and point $i$ in the reference frame. At each reweighting step, the weight of neighbor $j$ is multiplied by an additional factor $\min(1,\delta/\max(e_j,\epsilon))$. Here $e_j$ is the residual of neighbor $j$ in the previous fit, $\delta$ is 1.5 times the median residual of the valid neighbors, and $\epsilon$ is a small numerical floor. We reject a fit that has fewer than six valid neighbors. We also reject a fit whose reference Gram matrix, the Gram matrix of the weighted reference offsets $\boldsymbol a_{ij}$, is ill-conditioned: the smallest eigenvalue of this matrix must exceed $10^{-3}$ times its largest eigenvalue, and it must exceed $10^{-8}$ in crop-pixel units. We also require that the fitted map is finite and that the eigenvalues of $\boldsymbol A^\top\boldsymbol A$ lie between $10^{-6}$ and $10^4$. The same checks are applied in simulation and on real videos.

The local 2D logarithmic strain is
\begin{equation}
 \boldsymbol H_{i,t}^{(k)}=\tfrac12\log\left((\boldsymbol A_{i,t}^{(k)})^\top
                                  \boldsymbol A_{i,t}^{(k)}\right),
 \qquad \boldsymbol g^{\rm strain}_{i,t}
       =[H_{11},H_{22},H_{12}].
 \label{eq:app_strain2d}
\end{equation}
For $\boldsymbol A^\top\boldsymbol A=\boldsymbol V\operatorname{diag}(\lambda_1,\lambda_2)\boldsymbol V^\top$, the matrix logarithm is computed as $\boldsymbol H=\tfrac12\boldsymbol V\operatorname{diag}(\ln\lambda_1,\ln\lambda_2)\boldsymbol V^\top$. $\boldsymbol H$ depends on $\boldsymbol A$ only through $\boldsymbol A^\top\boldsymbol A$, so a pure in-plane rotation of the neighborhood does not change it. The three strain components are computed for $k=8$ and $k=24$, giving six channels.
This strain is an image-plane deformation cue, not a measurement of the full 3D strain tensor; perspective and out-of-plane motion can still affect it.

\paragraph{2D stretch.}
An edge is a pair of a point $i$ and one of its fixed neighbors $j$. For the same edges as in the affine fit, define the length $\ell_{ij,t}$ of an edge in frame $t$ and its logarithmic stretch $s_{ij,t}$ relative to the reference frame $r$:
\begin{equation}
 \ell_{ij,t}=\|\boldsymbol p_{j,t}-\boldsymbol p_{i,t}\|_2,
 \qquad s_{ij,t}=\ln\frac{\ell_{ij,t}}{\ell_{ij,r}}.
 \label{eq:app_edge_stretch}
\end{equation}
For each neighborhood size $k$, we summarize the stretches of the edges of point $i$ by their mean, their standard deviation, and their 10\% and 90\% quantiles:
\begin{equation}
 \boldsymbol S_{i,t}^{(k)}=
 [\operatorname{mean}(s),\operatorname{std}(s),Q_{0.1}(s),Q_{0.9}(s)]_{j\in\mathcal N_k(i)}.
\end{equation}
Concatenating $\boldsymbol S^{(8)}$ and $\boldsymbol S^{(24)}$ gives eight channels. Each summary requires at least $\max(4,k/2)$ valid edges. We use natural logarithms for deformation cues and base-10 logarithms for parameter errors.

\paragraph{2D non-affine residual.}
The affine fit does not explain every local motion. We measure the remaining motion with a dimensionless residual:
\begin{equation}
 r_{i,t}^{(k)}=
 \sqrt{\frac{\sum_{j\in\mathcal N_k(i)}w_{ij,t}
 \|\boldsymbol b_{ij,t}-\boldsymbol A_{i,t}^{(k)}\boldsymbol a_{ij}\|_2^2}
 {\max(\sum_{j\in\mathcal N_k(i)}w_{ij,t}\|\boldsymbol a_{ij}\|_2^2,\epsilon)}}.
 \label{eq:app_nonaffine}
\end{equation}
The non-affine cue is the pair $[r^{(8)},r^{(24)}]$. A model that reads only this cue receives neither the fitted affine map nor the strain channels, so it tests specifically the motion that remains after the local affine fit. The residual can contain non-affine deformation, correspondence error, and occlusion effects; a large value is not by itself evidence of a material response.

\paragraph{3D tracklets.}
We lift the same tracks to 3D with the camera intrinsics $\boldsymbol K$ and the calibrated depth $d_t$, that is, estimated depth scaled to metric (App.~\ref{sec:app_data_pipeline}):
\begin{equation}
 \boldsymbol X_{i,t}=d_t(\boldsymbol p_{i,t})\,
       \boldsymbol K^{-1}[u_{i,t},v_{i,t},1]^\top.
 \label{eq:lift}
\end{equation}
Here $(u_{i,t},v_{i,t})$ is the position of the point $\boldsymbol p_{i,t}$ in the original, uncropped image. We obtain it by undoing the crop and the resize, taking the pixel-center convention into account.
The 3D tracklets describe each point by its lifted reference position and its 3D displacement, in metres and normalized with statistics of the training set, in place of the 2D position and displacement of Eq.~\eqref{eq:raw_trajectory}.

\subsection{Branch architecture}
\label{sec:app_branch_architecture}
Our models have three branches: RGB, geometry, and shared context.
The geometry branch reads either the 2D tracklets (RGB + 2D tracklets and \model{}) or the deformation cues of App.~\ref{sec:app_cue_details} (the cue models). We describe the cue version here; the input of the tracklet models is given in App.~\ref{sec:app_impl_details}. The geometry branch reads the four tracklets (tracking windows), each with five time steps and up to 128 tracked points.
At each point and time, the combined 2D representation contains six strain, eight stretch, and two non-affine residual channels, computed at neighborhood sizes of 8 and 24.
We standardize each cue channel using fixed statistics computed from valid observations in the synthetic training split.
We then append three cue-validity indicators, one per cue family, and the five context values defined below. This gives a 24-dimensional input ($16+3+5$).
Unavailable cue families (\eg, not enough neighbors) are zeroed after standardization and identified by their validity indicators.
A pointwise linear layer followed by GELU maps this input to a 128-dimensional feature.
This width is fixed across cue variants; only the input projection size changes.

Three spatial-temporal blocks update these point features.
Within each block, EdgeConv~\cite{wang2019dgcnn} computes messages from the same 24 reference neighbors, aggregates them by channel-wise maximum over valid neighbors, and adds the result through a residual connection with GELU.
A one-dimensional convolution then processes each point along its own track, with kernel size three and dilations 1, 2, and 4 in successive blocks.
The output of this convolution is rescaled by the number of valid time steps among its inputs. LayerNorm, GELU, and a residual connection follow.
Invalid point outputs remain zero.
Let $\boldsymbol h_{w,t,i}\in\mathbb R^{128}$ be the final feature of point $i$ at time $t$ in window $w$, and let $\mathcal V$ contain the entries with at least one valid cue family.
We pool over all windows, times, and points:
\begin{equation}
 \boldsymbol f_G =
 \left[
 \frac{1}{|\mathcal V|}\sum_{(w,t,i)\in\mathcal V}\boldsymbol h_{w,t,i},\quad
 \max_{(w,t,i)\in\mathcal V}\boldsymbol h_{w,t,i}
 \right]\in\mathbb R^{256},
 \label{eq:geometry_pooling}
\end{equation}
where the maximum is channel-wise and an empty valid set gives a zero geometry feature.

The RGB branch processes the unmasked clip with V-JEPA~2 ViT-L~\cite{assran2025vjepa2} and LoRA adaptation.
Mean pooling over its output tokens, followed by LayerNorm, gives a 1024-dimensional video feature $\boldsymbol f_R$.
RGB uses its pretrained input normalization, independently of the cue statistics.
The context consists of five values: the time relative to the collision, the frame interval $\Delta t$, the ratio of crop size to focal length in each of the two image axes, and $g_{\mathrm{grav}}\Delta t^2$ with $g_{\mathrm{grav}}=9.81\,\mathrm{m/s^2}$.
The frame interval and gravity term are divided by fixed constants $1/30\,\mathrm{s}$ and $0.01\,\mathrm{m}$, respectively, before encoding.
The same five context values are supplied to all points at a given window and time in the geometry branch.
The context branch is separate from the geometry branch. It takes the five context values at each of the five time steps of the four tracklets ($4\times5\times5$ values), flattens them to 100 dimensions, and maps them with a linear layer and GELU to a 32-dimensional video feature $\boldsymbol f_{\mathrm{ctx}}$.

Late fusion concatenates the three video features into a 1312-dimensional vector for the regression MLP in Eq.~\eqref{eq:model_late_fusion}.
The constants in that equation are fixed target normalization parameters.
RGB-only omits the geometry feature while retaining the same RGB adaptation and context branch.

\subsection{Implementation details}
\label{sec:app_impl_details}

\paragraph{Image branches.}
V-JEPA 2 ViT-L supplies 1,024-dimensional tokens. Rank-16 LoRA with scale 32 adapts query/value projections in the last half of the backbone; the input projection and the LayerNorm parameters are also trainable. The image branches for dense depth and dense flow use the same backbone; they adapt its input projection to one depth channel or two flow channels. Each image branch has its own LoRA weights and its own trainable parameters. The frozen weights are identical in all branches and can share memory, but each branch computes its own adapted features. The tokens of each branch are mean-pooled and passed through a LayerNorm without learned scale and bias before concatenation. The two regression heads are MLPs with hidden widths 512, 256, and 128 and no dropout, one for $z_E$ and one for $z_Y$.

\paragraph{Track construction.}
Four five-frame tracklets start at window indices 0, 4, 8, and 11. Each starts with 128 points sampled on the object mask. We sample them from the mask eroded with a $3\times3$ kernel if at least 32 pixels remain after the erosion, and from the original mask otherwise. Each point is advanced to the next frame with the WAFT flow, sampled bilinearly at its position. A point counts as visible in a frame if its flow is finite, it lies inside the image, and it lies on the object mask. These tracklets are not exact physical correspondences, and we apply no forward--backward check. The graph connects each point to its 24 nearest neighbors in the reference frame and stays the same throughout the tracklet.

\paragraph{Robust local fits.}
For each neighborhood size $k\in\{8,24\}$, the affine fit of App.~\ref{sec:app_cue_details} starts from Gaussian weights in reference distance and refines them with three Huber reweighting steps. A fit is kept only if it has at least six valid neighbors, the ratio of the smallest to the largest eigenvalue of the reference Gram matrix is greater than $10^{-3}$, and the singular values of the fitted map $\boldsymbol A$ lie between $10^{-3}$ and $100$. These are the checks of App.~\ref{sec:app_cue_details}; the bounds on the singular values of $\boldsymbol A$ are the bounds given there on the eigenvalues of $\boldsymbol A^\top\boldsymbol A$. Invalid fits are excluded, not interpreted as zero deformation. Strain, stretch, and the non-affine residual are computed for both neighborhood sizes, giving 6+8+2=16 channels. The logarithm in Eq.~\eqref{eq:app_strain2d} is the matrix logarithm of the symmetric positive-definite matrix $\boldsymbol A^\top\boldsymbol A$, not an element-wise logarithm.

\paragraph{Geometry encoder and metadata.}
Each cue channel is normalized with a fixed mean and standard deviation computed on the synthetic training split. Each cue family (strain, stretch, non-affine residual) has its own validity flag. All cue variants receive the same timing and scale context: the time relative to the collision, the frame interval, the ratio of crop side to focal length in each image axis, and $g\Delta t^2$. The cue model projects 16 cue channels, three validity flags (one per cue family), and five context values; the tracklet models project the four tracklet channels of Eq.~\eqref{eq:raw_trajectory} instead. Three blocks of width 128 follow. Each applies EdgeConv and then a temporal convolution with kernel size 3; the dilations of the three blocks are 1, 2, and 4. Masked mean and max pooling produce 256 features; concatenating the RGB feature (1,024) and the context feature (32) gives 1,312 features before the output heads. The model does not sample RGB tokens at track locations.

\paragraph{Loss and optimizer.}
Let $B$ be a minibatch and $B_Y$ its yield-excited subset, the videos whose window contains plastic excitation (rule below). The loss of the direct predictors is
\begin{equation}
\mathcal L_{\rm FF}=\frac{1}{|B|}\sum_{i\in B}
 (\widehat{\widetilde z}_{E,i}-\widetilde z_{E,i})^2+
 \frac{1}{\max(1,|B_Y|)}\sum_{i\in B_Y}
 (\widehat{\widetilde z}_{Y,i}-\widetilde z_{Y,i})^2.
\end{equation}
Yield supervision requires the recorded yielded state and a positive plastic-energy increment in the selected window under the implemented normalized $10^{-8}$ threshold. This rule only detects plastic excitation in the window. It does not prove that the video determines the yield stress uniquely. The stiffness term and the yield term have equal weight.

We train with AdamW. The learning rate is $10^{-4}$ for the adapted backbone, the geometry encoder, and the context branch, and $3\times10^{-4}$ for the heads. Weight decay is 0.05, the betas are (0.9,0.999), and epsilon is $10^{-8}$. Gradients are clipped at 1, and computation uses bfloat16. The seed is 1701. A warm-up over 5\% of the steps is followed by cosine decay to 5\% of the initial learning rate over four epochs; the fifth epoch keeps this final rate. Each GPU holds four videos, and gradients are accumulated to an effective batch of 480 videos; the number of GPUs differs between runs (for example, 10 accumulation steps on 12 GPUs). All direct predictors share this schedule and are reported after the fifth epoch; no checkpoint is selected on real videos. The teacher and the students are described in App.~\ref{sec:app_oracle} and App.~\ref{sec:app_teacher_student}.

\section{Teacher on Exact Surface Deformation}
\label{sec:app_oracle}
The teacher reads the motion of points on the simulated object surface. For each video we take up to 1,024 surface particles of the simulation and their positions in the 16 frames of the window. The particles are the same in every frame, so the tracks are exact. Positions are centered and divided by the object size in the first frame. Each point has up to 16 fixed neighbors on the surface, chosen in the first frame with tangent and normal checks. From these positions we compute ten features per point and frame: the two principal logarithmic strains of the surface, the area change, the strain anisotropy, the rates of the strains and of the area change, bending, the residual of the local affine fit, and a validity flag. Edges carry the reference length, the stretch, the agreement of normals, bending, and validity. These features describe how the surface deforms; they are not the full volumetric deformation gradient. The teacher does not read the solver's $\boldsymbol F_e$, $\boldsymbol F_p$, stress, or energy, and it does not read RGB, absolute coordinates, density, or Poisson's ratio.

The network has four 128-wide blocks. Each block passes messages over the fixed neighbors and then applies a temporal convolution along each track, with dilations 1, 2, 4, and 8. Attention pooling and mean pooling over all points and frames give 256 features. A small branch encodes three context values into 32 features: the logarithm of the object size, the logarithm of the frame interval, and the vertical speed before impact. Concatenating the 256 and the 32 features gives a 288-dimensional vector. This vector is the feature $\boldsymbol f_T$ that the student regresses (Sec.~\ref{sec:physreader}). One 128-wide layer and two linear outputs map it to $z_E$ and $z_Y$. The teacher has 572,451 parameters.

We train the teacher with the Smooth L1 loss ($\beta=0.25$), with the yield term masked as in App.~\ref{sec:app_architecture}, using AdamW with learning rate $3\times10^{-4}$, weight decay $10^{-4}$, and gradient clipping at 5. The learning rate follows a cosine schedule to 10\% over five epochs and keeps that rate for five more. We keep the epoch with the lowest sum of $z_E$ and $z_Y$ errors on synthetic validation, epoch 8; no real video enters this choice. The training split provides 107,039 simulated trajectories with usable surface geometry, which cover 98.9\% of the training videos. On validation, 14,293 of the 14,588 videos have usable geometry, and the teacher is scored on those (Table~\ref{tab:oracle}).

Table~\ref{tab:oracle} also lists a teacher that sees only the camera-visible part of the surface, with neighborhoods rebuilt from the retained points. It is trained in the same way, and its best validation epoch is the tenth. It has usable geometry for 10,202 validation videos. Table~\ref{tab:error_budget} evaluates every one of its rows on these 10,202 videos. This is why the full-surface teacher has a $z_E$ error of 0.145\,dex in that table and of 0.161\,dex on all 14,293 videos with usable geometry.

The row with absolute targets predicts $E$ and $\sigma_Y$ in Pa. This is clearly worse than predicting the density-normalized ratios (0.272 against 0.161\,dex for stiffness), in line with App.~\ref{sec:app_identifiability}: a drop does not determine density, so the absolute parameters are not determined either. Exact points remove perception error, but not simulator bias, finite sampling, or the limits of the architecture, so the teacher's error is not an information-theoretic bound.

\begin{table}[t]
\centering\small
\caption{\textbf{Teachers on exact surface deformation.} $z_E$ / $z_Y$ error (dex) on synthetic validation; for the row with absolute targets, the error of $\log_{10}E$ / $\log_{10}\sigma_Y$. All rows are trained on \model{}-Sim for ten epochs and selected on synthetic validation.}\label{tab:oracle}
\begin{adjustbox}{max width=\linewidth}
\begin{tabular}{@{}llc@{}}\toprule
Input and target & Validation videos & $z_E$ / $z_Y$ $\downarrow$ \\\midrule
Full surface (teacher of \model{}) & 14,293 with geometry & 0.161 / 0.192 \\
 & 10,202 of Table~\ref{tab:error_budget} & 0.145 / 0.171 \\
Visible surface only & 10,202 & 0.217 / 0.252 \\
Full surface, absolute $E$, $\sigma_Y$ & 14,293 with geometry & 0.272 / 0.262 \\
\bottomrule\end{tabular}
\end{adjustbox}
\end{table}

\section{Optimization and Conditional Range Search}
\label{sec:app_optimization}
\paragraph{Variables and alternation.}
Each object shares $(E,\sigma_Y)$ across its five drops. Each drop has a 12-dimensional initial state: position, rotation, linear velocity, and angular velocity. Density, Poisson's ratio, geometry, cameras, and contact settings are not fit. They are fixed for each scene (Table~\ref{tab:fixed_reference} lists the density and Poisson's ratio of each object). The fit starts from a grid search whose best candidate was chosen by visual rating (App.~\ref{sec:app_reference_objective}), and it uses the scanned scene, the reconstructed mesh, and the calibrated camera. It is therefore not an automatic estimator from RGB alone, and the references are parameters fit in our simulator, not independent measurements.

The Bayesian search over the material uses a Gaussian process with Mat\'ern-$5/2$ covariance, a noise regularizer of $10^{-4}$, and normalized responses. It selects new candidates by expected improvement with an exploration offset of 0.002. Each round proposes 12 material candidates; three promising branches receive eight initial-state proposals per trial. At most four rounds are run. The search stops earlier after two rounds without improvement, provided that a minimum number of rounds has been completed. The candidate pool contains 3,000 proposals.

\begin{table}[t]\centering\small
\caption{\textbf{Fixed inputs of the reference fit.} Densities are our measurements where available (App.~\ref{sec:app_real_measure}) and common values otherwise. Poisson's ratios are estimates made before the tensile data were processed, or common values; App.~\ref{sec:app_real_measure} lists the apparent Poisson's ratios measured in the tensile tests; App.~\ref{sec:app_nu} explains why we keep the values of this table and shows the effect of Poisson's ratio on the simulation. Both are assumptions of the fit. $n_Y$: number of drops whose reference simulation yields.}\label{tab:fixed_reference}
\begin{tabular}{lccc}\toprule
Object & $\rho$ (kg/m$^3$) & $\nu$ & $n_Y$\\\midrule
belt & 700 & 0.35 & 0\\
bottle & 950 & 0.35 & 5\\
bowl & 900 & 0.35 & 5\\
box & 259 & 0.3 & 5\\
crocs & 288 & 0.271 & 5\\
diaper & 130 & 0.25 & 5\\
glove & 726 & 0.412 & 0\\
hanger & 482 & 0.35 & 5\\
kitty & 70 & 0.288 & 0\\
medkit & 300 & 0.3 & 5\\
napkin & 233 & 0.25 & 5\\
neck pillow & 80 & 0.25 & 0\\
pad & 396 & 0.294 & 0\\
pillow & 80 & 0.25 & 0\\
sneaker & 500 & 0.3 & 5\\
trash bin & 945 & 0.3124 & 5\\
\bottomrule\end{tabular}\end{table}

\paragraph{Observation terms.}
The reference fit uses SEA-RAFT flow~\cite{wang2024sea_raft} with a two-pixel forward--backward criterion. The trajectory term compares the 3D motion of the visible surface. On the real video, SEA-RAFT short tracks are lifted to 3D with DepthCrafter depth aligned to metric scale; in the simulation, the rendered surface points are used. The 3D parts of the trajectory and pose terms use only the rigid motion of the object, which is fit robustly to all tracked points in each short window (App.~\ref{sec:app_reference_objective}). This robust fit averages out the depth noise at individual points, so it describes the overall motion of the object well. It does not describe the local deformation, which the image-plane terms below compare. We did not repeat the reference fit with another depth estimator. Over short windows, the term compares the direction of the 3D velocity, its vertical part, and its horizontal speed, with weights 0.55, 0.30, and 0.15. The pose term combines the relative 3D rotation over short windows (weight 0.6, used only where it is reliable) with the distance between the silhouette boundaries (weight 0.4). The deformation and short-track terms compare image-plane cues.

The deformation term $\mathcal L_{\mathrm{deform}}$ works on the flow that remains after the dominant affine motion of the object in the image is removed. A sigmoid turns the magnitude of this residual flow into a soft support, that is, a soft mask of the pixels with residual motion. The sigmoid is centered at 0.01 object lengths and has width 0.0025. Its spatial extent is integrated in a boundary band from 2 pixels to $2+0.08L$, where $L$ is the image object scale, and normalized by boundary length/object scale. The simulated and real thickness curves are compared using a scale floor of 0.01. This is a projected residual-flow thickness, not a 3D material thickness or a per-particle strain measurement.

The short-track term $\mathcal L_{\mathrm{short}}$ starts new five-frame tracks every two frames around impact. For pairs of tracked points, it takes the absolute change of the logarithm of their distance after 2, 3, and 4 frames, and it uses the 75th percentile of these changes. Penalties on low coverage and on missing tracks prevent a simulation from scoring well simply because difficult tracks are lost. The prior term $\mathcal L_{\mathrm{prior}}$ penalizes the departure of each initial state from its initialization (App.~\ref{sec:app_reference_objective}). The reference fit can use frames beyond the predictor's 16-frame window; the replay metrics of Sec.~\ref{sec:benchmark} use the 16 frames from $C-3$ (App.~\ref{sec:app_behavior}).

\subsection{Local acceptance sets}
\label{sec:app_range}
The global material search domain is $\log_{10}(E/\mathrm{Pa})\in[4,9]$ and $\log_{10}(\sigma_Y/\mathrm{Pa})\in[5,\log_{10}(2\times10^7)]$. The local search starts from the reference material, which we call the center. It searches outward along 32 directions in the $(\log_{10}E,\log_{10}\sigma_Y)$ plane: first rays along the axes and the diagonals, then directions at intermediate angles. Along each direction, it advances in steps of 0.05\,dex for at most ten steps. At the first rejected sample, it bisects the interval between the last accepted and the first rejected sample twice. The search along a direction ends at the boundary between accepted and rejected samples, at the largest sampled radius, at the border of the global domain, or at a numerical limit of the simulation.

Let $L_0$ be the loss at the center. We perturb the observation in four ways: the masks are eroded or dilated by one pixel, and the flow-consistency threshold is set to 1.5 or 2.5 pixels. Let $\delta_j$ be the 90th percentile of the change of loss component $j$ under these four perturbations. The total-loss allowance is
\begin{equation}
\tau=\max\!\left(\operatorname{mean}_{\rm drops}2\sum_jw_j\delta_j,
0.05|L_0|,0.01\right).
\end{equation}
A candidate material is accepted if its loss satisfies $L\leq L_0+\tau$ and it passes three further checks. First, its trajectory loss and its pose loss may each exceed the value at the center, $L_{j,0}$, by at most $\max(0.05\max(L_{j,0},0.05),2\delta_j,0.01)$. Second, its object-center error before impact must stay below $\max(8\,\mathrm{px},1.05e_0+1\,\mathrm{px})$, where $e_0$ is this error at the center. Third, its fraction of missing tracks may exceed that of the center by at most 0.1; the limit is capped at 1. Initial states remain fixed during this local search.

\begin{figure*}[t]
    \centering
    \includegraphics[width=\textwidth]{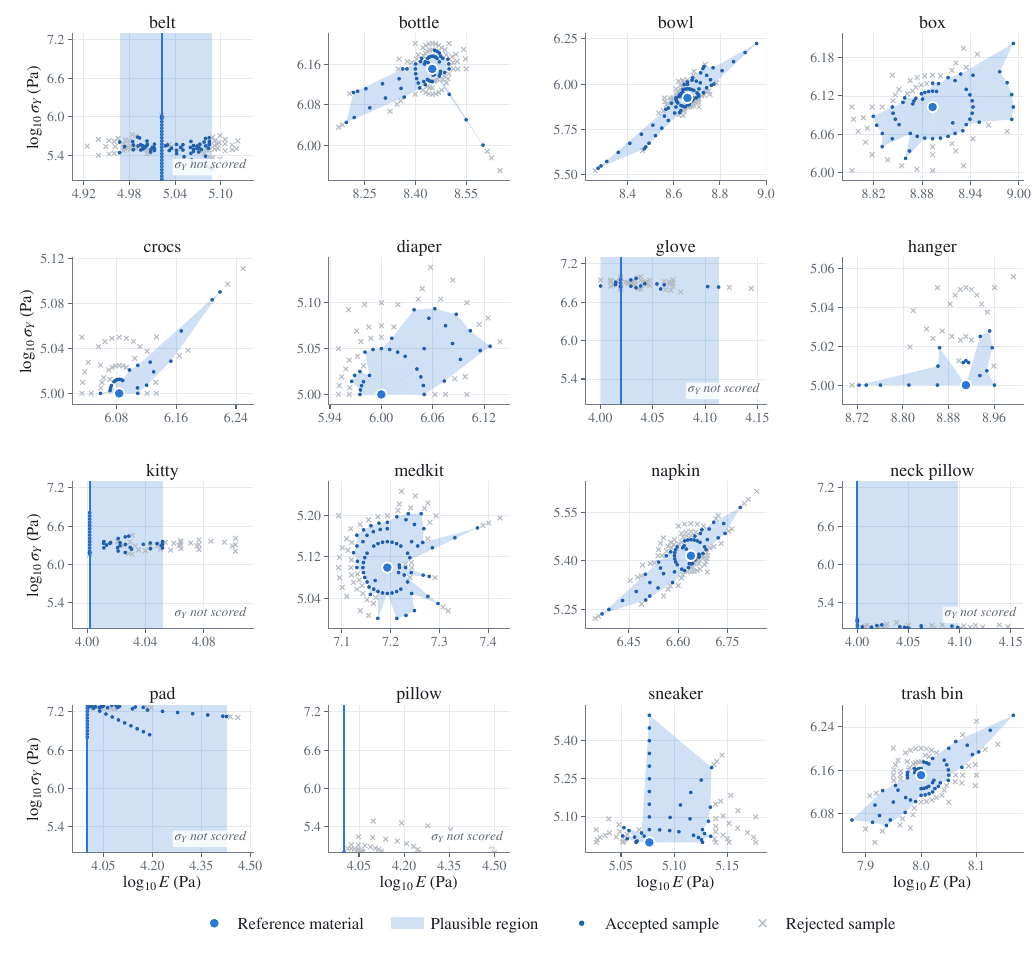}
    \caption{\textbf{Reference material and plausible region of every object.} Dots are sampled materials that pass the acceptance test, crosses are rejected samples, and the shaded area is the plausible region used by the parameter distance (Eq.~\ref{eq:plausible_region_distance}). For the six objects whose reference simulation does not yield, yield stress is not scored: the region is an interval of stiffness, drawn as a band. Axes differ between panels.}
    \label{fig:plausible_regions}
\end{figure*}

Fig.~\ref{fig:plausible_regions} shows the samples and the resulting region of every object, and Table~\ref{tab:ranges} lists their extent.

In total, the search evaluates 1,337 materials and accepts 675. Table~\ref{tab:ranges} reports the extent of the accepted samples, not a continuous rectangular feasible region or a confidence interval. For Pillow, every sample 0.1\,dex from the reference was rejected, so its region is the reference stiffness itself at that resolution.
\begin{table*}[t]\centering\small
\caption{\textbf{Reference materials and plausible regions.} All values are $\log_{10}$ of Pa. The region columns give the smallest and the largest accepted value of each parameter. The region itself is not this rectangle; Fig.~\ref{fig:plausible_regions} shows its shape. Yield stress is listed only for the ten objects whose reference simulation yields; for the other six it is not scored. A/N: number of accepted samples / number of evaluated samples. The search domain starts at $\log_{10}E=4$ and $\log_{10}\sigma_Y=5$; values at these bounds are limited by the domain (App.~\ref{sec:app_reference_free}).}\label{tab:ranges}
\begin{tabular}{lccccc}\toprule
Object & Reference $E$ & Region, $E$ & Reference $\sigma_Y$ & Region, $\sigma_Y$ & A/N \\\midrule
belt & 5.022 & [4.967, 5.088] & -- & -- & 84/144 \\
bottle & 8.450 & [8.196, 8.600] & 6.151 & [6.001, 6.187] & 56/115 \\
bowl & 8.659 & [8.272, 8.959] & 5.924 & [5.537, 6.224] & 80/132 \\
box & 8.893 & [8.820, 8.993] & 6.103 & [6.022, 6.203] & 51/92 \\
crocs & 6.083 & [6.058, 6.219] & 5.000 & [5.000, 5.091] & 21/55 \\
diaper & 6.000 & [5.965, 6.127] & 5.000 & [5.000, 5.094] & 31/66 \\
glove & 4.019 & [4.000, 4.113] & -- & -- & 17/69 \\
hanger & 8.911 & [8.723, 8.961] & 5.000 & [5.000, 5.028] & 16/39 \\
kitty & 4.002 & [4.002, 4.052] & -- & -- & 39/72 \\
medkit & 7.193 & [7.124, 7.378] & 5.099 & [5.001, 5.203] & 68/117 \\
napkin & 6.638 & [6.368, 6.788] & 5.415 & [5.235, 5.565] & 59/129 \\
neck pillow & 4.000 & [4.000, 4.098] & -- & -- & 16/36 \\
pad & 4.000 & [4.000, 4.427] & -- & -- & 44/55 \\
pillow & 4.000 & [4.000, 4.000] & -- & -- & 1/32 \\
sneaker & 5.077 & [5.052, 5.135] & 5.000 & [5.000, 5.500] & 38/75 \\
trash bin & 8.000 & [7.875, 8.166] & 6.151 & [6.057, 6.262] & 54/109 \\
\bottomrule\end{tabular}\end{table*}

\section{Replay Metrics}
\label{sec:app_behavior}
\paragraph{Replay with a predicted material.}
For a predicted pair $(\widehat z_E,\widehat z_Y)$, replay uses the fixed density of the object and decodes $E=\rho\,10^{\widehat z_E}$ and $\sigma_Y=\rho\,10^{\widehat z_Y}$. Mesh, camera, contact, and simulator settings are those of the reference. The simulation starts three frames before the collision frame and covers the 16 frames of the predictor's window.

\paragraph{Relative replay loss.}
With the material fixed, we refine the initial state of the drop: we try 25 candidates around the reference state and keep the one with the lowest objective. The candidates stay close to the reference state: position within 3\,cm, each rotation-vector component within $10^\circ$, velocity within 0.35\,m/s, and angular velocity within 0.8\,rad/s. For the reference material, the initial state is refined in the same way. The total replay loss is $0.25\mathcal L_{\mathrm{traj}}+0.25\mathcal L_{\mathrm{pose}}+0.45\mathcal L_{\mathrm{deform}}+0.05\mathcal L_{\mathrm{short}}$, and the deformation replay loss is $0.9\mathcal L_{\mathrm{deform}}+0.1\mathcal L_{\mathrm{short}}$, with the terms of App.~\ref{sec:app_reference_objective}. Eq.~\eqref{eq:replay_loss_difference} is evaluated per drop and then averaged over drops. A value below zero means that, after both initial states are refined, the predicted material matches the video better than the reference does. This can happen for individual drops, because the reference is fit to the five drops of an object jointly and with a different search.

\paragraph{Direct simulation difference.}
$\mathrm{Err}_{\mathrm{ref}}$ (App.~\ref{sec:app_err_ref}) uses no video and no refinement. The predicted and the reference material are simulated from the same reference initial state, and trajectory, pose, and strain are compared between the two simulations. Strain is computed from fitted local deformation gradients of the simulated particle tracks, so the comparison is in 3D and does not depend on optical flow. $\mathrm{Err}_{\mathrm{ref}}$ is averaged over the drops of each object and then over objects.

\section{Direct Predictor and Image-Plane Deformation Cues}
\label{sec:feedforward}
The physical model says that local deformation is an important source of information about stiffness. But recovering a deforming 3D mesh from a single collision video is difficult, for three reasons. From one view, depth gives at most a point cloud of the visible surface, which differs from the complete object surface. Rotation changes which points are visible. And depth errors can introduce false motion even when the 2D tracks are correct.
A simple approach trains a pretrained video backbone with prediction heads on synthetic RGB videos to predict $E/\rho$ and $\sigma_Y/\rho$. Can explicit motion and local deformation cues improve it? 
We study 2D trajectories, which describe motion, and three 2D cues that describe local deformation: strain, stretch, and the non-affine residual. 
We also lift these to 3D. 
We found that 2D motion and deformation, estimated from the video, retain useful material information. They are less sensitive to the estimation errors of the foundation models than their 3D versions. The pretrained RGB backbone supplies complementary context.
Physical intermediate representations have been studied before~\cite{ma2022risp,lin2026analyticdynamics}; we focus on which video-observable local cues remain useful after imperfect estimation.

\paragraph{Short correspondences and fixed neighborhoods.}
We derive short correspondences from optical flow rather than long ones from a point tracker, because the long-range trackers that we tested are not stable across our drops. Each of the eleven trackers and tracker modes covers less than half of the object at the end of the window in some drops, while our short tracklets cover more than half at that frame in every drop (App.~\ref{sec:app_cue_details} and~\ref{sec:app_more_trackers}). For example, CoTracker3 in its online mode reports most of the points that it starts before the collision as no longer visible about ten frames after the collision, as the object rotates and blurs (Fig.~\ref{fig:app_tracking}).
With forward optical flow $\boldsymbol{u}_t$ in crop-pixel units, a track starting at $\boldsymbol{p}_{i,r}$ is propagated by bilinear sampling (Eq.~\eqref{eq:short_tracks}).
We use four five-frame tracklets starting at clip indices $0,4,8,11$, each with up to 128 queries sampled from object masks.
The $k$ nearest neighbors $\mathcal N_k(i)$ are chosen in the reference image and kept fixed, with $k\in\{8,24\}$.

\paragraph{Image-plane deformation cues.}
With reference and current neighborhood offsets $\boldsymbol{a}_{ij}=\boldsymbol{p}_{j,r}-\boldsymbol{p}_{i,r}$ and $\boldsymbol{b}_{ij,t}=\boldsymbol{p}_{j,t}-\boldsymbol{p}_{i,t}$, we robustly fit a local affine map and compute three cues:
\begin{align}
\boldsymbol A_{i,t}
&=\operatorname*{arg\,min}_{\boldsymbol A}\sum_{j\in\mathcal N_k(i)} w_{ij,t}
  \|\boldsymbol b_{ij,t}-\boldsymbol A\boldsymbol a_{ij}\|_2^2,
\label{eq:local_affine}\\
\boldsymbol H_{i,t}
&=\tfrac12\log\bigl(\boldsymbol A_{i,t}^{\top}\boldsymbol A_{i,t}\bigr),
\qquad
s_{ij,t}=\ln\frac{\|\boldsymbol b_{ij,t}\|_2}{\|\boldsymbol a_{ij}\|_2},
\label{eq:strain2d}\\
r_{i,t}
&=\sqrt{
\frac{\sum_j w_{ij,t}
\|\boldsymbol b_{ij,t}-\boldsymbol A_{i,t}\boldsymbol a_{ij}\|_2^2}
{\sum_j w_{ij,t}\|\boldsymbol a_{ij}\|_2^2+\epsilon}}.
\label{eq:nonaffine}
\end{align}
Here, $\boldsymbol H$, $s$, and $r$ describe local logarithmic strain, pairwise stretch, and motion not explained by the fitted affine map. Stretch is summarized by its mean, standard deviation, and 10\% and 90\% quantiles over each neighborhood; strain, stretch, and residual are computed for both $k$. These image-plane cues are not the full 3D strain tensor: perspective and out-of-plane motion still affect them, and a large residual can also reflect correspondence error or occlusion (details in App.~\ref{sec:app_cue_details}).

\paragraph{2D and 3D tracklets.}
The 2D tracklets of the main paper are the input of \model{} and of RGB + 2D tracklets: each point is described by its normalized reference position and its displacement. The model that reads them uses the same graph and geometry encoder as the cue models. It thus receives the local geometry and motion needed to compute deformation. For 3D tracklets, we lift the same tracks with calibrated depth and intrinsics $\boldsymbol K$, $\boldsymbol X_{i,t}=d_t(\boldsymbol p_{i,t})\,\boldsymbol K^{-1}[u_{i,t},v_{i,t},1]^\top$, where $(u_{i,t},v_{i,t})$ is the position of $\boldsymbol p_{i,t}$ in the uncropped image (Eq.~\eqref{eq:lift}), and use 3D instead of 2D positions and displacements.

\paragraph{Model architecture.}
Our models have three branches for RGB, geometry, and shared context (Fig.~\ref{fig:method}).
In the cue models, the geometry branch reads the deformation cues; in the models on 2D tracklets, including \model{}, it reads the 2D tracklets. For every tracked point and time step, it maps the standardized cue channels, the cue-validity indicators, and five timing and scale context values to a 128-dimensional feature. Three blocks update these features, each with EdgeConv~\cite{wang2019dgcnn} over the 24 reference neighbors and a dilated temporal convolution along each track. Masked mean and max pooling then give the geometry feature $\boldsymbol f_G\in\mathbb R^{256}$.
The RGB branch encodes the unmasked clip with V-JEPA~2 ViT-L~\cite{assran2025vjepa2} and LoRA adaptation ($\boldsymbol f_R\in\mathbb R^{1024}$), and a context branch encodes the timing and scale values ($\boldsymbol f_{\mathrm{ctx}}\in\mathbb R^{32}$).
Late fusion feeds the concatenated features $\boldsymbol f=[\boldsymbol f_R,\boldsymbol f_G,\boldsymbol f_{\mathrm{ctx}}]$ to two regression heads for the targets of Eq.~\eqref{eq:ratio_target},
\begin{equation}
 \widehat z_E = 3.5 + 2.5\,\mathrm{MLP}_E(\boldsymbol f),\;\;
 \widehat z_Y = 3.1505 + 1.1505\,\mathrm{MLP}_Y(\boldsymbol f),
 \label{eq:model_late_fusion}
\end{equation}
with fixed target-normalization constants (App.~\ref{sec:app_identifiability}); the yield head is supervised only on windows with plastic excitation.
RGB-only omits $\boldsymbol f_G$ while retaining the same RGB adaptation and context branch. Further details are in App.~\ref{sec:app_architecture}.

\section{Evaluation Details}
\subsection{Synthetic Evaluation}
\label{sec:app_synthetic_eval}
Each synthetic video has ground-truth material parameters.
We report the mean absolute error (MAE) in log space, in dex, for
$q\in\{q_E,q_Y\}$ with $q_E=E/\rho$ and $q_Y=\sigma_Y/\rho$:
\begin{equation}
\operatorname{MAE}_{\log}(q)
=\frac{1}{\sum_n m_{n,q}}\sum_n m_{n,q}
    \left|\log_{10}\widehat q_n-\log_{10}q_n\right|,
\label{eq:masked_parameter_mae}
\end{equation}
where $m_{n,q_E}=1$ for every video and $m_{n,q_Y}=1$ only when the simulated observation
window contains plastic excitation, so other videos are not scored for yield stress.

\subsection{Reference Optimization Objective}
\label{sec:app_reference_objective}
This objective constructs the reference material of each object. The replay evaluation of a predicted material (App.~\ref{sec:app_behavior}) uses the same comparison terms but keeps the predicted material fixed: it refines only the initial state of each drop, with no alternation and no material search.
For each drop, the optimization objective is
\begin{equation}
\begin{aligned}
\mathcal L_{\mathrm{total}}
={}&0.25\mathcal L_{\mathrm{traj}}
   +0.25\mathcal L_{\mathrm{pose}}
   +0.45\mathcal L_{\mathrm{deform}}\\
 &+0.05\mathcal L_{\mathrm{short}}
   +0.02\mathcal L_{\mathrm{prior}}.
\end{aligned}
\label{eq:reference_optimization_loss}
\end{equation}
The first four terms compare the simulation with the real video. $\mathcal L_{\mathrm{traj}}$ compares the 3D motion of visible surface points;
$\mathcal L_{\mathrm{pose}}$ compares short-window 3D rotations and silhouette boundaries;
$\mathcal L_{\mathrm{deform}}$ compares projected residual-flow thickness near object boundaries after removing dominant affine motion;
$\mathcal L_{\mathrm{short}}$ compares local pair-distance changes along short tracks;
and $\mathcal L_{\mathrm{prior}}$ keeps the optimized parameters close to their initialization.
Sec.~\ref{sec:reference_protocol} calls the first two terms object-center trajectory and image-plane pose.
As described here, the trajectory term and the rotation part of the pose term use 3D motion; the rest of the pose term compares silhouette boundaries in the image (App.~\ref{sec:app_optimization}).
The 3D parts describe the rigid motion of the object: in each short window, one rigid displacement and one rotation are fit robustly to all tracked points.
This fit averages out the noise of the depth at individual points, but it also removes the local deformation, through which the material acts.
The deformation itself is measured by $\mathcal L_{\mathrm{deform}}$ and $\mathcal L_{\mathrm{short}}$, which compare image-plane cues. The initialization is the candidate of a grid search that a human rated best. The rater compared the real video with re-simulations of the grid candidates, which were shown without their material values.
The object-level objective averages this loss across its eligible drops.

To construct the reference, we alternate initial-state refinement and material search for up to four rounds:
with material parameters fixed, we refine each drop's initial state to improve trajectory and pose alignment; with initial states fixed, we optimize the shared material parameters by Bayesian optimization.
The alternation matters because the deformation loss is a projected 2D metric affected by 3D rotation and trajectory, so trajectory and pose must be aligned first.
The material and initial states with the lowest total loss form the reference
$(\theta_i^\star,\{\eta_{i,d}^\star\})$ (App.~\ref{sec:app_optimization}).

\subsection{Direct Simulation Difference}
\label{sec:app_err_ref}
We also directly compare the predicted-material simulation with the reference-material simulation, using the same reference initial state $\eta_{i,d}^\star$ for both, without further refinement.
We measure three errors. The trajectory error $e_{\mathrm{traj}}$ is the RMSE of the volume-weighted centroid, normalized by the initial bounding-box diagonal of the reference object. The pose error $e_{\mathrm{pose}}$ is the mean rotation error, normalized by $180^\circ$. The third is the deformation error $e_{\mathrm{strain}}$.
For deformation, we fit the local deformation gradient $F$ to the simulated particle trajectories and compare the Hencky strains
$H=\frac12\log(F^\top F)$.
The deformation error $e_{\mathrm{strain}}$ is the volume- and time-weighted Frobenius RMSE between the strains of the predicted-material simulation and those of the reference-material simulation.
The total simulation difference is
\begin{equation}
\label{eq:simulation_difference}
\mathrm{Err}_{\mathrm{ref}}
=
0.25\,e_{\mathrm{traj}}
+
0.25\,e_{\mathrm{pose}}
+
0.5\,e_{\mathrm{strain}}.
\end{equation}
We average over the drops of each object and then over objects.

\subsection{Random-Prediction Baseline}
\label{sec:app_random_baseline}
We also compute the expected parameter distance of a random prediction that draws $\log_{10}E$ and $\log_{10}\sigma_Y$ uniformly from the training ranges:
\begin{gather*}
x=\log_{10}E\sim U(4,9),\\
y=\log_{10}\sigma_Y\sim U\bigl(5,\log_{10}(2\times10^7)\bigr).
\end{gather*}
We average the parameter distance (Eq.~\eqref{eq:plausible_region_distance}) over these draws and over the $N$ real drops. For drop $d$, $\rho_d$ is the density of its object, which converts the draw to the ratios $z_E$ and $z_Y$, and $\mathcal A_d$ is the plausible region of its object:
\[
\begin{aligned}
\mathbb E[D]={}&\frac1{N}\sum_{d=1}^{N}\frac1{5\,(\log_{10}(2\times10^7)-5)}\int_4^9\int_5^{\log_{10}(2\times10^7)}\\
&\operatorname{dist}\!\left(\begin{bmatrix}x-\log_{10}\rho_d\\y-\log_{10}\rho_d\end{bmatrix},\mathcal A_d\right)dy\,dx=\boxed{2.071}.
\end{aligned}
\]

\subsection{Teacher--Student Details}
\label{sec:app_teacher_student}
\paragraph{Student loss.}
The student regresses the 288-dimensional feature of the teacher (App.~\ref{sec:app_oracle}). An MLP with one hidden layer of 512 units maps the student's fused feature (App.~\ref{sec:app_architecture}) to a prediction of the teacher's feature. The frozen material head of the teacher then maps this prediction to $z_E$ and $z_Y$.
The feature distance $d$ in Eq.~\eqref{eq:student_loss} is the mean Smooth L1 loss ($\beta=0.25$) between the predicted feature and the teacher's feature, where the difference in each feature dimension is divided by the standard deviation of that dimension over the training set. The regression loss $\ell$ is the squared error on normalized targets: the error of $z_E$ is divided by 2.5 and the error of $z_Y$ by 1.1505. We set $\lambda=1$. The student is trained for five epochs with the schedule of App.~\ref{sec:app_architecture}; the projection MLP uses the learning rate of the heads.

\paragraph{Learning intermediate 3D deformation features.}

\begin{table}[t]
\centering
\small
\caption{Teacher--student results (students at epoch~5; teacher at its selected epoch~8).
Synthetic validation errors are mean absolute errors of $\log_{10}(E/\rho)$ and $\log_{10}(\sigma_Y/\rho)$ in dex.
The last column is the parameter distance to the plausible region on the real drops.
The teacher is evaluated on the 14,293 validation videos with usable surface geometry (App.~\ref{sec:app_oracle});
both students are evaluated on all 14,588 videos.
The teacher needs exact surface deformation and cannot run on real videos (dash).
Lower is better for all metrics.}
\label{tab:teacher_student}
\begin{adjustbox}{max width=\linewidth}
\begin{tabular}{@{}lccc@{}}
\toprule
Model
& \makecell{Syn.\ val.\\$\log_{10}(E/\rho)$ MAE}
& \makecell{Syn.\ val.\\$\log_{10}(\sigma_Y/\rho)$ MAE}
& \makecell{Real parameter\\distance} \\
\midrule
Teacher                 & 0.161 & 0.192 & -- \\
Student, 2D tracklets (\model{})  & 0.481 & 0.356 & 0.677 \\
Student, 3D tracklets  & 0.443 & 0.336 & 1.164 \\
\bottomrule
\end{tabular}
\end{adjustbox}
\end{table}

Material acts through 3D deformation, but predicting material directly from video is hard. Can a video model predict an intermediate description of the 3D deformation instead?

During training in simulation, the teacher predicts material from exact 3D deformation of the object surface (App.~\ref{sec:app_oracle}): an encoder turns the deformation into a feature, and a material head turns the feature into $z_E$ and $z_Y$.
The student takes RGB and 2D tracklets, or RGB and 3D tracklets, and predicts the teacher's feature. It then applies the teacher's frozen material head. The student is the direct predictor of App.~\ref{sec:feedforward} with its regression heads replaced by the projection to the teacher's feature.
The student is supervised by the teacher's feature and by the material labels. At test time it sees a real video and its tracklets, predicts the feature, and then the material.

In this way the student can use 3D information without exact 3D deformation at test time.
With exact 3D deformation, the material can be predicted accurately: on synthetic validation, the teacher reaches an error of 0.161\,dex for $z_E$ and 0.192\,dex for $z_Y$ on the 14,293 videos with usable surface geometry (0.145\,dex for $z_E$ on the 10,202 videos of Table~\ref{tab:error_budget}; App.~\ref{sec:app_oracle}).
We ask whether 2D tracklets and RGB can recover this information, and whether tracklets lifted with estimated depth recover it better.
The student on 3D tracklets has the lower synthetic error, yet it is worse on real videos than the student on 2D tracklets.
Sec.~\ref{sec:3d_failure} examines this failure.

\paragraph{What do the learned features retain?}
We freeze the teacher and the student and train linear probes on their features. For each frame, a probe predicts the magnitude of the Hencky strain of the surface, averaged over a region of the object. This target is computed from the simulator's surface-point trajectories.
We compare two regions: the camera-visible surface, and the part of the surface near the ground ($y\leq0.07\,\mathrm{m}$), which approximates the part of the object that is in contact with the ground.
We first fit the probes on 271,134 synthetic training videos and select the regularization on the other 5,916 training videos, whose meshes do not appear among the 271,134. We then refit the probes on all 277,050 training videos.
We evaluate on the synthetic validation split and score both regions on the same videos and frames, those that are valid for both regions.
The student's features recover visible-surface strain with $R^2=0.625$, compared with $0.506$ near the ground.
The teacher's features reach a higher $R^2$ in both regions (Table~\ref{tab:strain_probe}). The student thus retains useful deformation information, but it recovers the deformation near the ground less accurately than that of the visible surface.

\begin{table}[t]
\centering
\small
\caption{\textbf{Linear readout of regional deformation.}
Both models are frozen, and the probes are fitted using the full
synthetic training split.
Probes predict regional mean surface-strain magnitude;
$R^2$ is computed after centering targets separately at each frame.
Evaluation uses 13,212 validation videos and 127,091 common valid
frames from $C$ to $C+12$.
The near-ground region is defined by height above the ground ($y\leq0.07\,\mathrm{m}$); it is not an exact contact label.
Higher is better.}
\label{tab:strain_probe}
\begin{tabular}{@{}lcc@{}}
\toprule
Features
& \makecell{Visible surface\\$R^2\uparrow$}
& \makecell{Near-ground region\\$R^2\uparrow$} \\
\midrule
Teacher & 0.726 & 0.679 \\
Student (\model{}) & 0.625 & 0.506 \\
\bottomrule
\end{tabular}
\end{table}

\subsection{Depth Estimator Comparison}
\label{sec:app_depth_comparison}
\paragraph{Depth estimation comparison.}
One might ask whether we chose the wrong depth estimation model.
We therefore test multiple recent and established video and image
depth estimation models, including DepthCrafter, Video Depth Anything,
NVDS, ChronoDepth, Depth Anything V2 and Depth Anything 3, MiDaS, ZoeDepth, and
the other models listed in Table~\ref{tab:depth_comparison}.

We evaluate these models on both real and synthetic videos.
For real videos, we measure the absolute relative error (AbsRel)
between the aligned estimated depth and our LiDAR depth within
the object mask. We also measure temporal depth jitter on the
static background.
For synthetic videos, we measure AbsRel against simulation-derived
depth within the object mask.

We estimate one scale from the reference frame and keep it fixed
throughout the 16-frame video:
\begin{equation}
s=\operatorname{median}_{p\in\mathcal C}
\frac{D^{\mathrm{ref}}_0(p)}{\widehat D_0(p)},
\qquad
\widetilde D_t(p)=s\widehat D_t(p),
\end{equation}
where $\mathcal C$ is the valid calibration support.
Calibration pixels are excluded from evaluation.
For valid object pixels $\mathcal V$, AbsRel is
\begin{equation}
\operatorname{AbsRel}
=
\frac{1}{|\mathcal V|}
\sum_{(t,p)\in\mathcal V}
\frac{|\widetilde D_t(p)-D^{\mathrm{ref}}_t(p)|}
   {D^{\mathrm{ref}}_t(p)}.
\end{equation}
Temporal jitter measures how far the depth of each static background
pixel deviates from its median over time. Depth that is stable over time is very important when we compute 3D trajectories:
\begin{equation}
J=
\frac{1}{|\mathcal B|}
\sum_{(t,p)\in\mathcal B}
\left|
\widetilde D_t(p)
-
\operatorname{median}_{\tau}\widetilde D_{\tau}(p)
\right|,
\end{equation}
where $\mathcal B$ contains valid static background observations.
We average clip-level measurements within each object and then
across objects.

4RC achieves the lowest real-video AbsRel and temporal jitter
among the tested models, supporting our choice of depth estimator.
We also repeat the comparison with a second alignment, a scale and a shift in inverse depth that are fit on the reference frame.
With it, 4RC again has the lowest real-video AbsRel (0.0526) and the lowest temporal jitter (0.00397\,m).

\begin{table}[t]
\centering
\small
\caption{Depth estimation comparison on 80 real videos and
200 synthetic validation videos.
All entries use one reference-frame scale fixed across the video.
AbsRel is dimensionless; temporal jitter is measured in meters.
Lower is better. Best results are in bold.}
\label{tab:depth_comparison}
\begin{adjustbox}{max width=\linewidth}
\begin{tabular}{@{}lccc@{}}
\toprule
Model
& \makecell{Real\\AbsRel $\downarrow$}
& \makecell{Real temporal\\jitter (m) $\downarrow$}
& \makecell{Synthetic\\AbsRel $\downarrow$} \\
\midrule
4RC                    & \textbf{0.0605} & \textbf{0.00401} & \textbf{0.1374} \\
DepthCrafter           & 0.2698 & 0.08097 & 0.2107 \\
GEMDepth               & 0.2168 & 0.01500 & 1.4715 \\
VGGT-$\Omega$          & 0.0763 & 0.00482 & 0.2693 \\
VDPM                   & 0.0675 & 0.00968 & 0.2422 \\
Point4D                & 0.1476 & 0.00669 & 0.1740 \\
Video Depth Anything   & 0.1684 & 0.01574 & 0.6692 \\
NVDS                   & 0.1098 & 3.69612 & 0.2787 \\
ChronoDepth            & 0.3427 & 0.03497 & 0.8076 \\
\midrule
MoGe-3                 & 0.0816 & 0.03730 & 0.3995 \\
DepthMaster            & 0.0946 & 0.05162 & 0.2977 \\
Depth Anything 3 Metric & 0.0883 & 0.02159 & 0.2577 \\
Depth Anything 3 Mono  & 0.1807 & 0.01778 & 1.2771 \\
UniDepth               & 0.0714 & 0.02040 & 0.3791 \\
Depth Anything V2      & 0.1982 & 135.17519 & 1.2547 \\
MiDaS                  & 0.1542 & 41.22725 & 0.4167 \\
ZoeDepth               & 0.1489 & 0.02009 & 0.2344 \\
\bottomrule
\end{tabular}
\end{adjustbox}
\end{table}

\subsection{Depth and Tracking Substitutions}
\label{sec:app_3d_cross_eval}
\begin{table}[t]
\centering
\small
\caption{\textbf{Depth and tracking substitutions} for RGB + 3D tracklets. Rows: training condition (A: simulator depth and tracks; B: simulator depth, WAFT tracks; C: 4RC depth, simulator tracks; D: 4RC depth, WAFT tracks). Columns: evaluation condition; real videos exist only for D. Entries: $E/\rho$ / $\sigma_Y/\rho$ log-MAE (dex). Bold: matched conditions.} %
\label{tab:3d_cross_eval}
\begin{adjustbox}{max width=\linewidth}
\begin{tabular}{@{}lccccc@{}}
\toprule
Train / Eval & Syn A & Syn B & Syn C & Syn D & Real D \\
\midrule
A & \textbf{0.414 / 0.327} & 0.532 / 0.449
  & 0.820 / 0.572 & 0.839 / 0.592 & 1.305 / 0.361 \\
B & 0.427 / 0.335 & \textbf{0.421 / 0.330}
  & 0.662 / 0.436 & 0.672 / 0.438 & 1.132 / 0.465 \\
C & 0.446 / 0.343 & 0.455 / 0.346
  & \textbf{0.451 / 0.346} & 0.460 / 0.348 & 1.178 / 0.589 \\
D & 0.446 / 0.342 & 0.446 / 0.342
  & 0.449 / 0.344 & \textbf{0.450 / 0.344} & 1.188 / 0.581 \\
\midrule
RGB + 2D tracklets & & & & 0.482 / 0.363 & 0.792 / 0.423 \\
\bottomrule
\end{tabular}
\end{adjustbox}
\end{table}

To separate the representation from its estimation, we train the lifted model with each combination of simulator or 4RC depth and simulator or WAFT tracks (A--D in Table~\ref{tab:3d_cross_eval}).
We evaluate each model under all four synthetic conditions and on real videos, where only D exists (Fig.~\ref{fig:sim_vs_real}, right).
D has the same inputs as RGB + 3D tracklets in Table~\ref{tab:representation_tradeoff}, but it is a separate training run (App.~\ref{sec:more_ablations}).
Model A, trained on exact geometry, is evaluated with one input replaced at a time. Its $E/\rho$ error rises by 0.12\,dex when WAFT tracks replace simulator tracks (column Syn B), but by 0.41\,dex when 4RC depth replaces simulator depth (column Syn C).
Training on estimated depth removes this cost in simulation: models C and D are within 0.04\,dex of A and B and insensitive to which depth they receive.
Yet all four lifted models remain 0.34--0.51\,dex worse than 2D tracklets on real drops (1.132--1.305 vs.\ 0.792\,dex).

\subsection{Stability Metrics}
\label{sec:app_stability_metrics}
Reliable sim-to-real transfer requires consistent predictions.
We therefore evaluate prediction stability on real data in three ways.

First, we evaluate stability across different drops of the same object. 
Each object has five drops with different orientations, and the model predicts $\log_{10}(E/\rho)$ and $\log_{10}(\sigma_Y/\rho)$ from each drop.
We compute the standard deviation across drops for each object and average across the 16 objects. For yield stress, we include only the drops whose reference simulation yields, and only the objects that have such drops:
\begin{equation}
\begin{aligned}
S_E^{\mathrm{drop}}
&=
\frac{1}{16}\sum_{i=1}^{16}
\operatorname{Std}_{d\in\mathcal D_i}
\left[\log_{10}\widehat q_{E,i,d}\right],\\
S_Y^{\mathrm{drop}}
&=
\frac{1}{|\mathcal I_Y|}
\sum_{i\in\mathcal I_Y}
\operatorname{Std}_{d\in\mathcal D_{i,Y}}
\left[\log_{10}\widehat q_{Y,i,d}\right].
\end{aligned}
\label{eq:cross_drop_stability}
\end{equation}
Here, $\mathcal D_i$ contains the evaluated drops of object $i$,
$\mathcal D_{i,Y}$ contains its yielded drops, and
$\mathcal I_Y$ indexes the yielded objects.

Second, we evaluate prediction stability and accuracy under substantial changes in recording conditions.
We use 40 additional real videos of Kitty with changes in camera viewing angle, distance, background, and lighting.
The benchmark Kitty does not yield in any of its five drops, so we report only stiffness results. The toy in these recordings is a second copy of the same product, because the toy of the benchmark was no longer available. The two copies differ only in appearance: the second copy wears a dress and has a different surface print. The tensile specimens of Kitty were cut from the head of this second copy (App.~\ref{sec:app_real_measure}). The log-MAE of Table~\ref{tab:recording_stability} uses the reference stiffness of Kitty.
We measure the standard deviation of predicted log stiffness ratios:
\begin{equation}
S_E =
\operatorname{Std}_{n=1,\ldots,40}
\left[\log_{10}\widehat q_{E,n}\right].
\label{eq:recording_stability}
\end{equation}

\paragraph{Consistency results.}
\label{sec:recording_stability}
\begin{table}[t]
\centering
\small
\caption{\textbf{Consistency} of the predictions in three settings. Cross-drop: standard deviation over the five drops of an object, averaged over the 16 objects ($S_E^{\mathrm{drop}}/S_Y^{\mathrm{drop}}$; $S_Y^{\mathrm{drop}}$ uses only yielded drops). Kitty: standard deviation and log-MAE of $z_E$ over the 40 Kitty recordings with changed viewpoint, distance, background, and lighting (these drops do not yield). Real--replay shift: mean $|\hat z_E^{\mathrm{real}}-\hat z_E^{\mathrm{replay}}|$ between each real drop and its rendered replay. All in dex; best in bold, second underlined.}
\label{tab:recording_stability}
\begin{adjustbox}{max width=\linewidth}
\begin{tabular}{@{}lcrrc@{}}
\toprule
Input
& \makecell{Cross-drop std.\\
$S_E^{\mathrm{drop}}/S_Y^{\mathrm{drop}}\downarrow$}
& \makecell{Kitty std.\\$S_E\downarrow$}
& \makecell{Kitty $E$\\log-MAE $\downarrow$}
& \makecell{Real--replay\\shift $\downarrow$} \\
\midrule
RGB & 0.362/0.234 & 0.309 & 0.494 & 0.584 \\
RGB + dense flow & 0.400/0.209 & 0.298 & 0.655 & 0.595 \\
RGB + depth & 0.341/0.198 & \underline{0.232} & 0.412 & 0.567 \\
RGB + 2D tracklets & \underline{0.324}/0.202 & 0.274 & 0.322 & \underline{0.541} \\
RGB + 3D tracklets & 0.411/0.187 & 1.018 & 0.825 & 0.704 \\
\textbf{\model{}} & \textbf{0.319}/0.189 & 0.241 & 0.304 & \textbf{0.530} \\
Student, 3D tracklets & 0.384/\underline{0.171} & 1.043 & 0.849 & 0.675 \\
\midrule
2D tracklets only & 0.482/\textbf{0.158} & 0.447 & 0.765 & 0.668 \\
\midrule
Vid2Sim & 0.590/0.223 & 0.277 & \underline{0.229} & 0.862 \\
Analytic Dynamics & 0.613/0.304 & \textbf{0.195} & \textbf{0.163} & 0.893 \\
\bottomrule
\end{tabular}
\end{adjustbox}
\end{table}

A predictor that reads material should give the same answer for the five drops of an object and for one object recorded under different conditions.
We report the spread of predictions across the drops of each object, across the 40 Kitty recordings, and between each real drop and its rendered replay (Table~\ref{tab:recording_stability}).
Lifting with estimated depth is the least consistent across the Kitty recordings, and among our inputs it has the largest real--replay shift.
The predictions of RGB + 3D tracklets have a standard deviation of 1.018\,dex across the Kitty recordings, against 0.241\,dex for \model{} and 0.309\,dex for RGB. This standard deviation is large under every recording change (Table~\ref{tab:kitty_conditions}).
We also compare the prediction from each real video with the prediction from the rendered replay of the same drop. The mean absolute difference between the two is 0.704\,dex for RGB + 3D tracklets and 0.530\,dex for \model{}.
\model{} also has the smallest standard deviation of predicted stiffness across the five drops of an object (0.319\,dex in Table~\ref{tab:recording_stability}; RGB + 2D tracklets, the direct 2D model, has 0.324\,dex).
Analytic Dynamics, which reads object masks, is the most stable across the Kitty recordings (standard deviation 0.195\,dex) and the closest to the reference stiffness of Kitty (log-MAE 0.163\,dex).
But it varies most across drops (0.613) and is much less accurate across objects (Sec.~\ref{sec:external_comparisons}).
Consistency alone does not establish accuracy, so we read these results together with Table~\ref{tab:representation_tradeoff}.

Table~\ref{tab:kitty_conditions} breaks the Kitty spread down by recording condition.
Viewpoint and distance are the most informative conditions. A predictor that reads 3D deformation should not change its prediction with the camera angle, and a predictor that reads metric motion should not change it with the distance to the camera.
For the two models on lifted tracklets, the standard deviation is 0.85--1.33\,dex under every condition, whereas for \model{} it stays within 0.17--0.25\,dex.
Analytic Dynamics, which uses object masks as input and is therefore less affected by changes in pixel appearance, is the most stable overall.

\begin{table}[t]
\centering
\small
\caption{\textbf{Kitty by recording condition:} standard deviation of predicted $z_E$ (dex) over the ten videos of each condition. Overall reports the standard deviation across all 40 videos. Best in bold, second underlined.}
\label{tab:kitty_conditions}
\begin{adjustbox}{max width=\linewidth}
\begin{tabular}{@{}lccccc@{}}
\toprule
Input & Angle & Distance & Background & Lighting & \textbf{Overall} \\
\midrule
RGB & 0.226 & 0.292 & 0.263 & 0.324 & 0.309 \\
RGB + dense flow & 0.206 & 0.343 & 0.342 & 0.216 & 0.298 \\
RGB + depth & 0.189 & 0.218 & \underline{0.201} & 0.280 & \underline{0.232} \\
RGB + 2D tracklets & \textbf{0.162} & 0.238 & 0.261 & 0.245 & 0.274 \\
RGB + 3D tracklets & 1.061 & 0.848 & 1.286 & 0.973 & 1.018 \\
\textbf{\model{}} & 0.170 & 0.254 & 0.215 & \underline{0.204} & 0.241 \\
Student, 3D tracklets & 1.046 & 0.880 & 1.325 & 1.004 & 1.043 \\
\midrule
2D tracklets only & 0.525 & 0.483 & 0.268 & 0.419 & 0.447 \\
\midrule
Vid2Sim & 0.283 & \underline{0.207} & \textbf{0.193} & 0.277 & 0.277 \\
Analytic Dynamics & \underline{0.166} & \textbf{0.136} & 0.209 & \textbf{0.187} & \textbf{0.195} \\
\bottomrule
\end{tabular}
\end{adjustbox}
\end{table}

\subsection{Adapting the External Baselines}
\label{sec:app_baseline_adaptation}
PhysDreamer and PhysFlow use our meshes, simulator, Neo-Hookean elasticity, von Mises plasticity, and fixed density and Poisson's ratio.
In the main paper (Table~\ref{tab:external_comparisons}), PhysDreamer is initialized from the predictions of \model{} (Fig.~\ref{fig:app_optimizer}). App.~\ref{sec:app_rev_init} also starts it from the predictions of our RGB + 2D tracklets + cues predictor (parameter distance 0.739; Table~\ref{tab:representation_tradeoff}).
PhysFlow is initialized either from the predictions of \model{} (our student) or from the predictions of the vision-language model given the first frame (VLM-image). The original paper also starts from a VLM prediction.
MonoPhysics retains its Gaussian representation, because all of its losses are computed on renderings of its Gaussians: the rendered alpha against the object mask, the optical flow rendered from the moving Gaussians against SEA-RAFT flow, and a silhouette loss. A regularizer also keeps the Gaussian centers evenly spaced. The Gaussians are initialized from the particles of our mesh. MonoPhysics uses Neo-Hookean elasticity and von Mises plasticity with our fixed density and Poisson's ratio.
It starts either from the predictions of \model{} or from its original initialization ($E=10^5$\,Pa, $\sigma_Y=10^4$\,Pa).
Its estimated material parameters are evaluated through replay on our benchmark meshes.
PhysTwin retains its mass--spring model and is compared through replay only.
PhysDreamer's material field is compared through its volume-weighted average.

PhysDreamer runs take two and a half to three hours per video, and App.~\ref{sec:app_rev_init} discusses the initializations.
Density and Poisson's ratio are fixed to the same values used in our reference simulations.

For PhysDreamer, the original version uses Gaussians and optimizes a Young's modulus field. We change the object representation to a mesh and use our corrected MPM implementation with differentiable optimization.
We use neo-Hookean elasticity with von Mises plasticity and optimize both $E$ and $\sigma_Y$ (Table~\ref{tab:external_comparisons}).
To handle the wide $E$ range of $[10^4,10^9]$ Pa, we optimize
$\log_{10}E$ and $\log_{10}\sigma_Y$.
We retain PhysDreamer's two-stage velocity-field and material-field optimization, image reconstruction and SSIM loss, and field smoothness regularization.

PhysDreamer represents material properties as spatial fields, whereas our reference parameters are object-level.
For parameter comparison, we compute volume-weighted averages:
\begin{equation}
\overline E=\frac{\sum_p V_p E_p}{\sum_p V_p},
\qquad
\overline{\sigma}_Y
=\frac{\sum_p V_p \sigma_{Y,p}}{\sum_p V_p},
\end{equation}
where $V_p$ is the fixed reference volume of particle $p$.
Replay retains the spatial fields rather than replacing them with these averages.

Every PhysDreamer run is evaluated after exactly 25 velocity-field updates followed by 25 material-field updates.

The released PhysDreamer code runs 300 velocity updates and 200 material updates, and it adds its learned field to a constant stiffness in Pa.
With this parameterization and the released material learning rate ($10^{-3}$), the stiffness hardly moved on our drops.
The first material update changed the mean $E$ by less than 1\,Pa, and after 200 material updates $E$ had changed by only 2.5--9\%.
With 100 velocity and 200 material updates, a run takes eight to nine hours per video.
We therefore changed two things.
First, we optimize $\log_{10}E$ and $\log_{10}\sigma_Y$ instead of the values in Pa, with a larger learning rate. The learning rate peaks at $4\times10^{-3}$ in the velocity stage and at $5\times10^{-3}$ in the material stage, each after a warm-up of three updates.
Second, we stop after 25 velocity and 25 material updates.
A run then takes two and a half to three hours per video.

For PhysFlow, we retain its object-level material parameterization and RAFT flow-matching loss.
We optimize $\log_{10}E$ and $\log_{10}\sigma_Y$ and keep the reference initial state fixed during material fitting. This initial state was fit together with the reference material, which favors PhysFlow. Table~\ref{tab:external_comparisons} reports neo-Hookean elasticity with von Mises plasticity.

PhysTwin uses a mass--spring model.
We use our mesh to construct its physical nodes and render the deformation while retaining its native mass--spring solver.
Depth-based initialization is used only when the mesh-only initialization fails.
Its spring parameters do not directly correspond to Young's modulus or yield stress, so we compare replay errors only.

For PhysDreamer and PhysTwin, replay retains each method's native material representation and physical model.
The replay losses that compare a simulation with the video (the total and deformation replay losses of Eq.~\eqref{eq:replay_loss_difference}) use the same objective as for our predictors. As there, the initial state is refined both for the method's material and for the reference material.
For $\mathrm{Err}_{\mathrm{ref}}$, both simulations use the reference initial state. 
Its deformation error fits local deformation gradients $F$ to the simulated point tracks and compares the resulting strains. It does not use the deformation gradients saved by the simulators, because the methods use different simulators.
PhysTwin simulates spring nodes and the reference simulation uses MPM particles, so the two discretizations do not share points. For PhysTwin, this comparison therefore uses a common set of material query points and a mapping from the spring nodes to these points that is fixed in the rest configuration.

\paragraph{Vid2Sim predictor.}
For Vid2Sim~\cite{Chen2025vid2sim}, we use its feed-forward predictor and follow the released training code.
The backbone is VideoMAE-base. It is fully fine-tuned on 16 frames at $448\times448$ pixels. Its tokens are mean-pooled and passed to regression heads with the released architecture.
The released model has three heads: one for Young's modulus, one for Poisson's ratio, and one for the weights of its linear-blend-skinning (LBS) simulator.
We remove the LBS head, which our simulator does not use, and the head for Poisson's ratio, which is fixed in our benchmark.
We keep two heads and train them on our two targets, $\log_{10}(E/\rho)$ and $\log_{10}(\sigma_Y/\rho)$. The yield term is masked as for our predictors.
The optimizer is the released one: Adam with a learning rate of $10^{-5}$, no weight decay, and no schedule.
We train on \model{}-Sim for five epochs with our global batch of 480 and report the fifth epoch.
The input is our crop of the video with its background, resized from 224 to 448 pixels. The released model was trained on objects rendered on a white background.
We do not run the per-scene refinement stage of Vid2Sim.

\paragraph{Analytic Dynamics.}
Analytic Dynamics~\cite{lin2026analyticdynamics} has no released code, so we implement it from the paper.
The teacher reads simulator particle trajectories: 1,024 points over 16 frames, with their displacements and a local deformation gradient fitted to neighboring points.
It is a point-cloud video transformer (P4Transformer) and regresses the same two targets as our predictors.
The student reads the 16 object masks at $128\times128$ pixels, without RGB, with a UniFormer backbone.
It is trained with the regression loss and with an alignment loss to the feature of the frozen teacher, the sum of a cosine distance and an L1 distance.
The paper leaves some details open, such as the weights of the alignment terms, the normalization of the targets, and the layer sizes. For these we made our own choices.
The student uses batch normalization synchronized across GPUs, because each GPU holds one clip.
Both networks are trained on \model{}-Sim for five epochs, and we report the fifth epoch.

\section{Additional Experiments}
\label{sec:app_additional_experiments}

\subsection{Material Sensitivity and Appearance Robustness}
\label{sec:material_appearance}

The real-object references are not evenly distributed over material
parameters. We therefore complement the real-video evaluation with
controlled simulations that change material parameters while keeping
each object's appearance fixed. This tests whether the models respond
to material-dependent motion, rather than only to differences between
object appearances.

\paragraph{Material sensitivity.}
Simulation is the right place to test whether predictors respond to material at all.
We sweep the stiffness in two sets of simulations that are not used for training.
The first re-simulates each of the 80 real drops in its reconstructed scene.
The second re-simulates 200 clips of the synthetic validation set from their source meshes, with the same pipeline; these are the 200 clips of Table~\ref{tab:depth_comparison}, from 199 held-out meshes.
Each drop or clip is simulated at eleven stiffness levels, $\log_{10}E\in\{4,4.5,\dots,9\}$ with $E$ in Pa, and with the yield stress fixed at $\sigma_Y=10^7$\,Pa.
Within a drop or clip only $E$ changes: mesh, initial state, camera, density, and Poisson's ratio are fixed.
A synthetic clip also keeps its background image, exposure, and compression setting.
Masks, flow, and depth are estimated again on every video.
Of the 880 simulations of the real scenes, we use 874: one at $\log_{10}E=8.5$ and five at 9 are numerically unstable.
Of the 2,200 synthetic simulations, we use 2,097, from 193 clips: 70 are numerically unstable, and 33, all simulations of three clips, failed for technical reasons.

\begin{figure}[t]
\centering
\includegraphics[width=\linewidth]{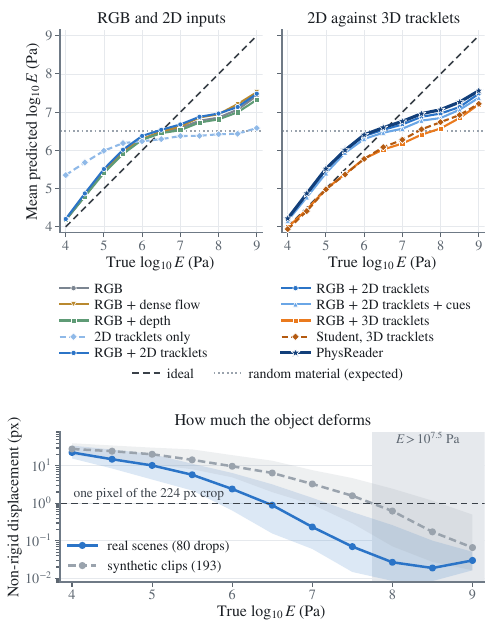}
\caption{\textbf{Material sensitivity in simulation.} Top: mean predicted $\log_{10}E$ against true $\log_{10}E$ on the re-simulated real scenes, with $\sigma_Y=10^7$\,Pa and mesh, initial state, camera, density, and Poisson's ratio fixed within each drop. Curves average over the 80 drops at each stiffness level; one simulation at $\log_{10}E=8.5$ and five at $9$ are numerically unstable and are left out. The dashed line is ideal recovery; the dotted line is the expected prediction of a uniformly random material. All predictors are evaluated after the fifth epoch. RGB + 2D tracklets is drawn in both top panels for comparison. Bottom: how much the object deforms in each simulation (non-rigid displacement, see text), in pixels of the $224\times224$ predictor crop. Lines: median over the real drops (blue) and over the synthetic clips (gray); bands: 25--75\%. Above $10^{7.5}$\,Pa (shaded), almost every simulation of the real scenes deforms by less than one pixel.}
\label{fig:material_sensitivity}
\end{figure}

Every RGB-based model orders stiffness correctly for 90--95\% of pairs in the real scenes and for 95--97\% in the synthetic clips (Table~\ref{tab:material_sensitivity} and Fig.~\ref{fig:material_sensitivity}).
For these models the slope of predicted against true $\log_{10}E$ is 0.56--0.64 in the real scenes and 0.69--0.74 in the synthetic clips.
A slope below one means that the predictions cover a narrower range than the true stiffness.
Without RGB, the model on 2D tracklets alone orders only 71.45\% and 81.76\% of pairs correctly.
The predictors therefore read material, not only appearance, and they differ mainly in how that reading survives the move to real video.
In the synthetic clips, the two models on tracklets lifted to 3D have the lowest log-$E$ error (0.54 and 0.52\,dex), as on synthetic validation. In the real scenes they do not (0.86 and 0.81\,dex).

\paragraph{High stiffness.}
Much of the error comes from the stiffest levels.
In these sweeps, $\sigma_Y=10^7$\,Pa, so the stiffest simulations stay elastic, and an elastic drop deforms less as $E$ grows.
We measure how much each simulation deforms. In every frame, we remove the best rigid motion of all particles and take the 95th percentile of the remaining displacement. We keep its largest value over the window, in pixels of the $224\times224$ crop given to the predictors.
In the real scenes, the median of this non-rigid displacement over the 80 drops is 10 pixels at $10^5$\,Pa, 2.4 at $10^6$\,Pa, 0.23 at $10^7$\,Pa, and at most 0.07 from $10^{7.5}$\,Pa on (Fig.~\ref{fig:material_sensitivity}, bottom). From $10^8$\,Pa on, 95--100\% of the drops deform by less than one pixel.
In the synthetic clips, the median of this non-rigid displacement over the 193 clips is 20 pixels at $10^5$\,Pa, 10 at $10^6$\,Pa, 3.3 at $10^7$\,Pa, and at most 1.6 from $10^{7.5}$\,Pa on. From $10^8$\,Pa on, 60--82\% of the clips deform by less than one pixel.
The synthetic clips deform more at the same $E$ because their objects are denser (median density about 1,100 against 348\,kg/m$^3$), and the deformation depends on $E/\rho$ (App.~\ref{sec:app_identifiability}).
In the real scenes, the drops at $10^8$\,Pa and above look like the drop of a rigid object, and $E=10^8$ and $10^9$\,Pa give nearly the same video (Fig.~\ref{fig:app_sweep_frames}).
Without the three stiffest levels, the log-$E$ error of the models with RGB falls from 0.71--0.86 to 0.52--0.58\,dex in the real scenes and from 0.53--0.60 to 0.42--0.47\,dex in the synthetic clips, and their slopes rise to 0.69--0.77 and 0.80--0.84 (Table~\ref{tab:material_sensitivity}).
Five objects of the benchmark have a reference stiffness of $10^8$\,Pa or more: the bottle, the bowl, the box, the hanger, and the trash bin (Table~\ref{tab:ranges}).
All five yield in their reference simulations, at $\sigma_Y\leq10^{6.2}$\,Pa, so their real drops deform plastically and visibly.

\begin{table*}[t]
\centering
\small
\caption{\textbf{Material sensitivity in simulation.} Two sets of held-out simulations, swept over eleven stiffness levels $\log_{10}E\in\{4,4.5,\dots,9\}$ with $\sigma_Y=10^7$\,Pa: the 80 real drops re-simulated in their reconstructed scenes (874 simulations), and 193 synthetic validation clips (2,097 simulations). Only $E$ changes; the appearance is fixed. Log-$E$ MAE: mean absolute error of the predicted $\log_{10}E$ (the predicted $E/\rho$ times the known density, so equal to the $z_E$ error), in dex. Ranking: share of pairs of simulations of the same drop or clip that the predictor orders correctly by stiffness, among the pairs whose visible surface strain differs (RMS difference of edge log-stretch at least 0.01). Slope: slope of predicted against true $\log_{10}E$; the ideal value is one. $E\leq10^{7.5}$: the same metrics without the three stiffest levels, at which most simulations deform by less than one pixel (Fig.~\ref{fig:material_sensitivity}, bottom).}
\label{tab:material_sensitivity}
\begin{adjustbox}{max width=\textwidth}
\begin{tabular}{@{}lrrrrr@{\hspace{1.4em}}rrrrr@{}}
\toprule
& \multicolumn{5}{c}{Re-simulated real scenes (80 drops)} & \multicolumn{5}{c}{Synthetic validation clips (193)} \\
\cmidrule(lr){2-6}\cmidrule(l){7-11}
& \multicolumn{3}{c}{All levels} & \multicolumn{2}{c}{$E\leq10^{7.5}$\,Pa} & \multicolumn{3}{c}{All levels} & \multicolumn{2}{c}{$E\leq10^{7.5}$\,Pa} \\
\cmidrule(lr){2-4}\cmidrule(lr){5-6}\cmidrule(lr){7-9}\cmidrule(l){10-11}
Input & Log-$E$ MAE $\downarrow$ & Ranking $\uparrow$ & Slope & MAE $\downarrow$ & Slope & Log-$E$ MAE $\downarrow$ & Ranking $\uparrow$ & Slope & MAE $\downarrow$ & Slope \\
\midrule
RGB & 0.753 & 91.36\% & 0.572 & 0.522 & 0.726 & 0.585 & 95.17\% & 0.694 & 0.443 & 0.809 \\
RGB + dense flow & 0.733 & 91.51\% & 0.595 & 0.536 & 0.739 & 0.572 & 95.17\% & 0.709 & 0.439 & 0.826 \\
RGB + depth & 0.776 & 90.61\% & 0.557 & 0.529 & 0.714 & 0.586 & 95.17\% & 0.695 & 0.447 & 0.809 \\
2D tracklets only & 1.187 & 71.45\% & 0.197 & 0.886 & 0.277 & 0.982 & 81.76\% & 0.371 & 0.735 & 0.459 \\
RGB + 2D tracklets & 0.735 & 91.30\% & 0.578 & 0.527 & 0.743 & 0.591 & 95.08\% & 0.689 & 0.454 & 0.803 \\
RGB + 2D tracklets + cues & 0.756 & 90.68\% & 0.572 & \textbf{0.518} & 0.734 & 0.600 & 94.79\% & 0.688 & 0.468 & 0.800 \\
RGB + 3D tracklets & 0.861 & 94.10\% & 0.602 & 0.584 & 0.693 & 0.536 & 96.40\% & 0.734 & 0.430 & 0.829 \\
Student, 3D tracklets & 0.814 & 95.34\% & 0.635 & 0.552 & 0.746 & \textbf{0.525} & 96.78\% & 0.740 & \textbf{0.421} & 0.840 \\
\model{} (Student, 2D tracklets) & \textbf{0.706} & 92.02\% & 0.605 & 0.523 & 0.769 & 0.580 & 95.16\% & 0.690 & 0.448 & 0.798 \\
Random material (analytic expectation) & 1.745 & 50.00\% & 0.000 & 1.625 & 0.000 & 1.740 & 50.00\% & 0.000 & 1.609 & 0.000 \\
\bottomrule
\end{tabular}
\end{adjustbox}
\end{table*}

\paragraph{Appearance robustness.}
We test appearance in two ways.
In simulation, we change only the texture.
Each of the 200 synthetic validation clips is simulated once with its own material and rendered six times over identical motion: with its original texture and with the textures of five other objects, the same five for every clip (Fig.~\ref{fig:app_texture}).
The texture can be replaced only if the mesh carries an image texture. This is the case for 182 of the 200 clips.
This gives 1,110 renders: one with the original texture for each of the 200 clips, and five with other textures for each of the 182 clips. We use 1,073 of them. For the others, the simulation is numerically unstable (19) or the run failed for technical reasons (18). This leaves 880 pairs of an original and a retextured render, from 176 clips.
Every render gets the background image, exposure, and compression setting of the original clip, and flow, depth, and masks are estimated again on every render.
Table~\ref{tab:texture} reports how much the predicted $z_E$ changes from the original render to a retextured render.
Replacing the texture changes the predicted $z_E$ by 0.22--0.26\,dex for the models with RGB and by 0.34\,dex for 2D tracklets alone.
The $z_E$ error changes little with the texture, by at most 0.032\,dex. For the models with RGB, it is 0.42--0.52\,dex on the original renders and 0.43--0.50\,dex on the retextured renders.
No model is invariant to texture: the predictions move. But no model is misled by it.
On real video, App.~\ref{sec:recording_stability} changes viewpoint, distance, background, and lighting over 40 recordings of one object (Fig.~\ref{fig:app_kitty}).
There the standard deviation of the predicted $z_E$ over the 40 recordings is 0.24\,dex for \model{} and more than 1\,dex for the two models on 3D tracklets.

\begin{table}[t]
\centering\small
\caption{\textbf{Texture replacement in simulation} on 176 synthetic validation clips (880 pairs of an original and a retextured render). Change: mean absolute change of the predicted $z_E$ from the original render to a retextured render. Std.: standard deviation of the predictions over the renders of a clip. MAE: $z_E$ error on the original and on the retextured renders. All in dex.}
\label{tab:texture}
\begin{adjustbox}{max width=\linewidth}
\begin{tabular}{@{}lcccc@{}}\toprule
Input & Change $\downarrow$ & Std.\ $\downarrow$ & \makecell{MAE,\\original} & \makecell{MAE,\\retextured} \\\midrule
RGB & 0.245 & 0.192 & 0.468 & 0.475 \\
RGB + dense flow & 0.245 & 0.198 & 0.474 & 0.483 \\
RGB + depth & 0.233 & 0.194 & 0.490 & 0.493 \\
2D tracklets only & 0.340 & 0.298 & 0.818 & 0.803 \\
RGB + 2D tracklets & 0.249 & 0.198 & 0.463 & 0.495 \\
RGB + 2D tracklets + cues & 0.256 & 0.200 & 0.517 & 0.503 \\
RGB + 3D tracklets & 0.239 & 0.195 & 0.438 & 0.441 \\
Student, 3D tracklets & 0.221 & 0.183 & 0.417 & 0.425 \\
\model{} (Student, 2D tracklets) & 0.224 & 0.178 & 0.449 & 0.474 \\
\bottomrule\end{tabular}
\end{adjustbox}
\end{table}

\subsection{More Ablations}
\label{sec:more_ablations}

All ablations reuse the training set, the schedule, and the evaluation of the main paper.
Tables~\ref{tab:ablation_inputs} and~\ref{tab:ablation_students} evaluate every model on all 80 real drops.

\paragraph{Geometric inputs.}
Table~\ref{tab:ablation_inputs} lists the direct predictors, that is, the predictors trained without a teacher, that differ in their geometric input. Table~\ref{tab:3d_cross_eval} adds the depth and tracking variants of the lifted model.
Combined with RGB, all 2D inputs reach nearly the same $z_E$ error on synthetic validation (0.48\,dex). The deformation cues are the local strain, the stretch, and the non-affine residual computed from the 2D tracklets. On real drops, RGB with each deformation cue alone, without the tracklets themselves, gives a parameter distance of 0.76--0.77, and RGB with all three cues gives 0.78. These values lie between RGB (0.80) and RGB + 2D tracklets (0.74).
Adding the cues to the 2D tracklets changes little: 0.739 with all three cues and 0.761 with the non-affine residual, against 0.745 for the tracklets alone.
The cues are computed from the tracklets, and they are relative quantities: strain, stretch, and the non-affine residual do not change when the whole object moves or rotates, so they carry no information about its motion.
The tracklets keep the position and the displacement of every point, so they contain both the motion of the object and the local deformation from which the cues are computed.
A likely reason why the cues add little is that the model can learn from the tracklets what it needs.

\paragraph{Repeated runs.}
Every model of the main paper is a single training run; App.~\ref{sec:app_seeds} retrains nine of them with two more seeds.
RGB + 3D tracklets was also trained a second time, in the study of Table~\ref{tab:3d_cross_eval}.
That study trains the model under four conditions, A--D. Condition D, with 4RC depth and WAFT tracks, has the same inputs as the main run, so it is a second run of the same model. We call it run D.

\paragraph{Teacher--student training.}
Table~\ref{tab:ablation_students} varies the teacher--student training.
With 2D tracklets, every student variant has a smaller parameter distance on real drops than the direct predictor, RGB + 2D tracklets: 0.68--0.73 against 0.74.
The $z_E$ error on synthetic validation is nearly the same.
\model{} uses a teacher that is trained for ten epochs and selected on synthetic validation.
With a teacher stopped after five epochs instead, the parameter distance of the student is 0.699 instead of 0.677.
Two variants change the student of the five-epoch teacher (0.699).
In the first, the material head is trained together with the student instead of being kept frozen; its parameter distance is 0.711.
In the second, the feature loss has a smaller weight, 0.25 (Table~\ref{tab:ablation_students}); its parameter distance is 0.726.
Neither helps.
With lifted tracklets, the two students have parameter distances of 1.09 and 1.16.
They stay far from every model on 2D tracklets, so training with the teacher does not repair the lifted input.

\paragraph{Regression target.}
Our models predict the density-normalized ratios $E/\rho$ and $\sigma_Y/\rho$.
App.~\ref{sec:app_oracle} compares this regression target with the absolute parameters $E$ and $\sigma_Y$, for the teacher on exact surface deformation (Table~\ref{tab:oracle}).
For this comparison, we train a second teacher on \model{}-Sim with absolute targets, $\log_{10}E$ and $\log_{10}\sigma_Y$, and select its checkpoint in the same way.
On the 14,293 synthetic validation videos with usable surface geometry, its error is 0.272\,dex for stiffness and 0.262\,dex for yield stress.
On the same videos, the teacher with density-normalized ratios reaches 0.161\,dex and 0.192\,dex.
No input carries the density, so an absolute target adds the spread of $\log_{10}\rho$ to the error.
We also train a student on the teacher with absolute targets, with everything else as for \model{}.
On synthetic validation it is nearly as accurate as \model{}: its error of $\log_{10}E$ / $\log_{10}\sigma_Y$ is 0.494 / 0.360\,dex, against 0.481 / 0.356\,dex for \model{}. For \model{} these equal its $z_E$ and $z_Y$ errors, because the density of each synthetic video is known.
On the real drops it is much worse.
Its parameter distance is 1.016, against 0.677 for \model{}.
Replay with its material increases the total loss by 13.6\% and the deformation loss by 19.7\% over the reference, against 10.1\% and 12.3\% for \model{}.
Its $\mathrm{Err}_{\mathrm{ref}}$ is 0.136, against 0.107.

\paragraph{Backbone and adaptation.}
All our predictors use V-JEPA~2 with the same adaptation: LoRA of rank 16 on the query and value projections of the last 12 layers, with the patch projection and the layer norms also trained (2.5\,M of the backbone's 305\,M parameters).
For RGB + 2D tracklets we trained the same model with the backbone frozen entirely, all else equal.
Its $z_E$ / $z_Y$ error on synthetic validation rises from 0.482 / 0.363 to 0.692 / 0.447\,dex.
Its parameter distance on the 80 real drops rises from 0.745 to 0.960.
We also fine-tune the whole backbone without LoRA, at the learning rate of the trained backbone parameters in our default setting ($10^{-4}$).
The synthetic error is then 0.374 / 0.317\,dex and the parameter distance 0.818.
We did not vary the backbone itself in a matched setting.
The Vid2Sim predictor of Table~\ref{tab:external_comparisons} is a fully fine-tuned VideoMAE trained on the same videos and targets.
Its synthetic error is 0.556 / 0.414\,dex and its parameter distance 0.960, against 0.478 / 0.362\,dex and 0.797 for our RGB-only predictor.

\section{Per-Object and Per-Metric Results}
\label{sec:app_epoch5_results}
All predictors are evaluated after the fifth epoch.
Lifted models are the models that read tracklets lifted to 3D.
The main paper reports the parameter distance for some of them (Table~\ref{tab:representation_tradeoff}) and the $z_E$ error for others (Table~\ref{tab:same_drops}, Fig.~\ref{fig:sim_vs_real}).
Table~\ref{tab:lifted_all_metrics} gives every parameter metric for all of them.
Every lifted model is worse than RGB + 2D tracklets in parameter distance and in both stiffness metrics, the $E$ distance and the $z_E$ error.
This holds for each depth source and each training condition.
For most of them the $z_Y$ error is higher as well.
The two exceptions are model A, trained on simulator depth and tracks (0.361\,dex against 0.423\,dex for RGB + 2D tracklets), and the model on 3D tracklets alone (0.408\,dex).

Tables~\ref{tab:objects_param_distance}--\ref{tab:objects_replay} break the real-video results down by object, for five of our predictors and three baselines.
On four soft objects (crocs, glove, pillow, and sneaker), the lifted models are closer to the references than the models on 2D tracklets. On the stiff objects, they are much farther from the references.
On real video, the stiffness that the lifted models predict is lower than the reference on 67 (RGB + 3D tracklets) and 66 (student on 3D tracklets) of the 80 drops, by 1.1\,dex on average.
The 2D models have no such shift (App.~\ref{sec:app_reference_free}).
\model{} has a smaller parameter distance than RGB + 2D tracklets on 12 of the 16 objects.
It also has a smaller parameter distance than RGB + 3D tracklets on 12 of the 16 objects.
No predictor is accurate on every object: for box, hanger, and medkit, all of ours are more than 1\,dex from the plausible region.

Table~\ref{tab:tensile_numbers} in App.~\ref{sec:app_real_measure} lists the numbers behind the tensile comparison of Fig.~\ref{fig:benchmark}; App.~\ref{sec:app_reference_free} discusses them.

\begin{table*}[t]\centering\small
\caption{\textbf{Geometric inputs of the direct predictors.} All rows share the training set and the schedule, and all rows except the last two (``only'') share the RGB branch. All are evaluated after the fifth epoch. Syn.: $z_E$ / $z_Y$ error on the 14,588 validation videos (dex). Real: parameter distance and $z_E$ error on the 80 real drops, and $z_Y$ error on the 50 drops whose reference simulation yields (dex).}\label{tab:ablation_inputs}
\begin{tabular}{@{}lcccc@{}}\toprule
Input & Syn.\ $z_E$ / $z_Y$ $\downarrow$ & Parameter distance $\downarrow$ & Real $z_E$ $\downarrow$ & Real $z_Y$ $\downarrow$ \\\midrule
RGB & 0.478 / 0.362 & 0.797 & 0.829 & 0.457 \\
RGB + dense flow & 0.472 / 0.361 & 0.833 & 0.862 & 0.421 \\
RGB + depth & 0.477 / 0.363 & 0.789 & 0.822 & 0.454 \\
\midrule
RGB + 2D strain & 0.484 / 0.364 & 0.773 & 0.829 & 0.436 \\
RGB + 2D stretch & 0.481 / 0.363 & 0.757 & 0.791 & 0.436 \\
RGB + 2D non-affine residual & 0.479 / 0.363 & 0.764 & 0.805 & 0.447 \\
RGB + three 2D cues & 0.483 / 0.364 & 0.778 & 0.831 & 0.402 \\
\midrule
RGB + 2D tracklets & 0.482 / 0.363 & 0.745 & 0.792 & 0.423 \\
RGB + 2D tracklets + non-affine residual & 0.484 / 0.363 & 0.761 & 0.804 & 0.440 \\
RGB + 2D tracklets + three cues & 0.487 / 0.367 & 0.739 & 0.792 & 0.426 \\
\midrule
RGB + 3D tracklets & 0.444 / 0.344 & 1.210 & 1.249 & 0.583 \\
\quad second training run (D of Table~\ref{tab:3d_cross_eval}) & 0.450 / 0.344 & 1.148 & 1.188 & 0.581 \\
\midrule
2D tracklets only & 0.764 / 0.461 & 1.080 & 1.110 & 0.492 \\
Three 2D cues only & 0.830 / 0.475 & 1.182 & 1.216 & 0.506 \\
\bottomrule\end{tabular}\end{table*}

\begin{table*}[t]\centering\small
\caption{\textbf{Teacher--student variants.} Columns as in Table~\ref{tab:ablation_inputs}; all 80 real drops. The two indented rows below ``Student, teacher after five epochs'' use the same five-epoch teacher. Each changes one thing from that row.}\label{tab:ablation_students}
\begin{tabular}{@{}lcccc@{}}\toprule
Model & Syn.\ $z_E$ / $z_Y$ $\downarrow$ & Parameter distance $\downarrow$ & Real $z_E$ $\downarrow$ & Real $z_Y$ $\downarrow$ \\\midrule
RGB + 2D tracklets, no teacher & 0.482 / 0.363 & 0.745 & 0.792 & 0.423 \\
\midrule
Student, teacher after five epochs & 0.482 / 0.357 & 0.699 & 0.745 & 0.397 \\
\quad material head trained with the student & 0.479 / 0.357 & 0.711 & 0.766 & 0.394 \\
\quad feature-loss weight 0.25 & 0.488 / 0.360 & 0.726 & 0.771 & 0.402 \\
Student, teacher selected from ten epochs (\model{}) & 0.481 / 0.356 & \textbf{0.677} & 0.731 & 0.381 \\
\midrule
RGB + 3D tracklets, no teacher & 0.444 / 0.344 & 1.210 & 1.249 & 0.583 \\
Student on 3D tracklets, teacher after five epochs & 0.440 / 0.336 & 1.091 & 1.143 & 0.514 \\
Student on 3D tracklets, teacher selected from ten epochs & 0.443 / 0.336 & 1.164 & 1.214 & 0.552 \\
\bottomrule\end{tabular}\end{table*}

\begin{table*}[t]\centering\small
\caption{\textbf{Every real-video parameter metric for the lifted models} of Tables~\ref{tab:representation_tradeoff} and~\ref{tab:same_drops} and Fig.~\ref{fig:sim_vs_real}, on the 80 real drops (dex). Syn.: $z_E$ error on synthetic validation, evaluated with the same depth and tracks as the model was trained on. Parameter distance: Eq.~\eqref{eq:plausible_region_distance}. $E$ distance: distance of the predicted stiffness to the range of stiffness that the plausible region covers, without the yield stress. $z_E$, $z_Y$: errors against the reference material ($z_Y$ on the 50 drops whose reference simulation yields). Clip scale and frame scale: the depth is made metric with one scale per clip or per frame. The main run is the RGB + 3D tracklets model of Table~\ref{tab:representation_tradeoff}. A--D are the training conditions of Table~\ref{tab:3d_cross_eval}.}\label{tab:lifted_all_metrics}
\begin{tabular}{@{}lccccc@{}}\toprule
Model & Syn.\ $z_E$ $\downarrow$ & Parameter distance $\downarrow$ & $E$ distance $\downarrow$ & $z_E$ $\downarrow$ & $z_Y$ $\downarrow$ \\\midrule
RGB + 2D tracklets & 0.482 & 0.745 & 0.650 & 0.792 & 0.423 \\
\model{} & 0.481 & 0.677 & 0.591 & 0.731 & 0.381 \\
\midrule
RGB + 3D tracklets: 4RC depth, clip scale (main run) & 0.444 & 1.210 & 1.140 & 1.249 & 0.583 \\
\quad Depth Anything 3, clip scale & 0.448 & 1.255 & 1.189 & 1.302 & 0.593 \\
\quad Depth Anything 3, frame scale & 0.426 & 1.223 & 1.155 & 1.267 & 0.557 \\
\quad 4RC depth with depth augmentation & 0.455 & 0.928 & 0.850 & 0.964 & 0.537 \\
Student, 3D tracklets & 0.443 & 1.164 & 1.105 & 1.214 & 0.552 \\
\midrule
A: simulator depth, simulator tracks & 0.414 & 1.260 & 1.200 & 1.305 & 0.361 \\
B: simulator depth, WAFT tracks & 0.421 & 1.090 & 1.026 & 1.132 & 0.465 \\
C: 4RC depth, simulator tracks & 0.451 & 1.142 & 1.065 & 1.178 & 0.589 \\
D: 4RC depth, WAFT tracks & 0.450 & 1.148 & 1.075 & 1.188 & 0.581 \\
\midrule
3D tracklets only (4RC depth) & 0.702 & 1.337 & 1.293 & 1.411 & 0.408 \\
\bottomrule\end{tabular}\end{table*}

\begin{table*}[t]\centering\small
\caption{\textbf{Parameter distance per object} on the real drops (dex; mean over the five drops of each object). The last row is the mean over all 80 drops, as in Tables~\ref{tab:representation_tradeoff} and~\ref{tab:external_comparisons}. Trk.: tracklets.}\label{tab:objects_param_distance}
\begin{adjustbox}{max width=\textwidth}
\begin{tabular}{@{}lcccccccc@{}}\toprule
Object & RGB & \makecell{RGB +\\2D trk.} & \makecell{RGB +\\3D trk.} & \makecell{Student,\\3D trk.} & \model{} & Vid2Sim & \makecell{Analytic\\Dynamics} & \makecell{VLM\\(video)} \\\midrule
belt & 0.138 & 0.165 & 0.434 & 0.595 & 0.116 & 0.366 & 0.501 & 2.417 \\
bottle & 1.138 & 1.116 & 2.567 & 2.475 & 0.946 & 1.505 & 2.385 & 0.562 \\
bowl & 0.406 & 0.331 & 2.000 & 2.022 & 0.477 & 1.379 & 1.397 & 1.033 \\
box & 1.601 & 1.380 & 2.234 & 2.163 & 1.356 & 1.875 & 1.870 & 2.036 \\
crocs & 1.146 & 1.046 & 0.645 & 0.572 & 0.838 & 0.550 & 0.526 & 0.646 \\
diaper & 0.416 & 0.345 & 1.453 & 1.286 & 0.150 & 0.692 & 0.791 & 0.844 \\
glove & 1.156 & 1.093 & 0.500 & 0.364 & 0.948 & 0.557 & 1.255 & 1.608 \\
hanger & 1.161 & 1.414 & 2.388 & 2.271 & 1.266 & 1.703 & 2.292 & 1.793 \\
kitty & 0.342 & 0.208 & 0.262 & 0.361 & 0.220 & 0.494 & 0.194 & 0.643 \\
medkit & 1.211 & 1.187 & 2.351 & 2.247 & 1.475 & 1.344 & 1.602 & 1.539 \\
napkin & 0.874 & 0.772 & 1.967 & 1.843 & 0.912 & 1.798 & 1.053 & 0.438 \\
neck pillow & 0.534 & 0.412 & 0.434 & 0.365 & 0.244 & 0.355 & 0.539 & 0.435 \\
pad & 0.094 & 0.147 & 0.172 & 0.175 & 0.003 & 0.130 & 0.275 & 1.780 \\
pillow & 1.172 & 1.123 & 0.354 & 0.479 & 0.966 & 1.147 & 0.876 & 0.253 \\
sneaker & 1.237 & 1.027 & 0.394 & 0.326 & 0.783 & 0.619 & 1.548 & 1.272 \\
trash bin & 0.119 & 0.150 & 1.198 & 1.085 & 0.136 & 0.848 & 0.233 & 0.568 \\
\midrule
All drops & 0.797 & 0.745 & 1.210 & 1.164 & 0.677 & 0.960 & 1.084 & 1.117 \\
\bottomrule\end{tabular}
\end{adjustbox}
\end{table*}

\begin{table*}[t]\centering\small
\caption{\textbf{Stiffness and yield-stress errors per object} on the real drops: $z_E$ / $z_Y$ error against the reference material (dex). The $z_Y$ error is defined for the ten objects whose reference simulation yields; a dash marks the others. Trk.: tracklets.}\label{tab:objects_zE_zY}
\begin{adjustbox}{max width=\textwidth}
\begin{tabular}{@{}lcccccccc@{}}\toprule
Object & RGB & \makecell{RGB +\\2D trk.} & \makecell{RGB +\\3D trk.} & \makecell{Student,\\3D trk.} & \model{} & Vid2Sim & \makecell{Analytic\\Dynamics} & \makecell{VLM\\(video)} \\\midrule
belt & 0.20 / -- & 0.21 / -- & 0.49 / -- & 0.65 / -- & 0.17 / -- & 0.43 / -- & 0.56 / -- & 2.48 / -- \\
bottle & 1.18 / 0.72 & 1.24 / 0.62 & 2.73 / 0.79 & 2.61 / 0.87 & 1.05 / 0.57 & 1.58 / 0.78 & 2.51 / 0.89 & 0.69 / 0.25 \\
bowl & 0.57 / 0.24 & 0.57 / 0.18 & 2.36 / 0.69 & 2.38 / 0.72 & 0.68 / 0.14 & 1.76 / 0.49 & 1.68 / 0.13 & 0.76 / 0.77 \\
box & 1.55 / 0.61 & 1.37 / 0.44 & 2.03 / 1.07 & 2.05 / 0.89 & 1.32 / 0.52 & 1.70 / 0.97 & 1.84 / 0.58 & 2.10 / 0.19 \\
crocs & 1.08 / 0.71 & 0.98 / 0.70 & 0.63 / 0.19 & 0.58 / 0.15 & 0.74 / 0.65 & 0.44 / 0.23 & 0.20 / 0.53 & 0.34 / 0.70 \\
diaper & 0.25 / 0.33 & 0.34 / 0.19 & 1.33 / 0.65 & 1.17 / 0.58 & 0.17 / 0.09 & 0.62 / 0.35 & 0.70 / 0.40 & 0.30 / 0.87 \\
glove & 1.25 / -- & 1.19 / -- & 0.59 / -- & 0.46 / -- & 1.04 / -- & 0.65 / -- & 1.35 / -- & 1.70 / -- \\
hanger & 0.73 / 0.96 & 0.91 / 1.13 & 2.52 / 0.36 & 2.36 / 0.51 & 0.89 / 1.02 & 1.57 / 0.68 & 1.67 / 1.27 & 0.45 / 1.77 \\
kitty & 0.39 / -- & 0.26 / -- & 0.26 / -- & 0.36 / -- & 0.26 / -- & 0.49 / -- & 0.21 / -- & 0.69 / -- \\
medkit & 1.25 / 0.23 & 1.24 / 0.17 & 2.38 / 0.42 & 2.28 / 0.35 & 1.53 / 0.17 & 1.40 / 0.24 & 1.68 / 0.15 & 1.40 / 0.80 \\
napkin & 1.06 / 0.39 & 0.95 / 0.37 & 2.14 / 0.74 & 2.05 / 0.67 & 1.13 / 0.30 & 1.98 / 0.73 & 1.26 / 0.37 & 0.07 / 0.53 \\
neck pillow & 0.61 / -- & 0.51 / -- & 0.45 / -- & 0.38 / -- & 0.32 / -- & 0.37 / -- & 0.56 / -- & 0.53 / -- \\
pad & 0.50 / -- & 0.52 / -- & 0.26 / -- & 0.26 / -- & 0.30 / -- & 0.14 / -- & 0.37 / -- & 2.21 / -- \\
pillow & 1.17 / -- & 1.12 / -- & 0.35 / -- & 0.48 / -- & 0.97 / -- & 1.15 / -- & 0.88 / -- & 0.25 / -- \\
sneaker & 1.29 / 0.23 & 1.08 / 0.32 & 0.40 / 0.13 & 0.33 / 0.11 & 0.83 / 0.26 & 0.66 / 0.28 & 1.52 / 0.83 & 1.20 / 0.90 \\
trash bin & 0.17 / 0.13 & 0.18 / 0.11 & 1.07 / 0.79 & 1.02 / 0.67 & 0.27 / 0.07 & 0.56 / 0.70 & 0.18 / 0.18 & 0.36 / 0.40 \\
\midrule
All drops & 0.83 / 0.46 & 0.79 / 0.42 & 1.25 / 0.58 & 1.21 / 0.55 & 0.73 / 0.38 & 0.97 / 0.54 & 1.07 / 0.53 & 0.97 / 0.72 \\
\bottomrule\end{tabular}
\end{adjustbox}
\end{table*}

\begin{table*}[t]\centering\small
\caption{\textbf{Replay per object:} relative increase of the deformation replay loss over the reference material (\%) / direct simulation difference $\mathrm{Err}_{\mathrm{ref}}$. A negative increase means that the predicted material reproduces the observed deformation better than the reference after both initial states are refined. Last row: mean over the 80 drops. Trk.: tracklets.}\label{tab:objects_replay}
\begin{adjustbox}{max width=\textwidth}
\begin{tabular}{@{}lcccccccc@{}}\toprule
Object & RGB & \makecell{RGB +\\2D trk.} & \makecell{RGB +\\3D trk.} & \makecell{Student,\\3D trk.} & \model{} & Vid2Sim & \makecell{Analytic\\Dynamics} & \makecell{VLM\\(video)} \\\midrule
belt & $-$5\% / 0.035 & 0\% / 0.034 & 0\% / 0.081 & 16\% / 0.109 & $-$6\% / 0.032 & $-$1\% / 0.050 & 0\% / 0.055 & 57\% / 0.117 \\
bottle & $-$2\% / 0.117 & $-$3\% / 0.104 & $-$12\% / 0.315 & $-$10\% / 0.292 & 6\% / 0.083 & $-$19\% / 0.154 & $-$14\% / 0.262 & 3\% / 0.169 \\
bowl & 4\% / 0.033 & $-$3\% / 0.030 & 72\% / 0.085 & 69\% / 0.090 & 5\% / 0.059 & 44\% / 0.086 & 71\% / 0.097 & 56\% / 0.158 \\
box & 15\% / 0.232 & 16\% / 0.215 & 31\% / 0.318 & 28\% / 0.301 & 21\% / 0.208 & 21\% / 0.265 & 22\% / 0.270 & 2\% / 0.339 \\
crocs & 21\% / 0.065 & 9\% / 0.054 & 30\% / 0.108 & 28\% / 0.099 & 10\% / 0.075 & 44\% / 0.120 & 32\% / 0.136 & 27\% / 0.141 \\
diaper & 45\% / 0.123 & 35\% / 0.110 & 107\% / 0.151 & 47\% / 0.129 & 8\% / 0.041 & 81\% / 0.084 & 28\% / 0.091 & 16\% / 0.090 \\
glove & 21\% / 0.310 & 19\% / 0.307 & 0\% / 0.241 & 1\% / 0.220 & 16\% / 0.295 & $-$3\% / 0.250 & 8\% / 0.323 & 24\% / 0.365 \\
hanger & 5\% / 0.054 & $-$5\% / 0.083 & 4\% / 0.074 & 8\% / 0.061 & $-$6\% / 0.073 & 3\% / 0.063 & 4\% / 0.111 & 7\% / 0.109 \\
kitty & 34\% / 0.117 & 14\% / 0.082 & 2\% / 0.105 & 9\% / 0.138 & 19\% / 0.084 & 4\% / 0.181 & $-$4\% / 0.083 & 44\% / 0.165 \\
medkit & 14\% / 0.133 & 18\% / 0.137 & 33\% / 0.220 & 26\% / 0.221 & 15\% / 0.176 & 18\% / 0.164 & 2\% / 0.187 & 40\% / 0.246 \\
napkin & 18\% / 0.137 & $-$20\% / 0.116 & 58\% / 0.223 & 75\% / 0.213 & 26\% / 0.154 & 24\% / 0.190 & 47\% / 0.160 & 9\% / 0.097 \\
neck pillow & 9\% / 0.081 & 10\% / 0.086 & 3\% / 0.120 & 2\% / 0.102 & 17\% / 0.060 & 12\% / 0.105 & 3\% / 0.141 & 27\% / 0.114 \\
pad & 14\% / 0.174 & 21\% / 0.185 & 11\% / 0.142 & 14\% / 0.150 & 14\% / 0.135 & 15\% / 0.105 & 24\% / 0.164 & 46\% / 0.347 \\
pillow & 23\% / 0.141 & 15\% / 0.140 & $-$1\% / 0.059 & $-$11\% / 0.082 & 27\% / 0.131 & 5\% / 0.136 & $-$2\% / 0.118 & 3\% / 0.068 \\
sneaker & 50\% / 0.145 & 37\% / 0.122 & 10\% / 0.090 & 8\% / 0.067 & 27\% / 0.095 & 18\% / 0.077 & 44\% / 0.104 & 38\% / 0.095 \\
trash bin & 21\% / 0.018 & 45\% / 0.024 & 150\% / 0.073 & 147\% / 0.062 & $-$1\% / 0.016 & 162\% / 0.076 & 33\% / 0.039 & 34\% / 0.079 \\
\midrule
All drops & 18\% / 0.120 & 13\% / 0.114 & 31\% / 0.150 & 28\% / 0.146 & 12\% / 0.107 & 27\% / 0.132 & 19\% / 0.146 & 27\% / 0.169 \\
\bottomrule\end{tabular}
\end{adjustbox}
\end{table*}

\clearpage
\onecolumn
\raggedbottom
\suppresetcolumn
\section{Material Parameter Measurements}
\label{sec:app_real_measure}
\begingroup\small
\setlength{\abovedisplayskip}{6pt}\setlength{\belowdisplayskip}{6pt}
The tensile tests were run on an Instron 68FM-100 universal testing machine. The specimens were photographed with a Teledyne FLIR Blackfly~S BFS-U3-200S6M monochrome camera (Sony IMX183 CMOS sensor, $5472\times3648$ pixels) through a Kowa LM75FC24M C-mount lens (focal length 75\,mm, aperture f/2.5--f/16). We analyze 38 tensile test records from eight objects. A record can contain several loading-unloading cycles. For the 11 records with four loading-unloading cycles, we use only the fourth loading data to calculate the stiffness. Each of the other records keeps all the loading branches it has. This gives 39 analyzed branches in total. %

\subsection{Computation and interval selection}
Each uniaxial tensile test records the displacement $d$ and the force $F$. For each specimen we also have the initial gauge length $L_0$, the full length $L_{\rm total}$, the initial cross-sectional area $A_0$ (mean width times mean thickness), and the mass $m$. From these we calculate three apparent quantities of the specimen: the engineering strain $\varepsilon$, the engineering stress $\sigma$, and the mass density $\rho_{\mathrm{env}}$,
\begin{equation}
\varepsilon=d/L_0,\qquad \sigma=F/A_0,\qquad
\qquad \rho_{\rm env}=\frac{m}{A_0L_{\rm total}}.
\end{equation}

For each retained loading branch, we select one approximately linear region of the stress--strain curve, below 20\% strain and before the peak stress. We fit a straight line with a free intercept to all data points in this region. Its slope is Young's modulus $E_{[a,b]}$, where $[a,b]$ is the range of engineering strain used for the fit. Its intercept $c$ comes from pre-stress in the uniaxial tensile tests. 
\begin{equation}
    (E_{[a,b]},c)=\arg\min_{E,c}\sum_{a\leq\varepsilon_i\leq b}(\sigma_i-E\varepsilon_i-c)^2,
\end{equation}

For each object and each loading direction, we report the arithmetic mean and sample standard deviation across specimens ($n$ is the number of tensile tests used for that object and loading direction):
\begin{equation}
\overline E=\frac1n\sum_{j=1}^{n}E_j,\qquad
s_E=\sqrt{\frac{\sum_{j=1}^{n}(E_j-\overline E)^2}{n-1}}.
\end{equation}
Pad measurement 1 has two fitted loading branches. We first average their two slopes and use this average as the $E_j$ of that specimen. For failure stress, density, and Poisson's ratio, we also report the mean and sample standard deviation, over the specimens for which the quantity is available.

\paragraph{Yield and failure stress.}
Different types of materials have different definitions of yield stress $\sigma_Y$, and the recorded stress--strain curves do not reliably identify one. The tables below therefore list $\sigma_Y$ as ``Not identified''. Instead, we mark and report the peak stress $\sigma_{\rm peak}=\max_i F_i/A_0$ over the retained branch(es). The object summary lists it as ``Peak stress''. Where rupture of the specimen is not documented, $\sigma_{\rm peak}$ is the highest recorded nominal stress, not a confirmed fracture stress.
For display, a curve with negative stress is shifted upward so its minimum is zero. Fits and statistics use the original stresses. In the plots, the selected fit intervals are drawn darker, and the dashed fitted lines extend beyond these intervals so that they are visible. Some plots are cut at a maximum strain, which their captions state. The reported peaks $\sigma_{\rm peak}$ are still taken over the full record. On the pages of the objects, M$j$ denotes measurement $j$ and L$k$ its $k$-th loading.

\paragraph{Transverse contraction.}
We measure the transverse contraction from pairs of photos taken during the first loading. The first photo of a pair shows the specimen at 0\% strain. The second shows it at the available strain nearest to 5\%, or at the largest valid strain below 5\%. In each photo, we annotate one horizontal section across the specimen by hand. Its left and right endpoints $x_L$ and $x_R$ give the width $w$ in pixels of the original image. The two photos of a pair are displayed with the same crop and scale. The photos are taken at 10\,fps, and we interpolate the recorded displacement $d$ to the times $t_0$ and $t_i$ of the two photos. We then compute the apparent Poisson's ratio~\cite{astmE132}:
\begin{equation}
w_0=x_{R,0}-x_{L,0},\quad w_i=x_{R,i}-x_{L,i},\quad
\varepsilon_{\parallel}=\frac{d(t_i)-d(t_0)}{L_0+d(t_0)},\quad
\varepsilon_{\perp}=\frac{w_i}{w_0}-1,\quad
\nu_{\rm app}=-\frac{\varepsilon_{\perp}}{\varepsilon_{\parallel}}.
\end{equation}
All 37 available pairs were confirmed; diaper transverse measurement 1 has missing photos. Napkin's frayed boundaries make its ratios noisy, so they are excluded from Poisson's ratio summaries. The silicone pad has a honeycomb structure. Its contraction therefore describes this structure, not the Poisson's ratio of the silicone itself. The two loading directions of the diaper give apparent ratios of opposite sign, so we keep them separate. We do not clip the measured ratios. We also do not change the simulation inputs to match them: each object keeps its fixed Poisson's ratio.

\clearpage
\subsection{Object summary}
Entries give mean $\pm$ sample standard deviation values over available specimens (NA: missing or excluded measurements). Hanger, crocs, Kitty, and the silicone pad were tested in one loading direction, which we list as ``Isotropic''.
\par\nopagebreak\smallskip
{\centering\small\setlength{\tabcolsep}{4pt}
\begin{tabular}{@{}llrrrr@{}}\toprule
Object & Direction & \makecell{$E$\\(MPa)} & \makecell{Peak stress\\(MPa)} & \makecell{Poisson's ratio\\$\nu_{\rm app}$} & \makecell{Density\\(kg/m$^3$)}\\\midrule
Box & Axial & 285 $\pm$ 57 & 6.75 $\pm$ 0.47 & 0.127 $\pm$ 0.009 & 254.5 $\pm$ 8.6 \\
Box & Transverse & 541 $\pm$ 81 & 8.49 $\pm$ 0.67 & 0.178 $\pm$ 0.036 & 264.0 $\pm$ 11.0 \\
Trash bin & Axial & 785 $\pm$ 101 & 17.5 $\pm$ 1.9 & 0.325 $\pm$ 0.005 & NA\\
Trash bin & Transverse & 701 $\pm$ 105 & 20.17 $\pm$ 0.74 & 0.304 $\pm$ 0.037 & 945.0 $\pm$ 3.0 \\
Hanger & Isotropic & 963 $\pm$ 250 & 22.7 $\pm$ 10.4 & 0.266 $\pm$ 0.048 & 481.8 $\pm$ 51.1\\
Crocs & Isotropic & 1.62 $\pm$ 0.07 & 2.25 $\pm$ 0.25 & 0.39 $\pm$ 0.07 & 287.9 $\pm$ 16.6\\
Kitty & Isotropic & 0.088 $\pm$ 0.040 & 0.55 $\pm$ 0.19 & 0.29 $\pm$ 0.07 & 69.61 $\pm$ 13\\
Silicone pad & Isotropic & 0.175 $\pm$ 0.063 & 0.33 $\pm$ 0.05 & 0.58 $\pm$ 0.26 & 376.2 $\pm$ 43.8\\
Napkin & Axial & 11.3 $\pm$ 2.2 & 1.56 $\pm$ 0.28 & NA & 199.8 $\pm$ 21.2\\
Napkin & Transverse & 33.2 $\pm$ 8.0 & 1.82 $\pm$ 0.60 & NA & 265.5 $\pm$ 48.7\\
Diaper & Axial & 0.43 $\pm$ 0.31 & 1.16 $\pm$ 0.44 & -0.25 $\pm$ 0.16 & 124.2 $\pm$ 24.1\\
Diaper & Transverse & 0.050 $\pm$ 0.046 & 0.31 $\pm$ 0.03 & 0.37 $\pm$ 0.35 & 137.8 $\pm$ 13.4\\
\bottomrule\end{tabular}\par}
\begin{table*}[!htb]\centering\small
\caption{\textbf{Reference stiffness against tensile measurements} for the eight objects we cut specimens from (the numbers behind Fig.~\ref{fig:benchmark}). Specimen $E$: mean $\pm$ sample standard deviation of Young's modulus for each loading direction, in MPa (App.~\ref{sec:app_real_measure}). The next four columns give $\log_{10}E$ with $E$ in Pa. Specimen means: the means of the loading directions, in increasing order. Mean $\pm$ 3 std: the band drawn in Fig.~\ref{fig:benchmark}, from the lowest mean minus 3 std to the highest mean plus 3 std over the loading directions; it has no lower bound when the mean minus 3 std is negative. Reference: the stiffness of our reference material. Plausible region: the range of stiffness covered by the plausible region of this reference. Last column: the reference minus the nearest specimen mean, in dex; ``within'' if the reference lies between the two directional means. Region to band: distance in dex from the plausible region to the mean $\pm$ 3 std band, 0 if they overlap, negative if the region is softer.}\label{tab:tensile_numbers}
\begin{adjustbox}{max width=\textwidth}
\begin{tabular}{@{}llcccccc@{}}\toprule
Object & Specimen $E$ (MPa) & Specimen means & Mean $\pm$ 3 std & Reference & Plausible region & Ref.\ $-$ specimen & Region to band \\\midrule
box & axial 285 $\pm$ 57; transverse 541 $\pm$ 81 & 8.45, 8.73 & [8.06, 8.89] & 8.89 & [8.82, 8.99] & $+0.16$ & 0 \\
trash bin & axial 785 $\pm$ 101; transverse 701 $\pm$ 105 & 8.85, 8.89 & [8.59, 9.04] & 8.00 & [7.88, 8.17] & $-0.85$ & $-0.42$ \\
hanger & 963 $\pm$ 250 & 8.98 & [8.33, 9.23] & 8.91 & [8.72, 8.96] & $-0.07$ & 0 \\
napkin & axial 11.3 $\pm$ 2.2; transverse 33.2 $\pm$ 8 & 7.05, 7.52 & [6.67, 7.76] & 6.64 & [6.37, 6.79] & $-0.41$ & 0 \\
crocs & 1.62 $\pm$ 0.07 & 6.21 & [6.15, 6.26] & 6.08 & [6.06, 6.22] & $-0.13$ & 0 \\
diaper & axial 0.43 $\pm$ 0.31; transverse 0.05 $\pm$ 0.046 & 4.70, 5.63 & ($-\infty$, 6.13] & 6.00 & [5.97, 6.13] & $+0.37$ & 0 \\
pad & 0.175 $\pm$ 0.063 & 5.24 & ($-\infty$, 5.56] & 4.00 & [4.00, 4.43] & $-1.24$ & 0 \\
kitty & 0.088 $\pm$ 0.04 & 4.94 & ($-\infty$, 5.32] & 4.00 & [4.00, 4.05] & $-0.94$ & 0 \\
\bottomrule\end{tabular}
\end{adjustbox}
\end{table*}

\endgroup

\clearpage
\subsection{Box (Axial)}
\noindent\includegraphics[width=\textwidth]{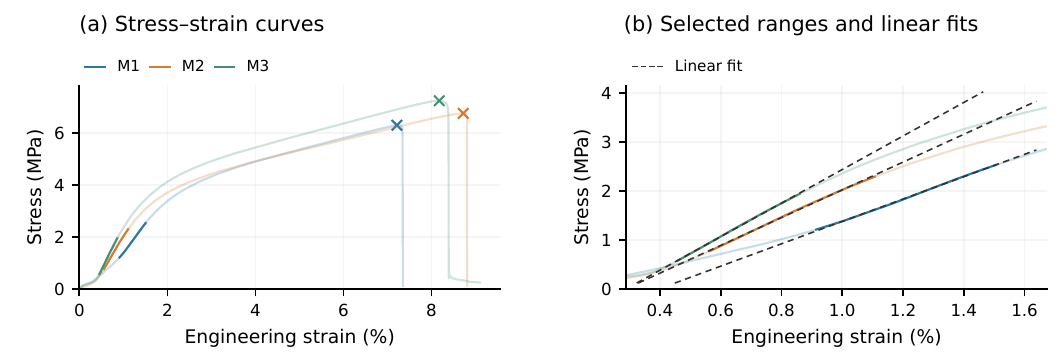}\par
{\footnotesize Left: the loading curves (stress against engineering strain). Right: the strain ranges selected for the fits, drawn darker, and the linear fits, drawn dashed and extended beyond the selected ranges. Each measurement has the same color in both panels. A cross marks the highest stress recorded within the displayed range; it is a peak, not a yield point.}\par\smallskip
\noindent\includegraphics[width=\textwidth]{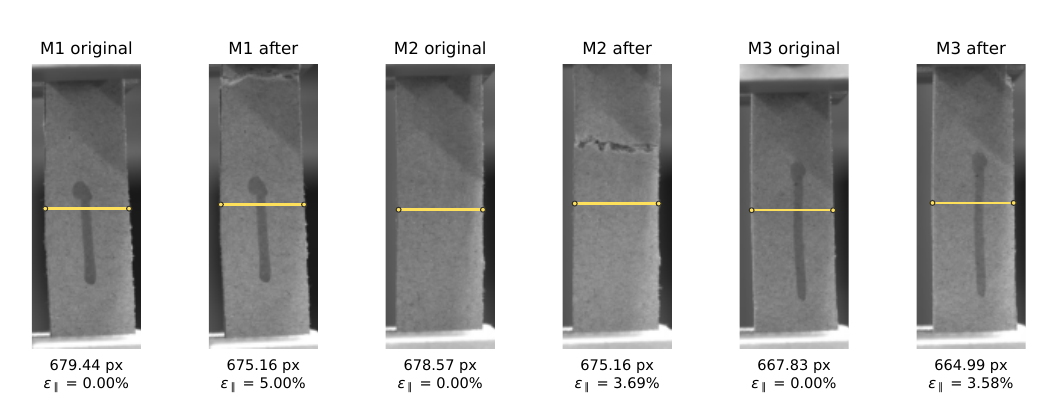}\par
{\footnotesize Photo pairs of the first loading, for the measurements that have them. ``Original'' is the specimen at 0\% strain, and ``after'' is the same specimen under load. The yellow line and its endpoints show the width measured by hand across the specimen. The axial strain reached in each photo is printed below it.}\par\medskip
{\small\setlength{\tabcolsep}{5pt}\centering\begin{tabular}{@{}ccccccc@{}}\toprule
Meas./load & Fit strain (\%) & $E$ (MPa) & $c$ (MPa) & $R^2$ & $\sigma_{\rm peak}$ (MPa) & $\sigma_Y$\\\midrule
M1/L1 & 0.913--1.506 & 228.47 & -0.9014 & 0.9994 & 6.286 & Not identified\\
M2/L1 & 0.564--1.109 & 283.54 & -0.8096 & 0.9993 & 6.74 & Not identified\\
M3/L1 & 0.457--0.862 & 342.52 & -0.9824 & 0.9998 & 7.226 & Not identified\\
\bottomrule\end{tabular}\par}\smallskip
{\footnotesize\noindent\textbf{Per-measurement calculations.} In $\sigma=E\varepsilon+c$, $\varepsilon$ is a fraction, not a percentage; the interval, slope, and intercept are given above. Lengths are mm, area is mm$^2$, mass is g, and density is kg/m$^3$.
\par\smallskip\noindent\textbf{M1.} $L_0=75.48,\ A_0=21.343\times 1.8867=40.2678.$ $\rho=\frac{10^6\times 1.24}{40.2678\times 121.143}=254.19.$ $\nu_{\rm app}=-\frac{675.163/679.4375-1}{0.0500411}=0.1257.$ 
\par\smallskip\noindent\textbf{M2.} $L_0=72.25,\ A_0=21.463\times 1.9567=41.9966.$ $\rho=\frac{10^6\times 1.243}{41.9966\times 120.33}=245.97.$ $\nu_{\rm app}=-\frac{675.1625/678.57-1}{0.0368664}=0.1362.$ 
\par\smallskip\noindent\textbf{M3.} $L_0=70.84,\ A_0=20.473\times 1.83=37.4662.$ $\rho=\frac{10^6\times 1.18}{37.4662\times 119.65}=263.23.$ $\nu_{\rm app}=-\frac{664.9934/667.8252-1}{0.0358329}=0.1183.$ 
}
\par\medskip
\noindent\textbf{Box --- Axial summary (mean $\pm$ std).} $E$: 284.8 $\pm$ 57 MPa ($n=3$); $\sigma_{\rm peak}$: 6.751 $\pm$ 0.47 MPa ($n=3$); $\nu_{\rm app}$: 0.1268 $\pm$ 0.00898 ($n=3$); $\rho$: 254.5 $\pm$ 8.63 kg/m$^3$ ($n=3$).

\clearpage
\subsection{Box (Transverse)}
\noindent\includegraphics[width=\textwidth]{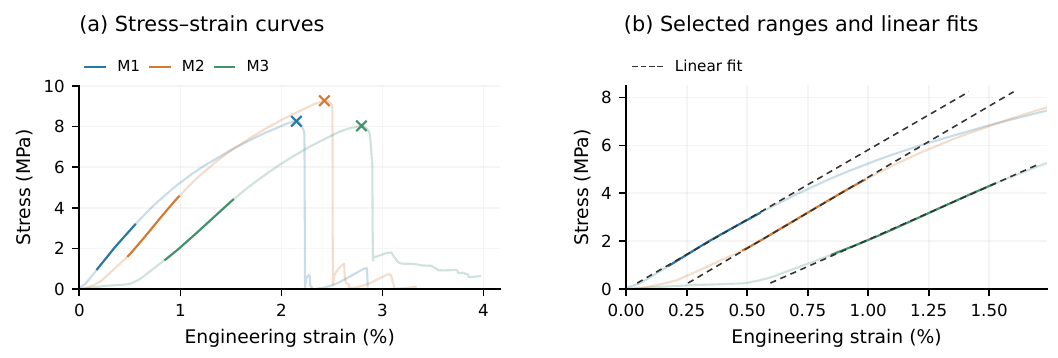}\par
{\footnotesize Left: the loading curves (stress against engineering strain). Right: the strain ranges selected for the fits, drawn darker, and the linear fits, drawn dashed and extended beyond the selected ranges. Each measurement has the same color in both panels. A cross marks the highest stress recorded within the displayed range; it is a peak, not a yield point.}\par\smallskip
\noindent\includegraphics[width=\textwidth]{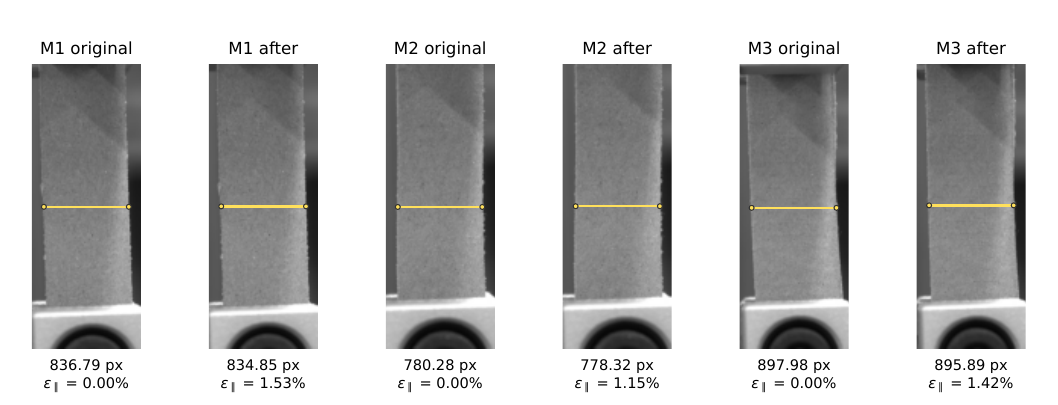}\par
{\footnotesize Photo pairs of the first loading, for the measurements that have them. ``Original'' is the specimen at 0\% strain, and ``after'' is the same specimen under load. The yellow line and its endpoints show the width measured by hand across the specimen. The axial strain reached in each photo is printed below it.}\par\medskip
{\small\setlength{\tabcolsep}{5pt}\centering\begin{tabular}{@{}ccccccc@{}}\toprule
Meas./load & Fit strain (\%) & $E$ (MPa) & $c$ (MPa) & $R^2$ & $\sigma_{\rm peak}$ (MPa) & $\sigma_Y$\\\midrule
M1/L1 & 0.176--0.553 & 582.27 & -0.07385 & 0.9993 & 8.189 & Not identified\\
M2/L1 & 0.481--0.986 & 593.27 & -1.262 & 0.9999 & 9.261 & Not identified\\
M3/L1 & 0.849--1.523 & 447.13 & -2.409 & 0.9996 & 8.028 & Not identified\\
\bottomrule\end{tabular}\par}\smallskip
{\footnotesize\noindent\textbf{Per-measurement calculations.} In $\sigma=E\varepsilon+c$, $\varepsilon$ is a fraction, not a percentage; the interval, slope, and intercept are given above. Lengths are mm, area is mm$^2$, mass is g, and density is kg/m$^3$.
\par\smallskip\noindent\textbf{M1.} $L_0=86.28,\ A_0=26.397\times 1.9067=50.3296.$ $\rho=\frac{10^6\times 1.693}{50.3296\times 133.82}=251.37.$ $\nu_{\rm app}=-\frac{834.8515/836.7884-1}{0.0153048}=0.1512.$ 
\par\smallskip\noindent\textbf{M2.} $L_0=84.43,\ A_0=24.823\times 1.7833=44.2683.$ $\rho=\frac{10^6\times 1.595}{44.2683\times 133.72}=269.45.$ $\nu_{\rm app}=-\frac{778.3208/780.2848-1}{0.0115184}=0.2185.$ 
\par\smallskip\noindent\textbf{M3.} $L_0=84.4,\ A_0=28.777\times 1.7833=51.3184.$ $\rho=\frac{10^6\times 1.852}{51.3184\times 133.06}=271.22.$ $\nu_{\rm app}=-\frac{895.8884/897.9842-1}{0.0141742}=0.1647.$ 
}
\par\medskip
\noindent\textbf{Box --- Transverse summary (mean $\pm$ std).} $E$: 540.9 $\pm$ 81.4 MPa ($n=3$); $\sigma_{\rm peak}$: 8.493 $\pm$ 0.671 MPa ($n=3$); $\nu_{\rm app}$: 0.1781 $\pm$ 0.0356 ($n=3$); $\rho$: 264 $\pm$ 11 kg/m$^3$ ($n=3$).

\clearpage
\subsection{Trash bin (Axial)}
The masses of these specimens were not recorded, so no density is reported.
\par\smallskip
\noindent\includegraphics[width=\textwidth]{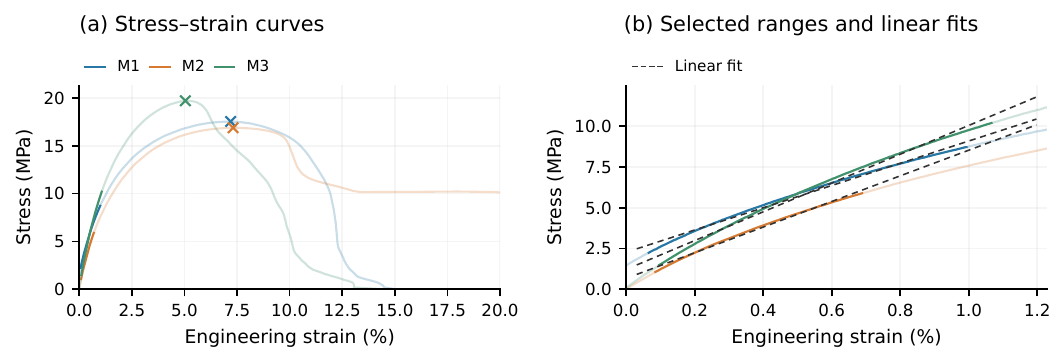}\par
{\footnotesize Left: the loading curves (stress against engineering strain). Right: the strain ranges selected for the fits, drawn darker, and the linear fits, drawn dashed and extended beyond the selected ranges. Each measurement has the same color in both panels. A cross marks the highest stress recorded within the displayed range; it is a peak, not a yield point. Curves are shown up to 20\% strain.}\par\smallskip
\noindent\includegraphics[width=\textwidth]{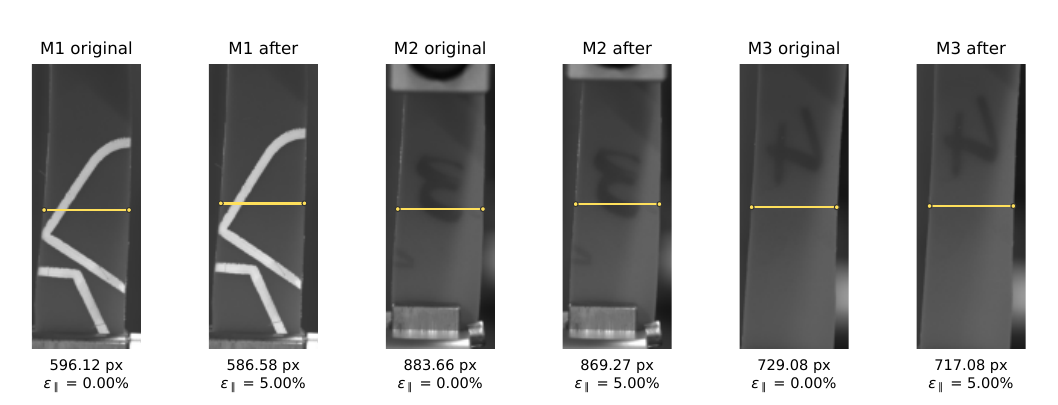}\par
{\footnotesize Photo pairs of the first loading, for the measurements that have them. ``Original'' is the specimen at 0\% strain, and ``after'' is the same specimen under load. The yellow line and its endpoints show the width measured by hand across the specimen. The axial strain reached in each photo is printed below it.}\par\medskip
{\small\setlength{\tabcolsep}{5pt}\centering\begin{tabular}{@{}ccccccc@{}}\toprule
Meas./load & Fit strain (\%) & $E$ (MPa) & $c$ (MPa) & $R^2$ & $\sigma_{\rm peak}$ (MPa) & $\sigma_Y$\\\midrule
M1/L1 & 0.065--0.994 & 682.92 & 0.8031 & 0.9914 & 16.08 & Not identified\\
M2/L1 & 0.084--0.689 & 787.03 & 0.6522 & 0.9945 & 16.88 & Not identified\\
M3/L1 & 0.095--1.068 & 883.98 & 1.126 & 0.9919 & 19.63 & Not identified\\
\bottomrule\end{tabular}\par}\smallskip
{\footnotesize\noindent\textbf{Per-measurement calculations.} In $\sigma=E\varepsilon+c$, $\varepsilon$ is a fraction, not a percentage; the interval, slope, and intercept are given above. Lengths are mm, area is mm$^2$, mass is g, and density is kg/m$^3$.
\par\smallskip\noindent\textbf{M1.} $L_0=81.4,\ A_0=20.93\times 1.58=33.0694.$ $m=\mathrm{NA},\ L_{\rm total}=\mathrm{NA},\ \rho=\mathrm{NA}.$ $\nu_{\rm app}=-\frac{586.581/596.1181-1}{0.0500037}=0.3199.$ 
\par\smallskip\noindent\textbf{M2.} $L_0=85.98,\ A_0=31.83\times 1.59=50.6097.$ $m=\mathrm{NA},\ L_{\rm total}=\mathrm{NA},\ \rho=\mathrm{NA}.$ $\nu_{\rm app}=-\frac{869.2686/883.664-1}{0.0499674}=0.326.$ 
\par\smallskip\noindent\textbf{M3.} $L_0=105.8,\ A_0=25.75\times 1.59=40.9425.$ $m=\mathrm{NA},\ L_{\rm total}=\mathrm{NA},\ \rho=\mathrm{NA}.$ $\nu_{\rm app}=-\frac{717.08/729.08-1}{0.0499641}=0.3294.$ 
}
\par\medskip
\noindent\textbf{Trash bin --- Axial summary (mean $\pm$ std).} $E$: 784.6 $\pm$ 101 MPa ($n=3$); $\sigma_{\rm peak}$: 17.53 $\pm$ 1.86 MPa ($n=3$); $\nu_{\rm app}$: 0.3251 $\pm$ 0.0048 ($n=3$); $\rho$: NA kg/m$^3$ ($n=0$).

\clearpage
\subsection{Trash bin (Transverse)}
Measurement 1 has no recorded full length, so its density is unavailable.
\par\smallskip
\noindent\includegraphics[width=\textwidth]{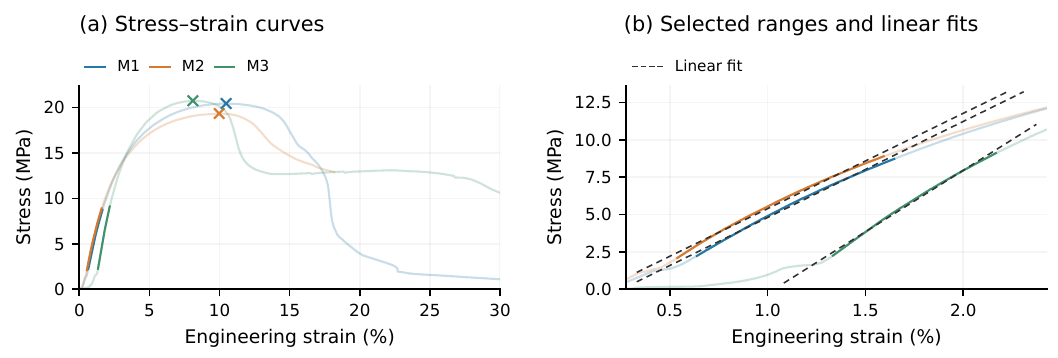}\par
{\footnotesize Left: the loading curves (stress against engineering strain). Right: the strain ranges selected for the fits, drawn darker, and the linear fits, drawn dashed and extended beyond the selected ranges. Each measurement has the same color in both panels. A cross marks the highest stress recorded within the displayed range; it is a peak, not a yield point. Curves are shown up to 30\% strain.}\par\smallskip
\noindent\includegraphics[width=\textwidth]{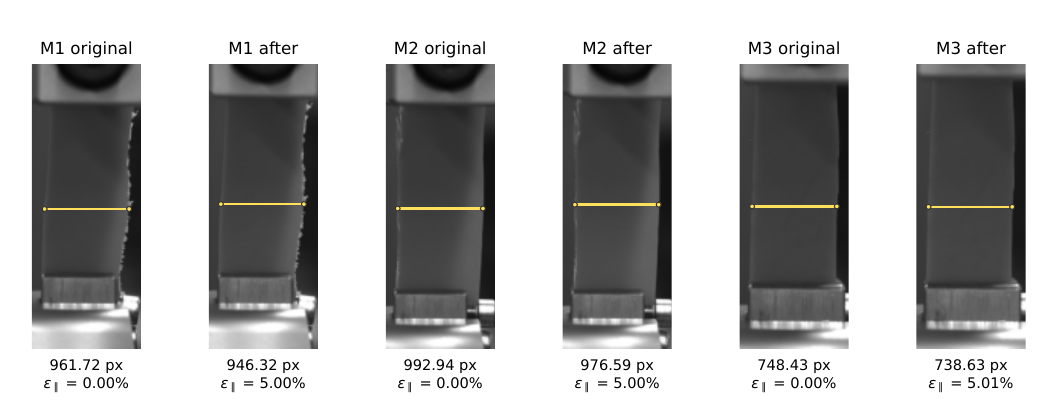}\par
{\footnotesize Photo pairs of the first loading, for the measurements that have them. ``Original'' is the specimen at 0\% strain, and ``after'' is the same specimen under load. The yellow line and its endpoints show the width measured by hand across the specimen. The axial strain reached in each photo is printed below it.}\par\medskip
{\small\setlength{\tabcolsep}{5pt}\centering\begin{tabular}{@{}ccccccc@{}}\toprule
Meas./load & Fit strain (\%) & $E$ (MPa) & $c$ (MPa) & $R^2$ & $\sigma_{\rm peak}$ (MPa) & $\sigma_Y$\\\midrule
M1/L1 & 0.637--1.649 & 643.09 & -1.639 & 0.9964 & 20.45 & Not identified\\
M2/L1 & 0.538--1.594 & 638.17 & -1.004 & 0.9947 & 19.33 & Not identified\\
M3/L1 & 1.332--2.167 & 822.79 & -8.525 & 0.9970 & 20.73 & Not identified\\
\bottomrule\end{tabular}\par}\smallskip
{\footnotesize\noindent\textbf{Per-measurement calculations.} In $\sigma=E\varepsilon+c$, $\varepsilon$ is a fraction, not a percentage; the interval, slope, and intercept are given above. Lengths are mm, area is mm$^2$, mass is g, and density is kg/m$^3$.
\par\smallskip\noindent\textbf{M1.} $L_0=65.12,\ A_0=27.64\times 1.6=44.224.$ $m=5.242,\ L_{\rm total}=\mathrm{NA},\ \rho=\mathrm{NA}.$ $\nu_{\rm app}=-\frac{946.3172/961.72-1}{0.0499693}=0.3205.$ 
\par\smallskip\noindent\textbf{M2.} $L_0=66.479,\ A_0=27.87\times 1.85=51.5595.$ $\rho=\frac{10^6\times 5.686}{51.5595\times 116.96}=942.89.$ $\nu_{\rm app}=-\frac{976.5856/992.944-1}{0.0500158}=0.3294.$ 
\par\smallskip\noindent\textbf{M3.} $L_0=57.47,\ A_0=22.39\times 1.62=36.2718.$ $\rho=\frac{10^6\times 4.585}{36.2718\times 133.47}=947.08.$ $\nu_{\rm app}=-\frac{738.6292/748.4293-1}{0.0500539}=0.2616.$ 
}
\par\medskip
\noindent\textbf{Trash bin --- Transverse summary (mean $\pm$ std).} $E$: 701.3 $\pm$ 105 MPa ($n=3$); $\sigma_{\rm peak}$: 20.17 $\pm$ 0.736 MPa ($n=3$); $\nu_{\rm app}$: 0.3038 $\pm$ 0.0368 ($n=3$); $\rho$: 945 $\pm$ 2.96 kg/m$^3$ ($n=2$).

\clearpage
\subsection{Hanger (Isotropic)}
For measurement 1, $A_0$ is the recorded area of the I-shaped cross-section. For measurements 2--3, $A_0$ is the area of the outer envelope of the cross-section (width $\times$ thickness). The mass of measurement 1 was not recorded.
\par\smallskip
\noindent\includegraphics[width=\textwidth]{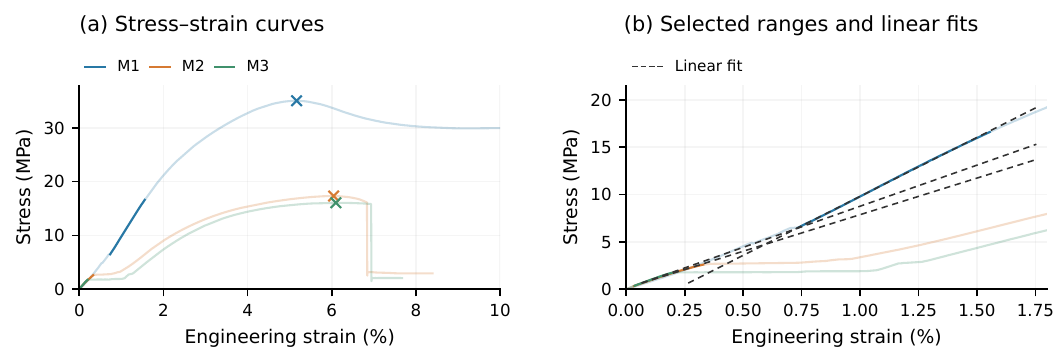}\par
{\footnotesize Left: the loading curves (stress against engineering strain). Right: the strain ranges selected for the fits, drawn darker, and the linear fits, drawn dashed and extended beyond the selected ranges. Each measurement has the same color in both panels. A cross marks the highest stress recorded within the displayed range; it is a peak, not a yield point. Curves are shown up to 10\% strain.}\par\smallskip
\noindent\includegraphics[width=\textwidth]{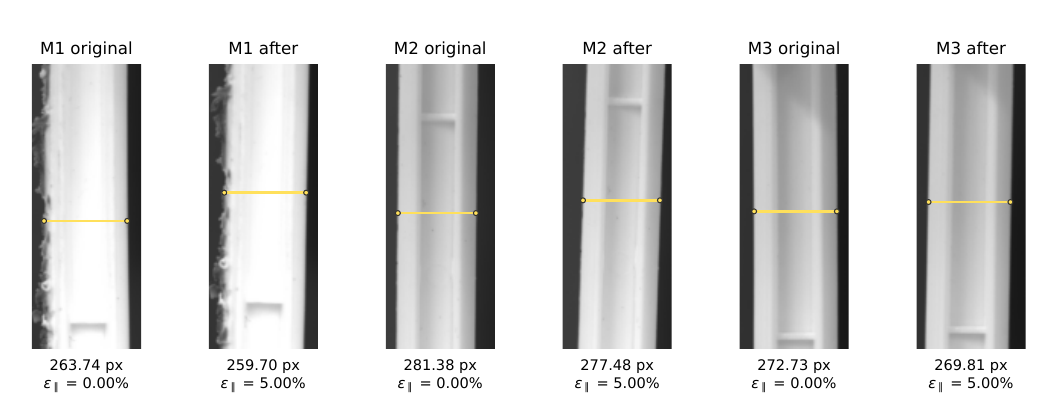}\par
{\footnotesize Photo pairs of the first loading, for the measurements that have them. ``Original'' is the specimen at 0\% strain, and ``after'' is the same specimen under load. The yellow line and its endpoints show the width measured by hand across the specimen. The axial strain reached in each photo is printed below it.}\par\medskip
{\small\setlength{\tabcolsep}{5pt}\centering\begin{tabular}{@{}ccccccc@{}}\toprule
Meas./load & Fit strain (\%) & $E$ (MPa) & $c$ (MPa) & $R^2$ & $\sigma_{\rm peak}$ (MPa) & $\sigma_Y$\\\midrule
M1/L1 & 0.733--1.555 & 1245.9 & -2.95 & 0.9998 & 34.73 & Not identified\\
M2/L1 & 0.032--0.330 & 773.08 & 0.1378 & 0.9975 & 17.31 & Not identified\\
M3/L1 & 0.034--0.195 & 868.52 & 0.07036 & 0.9992 & 16.06 & Not identified\\
\bottomrule\end{tabular}\par}\smallskip
{\footnotesize\noindent\textbf{Per-measurement calculations.} In $\sigma=E\varepsilon+c$, $\varepsilon$ is a fraction, not a percentage; the interval, slope, and intercept are given above. Lengths are mm, area is mm$^2$, mass is g, and density is kg/m$^3$.
\par\smallskip\noindent\textbf{M1.} $L_0=131.18,\ A_0=32.8072$ (recorded area of the I-shaped cross-section). $m=\mathrm{NA},\ L_{\rm total}=\mathrm{NA},\ \rho=\mathrm{NA}.$ $\nu_{\rm app}=-\frac{259.6973/263.7395-1}{0.0500114}=0.3065.$ 
\par\smallskip\noindent\textbf{M2.} $L_0=70.71,\ A_0=8.31\times 8.21=68.2251.$ $\rho=\frac{10^6\times 3.254}{68.2251\times 107.01}=445.71.$ $\nu_{\rm app}=-\frac{277.476/281.3822-1}{0.0499731}=0.2778.$ 
\par\smallskip\noindent\textbf{M3.} $L_0=53.6,\ A_0=6.91\times 8.29=57.2839.$ $\rho=\frac{10^6\times 2.551}{57.2839\times 85.98}=517.94.$ $\nu_{\rm app}=-\frac{269.8127/272.726-1}{0.0500373}=0.2135.$ 
}
\par\medskip
\noindent\textbf{Hanger --- Isotropic summary (mean $\pm$ std).} $E$: 962.5 $\pm$ 250 MPa ($n=3$); $\sigma_{\rm peak}$: 22.7 $\pm$ 10.4 MPa ($n=3$); $\nu_{\rm app}$: 0.2659 $\pm$ 0.0476 ($n=3$); $\rho$: 481.8 $\pm$ 51.1 kg/m$^3$ ($n=2$).

\clearpage
\subsection{Crocs (Isotropic)}
Only the fourth recorded loading is used for stiffness; the photographs show the first loading.
\par\smallskip
\noindent\includegraphics[width=\textwidth]{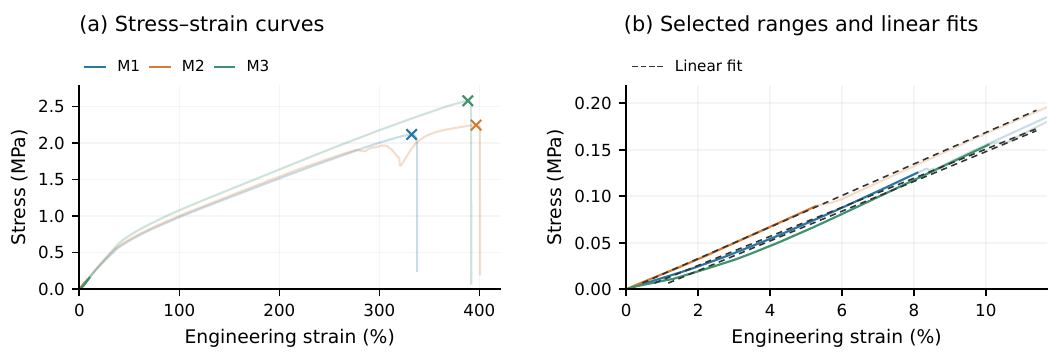}\par
{\footnotesize Left: the loading curves (stress against engineering strain). Right: the strain ranges selected for the fits, drawn darker, and the linear fits, drawn dashed and extended beyond the selected ranges. Each measurement has the same color in both panels. A cross marks the highest stress recorded within the displayed range; it is a peak, not a yield point.}\par\smallskip
\noindent\includegraphics[width=\textwidth]{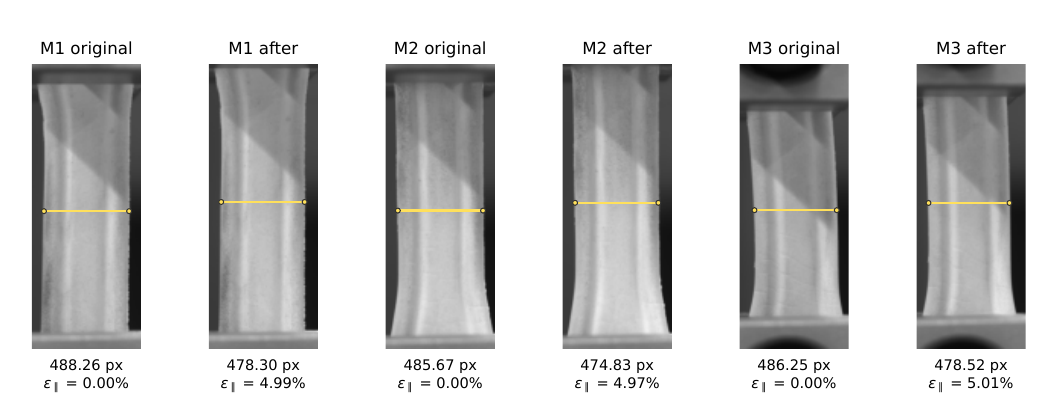}\par
{\footnotesize Photo pairs of the first loading, for the measurements that have them. ``Original'' is the specimen at 0\% strain, and ``after'' is the same specimen under load. The yellow line and its endpoints show the width measured by hand across the specimen. The axial strain reached in each photo is printed below it.}\par\medskip
{\small\setlength{\tabcolsep}{5pt}\centering\begin{tabular}{@{}ccccccc@{}}\toprule
Meas./load & Fit strain (\%) & $E$ (MPa) & $c$ (MPa) & $R^2$ & $\sigma_{\rm peak}$ (MPa) & $\sigma_Y$\\\midrule
M1/L4 & 0.014--8.110 & 1.565 & -0.06208 & 0.9956 & 2.061 & Not identified\\
M2/L4 & 0.014--5.278 & 1.6968 & -0.07815 & 0.9996 & 2.168 & Not identified\\
M3/L4 & 0.015--10.087 & 1.6015 & -0.05561 & 0.9906 & 2.535 & Not identified\\
\bottomrule\end{tabular}\par}\smallskip
{\footnotesize\noindent\textbf{Per-measurement calculations.} In $\sigma=E\varepsilon+c$, $\varepsilon$ is a fraction, not a percentage; the interval, slope, and intercept are given above. Lengths are mm, area is mm$^2$, mass is g, and density is kg/m$^3$.
\par\smallskip\noindent\textbf{M1.} $L_0=62.57,\ A_0=19.64\times 7.03=138.069.$ $\rho=\frac{10^6\times 3.946}{138.069\times 102.49}=278.86.$ $\nu_{\rm app}=-\frac{478.3022/488.2619-1}{0.0498849}=0.4089.$ 
\par\smallskip\noindent\textbf{M2.} $L_0=59.74,\ A_0=19.437\times 7.1633=139.231.$ $\rho=\frac{10^6\times 4.031}{139.231\times 104.26}=277.69.$ $\nu_{\rm app}=-\frac{474.8261/485.6663-1}{0.0497422}=0.4487.$ 
\par\smallskip\noindent\textbf{M3.} $L_0=53.47,\ A_0=18.987\times 6.6833=126.894.$ $\rho=\frac{10^6\times 4.0325}{126.894\times 103.49}=307.07.$ $\nu_{\rm app}=-\frac{478.5216/486.253-1}{0.0500636}=0.3176.$ 
}
\par\medskip
\noindent\textbf{Crocs --- Isotropic summary (mean $\pm$ std).} $E$: 1.621 $\pm$ 0.0681 MPa ($n=3$); $\sigma_{\rm peak}$: 2.254 $\pm$ 0.248 MPa ($n=3$); $\nu_{\rm app}$: 0.3917 $\pm$ 0.0672 ($n=3$); $\rho$: 287.9 $\pm$ 16.6 kg/m$^3$ ($n=3$).

\clearpage
\subsection{Kitty (Isotropic)}
We cut the specimens from the Kitty of the 40 additional recordings (App.~\ref{sec:app_stability_metrics}), only from the fabric of the head, not from the dress. The specimens were stitched closed again with filling inside.
\par\smallskip
\noindent\includegraphics[width=0.62\textwidth]{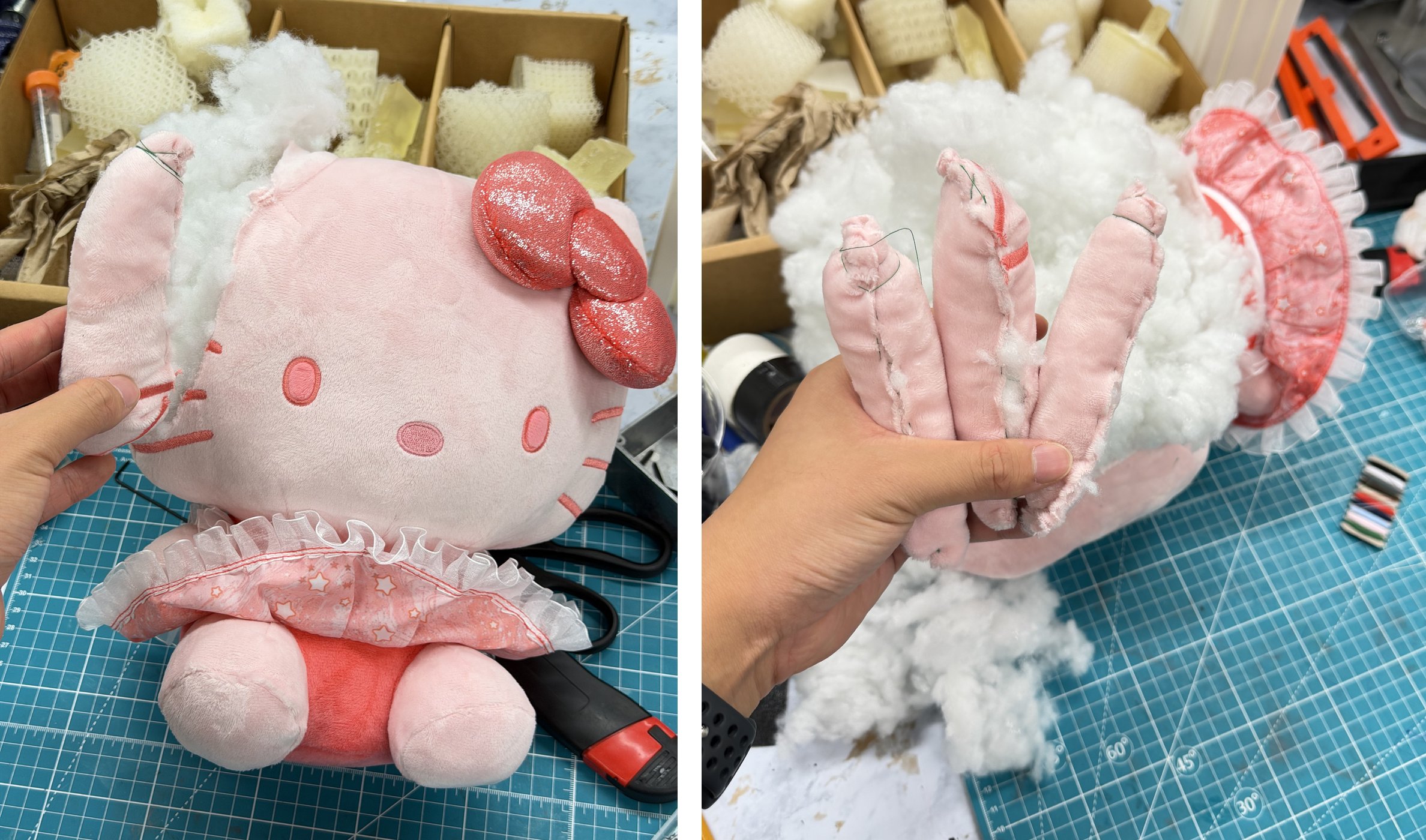}\par
{\footnotesize Left: Kitty, with the fabric of its head cut open. Right: the three specimens, sewn from the head fabric and filled.\par}
\smallskip The measured properties are computed from the outer dimensions (the envelope) of these filled specimens.
\par\smallskip
\noindent\includegraphics[width=\textwidth]{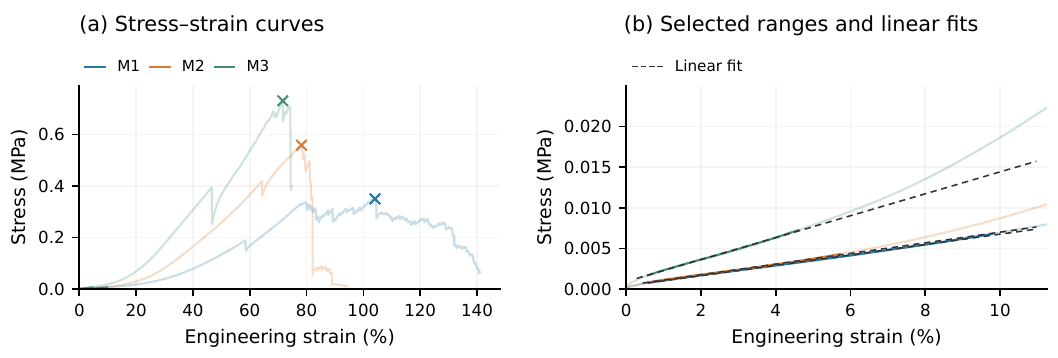}\par
{\footnotesize Left: the loading curves (stress against engineering strain). Right: the strain ranges selected for the fits, drawn darker, and the linear fits, drawn dashed and extended beyond the selected ranges. Each measurement has the same color in both panels. A cross marks the highest stress recorded within the displayed range; it is a peak, not a yield point.}\par\smallskip
\noindent\includegraphics[width=\textwidth]{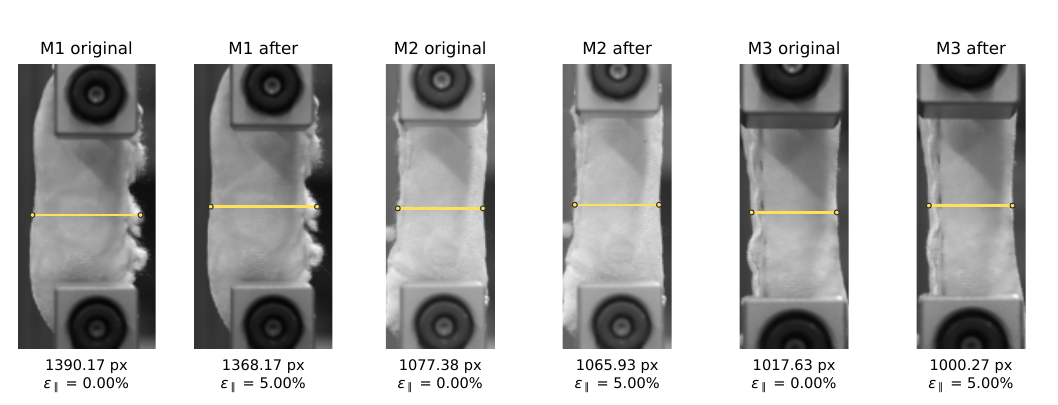}\par
{\footnotesize Photo pairs of the first loading, for the measurements that have them. ``Original'' is the specimen at 0\% strain, and ``after'' is the same specimen under load. The yellow line and its endpoints show the width measured by hand across the specimen. The axial strain reached in each photo is printed below it.}\par\medskip
{\small\setlength{\tabcolsep}{5pt}\centering\begin{tabular}{@{}ccccccc@{}}\toprule
Meas./load & Fit strain (\%) & $E$ (MPa) & $c$ (MPa) & $R^2$ & $\sigma_{\rm peak}$ (MPa) & $\sigma_Y$\\\midrule
M1/L1 & 0.739--9.780 & 0.063733 & 0.0003823 & 0.9990 & 0.3503 & Not identified\\
M2/L1 & 0.574--5.712 & 0.065383 & 0.0004636 & 0.9989 & 0.5584 & Not identified\\
M3/L1 & 0.576--4.605 & 0.1346 & 0.0009467 & 0.9994 & 0.7308 & Not identified\\
\bottomrule\end{tabular}\par}\smallskip
{\footnotesize\noindent\textbf{Per-measurement calculations.} In $\sigma=E\varepsilon+c$, $\varepsilon$ is a fraction, not a percentage; the interval, slope, and intercept are given above. Lengths are mm, area is mm$^2$, mass is g, and density is kg/m$^3$.
\par\smallskip\noindent\textbf{M1.} $L_0=71,\ A_0=36.017\times 28.517=1027.08.$ $\rho=\frac{10^6\times 9.457}{1027.08\times 128.65}=71.572.$ $\nu_{\rm app}=-\frac{1368.168/1390.168-1}{0.0500465}=0.3162.$ 
\par\smallskip\noindent\textbf{M2.} $L_0=75.16,\ A_0=33.257\times 27.49=914.226.$ $\rho=\frac{10^6\times 6.706}{914.226\times 131.48}=55.789.$ $\nu_{\rm app}=-\frac{1065.928/1077.381-1}{0.0499787}=0.2127.$ 
\par\smallskip\noindent\textbf{M3.} $L_0=71.3,\ A_0=27.123\times 22.13=600.239.$ $\rho=\frac{10^6\times 5.752}{600.239\times 117.63}=81.466.$ $\nu_{\rm app}=-\frac{1000.268/1017.632-1}{0.0500252}=0.3411.$ 
}
\par\medskip
\noindent\textbf{Kitty --- Isotropic summary (mean $\pm$ std).} $E$: 0.0879 $\pm$ 0.0404 MPa ($n=3$); $\sigma_{\rm peak}$: 0.5465 $\pm$ 0.191 MPa ($n=3$); $\nu_{\rm app}$: 0.29 $\pm$ 0.0681 ($n=3$); $\rho$: 69.61 $\pm$ 13 kg/m$^3$ ($n=3$).

\clearpage
\subsection{Silicone pad (Isotropic)}
Measurement 1 has two fitted loadings (L1 and L2); its $E$ is the average of their two slopes. The apparent Poisson's ratios describe the honeycomb structure of the pad, not the silicone itself.
\par\smallskip
\noindent\includegraphics[width=\textwidth]{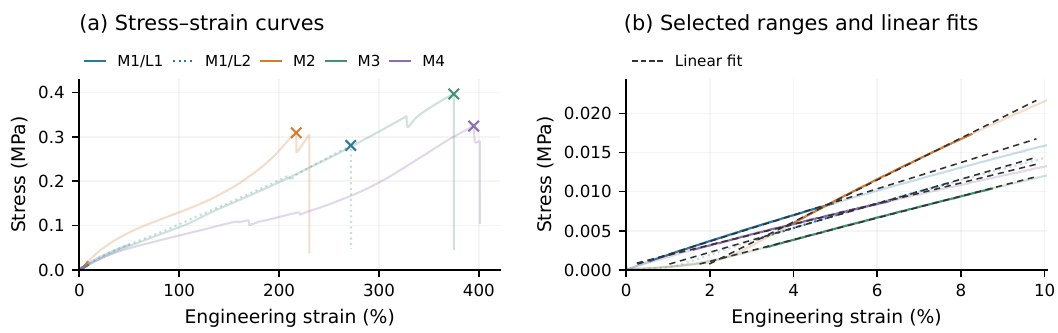}\par
{\footnotesize Left: the loading curves (stress against engineering strain). Right: the strain ranges selected for the fits, drawn darker, and the linear fits, drawn dashed and extended beyond the selected ranges. Each measurement has the same color in both panels. A cross marks the highest stress recorded within the displayed range; it is a peak, not a yield point.}\par\smallskip
\noindent\includegraphics[width=\textwidth]{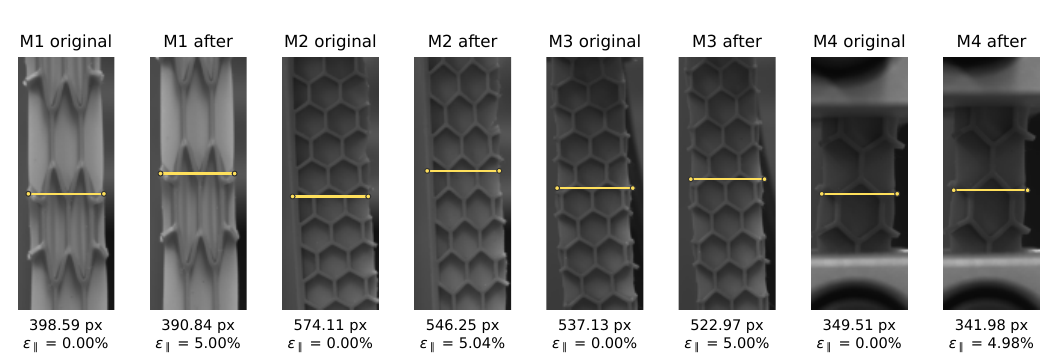}\par
{\footnotesize Photo pairs of the first loading, for the measurements that have them. ``Original'' is the specimen at 0\% strain, and ``after'' is the same specimen under load. The yellow line and its endpoints show the width measured by hand across the specimen. The axial strain reached in each photo is printed below it.}\par\medskip
{\small\setlength{\tabcolsep}{5pt}\centering\begin{tabular}{@{}ccccccc@{}}\toprule
Meas./load & Fit strain (\%) & $E$ (MPa) & $c$ (MPa) & $R^2$ & $\sigma_{\rm peak}$ (MPa) & $\sigma_Y$\\\midrule
M1/L1 & 0.561--4.955 & 0.1682 & 0.0002735 & 0.9994 & 0.2798 & Not identified\\
M1/L2 & 3.103--7.800 & 0.15592 & -0.001436 & 0.9996 & see M1/L1 & Not identified\\
M2/L4 & 3.869--8.267 & 0.26815 & -0.005525 & 0.9994 & 0.3082 & Not identified\\
M3/L4 & 3.340--8.746 & 0.13912 & -0.004259 & 0.9998 & 0.394 & Not identified\\
M4/L4 & 1.591--6.313 & 0.13251 & -0.005429 & 0.9995 & 0.3182 & Not identified\\
\bottomrule\end{tabular}\par}\smallskip
{\footnotesize\noindent\textbf{Per-measurement calculations.} In $\sigma=E\varepsilon+c$, $\varepsilon$ is a fraction, not a percentage; the interval, slope, and intercept are given above. Lengths are mm, area is mm$^2$, mass is g, and density is kg/m$^3$.
\par\smallskip\noindent\textbf{M1.} $L_0=126.85,\ A_0=21.933\times 7.74=169.764.$ $\rho=\frac{10^6\times 10.012}{169.764\times 167.75}=351.57.$ $E=(0.1682+0.15592)/2=0.16206\ \mathrm{MPa}.$ $\nu_{\rm app}=-\frac{390.8358/398.5937-1}{0.05003}=0.389.$ 
\par\smallskip\noindent\textbf{M2.} $L_0=118.4,\ A_0=20.273\times 7.57=153.469.$ $\rho=\frac{10^6\times 11.397}{153.469\times 168.84}=439.84.$ $\nu_{\rm app}=-\frac{546.2537/574.112-1}{0.0503742}=0.9633.$ 
\par\smallskip\noindent\textbf{M3.} $L_0=83.45,\ A_0=19.647\times 7.67=150.69.$ $\rho=\frac{10^6\times 6.759}{150.69\times 130.51}=343.68.$ $\nu_{\rm app}=-\frac{522.9701/537.132-1}{0.0500395}=0.5269.$ 
\par\smallskip\noindent\textbf{M4.} $L_0=30.6,\ A_0=13.45\times 7.39=99.3955.$ $\rho=\frac{10^6\times 2.571}{99.3955\times 69.94}=369.84.$ $\nu_{\rm app}=-\frac{341.9839/349.5132-1}{0.0498203}=0.4324.$ 
}
\par\medskip
\noindent\textbf{Silicone pad --- Isotropic summary (mean $\pm$ std).} $E$: 0.1755 $\pm$ 0.0631 MPa ($n=4$); $\sigma_{\rm peak}$: 0.3251 $\pm$ 0.0488 MPa ($n=4$); $\nu_{\rm app}$: 0.5779 $\pm$ 0.263 ($n=4$); $\rho$: 376.2 $\pm$ 43.8 kg/m$^3$ ($n=4$).

\clearpage
\subsection{Napkin (Axial)}
Frayed boundaries make the lateral width ratios noisy; they are not included in a Poisson-ratio mean.
\par\smallskip
\noindent\includegraphics[width=\textwidth]{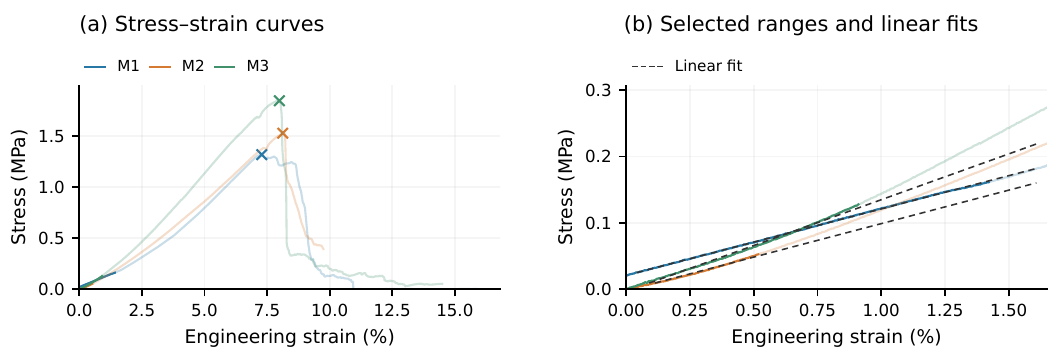}\par
{\footnotesize Left: the loading curves (stress against engineering strain). Right: the strain ranges selected for the fits, drawn darker, and the linear fits, drawn dashed and extended beyond the selected ranges. Each measurement has the same color in both panels. A cross marks the highest stress recorded within the displayed range; it is a peak, not a yield point.}\par\smallskip
\noindent\includegraphics[width=\textwidth]{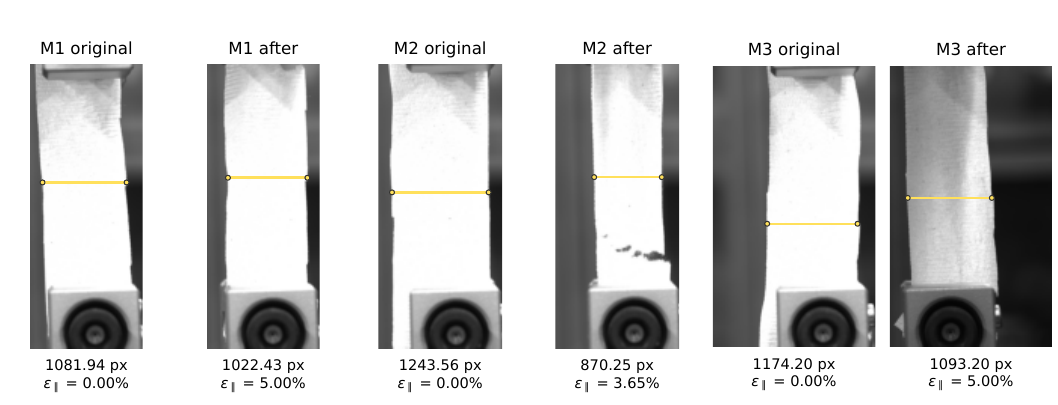}\par
{\footnotesize Photo pairs of the first loading, for the measurements that have them. ``Original'' is the specimen at 0\% strain, and ``after'' is the same specimen under load. The yellow line and its endpoints show the width measured by hand across the specimen. The axial strain reached in each photo is printed below it.}\par\medskip
{\small\setlength{\tabcolsep}{5pt}\centering\begin{tabular}{@{}ccccccc@{}}\toprule
Meas./load & Fit strain (\%) & $E$ (MPa) & $c$ (MPa) & $R^2$ & $\sigma_{\rm peak}$ (MPa) & $\sigma_Y$\\\midrule
M1/L1 & 0.000--1.422 & 10.01 & 0.0001515 & 0.9998 & 1.295 & Not identified\\
M2/L1 & 0.000--0.520 & 10.088 & -0.002419 & 0.9940 & 1.526 & Not identified\\
M3/L1 & 0.000--0.912 & 13.805 & -0.003446 & 0.9963 & 1.844 & Not identified\\
\bottomrule\end{tabular}\par}\smallskip
{\footnotesize\noindent\textbf{Per-measurement calculations.} In $\sigma=E\varepsilon+c$, $\varepsilon$ is a fraction, not a percentage; the interval, slope, and intercept are given above. Lengths are mm, area is mm$^2$, mass is g, and density is kg/m$^3$.
\par\smallskip\noindent\textbf{M1.} $L_0=90.64,\ A_0=35.627\times 0.2=7.12533.$ $\rho=\frac{10^6\times 0.186}{7.12533\times 144.19}=181.04.$ $\nu_{\rm app}=-\frac{1022.429/1081.937-1}{0.05}=1.1.$ (Noisy-boundary diagnostic only.) 
\par\smallskip\noindent\textbf{M2.} $L_0=93.83,\ A_0=41.087\times 0.18667=7.66951.$ $\rho=\frac{10^6\times 0.216}{7.66951\times 143.96}=195.63.$ $\nu_{\rm app}=-\frac{870.2465/1243.56-1}{0.0365096}=8.222.$ (Noisy-boundary diagnostic only.) 
\par\smallskip\noindent\textbf{M3.} $L_0=92.99,\ A_0=38.377\times 0.16333=6.26819.$ $\rho=\frac{10^6\times 0.201}{6.26819\times 143.92}=222.81.$ $\nu_{\rm app}=-\frac{1093.2/1174.2-1}{0.0500022}=1.38.$ (Noisy-boundary diagnostic only.) 
}
\par\medskip
\noindent\textbf{Napkin --- Axial summary (mean $\pm$ std).} $E$: 11.3 $\pm$ 2.17 MPa ($n=3$); $\sigma_{\rm peak}$: 1.555 $\pm$ 0.275 MPa ($n=3$); $\nu_{\rm app}$: not reported (noisy boundaries); $\rho$: 199.8 $\pm$ 21.2 kg/m$^3$ ($n=3$).

\clearpage
\subsection{Napkin (Transverse)}
Frayed boundaries make the lateral width ratios noisy; they are not included in a Poisson-ratio mean.
\par\smallskip
\noindent\includegraphics[width=\textwidth]{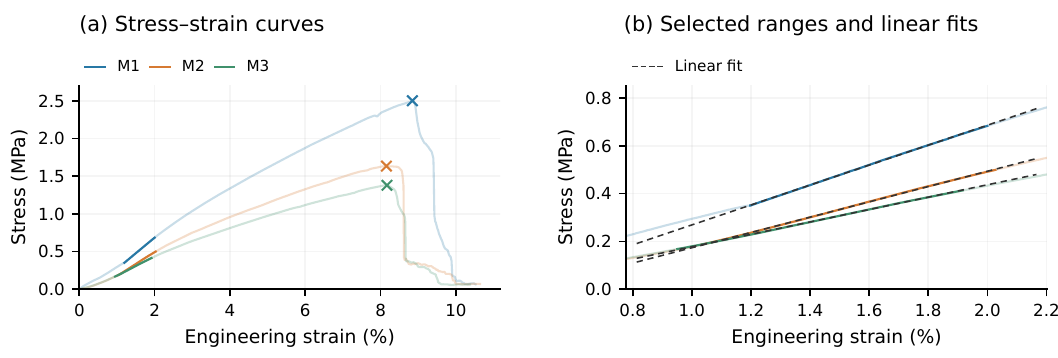}\par
{\footnotesize Left: the loading curves (stress against engineering strain). Right: the strain ranges selected for the fits, drawn darker, and the linear fits, drawn dashed and extended beyond the selected ranges. Each measurement has the same color in both panels. A cross marks the highest stress recorded within the displayed range; it is a peak, not a yield point.}\par\smallskip
\noindent\includegraphics[width=\textwidth]{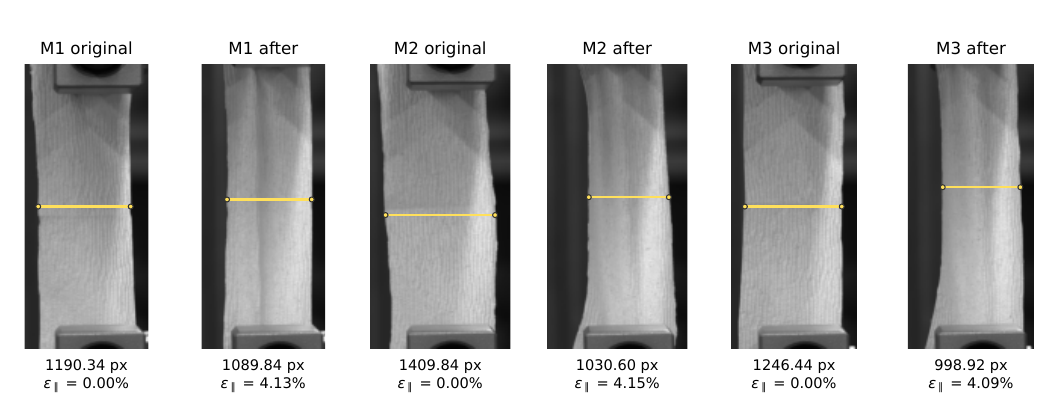}\par
{\footnotesize Photo pairs of the first loading, for the measurements that have them. ``Original'' is the specimen at 0\% strain, and ``after'' is the same specimen under load. The yellow line and its endpoints show the width measured by hand across the specimen. The axial strain reached in each photo is printed below it.}\par\medskip
{\small\setlength{\tabcolsep}{5pt}\centering\begin{tabular}{@{}ccccccc@{}}\toprule
Meas./load & Fit strain (\%) & $E$ (MPa) & $c$ (MPa) & $R^2$ & $\sigma_{\rm peak}$ (MPa) & $\sigma_Y$\\\midrule
M1/L1 & 1.197--2.000 & 41.744 & -0.1488 & 0.9999 & 2.5 & Not identified\\
M2/L1 & 1.028--2.029 & 32.036 & -0.1967 & 0.9999 & 1.584 & Not identified\\
M3/L1 & 0.951--1.917 & 25.947 & -0.08205 & 0.9998 & 1.379 & Not identified\\
\bottomrule\end{tabular}\par}\smallskip
{\footnotesize\noindent\textbf{Per-measurement calculations.} In $\sigma=E\varepsilon+c$, $\varepsilon$ is a fraction, not a percentage; the interval, slope, and intercept are given above. Lengths are mm, area is mm$^2$, mass is g, and density is kg/m$^3$.
\par\smallskip\noindent\textbf{M1.} $L_0=102.72,\ A_0=41.223\times 0.13667=5.63386.$ $\rho=\frac{10^6\times 0.245}{5.63386\times 158.71}=274.$ $\nu_{\rm app}=-\frac{1089.842/1190.337-1}{0.0413386}=2.042.$ (Noisy-boundary diagnostic only.) 
\par\smallskip\noindent\textbf{M2.} $L_0=104.78,\ A_0=45.663\times 0.17333=7.91498.$ $\rho=\frac{10^6\times 0.268}{7.91498\times 158.87}=213.13.$ $\nu_{\rm app}=-\frac{1030.598/1409.84-1}{0.0414812}=6.485.$ (Noisy-boundary diagnostic only.) 
\par\smallskip\noindent\textbf{M3.} $L_0=106.89,\ A_0=42.48\times 0.21333=9.0624.$ $\rho=\frac{10^6\times 0.44}{9.0624\times 156.92}=309.41.$ $\nu_{\rm app}=-\frac{998.9158/1246.435-1}{0.0408504}=4.861.$ (Noisy-boundary diagnostic only.) 
}
\par\medskip
\noindent\textbf{Napkin --- Transverse summary (mean $\pm$ std).} $E$: 33.24 $\pm$ 7.97 MPa ($n=3$); $\sigma_{\rm peak}$: 1.821 $\pm$ 0.597 MPa ($n=3$); $\nu_{\rm app}$: not reported (noisy boundaries); $\rho$: 265.5 $\pm$ 48.7 kg/m$^3$ ($n=3$).

\clearpage
\subsection{Diaper (Axial)}
\par\smallskip
\noindent\includegraphics[width=\textwidth]{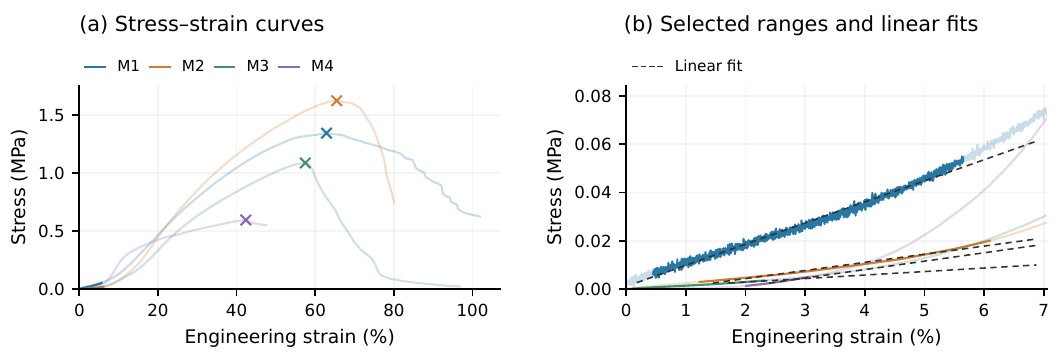}\par
{\footnotesize Left: the loading curves (stress against engineering strain). Right: the strain ranges selected for the fits, drawn darker, and the linear fits, drawn dashed and extended beyond the selected ranges. Each measurement has the same color in both panels. A cross marks the highest stress recorded within the displayed range; it is a peak, not a yield point.}\par\smallskip
\noindent\includegraphics[width=\textwidth]{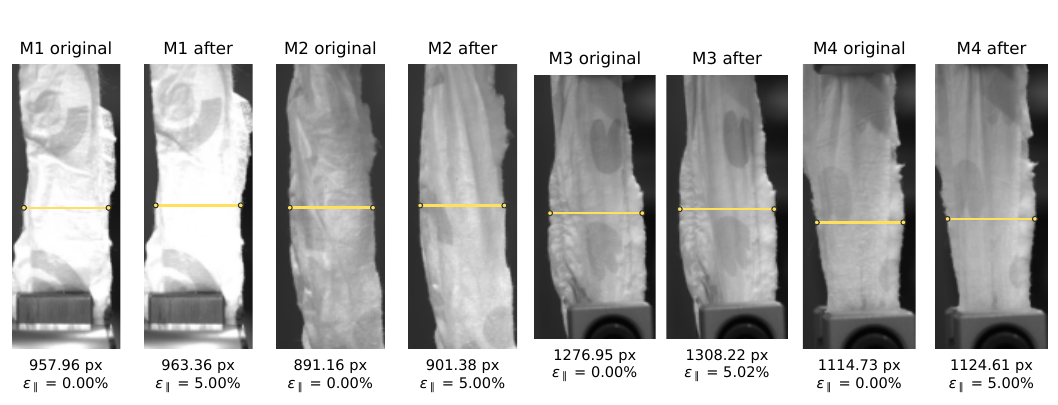}\par
{\footnotesize Photo pairs of the first loading, for the measurements that have them. ``Original'' is the specimen at 0\% strain, and ``after'' is the same specimen under load. The yellow line and its endpoints show the width measured by hand across the specimen. The axial strain reached in each photo is printed below it.}\par\medskip
{\small\setlength{\tabcolsep}{5pt}\centering\begin{tabular}{@{}ccccccc@{}}\toprule
Meas./load & Fit strain (\%) & $E$ (MPa) & $c$ (MPa) & $R^2$ & $\sigma_{\rm peak}$ (MPa) & $\sigma_Y$\\\midrule
M1/L1 & 0.451--5.651 & 0.87349 & -0.001535 & 0.9888 & 1.341 & Not identified\\
M2/L1 & 1.230--6.095 & 0.33655 & -0.002361 & 0.9798 & 1.623 & Not identified\\
M3/L4 & 0.097--2.321 & 0.14499 & -0.002248 & 0.9952 & 1.085 & Not identified\\
M4/L4 & 2.014--3.058 & 0.34806 & -0.007378 & 0.9933 & 0.5944 & Not identified\\
\bottomrule\end{tabular}\par}\smallskip
{\footnotesize\noindent\textbf{Per-measurement calculations.} In $\sigma=E\varepsilon+c$, $\varepsilon$ is a fraction, not a percentage; the interval, slope, and intercept are given above. Lengths are mm, area is mm$^2$, mass is g, and density is kg/m$^3$.
\par\smallskip\noindent\textbf{M1.} $L_0=99.6,\ A_0=25.52\times 8.7333=222.875.$ $\rho=\frac{10^6\times 3.176}{222.875\times 157.02}=90.754.$ $\nu_{\rm app}=-\frac{963.3569/957.9612-1}{0.0499558}=-0.1127.$ 
\par\smallskip\noindent\textbf{M2.} $L_0=108.84,\ A_0=19.14\times 8.2133=157.203.$ $\rho=\frac{10^6\times 3.161}{157.203\times 156}=128.9.$ $\nu_{\rm app}=-\frac{901.3766/891.1593-1}{0.0500147}=-0.2292.$ 
\par\smallskip\noindent\textbf{M3.} $L_0=104.19,\ A_0=32.563\times 7.9567=259.096.$ $\rho=\frac{10^6\times 5.349}{259.096\times 160.31}=128.78.$ $\nu_{\rm app}=-\frac{1308.219/1276.954-1}{0.0502198}=-0.4875.$ 
\par\smallskip\noindent\textbf{M4.} $L_0=105.25,\ A_0=30.83\times 8.4767=261.336.$ $\rho=\frac{10^6\times 6.271}{261.336\times 161.65}=148.44.$ $\nu_{\rm app}=-\frac{1124.607/1114.726-1}{0.049999}=-0.1773.$ 
}
\par\medskip
\noindent\textbf{Diaper --- Axial summary (mean $\pm$ std).} $E$: 0.4258 $\pm$ 0.313 MPa ($n=4$); $\sigma_{\rm peak}$: 1.161 $\pm$ 0.437 MPa ($n=4$); $\nu_{\rm app}$: -0.2517 $\pm$ 0.164 ($n=4$); $\rho$: 124.2 $\pm$ 24.1 kg/m$^3$ ($n=4$).

\clearpage
\subsection{Diaper (Transverse)}
Measurement 1 has a tensile record but no photographs.
\par\smallskip
\noindent\includegraphics[width=\textwidth]{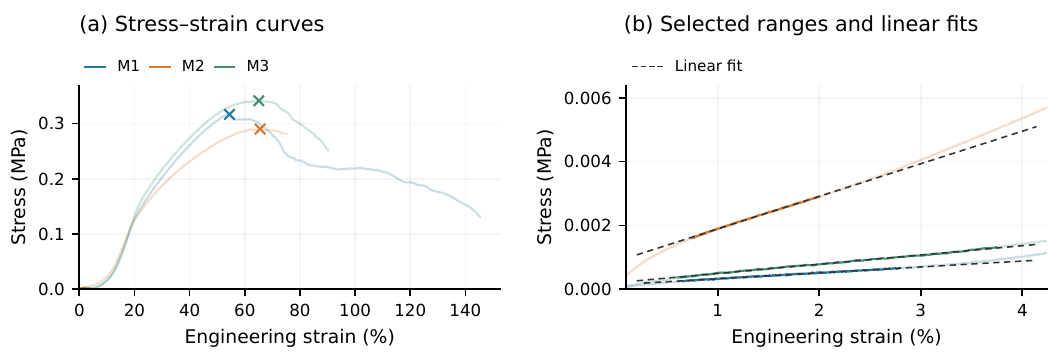}\par
{\footnotesize Left: the loading curves (stress against engineering strain). Right: the strain ranges selected for the fits, drawn darker, and the linear fits, drawn dashed and extended beyond the selected ranges. Each measurement has the same color in both panels. A cross marks the highest stress recorded within the displayed range; it is a peak, not a yield point.}\par\smallskip
\noindent\includegraphics[width=\textwidth]{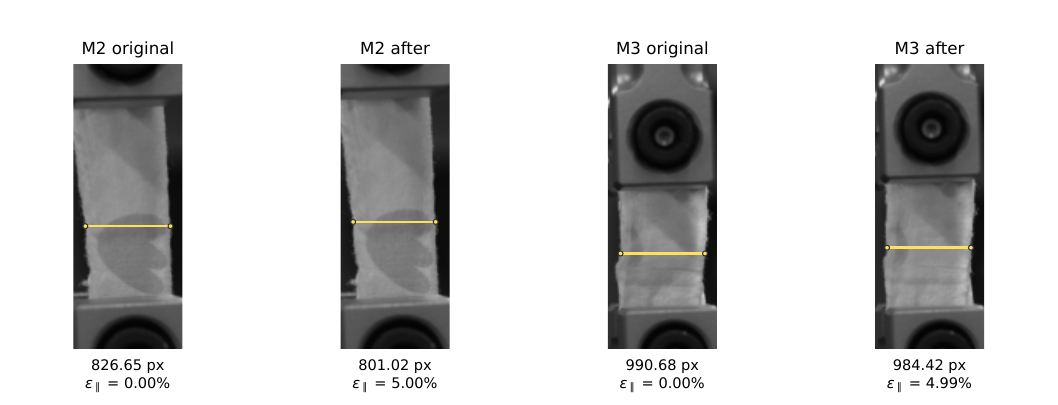}\par
{\footnotesize Photo pairs of the first loading, for the measurements that have them. ``Original'' is the specimen at 0\% strain, and ``after'' is the same specimen under load. The yellow line and its endpoints show the width measured by hand across the specimen. The axial strain reached in each photo is printed below it.}\par\medskip
{\small\setlength{\tabcolsep}{5pt}\centering\begin{tabular}{@{}ccccccc@{}}\toprule
Meas./load & Fit strain (\%) & $E$ (MPa) & $c$ (MPa) & $R^2$ & $\sigma_{\rm peak}$ (MPa) & $\sigma_Y$\\\midrule
M1/L4 & 0.654--2.797 & 0.018512 & -0.00118 & 0.9941 & 0.3153 & Not identified\\
M2/L4 & 0.749--2.002 & 0.10218 & -0.003996 & 0.9990 & 0.285 & Not identified\\
M3/L4 & 0.602--3.740 & 0.028983 & -0.001629 & 0.9979 & 0.3394 & Not identified\\
\bottomrule\end{tabular}\par}\smallskip
{\footnotesize\noindent\textbf{Per-measurement calculations.} In $\sigma=E\varepsilon+c$, $\varepsilon$ is a fraction, not a percentage; the interval, slope, and intercept are given above. Lengths are mm, area is mm$^2$, mass is g, and density is kg/m$^3$.
\par\smallskip\noindent\textbf{M1.} $L_0=57.33,\ A_0=28.133\times 7.5667=212.876.$ $\rho=\frac{10^6\times 3.271}{212.876\times 108.52}=141.59.$ $\nu_{\rm app}=\mathrm{NA}$ (photographs missing). 
\par\smallskip\noindent\textbf{M2.} $L_0=57.28,\ A_0=26.11\times 9.32=243.345.$ $\rho=\frac{10^6\times 3.108}{243.345\times 103.91}=122.91.$ $\nu_{\rm app}=-\frac{801.0156/826.6474-1}{0.05}=0.6201.$ 
\par\smallskip\noindent\textbf{M3.} $L_0=52.42,\ A_0=28.987\times 8.41=243.778.$ $\rho=\frac{10^6\times 3.749}{243.778\times 103.21}=149.$ $\nu_{\rm app}=-\frac{984.4158/990.68-1}{0.0499485}=0.1266.$ 
}
\par\medskip
\noindent\textbf{Diaper --- Transverse summary (mean $\pm$ std).} $E$: 0.04989 $\pm$ 0.0456 MPa ($n=3$); $\sigma_{\rm peak}$: 0.3132 $\pm$ 0.0272 MPa ($n=3$); $\nu_{\rm app}$: 0.3734 $\pm$ 0.349 ($n=2$); $\rho$: 137.8 $\pm$ 13.4 kg/m$^3$ ($n=3$).

\clearpage\flushbottom\twocolumn

\clearpage
\section{Additional Qualitative Analysis}
\label{sec:app_qualitative}
\newcommand{\suppfig}[2][\textwidth]{\IfFileExists{#2}{\includegraphics[width=#1]{#2}}{\fbox{\parbox[c][2cm][c]{.9\linewidth}{\centering missing figure file \detokenize{#2}}}}}

\begin{figure*}[t]
\centering
\suppfig[\textwidth]{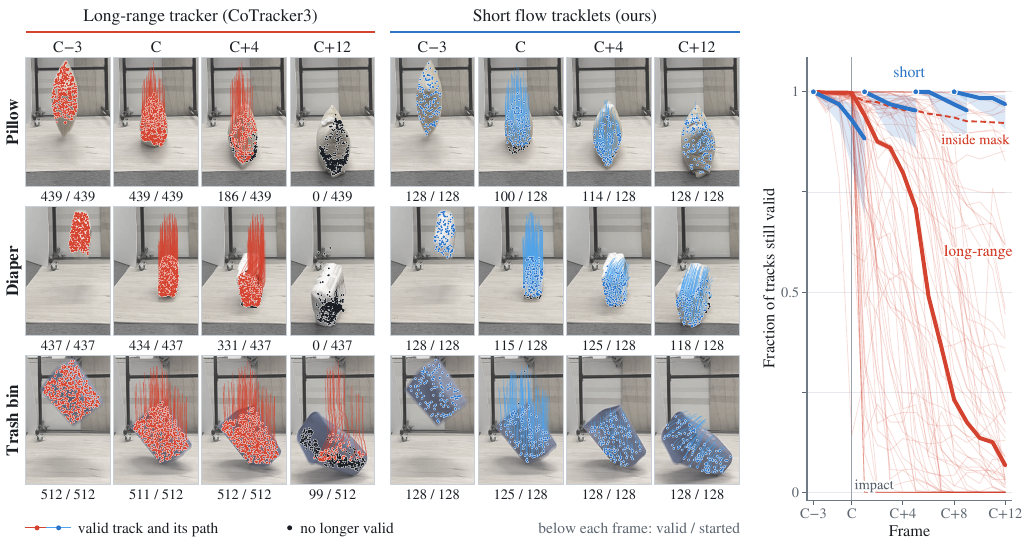}
\caption{\textbf{Long-range tracks through impact.} Left: three real drops at $C-3$, $C$, $C+4$, and $C+12$. Red: CoTracker3 (online), one 16-frame track per query started at $C-3$. Blue: our five-frame flow tracklets, restarted at $C-3$, $C+1$, $C+5$, and $C+8$. Dark dots are tracks that are no longer valid, drawn where the tracker still places them. The numbers below each frame give the valid tracks / the started tracks. Right: the fraction of long-range tracks that are still valid, on 77 drops (thin red lines: single drops; bold red line: median). Blue: the median fraction of our tracklets that are still inside the mask, for each of the four tracklet windows on the same 77 drops (band: interquartile range). A long-range track is valid while the tracker reports it visible and it lies inside the object mask. The dashed line counts long-range tracks by position alone (inside the mask), as the tracklets are counted.}
\label{fig:app_tracking}
\end{figure*}

\paragraph{Long-range tracks through impact.}
Fig.~\ref{fig:app_tracking} compares a long-range point tracker with our short tracklets on the real drops.
We ran CoTracker3 in its online mode.
Each query point is started three frames before the collision and tracked through the 16 frames of the clip.
This gave usable tracks on 77 of the 80 drops; on the other three it gave none.
A track counts as valid while the tracker reports the point as visible and the point lies inside the object mask.
In the median drop, 94\% of the tracks are still valid one frame after the collision, 80\% four frames after it, and 7\% at the end of the window.
In 60 of the 77 drops, fewer than half of the tracks are still valid at the end of the window.
The tracks do not become invalid because they leave the object.
Of the points that are no longer valid, 94\% still lie inside the mask.
They are invalid because the tracker itself reports them as not visible, as the object rotates and blurs.
Our tracklets carry no visibility estimate, so we count them by position alone: a tracklet counts while its point lies inside the object mask.
Counted in the same way, a median of 92\% of the long-range tracks remain inside the mask at the end of the window.
The long-range tracks that remain valid cover little of the object.
We measure coverage as the share of the object mask that lies within 8 pixels of a valid track, on the $224\times224$ crop.
At the end of the window, the median coverage of the long-range tracks is 17\%.
For our tracklets it is 91--95\% at the end of each of their four windows.
Our tracklets restart on the visible surface at $C-3$, $C+1$, $C+5$, and $C+8$, and a median of 88--97\% of them are still on the object when a window ends.
App.~\ref{sec:app_more_trackers} runs ten more long-range trackers on the same drops and compares how consistently they and our tracklets cover the object.

\begin{figure*}[t]
\centering
\suppfig[\textwidth]{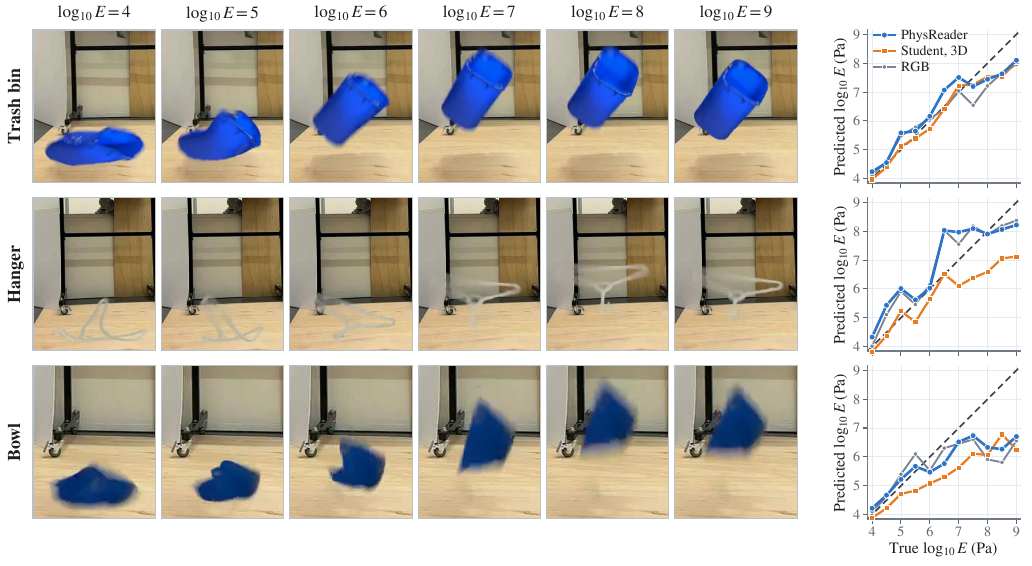}
\caption{\textbf{One drop at different stiffness.} Three real drops (trash bin, drop 5; hanger, drop 1; bowl, drop 4) are re-simulated in their reconstructed scenes with only $E$ changed; yield stress ($10^7$\,Pa), mesh, initial state, camera, and density are fixed. Left: for $\log_{10}E=4$ to $9$, the frame of largest deformation after impact, taken from the clip given to the predictors. All frames of a row use the same crop. Right: predicted against true $\log_{10}E$ at all eleven levels for \model{} (blue), the student on 3D tracklets (orange), and RGB (gray); the dashed line is ideal recovery. The $\log_{10}E$ error of \model{} and of the student on 3D tracklets is 0.45 and 0.32\,dex on the trash bin, 0.62 and 0.75\,dex on the hanger, and 0.86 and 1.15\,dex on the bowl.}
\label{fig:app_sweep_frames}
\end{figure*}

\paragraph{One drop at different stiffness.}
Fig.~\ref{fig:app_sweep_frames} shows three of the 80 drops of the stiffness sweep of App.~\ref{sec:material_appearance}, as the predictors see them.
At the lowest stiffness levels the object visibly changes shape; at the highest it moves almost rigidly, and the frames differ mainly in pose.
On the trash bin, the predictions of all three predictors shown (\model{}, the student on 3D tracklets, and RGB) follow the true stiffness, and the student on 3D tracklets has the lowest error.
On the hanger and the bowl, the predictions of the student on 3D tracklets stay below the true stiffness and below the predictions of \model{} for the stiffer materials.
This is the same shift toward soft materials as on real video (App.~\ref{sec:app_reference_free}).
Over all 80 drops of the sweep, \model{} has the lower error on 46.
Single drops vary; the averages over all 80 drops are in Table~\ref{tab:material_sensitivity}.

\begin{figure*}[t]
\centering
\suppfig[\textwidth]{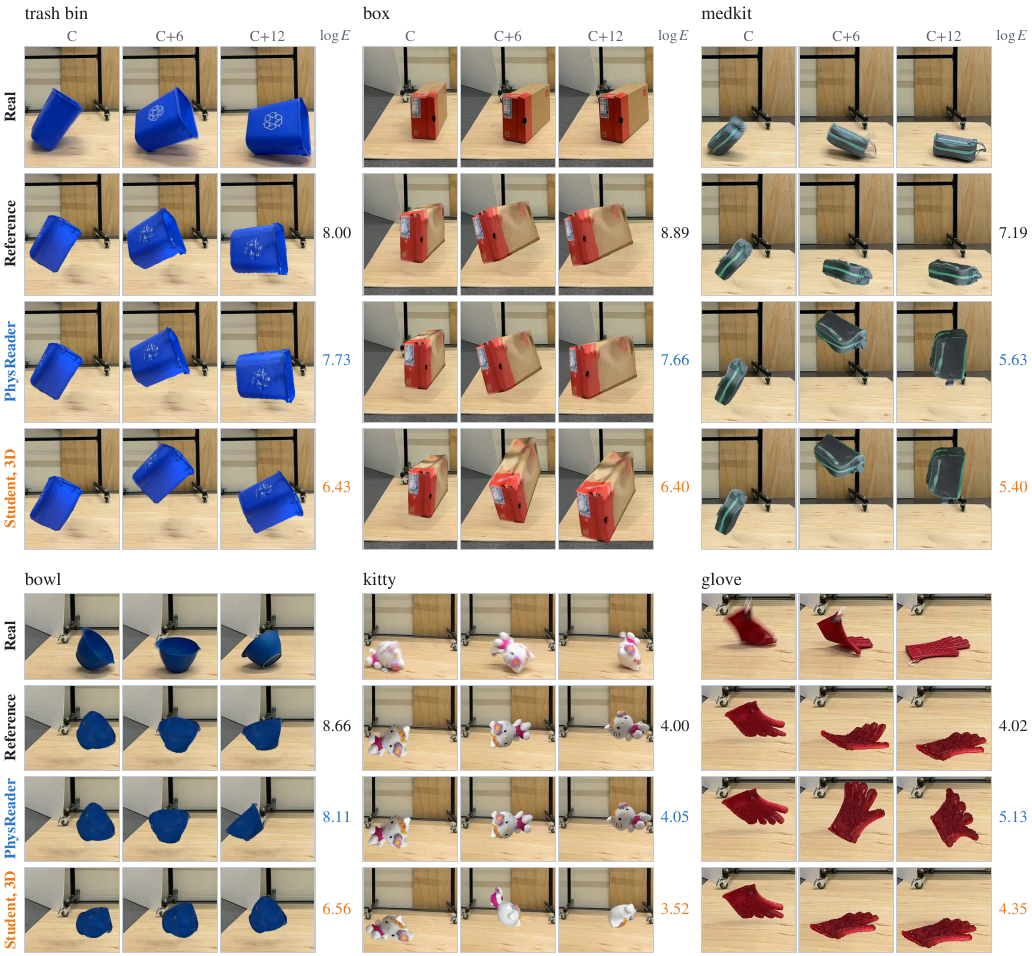}
\caption{\textbf{Replays with predicted materials.} Six real drops (trash bin, drop 3; box, drop 4; medkit, drop 1; bowl, drop 5; Kitty, drop 5; glove, drop 4) at the collision frame $C$ and 6 and 12 frames later. Rows: the real video, and re-simulations with the reference material, with the material predicted by \model{} (blue), and with the material predicted by the student on 3D tracklets (orange). Each simulation starts from its own refined initial state. Numbers are $\log_{10}E$ in Pa. The right column shows two drops on which \model{} fails.}
\label{fig:app_replays}
\end{figure*}

\paragraph{Replays with predicted materials.}
Fig.~\ref{fig:app_replays} replays six real drops with the reference material and with the materials predicted by \model{} and by the student on 3D tracklets.
Each simulation has its own refined initial state (App.~\ref{sec:app_behavior}).
We chose drops in which the difference between the two predicted materials is visible in still frames.
We also include two drops on which \model{} fails.
Over all 80 drops, \model{} has the smaller parameter distance on 61 and the student on 3D tracklets on 19.
For the trash bin, the bowl, and Kitty, the replay with \model{}'s material moves like the reference, while the material of the 3D student is too soft: the bin ends tilted, the bowl flattens, and the toy collapses.
For the box, both predictions are too soft, and that of the 3D student more so.
Only the box replayed with the 3D student's material shears.
For the medkit, both predictions are 1.6--1.8\,dex too soft and both replays rebound, unlike the real bag.
For the glove, \model{} is 1.1\,dex too stiff and its replay springs up, while the 3D student is close to the reference.
The reference replay does not match the real video exactly either; for example, the real box stays upright and the reference box tilts.

\begin{figure*}[t]
\centering
\suppfig[\textwidth]{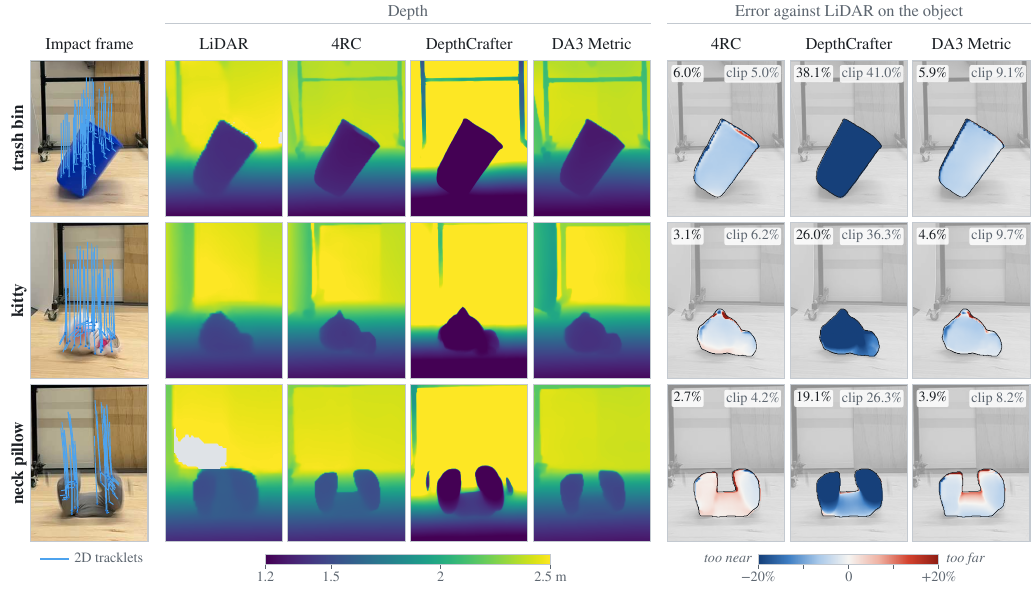}
\caption{\textbf{Estimated depth on real drops.} Three drops (trash bin, drop 3; Kitty, drop 1; neck pillow, drop 2) at the collision frame $C$. Left: the frame, with every third 2D tracklet of the first tracklet window drawn on it. Middle: LiDAR depth and the depth of 4RC, DepthCrafter, and Depth Anything~3 Metric (DA3 Metric), each with one scale per clip fitted to LiDAR at $C-3$ (for DepthCrafter the scale is applied to its inverse depth); the color scale of all depth panels spans 1.2--2.5\,m, and gray marks pixels without LiDAR. Right: signed error (estimate $-$ LiDAR)/LiDAR on the object, clipped at $\pm20\%$. In each error map, the first number is the absolute relative error on the object at this frame. The number after ``clip'' is the same error over the 16 frames of the clip, as in Table~\ref{tab:depth_comparison}.}
\label{fig:app_depth}
\end{figure*}

\paragraph{Estimated depth on real drops.}
Fig.~\ref{fig:app_depth} shows, on three real drops at the collision frame, the estimated depth that Table~\ref{tab:depth_comparison} evaluates.
Each estimate is made metric with one scale per clip, fitted to LiDAR in the first frame of the window.
At the collision frame, the absolute relative error of 4RC against LiDAR on the object is 3--6\% on the three drops, and that of Depth Anything~3 (metric) is 4--6\%.
Over the 16 frames of the clip, the error of 4RC stays at 4--6\%.
The error of Depth Anything~3, which is run on single images, rises to 8--10\%, because it is larger in the frames before the collision.
DepthCrafter outputs relative inverse depth, and with a single scale its error is 19--41\%.
LiDAR is coarse at silhouettes, so the colors at the rim of the error maps partly reflect the sensor.
Depth of this accuracy does not make the lifted tracklets transfer, and neither does the LiDAR depth itself (Table~\ref{tab:same_drops}).

\begin{figure*}[t]
\centering
\suppfig[\textwidth]{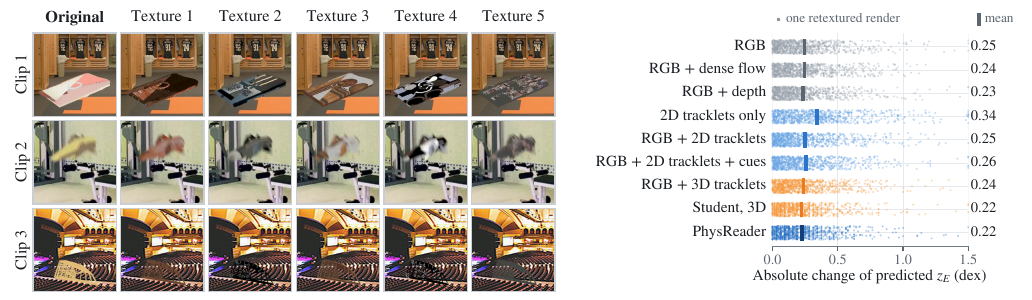}
\caption{\textbf{Same motion, different texture.} Left: three synthetic validation clips at the collision frame as the predictors see them (with motion blur and compression, cut to the predictor crop), rendered with the original texture and with five replacement textures; motion, camera, background, and lighting are identical. Right: absolute change of the predicted $z_E$ from the original render; each dot is one of the 880 pairs (changes above 1.5\,dex, at most nine per model, are drawn at 1.5), and the bar is the mean (value at the right). Gray: no tracklets; blue: 2D tracklets; orange: 3D tracklets.}
\label{fig:app_texture}
\end{figure*}

\paragraph{Texture replacement.}
Fig.~\ref{fig:app_texture} shows three clips of the texture test of App.~\ref{sec:material_appearance}, as the predictors see them, and the spread of the prediction changes over all 880 pairs.
The replacement textures are the texture images of five other objects, the same five for every clip.
For the models with RGB, three quarters of the changes are below 0.36\,dex, and one in ten exceeds 0.47--0.61\,dex.

\begin{figure*}[t]
\centering
\suppfig[0.93\textwidth]{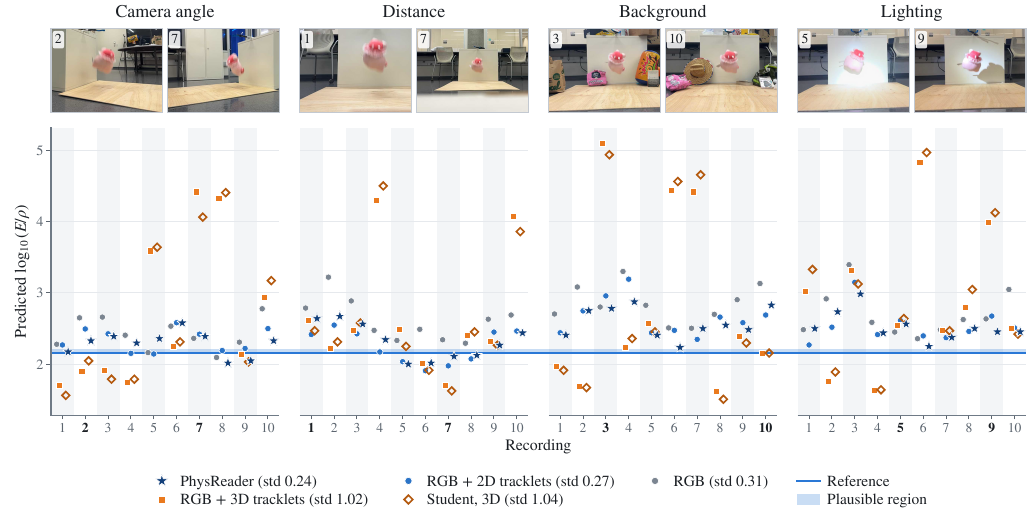}
\caption{\textbf{One object, forty recordings.} Kitty was dropped and recorded 40 more times, ten times under each of four changed conditions. Top: two of the ten recordings per condition. The number in each image is the index of the recording; the same index is bold on the axis below. Each image is the first of the 16 frames given to the predictors, three frames before the collision. Bottom: every predicted $z_E=\log_{10}(E/\rho)$, one column per recording. The line is the reference stiffness of Kitty ($z_E=2.15$) and the thin band its plausible region. The 40 predictions of \model{} span 1.0\,dex, with a standard deviation of 0.24\,dex. Those of RGB + 2D tracklets and of RGB each span 1.3\,dex (standard deviations 0.27 and 0.31\,dex). Those of RGB + 3D tracklets and of the student on 3D tracklets each span 3.5\,dex (standard deviations 1.02 and 1.04\,dex).}
\label{fig:app_kitty}
\end{figure*}

\paragraph{One object, forty recordings.}
Fig.~\ref{fig:app_kitty} shows the 40 additional recordings of Kitty that Tables~\ref{tab:recording_stability} and~\ref{tab:kitty_conditions} summarize, with every single prediction.
The predictions of \model{}, RGB + 2D tracklets, and RGB span 1.0 to 1.3\,dex.
Those of the two models on 3D tracklets span 3.5\,dex across recordings of the same toy.
This happens under each of the four changed conditions (camera angle, distance, background, and lighting).
The line in the figure is the reference stiffness of Kitty.

\begin{figure*}[t]
\centering
\suppfig[0.93\textwidth]{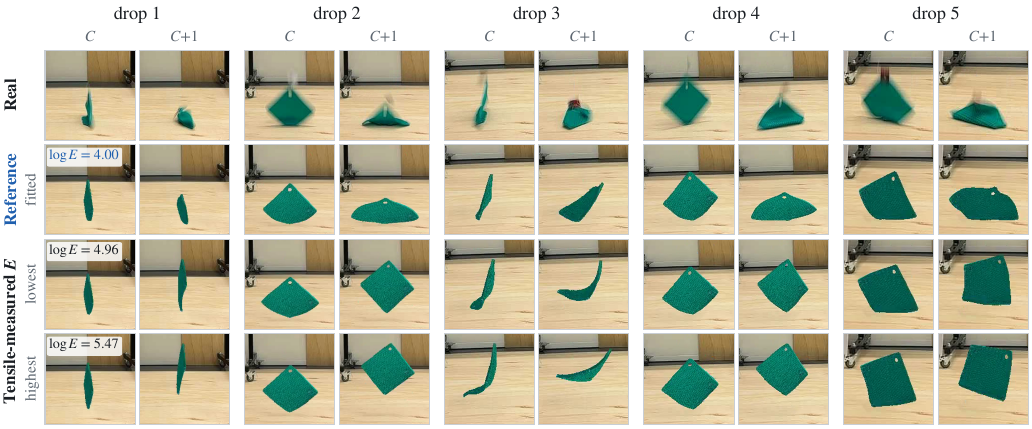}
\caption{\textbf{The silicone pad replayed with tensile-measured stiffness.} The five drops at the collision frame $C$ and one frame later. Rows: the real video; the replay with the reference material ($\log_{10}E=4.00$, $E$ in Pa); and replays with the lowest and the highest local stiffness of the pad's tensile curves, that is, the lowest and the highest slope found on its stress--strain curves ($\log_{10}E=4.96$ and $5.47$; the specimen mean is $5.24$). Yield stress, density, and Poisson's ratio are unchanged. At $C+1$ the real pad and the reference replay lie folded on the floor, while the replays with measured stiffness are flat plates that rebound. All replays start from the reference initial state of the drop.}
\label{fig:app_measured_replay}
\end{figure*}

\begin{figure*}[t]
\centering
\suppfig[\textwidth]{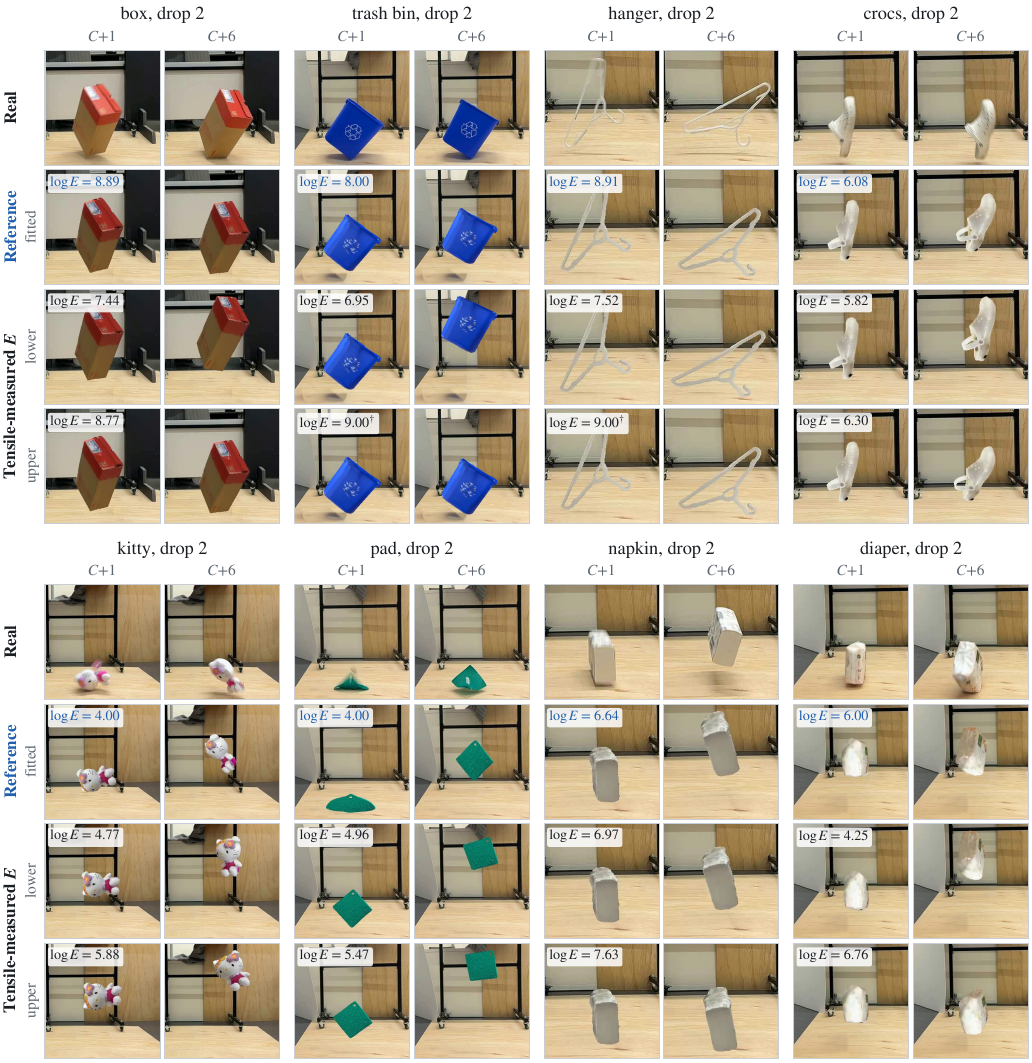}
\caption{\textbf{All eight tensile-tested objects replayed with tensile-measured stiffness.} Drop 2 of each object, one and six frames after the collision. Rows: the real video; the replay with the reference material; and replays with the lowest and the highest local stiffness of the object's tensile curves, that is, the lowest and the highest slope found on its stress--strain curves. Yield stress, density, and Poisson's ratio are unchanged. Numbers are $\log_{10}E$ in Pa. All replays start from the reference initial state of the drop, so only the stiffness differs. A dagger marks the trash bin and the hanger: their highest tensile stiffness lies above 1\,GPa, the largest stiffness our simulator accepts, so they are replayed at 1\,GPa.}
\label{fig:app_measured_replay_all}
\end{figure*}

\paragraph{Tensile-measured stiffness in replay.}
Fig.~\ref{fig:app_measured_replay} illustrates why the reference of a structured object can be softer than its specimens (App.~\ref{sec:app_reference_free}).
The silicone pad has the largest gap: its reference stiffness is 1.24\,dex below the mean stiffness of its tensile specimens.
One frame after the collision the real pad lies folded on the floor, and so does the replay with the reference material.
With the stiffness measured on its specimens, the simulated pad is a flat plate again and rebounds.
This is clearest in drops 2, 4, and 5; in drops 1 and 3 the pad is seen edge-on.
All replays start from the reference initial state of the drop, so only the stiffness differs.
Later in the window every replay, including the reference replay, rebounds more than the real pad, which stays on the floor. The reference stiffness of the pad lies at the lower bound of our search range ($E=10^4$\,Pa).
Fig.~\ref{fig:app_measured_replay_all} shows the same comparison for one drop of each of the eight tensile-tested objects.

\clearpage\suppresetcolumn
\paragraph{Tensile-measured stiffness for all eight objects.}
Fig.~\ref{fig:app_measured_replay_all} repeats the replay with tensile-measured stiffness for one drop of each of the eight objects with tensile tests.
The tensile curves do not give one stiffness per object.
Their local slope, which is the stiffness, varies between specimens, loading directions, and strain levels.
Within an object it varies by 0.5 to 2.5\,dex; for the trash bin, it ranges from 8.9\,MPa to 1.1\,GPa.
We replay each drop at the lowest and at the highest of these values.
The reference lies inside this interval for four objects (trash bin, hanger, crocs, and diaper), 0.12\,dex above it for the box, and below it for Kitty, the pad, and the napkin pack.
This interval runs from the lowest to the highest local slope. It differs from the band of Fig.~\ref{fig:benchmark}, which is the mean $\pm$ 3 standard deviations of the specimens' Young's moduli; App.~\ref{sec:app_reference_free} compares the references with that band.
The replays at tensile stiffness differ visibly from the reference replay for the pad, which becomes a rigid plate at both values.
At the lowest value, the trash bin, the crocs, and the diaper also rebound higher than in the reference replay.
For the box, the hanger, Kitty, and the napkin pack, the replays at tensile stiffness look like the reference replay.
For Kitty, the pad, and the diaper, the reference replay itself rebounds more than the real object, and every replay of the napkin pack rebounds less than the real pack.
The comparison is qualitative. All replays start from the reference initial state of the drop, so only the stiffness differs.
The highest tensile stiffness of the trash bin and the hanger lies above 1\,GPa, the largest stiffness our simulator accepts; we replay them at 1\,GPa.

\section{Vision-Language Model Baselines}
\label{sec:app_vlm_baselines}

We evaluate OpenAI GPT-6 Astra (\texttt{gpt-6-astra}) on all 80 real drop recordings under two input conditions: a single image and a sequence of 16 frames. Each drop and input condition is processed in a separate request, without previous responses or conversation history. The two baselines share the system prompt, parameter bounds, and model settings.

\paragraph{Image input.}
We use frame~0 of the original recording. From its SAM~3.1 object mask, we compute a bounding box and construct a square crop centered at the box center, with side length $\lceil1.15\max(w,h)\rceil$, where $w$ and $h$ are the box dimensions in pixels. Pixels outside the image are padded by edge replication. The crop is resized to $512\times512$ using Lanczos interpolation. We retain the natural background inside the crop; the mask is used only to determine the crop. No additional frames or observations of later motion are supplied.

\paragraph{Video input.}
We use exactly the same 16 frames of the original recording as our predictors. This is the window from $C-3$ through $C+12$ around the collision frame $C$, so $C$ is the fourth frame of the sequence. We compute a single square crop from the union of the SAM~3.1 object masks over these frames, with the same margin, padding, and resizing as the image baseline. This crop remains fixed throughout the sequence. We send the frames as 16 separate image inputs within one request, in chronological order, interleaved with text specifying their original frame indices and timestamps. The timestamp of frame $i$ comes from its presentation time $T_i$, decoded from the source video. We express it relative to frame $C$:
\begin{equation}
 t_i = T_i-T_C.
\end{equation}
For a constant-rate recording, this equals $(i-C)/f$, where $f$ is its original frame rate. We do not draw timestamps on the RGB images.

\paragraph{Shared context and parameter outputs.}
The prompts specify the benchmark's homogeneous, isotropic Neo-Hookean elastic model with von Mises plasticity. We supply the fixed density and Poisson's ratio of the object in the recording, gravity, original image dimensions, crop coordinates, processed image dimensions, padding convention, and available original camera intrinsics. Missing intrinsics are recorded as unavailable. Reference materials, plausible regions, tensile measurements, and predictions from other recordings are not included in the prompts.
The permitted absolute parameter ranges are
\begin{equation}
 E\in[10^4,10^9]\,\mathrm{Pa},\qquad
 \sigma_Y\in[10^5,2\times10^7]\,\mathrm{Pa}.
\end{equation}
For the supplied density $\rho$ in $\mathrm{kg/m^3}$, the corresponding permitted log-ratio bounds are
\begin{align}
 z_E &\in[4-\log_{10}\rho,\;9-\log_{10}\rho],\\
 z_Y &\in[5-\log_{10}\rho,\;\log_{10}(2\times10^7)-\log_{10}\rho].
\end{align}
The model returns $z_E$ and $z_Y$, a categorical description of yield evidence, and a short evidence summary. Absolute parameters are calculated in code as
\begin{equation}
 \widehat E=\rho\,10^{\widehat z_E},\qquad
 \widehat\sigma_Y=\rho\,10^{\widehat z_Y}.
\end{equation}
The model does not predict density separately.

\paragraph{API settings and evaluation.}
We use the official OpenAI Python SDK (version 2.48.0) and Responses API, with image detail set to \texttt{high}, reasoning effort to \texttt{medium}, and a maximum output budget of 4,096 tokens, including reasoning tokens. Temperature, top-$p$, and seed are left unspecified for both baselines. Responses use the API's strict JSON-schema output format with exactly the four fields listed below. We retain all raw responses, processed input images, frame indices, timestamps, crop metadata, prompts, and token usage. We do not clamp the predictions to the permitted bounds, and we do not repeat a request because of the accuracy of its prediction. All 160 requests completed successfully on their first attempt. Parameter distance and replay are computed on all 80 drops with the references of Sec.~\ref{sec:benchmark}. Yield stress is scored on the 50 drops whose reference simulation yields, independently of the yield evidence that the model reports.

\paragraph{Shared system prompt.}
The following prompt is identical for the two baselines.
\begin{quote}
\small\raggedright
You estimate effective mechanical parameters of a whole deformable object from visual observations.

The benchmark represents the entire object using one homogeneous, isotropic Neo-Hookean elastic material with von Mises plasticity. Estimate parameters for this effective whole-object model, not merely textbook properties of a visible constituent material.

The requested outputs are:

z\_E = log10(E / rho)

z\_Y = log10(sigma\_Y / rho)

Here E and sigma\_Y are in pascals and rho is in kg/m\textasciicircum{}3. The ratios are expressed numerically in SI units, m\textasciicircum{}2/s\textasciicircum{}2. Do not estimate density separately.

Use the supplied observation and calibration information. Do not assume unavailable measurements, hidden internal structure, or access to other recordings of the object.

When temporal observations are available, use deformation magnitude, deformation timing, recovery, and residual shape change as evidence. Distinguish deformation from rigid translation, rotation, perspective change, and occlusion.

A short interval without visible recovery does not by itself prove plasticity. A recording with no visible plasticity does not uniquely identify yield stress. When evidence is insufficient, still return a best prior-based point estimate, and mark the evidence limitation.

Return only one JSON object with these fields:

- "log10\_E\_over\_rho": a finite number\\
- "log10\_sigmaY\_over\_rho": a finite number\\
- "yield\_evidence": one of "observed", "not\_observed", "unclear", or "not\_assessable"\\
- "evidence\_summary": at most two sentences describing the visible evidence and important uncertainty

Do not output ranges instead of the required point estimates.
\end{quote}

\paragraph{Image user prompt.}
\begin{quote}
\small\raggedright
Observation type: one RGB image.

This is frame 0 of the original drop recording, cropped around the object with a fixed 15\% margin.

No temporal observations are supplied. Do not infer that a particular deformation, rebound, or recovery event was observed.

Shared benchmark context:

\{COMMON\_CALIBRATION\_AND\_CONTEXT\}

Benchmark-wide permitted log-parameter bounds:

\{FROZEN\_LOG\_PARAMETER\_BOUNDS\}

Estimate the two effective material log-ratios for the object.
\end{quote}

\paragraph{Video user prompt.}
\begin{quote}
\small\raggedright
Observation type: a chronological RGB sequence from one drop.

All frames depict the same object in one fixed crop.

The timestamps describe the original recording speed. Time zero corresponds to the contact frame.

Frame timestamps in seconds:

\{ACTUAL\_TIMESTAMPS\}

Shared benchmark context:

\{COMMON\_CALIBRATION\_AND\_CONTEXT\}

Benchmark-wide permitted log-parameter bounds:

\{FROZEN\_LOG\_PARAMETER\_BOUNDS\}

Use the observed temporal behavior to estimate the two effective material log-ratios. Appearance may provide a prior when motion is uninformative, but do not replace observed dynamics with a material name lookup.
\end{quote}

The names in braces are placeholders that we fill in for each request. \{COMMON\_CALIBRATION\_AND\_CONTEXT\} is replaced by the calibration fields described above. \{FROZEN\_LOG\_PARAMETER\_BOUNDS\} is replaced by both the absolute ranges and the density-normalized log bounds. \{ACTUAL\_TIMESTAMPS\} is replaced by a list that gives the original index, the relative frame index, and the relative timestamp of every input frame. After the video user prompt, the request contains the 16 frames in order. Each frame consists of a text entry of the form ``Frame $r$ (original frame $i$), time $=t$ seconds'' and then its RGB image. Actual timestamp values are written to nine decimal places.

\subsection{Agent-Based Material Estimation}
\label{sec:app_agent_baseline}

We evaluate whether a general-purpose coding agent can recover
material parameters by autonomously interacting with a physics
simulator. We use Codex as the agent and provide it with the
full 60-frame real drop video, the reconstructed object mesh,
and our MPM simulator.

\paragraph{Simulation environment.}
The agent uses the same MPM simulator as our other
simulation-based baselines, with Neo-Hookean elasticity and
von Mises plasticity. For each object, density and Poisson's
ratio are fixed to the values in Table~\ref{tab:fixed_reference}, while Young's
modulus $E$ and yield stress $\sigma_Y$ are unknown.
The agent may also estimate drop-specific initial conditions,
including position, orientation, linear velocity, and angular
velocity. The constitutive model and other fixed simulation
settings cannot be changed.

\paragraph{Agent autonomy.}
We do not prescribe a loss function, optimization algorithm,
parameter initialization, or search strategy.
The agent can inspect the video and mesh, execute simulations,
write and run code, analyze simulated results, and iteratively
refine its estimates. It independently determines how to
compare simulated and observed motion and deformation.
The final prediction must consist of one pair of material
parameters $(E,\sigma_Y)$, expressed in Pa, together with the
initial conditions and configuration needed to reproduce its
simulation.

\paragraph{Independent execution.}
Each drop is processed by a fresh Codex session, launched
using \texttt{codex exec}. Each session runs in an isolated
workspace containing only its corresponding video, object
mesh, fixed simulation configuration, and the shared simulator.
The agent cannot access other drop videos, reference material
parameters, PhysReader predictions, or previous optimization
results. All sessions use an identical task prompt, with no
object-specific instructions or optimization hints.

\paragraph{Compute budget.}
Each agent is allocated one NVIDIA A100 GPU and a maximum
wall-clock time of one hour, including code execution,
simulation, and analysis. The time limit is enforced by an
external process supervisor. We retain the agent's final
submitted parameters and execution logs, including simulation
runs and generated code. Runs that do not produce valid
parameters within the budget are recorded as incomplete.

\paragraph{Evaluation.}
We convert the predicted parameters to the density-normalized
quantities $\log_{10}(E/\rho)$ and
$\log_{10}(\sigma_Y/\rho)$ using the fixed object density.
We evaluate the predictions with the same material-parameter
metrics as the other baselines. As for them, yield stress counts
only for drops with yield excitation in the reference simulation
(Sec.~\ref{sec:benchmark}). The replay metrics follow the same
protocol as for all other methods. They do not use the matching
objective that the agent defined for itself.

Unlike our feed-forward predictor, the agent receives the
complete video, object mesh, and simulator, and performs
per-drop test-time computation. Its results therefore show what an autonomous agent with a
simulator can identify. They are not a comparison with our
predictor under the same inputs and the same computational cost.

\paragraph{Agent prompt.}
All sessions receive the following identical task instruction.

\begin{quote}
\small\raggedright
Task: Material Estimation from a Real Drop Video

You are given:\\
-- \texttt{input/real.mp4}: a 60-frame real drop video.\\
-- \texttt{input/mesh.obj}: the reconstructed object mesh.\\
-- \texttt{input/config.json}: the fixed scene configuration.\\
-- \texttt{simulator/}: an MPM simulator and documentation.

Objective: Estimate Young's modulus E and yield stress sigmaY such that the simulation reproduces the real drop as closely as possible.

Density and Poisson's ratio are fixed and must not change. Unknown initial conditions may be estimated.

You may inspect the inputs, write and execute code, run simulations, analyze results, and determine how to estimate the material parameters.

You may choose your own loss functions, optimization algorithms, and search strategies. No specific method is prescribed.

The final prediction must use the provided MPM simulator and constitutive model. Do not replace simulation with manually specified motion or deformation.

Save your estimates in \texttt{output/result.json}, including \texttt{E\_Pa}, \texttt{sigmaY\_Pa}, \texttt{initial\_state}, and notes. Save the code and configuration required for reproduction.

You have a maximum of one hour of wall-clock time. Work autonomously and submit your best estimate.
\end{quote}

\section{Statistics, Validation Protocols, and Robustness}
\label{sec:app_revision}

This section tests how far the main results generalize: their uncertainty over objects, which validation data predict real accuracy, where the lifting failure comes from, and how much training choices matter.

\subsection{Object-Level Uncertainty}
\label{sec:app_rev_bootstrap}
\model{}-Real has 16 objects, so we treat the object as the unit of resampling.
We draw 16 objects with replacement 10,000 times, keep all drops of each drawn object, and report the mean parameter distance with its 95\% percentile interval.
For the difference between two predictors, each draw uses the same objects for both predictors, so the difference is paired.
An object counts as a win for a predictor when the predictor's mean parameter distance over the drops of that object is lower than that of the other predictor (Table~\ref{tab:app_bootstrap}).

\begin{table}[h]
\centering
\small
\caption{\textbf{Object-level intervals} for Table~\ref{tab:representation_tradeoff}: parameter distance on real drops with its 95\% interval, the paired difference from RGB + 2D tracklets ($\Delta$: the row minus RGB + 2D tracklets, so a positive value means that the row is farther from the plausible region), and the number of objects (of 16) on which the row has a lower mean distance than RGB + 2D tracklets (Wins).}
\label{tab:app_bootstrap}
\begin{adjustbox}{max width=\linewidth}
\begin{tabular}{@{}lccc@{}}
\toprule
Input & Distance [95\%] & $\Delta$ vs.\ 2D [95\%] & Wins \\
\midrule
RGB & 0.797 [0.563, 1.026] & $+0.052$ [$ -0.005, 0.102$] & 4 \\
RGB + dense flow & 0.833 [0.610, 1.057] & $+0.089$ [$ -0.027, 0.207$] & 7 \\
RGB + depth & 0.789 [0.551, 1.020] & $+0.044$ [$ 0.003, 0.092$] & 7 \\
RGB + 2D tracklets & 0.745 [0.520, 0.966] & -- & -- \\
RGB + 2D tracklets + cues & 0.739 [0.488, 0.993] & $-0.005$ [$ -0.080, 0.073$] & 10 \\
RGB + 3D tracklets (est.\ depth) & 1.210 [0.779, 1.632] & $+0.465$ [$ 0.073, 0.843$] & 4 \\
Student, 3D tracklets & 1.164 [0.756, 1.572] & $+0.419$ [$ 0.042, 0.785$] & 5 \\
\model{} & 0.677 [0.448, 0.907] & $-0.068$ [$ -0.136, 0.009$] & 12 \\
\bottomrule
\end{tabular}
\end{adjustbox}
\end{table}

\subsection{Validation Protocols}
\label{sec:app_rev_protocols}
Table~\ref{tab:validation_protocols} compares several kinds of validation data.
We call each of them a protocol.
Each protocol gives one validation error to each of the \tauN{} predictors of Table~\ref{tab:representation_tradeoff} that share the RGB branch; the depth-augmentation variant is left out.
We report Kendall's $\tau$ between these validation errors and the real parameter distances of the same predictors.
\emph{Synthetic validation} is the $z_E$ error on the 14,588 held-out synthetic videos.
For \emph{Real scenes, $E$ swept}, we re-simulate each real drop in its reconstructed scene once for each stiffness value $\log_{10}E\in\{4,4.5,\dots,9\}$ (App.~\ref{sec:material_appearance}).
The validation error is the log-$E$ error of a predictor on these re-simulated drops, measured against the stiffness used in the simulation.
This protocol needs scanned objects and scenes and fitted initial states, but no material labels.
\emph{Replays of real drops} is the $z_E$ error on the same drops re-simulated with the reference materials (Table~\ref{tab:same_drops}).
\emph{$k$ real objects} uses $k$ of the 16 real objects as validation data.
The validation error of a predictor is its mean parameter distance on these $k$ objects, and the protocol selects the predictor with the lowest one.
We compute $\tau$ between these validation errors and the parameter distances on the other $16-k$ objects.
We average over all subsets of $k$ objects.
Intervals for $\tau$ resample real objects as in App.~\ref{sec:app_rev_bootstrap}.
\emph{Real scenes, $E$ swept} and \emph{Replays of real drops} use renders of the same objects and scenes as the real test videos.
We therefore add a held-out variant of each (rows ``held-out objects'' in Table~\ref{tab:app_protocols}).
It computes the validation error on the renders of half of the objects and compares it with the real error on the other half.
We average over 1,000 random splits.

\begin{table}[h]
\centering
\small
\caption{\textbf{Validation protocols} with 95\% intervals for Kendall's $\tau$, and how often $k$ real objects select the predictor that is best on the remaining objects.}
\label{tab:app_protocols}
\begin{adjustbox}{max width=\linewidth}
\begin{tabular}{@{}lcc@{}}
\toprule
Protocol & $\tau$ [95\%] & Selects best \\
\midrule
Synthetic validation & $-0.714$ [$-0.786$, $-0.214$] & No \\
Real scenes, $E$ swept & 0.500 [0.214, 0.857] & Yes \\
\quad held-out objects & 0.554 & 90.5\% \\
Replays of real drops & 0.929 [0.429, 1.000] & Yes \\
\quad held-out objects & 0.688 & 52.6\% \\
1 real object & 0.299 & 12.5\% \\
2 real objects & 0.428 & 42.5\% \\
4 real objects & 0.568 & 65.8\% \\
\bottomrule
\end{tabular}
\end{adjustbox}
\par\vspace{2pt}\footnotesize Selects best: for the protocols that use all 16 objects (the rows with an interval), Yes or No says whether the protocol selects the predictor with the lowest parameter distance on all real objects. For the other rows, the percentage is the share of splits or subsets in which the protocol selects the predictor that is best on the held-out objects. Intervals: for synthetic validation and the sweep, we resample the real objects and keep the validation errors fixed. For the replays, each drawn object brings both its replays and its real drops. The held-out and $k$-object rows are means over splits or subsets.
\end{table}

\subsection{Rendered versus Estimated Depth on Replays}
\label{sec:app_rev_depth}
The replays are the real drops re-simulated with the reference materials and rendered.
On the replays, a lifted model can fail for two reasons: the errors of depth estimation, or the change from the training scenes to the scenes of the real drops.
To separate the two, we evaluate the lifted models on the replays twice (Table~\ref{tab:same_drops}).
The column ``Re-sim.'' gives each model depth estimated on the rendered replay, in the same way as on its real videos.
The column ``Re-sim., rendered'' gives the renderer's metric depth.
Both columns use WAFT tracks.
The model trained on exact geometry (condition A of Table~\ref{tab:3d_cross_eval}) scores 0.820\,dex with estimated depth and 1.310\,dex with the renderer's depth.
We also evaluate it with the simulator's depth and correspondences, so that its replay input contains no estimation error at all.
With exact depth and tracks, its $z_E$ error on the replays is 1.118\,dex, against 0.414\,dex on synthetic validation.
The replays therefore differ from the training scenes in their 3D geometry, and not only in the errors of depth estimation.

\subsection{Depth Augmentation and Geometry-Only Models}
\label{sec:app_rev_aug}
The depth-augmented model is the row ``+ depth augmentation'' of Table~\ref{tab:representation_tradeoff}.
It is trained like RGB + 3D tracklets (est.\ depth), the row above it, except that we perturb the 4RC depth of the training videos before lifting.
The perturbations are:
one scale per clip drawn log-uniformly from $[0.8,1.25]$, a per-frame multiplicative drift following a random walk with 1\% steps, Gaussian noise with a standard deviation of 0.5\% of the depth, and depth removed within 3 pixels of the mask boundary with probability 0.3.
The geometry-only model is the row ``3D tracklets only (est.\ depth)'' of Table~\ref{tab:representation_tradeoff}.
It uses the tracklet encoder and the context branch of the ``2D tracklets only'' model, and it takes lifted tracklets as input.
Both are trained for five epochs with the schedule of App.~\ref{sec:app_impl_details}; their results are in Table~\ref{tab:representation_tradeoff}.

\subsection{Training Epochs}
\label{sec:app_rev_epochs}
The main paper reports a reversal: tracklets lifted to 3D have a lower error than 2D tracklets in simulation and a higher error on real drops.
The models of the main paper are trained with one seed (App.~\ref{sec:app_seeds} adds two more).
As a further check, we test here that the reversal does not depend on the epoch at which the models are evaluated.
Table~\ref{tab:app_epochs} lists the $z_E$ error after every epoch, on synthetic validation and on the 80 real drops.
At every epoch, lifting lowers the synthetic error and raises the real error.
This holds for the direct predictors (RGB + 2D tracklets against RGB + 3D tracklets) and for the students (\model{} against the student on 3D tracklets).

\begin{table}[h]
\centering
\small
\caption{\textbf{The reversal at every epoch:} $z_E$ error (dex) on synthetic validation / on the 80 real drops, against the reference material (not the parameter distance of Table~\ref{tab:representation_tradeoff}; see Table~\ref{tab:app_reading}).}
\label{tab:app_epochs}
\begin{adjustbox}{max width=\linewidth}
\begin{tabular}{@{}lccccc@{}}
\toprule
Input & Epoch 1 & 2 & 3 & 4 & 5 \\
\midrule
RGB + 2D tracklets & 0.554 / 0.790 & 0.501 / 0.736 & 0.487 / 0.789 & 0.483 / 0.774 & 0.482 / 0.792 \\
RGB + 3D tracklets & 0.520 / 1.303 & 0.493 / 1.418 & 0.450 / 1.282 & 0.446 / 1.245 & 0.444 / 1.249 \\
\model{} & 0.542 / 0.751 & 0.501 / 0.709 & 0.481 / 0.722 & 0.481 / 0.738 & 0.481 / 0.731 \\
Student, 3D tracklets & 0.507 / 1.113 & 0.480 / 1.326 & 0.447 / 1.288 & 0.447 / 1.207 & 0.443 / 1.214 \\
\bottomrule
\end{tabular}
\end{adjustbox}
\end{table}

\subsection{Per-Scene Optimizers with Other Initializations}
\label{sec:app_rev_init}
\begin{figure*}[t]
\centering
\includegraphics[width=\textwidth]{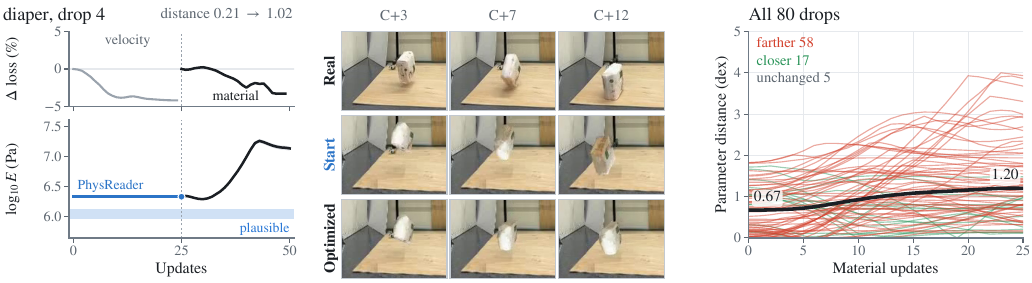}
\caption{\textbf{Inside per-scene optimization.} PhysDreamer is started from \model{}'s predictions and runs 25 velocity updates and 25 material updates. Left: one example, the drop at the upper quartile of the change in parameter distance over the 80 drops (diaper, drop 4). The top plot shows the optimizer's image loss, relative to its value at the first update of each stage (grey: velocity stage; black: material stage). The bottom plot shows the volume mean of the stiffness field (black; the band spans the 10th to 90th volume percentile), \model{}'s prediction (blue), and the plausible interval of the object (light blue). Middle: observed frames, and the optimizer's own renderings with the starting material and in its last rendered update, with the material after 24 of the 25 updates (the optimizer does not render the final material). Right: parameter distance of all 80 drops against the number of material updates. The black line is the mean. Red lines are the 58 drops that end farther from the plausible region, green lines the 17 drops that end closer, and grey lines the 5 drops whose distance is unchanged.}
\label{fig:app_optimizer}
\end{figure*}
PhysFlow and MonoPhysics are each run from two initializations on all 80 drops (Table~\ref{tab:external_comparisons}).
Started from \model{}'s predictions, which have a parameter distance of 0.677, PhysFlow ends at 0.668 and MonoPhysics at 0.698.
Started from the predictions of the image-input VLM (1.267), PhysFlow ends at 1.266.
Started from the constants of its original implementation (parameter distance 1.972; ``Original init'' in Table~\ref{tab:external_comparisons}), MonoPhysics ends at 1.933.
PhysDreamer in Table~\ref{tab:external_comparisons} starts from \model{}'s predictions and runs 25 velocity and 25 material updates on all 80 drops.
The mean parameter distance rises from 0.677 to 1.203.
PhysDreamer clips the predictions to its parameter bounds before it starts; after this clipping, the starting distance is 0.673.
Of the 80 drops, 58 end farther from their plausible region, 17 end closer, and 5 stay inside it.
We then replay the drops with the optimized materials.
Relative to the reference material, the total replay loss increases by 21.0\% and the deformation replay loss by 40.2\%.
For \model{}'s predictions, from which the optimization started, the increases are 10.1\% and 12.3\%.
$\mathrm{Err}_{\mathrm{ref}}$ rises from 0.107 for the predictions to 0.147 for the optimized materials.
We also started PhysDreamer from the predictions of our RGB + 2D tracklets + cues predictor (0.739; Table~\ref{tab:representation_tradeoff}), with the same 25 velocity and 25 material updates.
That run ends at a parameter distance of 1.164, with increases of 22.9\% in the total and 43.3\% in the deformation replay loss and an $\mathrm{Err}_{\mathrm{ref}}$ of 0.143, close to the values of the run from \model{}'s predictions.
PhysDreamer's result therefore does not come from its initialization.
For every optimizer and every initialization above, the optimized material either stays close to its initialization or moves away from the reference.

Fig.~\ref{fig:app_optimizer} looks inside the PhysDreamer runs that start from \model{}'s predictions.
Over the 25 material updates, the volume mean of the stiffness field moves by 0.73\,dex on average, and the mean parameter distance grows from 0.67 to 1.20.
The optimizer's own image loss does not show this.
In the example of the figure, the loss falls while the stiffness moves away from the plausible region.
Over all drops, the change of the loss and the change of the distance are not correlated.
A lower loss therefore does not mean a better material: the optimizer drifts along directions of the material parameters that its loss does not see.
The parameter distance is computed from the volume means of PhysDreamer's material field, and the loss shown is the optimizer's training loss, on the frames that supervise it.

\section{Pixie: A Baseline on the Static Asset}
\label{sec:app_pixie}

Pixie~\cite{le2025pixie} predicts a material field from the appearance of a static 3D asset.
It does not see the drop and predicts no yield stress; of its predictions we use Young's modulus $E$ only, so we report it here and not in Table~\ref{tab:external_comparisons}.

\paragraph{Inputs and material prediction.}
We use the released Pixie implementation and pretrained material
networks, without training them on our synthetic dataset.
The input is each object's reconstructed textured mesh.
Material prediction uses no real drop video, reference material
parameters, plausible regions, tensile measurements, or
\model{} predictions.
Each of the 16 reconstructed objects produces one material field,
which is subsequently evaluated on its five real drops.

For each asset, we render 200 calibrated RGB views at
$512\times512$ resolution.
On these views, we fit the feature field of the released pipeline (F3RM) for 5,000 steps.
This field reconstructs CLIP features in 3D.
From it we extract a $64^3$ grid of 768-dimensional CLIP features,
together with its occupancy mask.
These steps fit the visual feature field; they do not optimize
material parameters against the observed drop.
The released pretrained 3D material network then
predicts spatial fields of Young's modulus, density, and Poisson's
ratio.
We keep Pixie's output normalization and decoding, including the
clipping that its decoder applies.
We do not clip its predictions further to the parameter ranges of
our benchmark.
Pixie has no yield-stress prediction head.

\paragraph{Mapping to simulation particles.}
We map the predicted Young's modulus field onto the benchmark's
MPM particles. To do so, we apply the coordinate transformations between the
reconstructed mesh, the rendered asset, the feature grid, and the simulation.
We use the rule that the released code applies to continuous properties: each particle receives the unweighted average of the predicted modulus over its ten nearest support points. The support points are the occupied cells of the feature grid.
The released code uses a distance threshold of 0.1 in normalized asset
coordinates.
A particle that is farther than this threshold from the support
points receives the global mean of the predicted modulus.
The release requires that such particles are no more than 10\% of
all particles, and we keep this requirement.
We preserve the resulting per-particle moduli during simulation.

For parameter evaluation, we compute a volume-weighted equivalent
modulus in linear units,
\begin{equation}
E_{\mathrm{eq}}
=
\frac{\sum_p V_p E_p}{\sum_p V_p},
\end{equation}
where $E_p$ and $V_p$ denote the modulus and volume of particle $p$.
The logarithm is taken after this averaging.
We compare $E_{\mathrm{eq}}$ with the $E$ projection of the
plausible region, that is, the range of stiffness that the region
covers when yield stress is ignored.
We do not assign a yield-stress prediction or report a joint
$E$--$\sigma_Y$ parameter distance for Pixie.

\paragraph{Reconstruction failures and recovery.}
With the released settings, the first reconstruction gave an
occupancy mask with no usable cells for bottle, pad, belt, and
hanger.
For bottle and pad, we applied one fixed square crop shared across
all rendered views, enclosing the union of the rendered object
bounds with a 15\% margin.
The cropped images were resized to $512\times512$, and the camera
intrinsics were updated accordingly.
Camera extrinsics and mesh normalization were preserved.

For belt and hanger, cropping alone did not recover usable
occupancy.
We additionally gave rendered object masks, dilated by eight
pixels, to the pixel sampler of the released reconstruction code,
which selects the pixels used in the fit.
Bottle and pad thus took two attempts each, and belt
and hanger three.
Each attempt used the same 5,000-step budget.
These adaptations were triggered by reconstruction failure,
not by material accuracy or replay error.
The material-network weights, decoder, occupancy thresholds,
and mapping acceptance criteria remained unchanged.
We kept the records of all failed attempts, and their runtimes are
counted in the timing below.

\paragraph{Re-simulation evaluation.}
We use our benchmark MPM implementation with Neo-Hookean
elasticity and explicitly disable plasticity, since Pixie does
not predict yield stress.
Density and Poisson's ratio are fixed to the benchmark values.
Pixie also predicts these two quantities.
We keep its predictions in the saved outputs but do not use them
in the replay.
The predicted Young's-modulus field remains fixed throughout
evaluation.

For the relative replay losses, we refine the initial state of
each drop with the same search over 25 candidates that the
benchmark uses.
The search uses the bounds, the proposal procedure, and the
objective of the benchmark
(Eq.~\eqref{eq:reference_optimization_loss};
App.~\ref{sec:app_behavior}).
This objective contains a prior on the initial state; the replay
losses that we report do not include this term.
The reference material is replayed with its own initial-state
search.
For each drop, we compute the relative increase of the total and
of the deformation loss over this reference replay, and we then
average the percentages over the drops.
Alignment-quality failures are retained and flagged rather than
excluded from the aggregate.

For $\mathrm{Err}_{\mathrm{ref}}$, prediction and reference
simulations instead share the reference initial state.
Both use the 16-frame window from $C-3$ to $C+12$.
We apply the same trajectory, pose, and deformation comparison
as for the other methods.
The deformation comparison uses local deformation gradients that
we fit to the tracks of the simulated particles, not the
deformation gradients that the simulator stores.

\paragraph{Timing and interpretation.}
The pretrained material network performs feed-forward prediction,
but the complete pipeline also includes per-asset rendering and
5,000 steps of F3RM reconstruction.
These preprocessing costs must be included in an end-to-end
runtime comparison.
They are incurred once per object and shared across its five
drops.
Rendering and reconstruction take 13.5 minutes per object on
average (from 10.5 to 25.9 minutes, depending on the object),
including the failed attempts.
The five drops of an object share one Pixie prediction.
Evaluating this prediction on five drops is therefore not an
independent test of how stable a prediction from video is across
drops.

\paragraph{Results.}
Table~\ref{tab:pixie} gives the results on all 80 drops.
All 2,000 simulations of the initial-state search (25 candidates
for each of the 80 drops) ran without execution failures.
For comparison, the table repeats \model{} and the image-input
vision-language model from Table~\ref{tab:external_comparisons},
with their distance to the same $E$ projection.
Pixie is more accurate than the image-input vision-language model on every metric
and less accurate than \model{} on every metric.

\begin{table}[t]
\centering
\small
\caption{\textbf{Pixie on the 80 real drops.}
$E$ dist.: distance from the predicted stiffness to the $E$ projection of the plausible region (dex); it is not the joint distance of Table~\ref{tab:external_comparisons}.
Pixie is replayed with elasticity only, since it predicts no yield stress.
The other two rows use their predicted yield stress.}
\label{tab:pixie}
\begin{adjustbox}{max width=\linewidth}
\begin{tabular}{@{}lcccc@{}}
\toprule
Method
& \makecell{$E$\\dist.\ $\downarrow$}
& $\Delta_{\mathrm{deform}}^{\mathrm{replay}}\downarrow$
& $\Delta_{\mathrm{total}}^{\mathrm{replay}}\downarrow$
& $\mathrm{Err}_{\mathrm{ref}}\downarrow$ \\
\midrule
Pixie ($E$ only, static asset)
& 0.950 & 31.82\% & 23.29\% & 0.1533 \\
VLM (GPT-6 Astra, image input)
& 1.114 & 34.95\% & 30.54\% & 0.1683 \\
\model{}
& \textbf{0.591} & \textbf{12.32\%} & \textbf{10.15\%} & \textbf{0.1072} \\
\bottomrule
\end{tabular}
\end{adjustbox}
\end{table}

\section{Stiffness and Yield Stress Separately, and What the References Measure}
\label{sec:app_reference_free}

The real-video numbers of the main paper are measured against reference materials that we fit in our own simulator.
This section states what that implies.
All numbers are computed from the predictions behind Tables~\ref{tab:representation_tradeoff} and~\ref{tab:external_comparisons}; no model is retrained.

\paragraph{What the real-video metrics measure.}
A reference is one homogeneous elastoplastic material that, in our simulator, reproduces the five drops of an object best (App.~\ref{sec:app_reference_objective}).
It is an effective parameter of the whole object under this simulator, not an intrinsic constant.
The parameter distance, the $z_E$ and $z_Y$ errors, and $\mathrm{Err}_{\mathrm{ref}}$ therefore measure agreement with these fitted references.
The replay losses compare a simulation with the observed video and do not use the reference material, but their search for the initial state starts from the reference initial state, and they use the same simulator and the same observation terms as the fit.
This is what a predictor is asked for: the material that reproduces the drop under our physical model, which is Neo-Hookean elasticity with von Mises plasticity in an MPM simulator.
We therefore evaluate every prediction against the references, and not against the tensile tests.
The tensile tests are used only to check the references themselves. They cover stiffness for eight of the 16 objects (Fig.~\ref{fig:benchmark}). A tensile test destroys the object, because the specimens are cut from it, and it is expensive and takes much time and labor, so we tested half of the objects. The eight tested objects cover very different materials: their specimen moduli range from 0.05\,MPa (the diaper) to 963\,MPa (the hanger), more than four orders of magnitude (Table~\ref{tab:tensile_numbers}).

\paragraph{Stiffness and yield stress separately.}
The parameter distance combines stiffness and yield stress in one number.
Table~\ref{tab:app_reference_free} reports them separately.
The $z_E$ error is the mean absolute difference between the predicted and the reference $\log_{10}(E/\rho)$ over the 80 drops.
The $z_Y$ error is the same for $\log_{10}(\sigma_Y/\rho)$, over the 50 drops of the ten objects whose reference simulation yields.
The main paper reports a reversal for stiffness: lifting the tracklets to 3D lowers the error in simulation and raises it on real drops.
The same holds for yield stress.
From RGB + 2D tracklets to RGB + 3D tracklets, the synthetic $z_Y$ error falls from 0.363 to 0.344\,dex, and the real $z_Y$ error rises from 0.423 to 0.583\,dex.
Among the predictors of Table~\ref{tab:app_reference_free}, \model{} has the lowest real $z_Y$ error (0.381\,dex).
Yield references come from the simulator fit only.
The stress--strain curves of our tensile tests do not identify a yield stress (App.~\ref{sec:app_real_measure}), so no yield reference is checked by a measurement.

\begin{table}[t]
\centering
\small
\caption{\textbf{Stiffness-only, yield-only, and signed errors on real drops} (dex). $z_E$: mean absolute error of $\log_{10}(E/\rho)$ against the reference material, over the 80 drops. $z_Y$: the same for $\log_{10}(\sigma_Y/\rho)$, over the 50 drops whose reference simulation yields. Signed: mean of predicted minus reference $z_E$; a negative value means that the predictions are too soft. Softer: number of drops (of 80) predicted softer than the reference.}
\label{tab:app_reference_free}
\begin{adjustbox}{max width=\linewidth}
\begin{tabular}{@{}lcccc@{}}
\toprule
Model & $z_E\downarrow$ & $z_Y\downarrow$ & Signed & Softer \\
\midrule
RGB & 0.829 & 0.457 & $+0.03$ & 36 \\
RGB + dense flow & 0.862 & 0.421 & $+0.12$ & 36 \\
RGB + depth & 0.822 & 0.454 & $+0.00$ & 35 \\
2D tracklets only & 1.110 & 0.492 & $+0.02$ & 43 \\
RGB + 2D tracklets & 0.792 & 0.423 & $+0.00$ & 31 \\
RGB + 2D tracklets + cues & 0.792 & 0.426 & $-0.19$ & 39 \\
RGB + 3D tracklets (est.\ depth) & 1.249 & 0.583 & $-1.08$ & 67 \\
\midrule
Student, 3D tracklets & 1.214 & 0.552 & $-1.06$ & 66 \\
\model{} (student, 2D tracklets) & 0.731 & 0.381 & $-0.11$ & 35 \\
\midrule
Vid2Sim predictor & 0.968 & 0.545 & $-0.61$ & 59 \\
Analytic Dynamics & 1.071 & 0.533 & $-0.46$ & 51 \\
VLM (GPT-6 Astra, image input) & 1.259 & 0.531 & $+0.26$ & 36 \\
VLM (GPT-6 Astra, video input) & 0.972 & 0.719 & $+0.23$ & 35 \\
\bottomrule
\end{tabular}
\end{adjustbox}
\end{table}

\begin{table*}[t]
\centering
\small
\caption{\textbf{Why the same model can have two numbers.} All values in dex.}
\label{tab:app_reading}
\begin{tabular}{@{}p{3.6cm}p{3.5cm}p{3.5cm}p{5.6cm}@{}}
\toprule
Model and quantity & First value & Second value & Reason \\
\midrule
\model{}, real error & 0.677 (Tables~\ref{tab:representation_tradeoff}, \ref{tab:validation_protocols}, \ref{tab:external_comparisons}; Fig.~\ref{fig:teaser}) & 0.731 (Table~\ref{tab:same_drops}; Fig.~\ref{fig:sim_vs_real}) & Two metrics. First value: parameter distance to the plausible region. Second value: $z_E$ error against the reference material. \\
RGB + 2D tracklets, real error & 0.745 (Table~\ref{tab:representation_tradeoff}) & 0.792 (Tables~\ref{tab:error_budget}, \ref{tab:same_drops}; Fig.~\ref{fig:sim_vs_real}) & Two metrics. First value: parameter distance to the plausible region. Second value: $z_E$ error against the reference material. \\
RGB + 3D tracklets, real error & 1.210 (Table~\ref{tab:representation_tradeoff}) & 1.249 (Table~\ref{tab:same_drops}) & Two metrics. First value: parameter distance to the plausible region. Second value: $z_E$ error against the reference material. \\
RGB + 3D tracklets, real $z_E$ error & 1.249 (Table~\ref{tab:same_drops}) & 1.188 (Tables~\ref{tab:error_budget}, \ref{tab:3d_cross_eval}; ``D'' in Fig.~\ref{fig:sim_vs_real}) & Two training runs with the same inputs. First value: the main run, reported in Table~\ref{tab:representation_tradeoff}. Second value: run D, the model trained under condition D in the study of Table~\ref{tab:3d_cross_eval}. \\
RGB + 3D tracklets, synthetic $z_E$ error & 0.444 (Tables~\ref{tab:representation_tradeoff}, \ref{tab:same_drops}) & 0.444 (Table~\ref{tab:error_budget}) & A coincidence. The first value is the main run on all 14,588 validation videos; on the 10,202-video subset of Table~\ref{tab:error_budget}, this run scores 0.441. The second value is run D on that subset; on all videos, run D scores 0.450. \\
RGB + 2D tracklets, synthetic $z_E$ error & 0.482 (Tables~\ref{tab:representation_tradeoff}, \ref{tab:same_drops}) & 0.478 (Table~\ref{tab:error_budget}) & Two sets of validation videos. First value: all 14,588. Second value: the 10,202 on which every row of Table~\ref{tab:error_budget} is defined (those with usable visible-surface geometry). \\
Teacher, synthetic $z_E$ error & 0.161 (Tables~\ref{tab:oracle}, \ref{tab:teacher_student}) & 0.145 (Table~\ref{tab:error_budget}) & Two sets of validation videos. First value: the 14,293 with usable surface geometry. Second value: the same 10,202-video subset. \\
\model{}, distance to the plausible region & 0.677 (Table~\ref{tab:representation_tradeoff}) & 0.591 (Table~\ref{tab:pixie}) & Two metrics. First value: joint distance in stiffness and yield stress. Second value: distance in stiffness only, which Table~\ref{tab:pixie} uses because Pixie predicts no yield stress. \\
Random material & 2.071 (Table~\ref{tab:representation_tradeoff}, real) & 1.661 (Table~\ref{tab:representation_tradeoff}, Syn.); 1.745 (Table~\ref{tab:material_sensitivity}) & Expected error of a uniformly random prediction, scored against three sets of labels: the plausible regions of the real drops, the synthetic validation labels, and the labels of the stiffness sweep. \\
\bottomrule
\end{tabular}
\end{table*}

\paragraph{Signed errors.}
The two lifted models, RGB + 3D tracklets and the student on 3D tracklets, do not fail by random scatter.
They fail by a shift in one direction: their predictions are too soft.
On real videos, their predicted stiffness is lower than the reference on 67 and 66 of the 80 drops, respectively.
On average over the 80 drops, their predictions are 1.1\,dex below the reference (columns ``Softer'' and ``Signed'' of Table~\ref{tab:app_reference_free}).
The predictors on 2D tracklets have almost no mean shift.
This shift toward soft materials is also why the lifted models are closer to the references than the 2D models on a few soft objects (App.~\ref{sec:app_epoch5_results}).
All predictors compress the range of stiffness: they predict soft objects too stiff and stiff objects too soft.
The same happens in the simulated sweep in which only $E$ changes (Fig.~\ref{fig:material_sensitivity}).

\paragraph{How close are the references to the tensile tests?}
Table~\ref{tab:tensile_numbers} gives the numbers behind Fig.~\ref{fig:benchmark}.
Five references are within 0.41\,dex of the specimen means: hanger ($-0.07$), crocs ($-0.13$), box ($+0.16$), diaper ($+0.37$), and napkin ($-0.41$).
Each number is $\log_{10}E$ of the reference minus that of the nearest specimen mean, so a negative value means that the reference is softer than the specimens.
The other three references are farther from their specimens, and softer: the trash bin by 0.85\,dex, Kitty by 0.94\,dex, and the silicone pad by 1.24\,dex.
Five references lie at or just above the lower bound of our search range, $10^4$\,Pa: the glove, Kitty, the neck pillow, the pad, and the pillow (Table~\ref{tab:ranges}).
Their true stiffness may be lower, and their plausible regions end at the bound.
Four yield-stress references lie at the lower bound of $10^5$\,Pa: those of the crocs, the diaper, the hanger, and the sneaker.
The bound does not create the gap between 2D and 3D tracklets.
Without the five objects at the stiffness bound, the parameter distance is 0.769\,dex for \model{} and 1.535\,dex for the student on 3D tracklets, against 0.677 and 1.164 on all 80 drops.
The gray band of Fig.~\ref{fig:benchmark} is the mean $\pm$ 3 standard deviations of the specimens.
For Kitty, the pad, and the diaper, the mean minus 3 standard deviations is negative.
On the logarithmic axis of the figure, the band of these three objects therefore has no lower end.
The agreement of these three objects in the figure is thus weak evidence.
The plausible region overlaps the band for seven of the eight objects; the region of the trash bin lies 0.42\,dex below its band (last column of Table~\ref{tab:tensile_numbers}).
The reference itself lies inside the band for five objects.
The trash bin, Kitty, and the pad, whose references are softer than their specimens by 0.85\,dex or more, are structured objects: a hollow bin, a stuffed plush toy, and a honeycomb pad.
A whole structured object can respond more softly than a specimen cut from its material.
Fig.~\ref{fig:app_measured_replay} shows this for the silicone pad: replayed with the stiffness of its specimens, the simulated pad is a nearly rigid plate one frame after the collision, when the real pad lies folded on the floor.

\paragraph{Why not use the tensile tests as labels?}
The tensile tests cannot replace the fitted references as labels, for three reasons.
First, they exist for eight of the 16 objects, and they give no yield stress.
Second, they do not give one stiffness per object.
The mean Young's moduli of the two loading directions differ by up to 0.93\,dex (for the diaper).
The local slope of the stress--strain curves varies within an object by 0.5 to 2.5\,dex, depending on the object (App.~\ref{sec:app_qualitative}).
Third, a specimen measures the material that it is cut from, not the response of the whole object.
Figs.~\ref{fig:app_measured_replay} and~\ref{fig:app_measured_replay_all} replay the drops with tensile-measured stiffness.
In these replays, some objects move visibly differently from the real video.
The silicone pad stays a rigid plate where the real pad folds.
The trash bin, the crocs, and the diaper rebound higher than in the reference replay when they are replayed with their lowest tensile stiffness.
Table~\ref{tab:app_tensile_replay} repeats this comparison with numbers.
We replay all drops of the eight objects with three tensile values, the lowest and the highest local slope and the mean Young's modulus, with the same initial-state search as for predicted materials.
Depending on which of the three tensile values is used, the total replay loss is 13.1--24.4\% higher than with the reference material, and the deformation replay loss is 36.0--59.4\% higher.
For \model{}'s predictions on the same drops, the increases are 5.5\% and 11.4\%.
By this measure the tensile values describe the drops less well than the predictions do.
\begin{table}[t]
\centering\small
\caption{\textbf{Replays at tensile stiffness.} We replay the 40 drops of the eight tensile-tested objects with three stiffness values from the tensile tests: the lowest and the highest local slope of the loading curves before the peak, and the geometric mean of the specimens' Young's moduli (App.~\ref{sec:app_real_measure}). The yield stress is that of the reference, and the initial state is searched as for a predicted material (Sec.~\ref{sec:benchmark}, App.~\ref{sec:app_behavior}). All numbers are relative to the reference material; the replay measures are those of Table~\ref{tab:external_comparisons}. For the trash bin and the hanger, the highest local slope lies above 1\,GPa, the largest stiffness our simulator accepts, and is replayed at 1\,GPa. The last two rows are predictions on the same 40 drops.}
\label{tab:app_tensile_replay}
\begin{adjustbox}{max width=\linewidth}
\begin{tabular}{@{}lcccc@{}}\toprule
Material & $\Delta_{\mathrm{total}}^{\mathrm{replay}}\downarrow$ & $\Delta_{\mathrm{deform}}^{\mathrm{replay}}\downarrow$ & $\mathrm{Err}_{\mathrm{ref}}\downarrow$ & Drops \\\midrule
Tensile, lowest slope & 24.4\% & 36.0\% & 0.138 & 40 \\
Tensile, mean modulus & 13.1\% & 48.5\% & 0.106 & 40 \\
Tensile, highest slope & 14.8\% & 59.4\% & 0.116 & 40 \\
\midrule
\model{} & 5.5\% & 11.4\% & 0.098 & 40 \\
VLM (image) & 20.4\% & 23.6\% & 0.159 & 40 \\
\bottomrule\end{tabular}
\end{adjustbox}
\end{table}

We therefore use the tensile tests to check the references, and not as labels.

\paragraph{Does the benchmark favor 2D inputs?}
The deformation terms of the reference fit and of the replay losses compare image-plane cues, the kind of signal that the 2D predictors read. Their trajectory term and the rotation part of their pose term compare 3D motion (App.~\ref{sec:app_optimization}).
Three results do not rely on these image-plane deformation terms.
First, the fit uses SEA-RAFT flow~\cite{wang2024sea_raft}, not the WAFT flow of the predictors (App.~\ref{sec:app_optimization}).
Second, $\mathrm{Err}_{\mathrm{ref}}$ compares trajectory, pose, and 3D strain between two simulations and uses no optical flow.
Third, on the rendered replays of the real drops (Table~\ref{tab:same_drops}), the label is the material that the video was simulated with, so the label is exact.
Lifted tracklets lose there as well: their $z_E$ error is 0.808\,dex, against 0.453\,dex for 2D tracklets.

\paragraph{Rank correlations with matched metrics.}
In Table~\ref{tab:validation_protocols}, the two sides of each correlation use different metrics: the validation error is a $z_E$ error, and the real error is the parameter distance.
Here we use the $z_E$ error on both sides.
Kendall's $\tau$ over the same \tauN{} predictors is then $-0.64$ for synthetic validation, $+0.57$ for the real scenes with $E$ swept, and $+0.86$ for replays of the real drops.
With the real parameter distance, as in Table~\ref{tab:validation_protocols}, the three values are $-0.71$, $+0.50$, and $+0.93$.
We also replace the real error by the distance in stiffness only, from the predicted stiffness to the range of stiffness that the plausible region covers.
It gives the same three values as the real $z_E$ error.
Each validation set selects the same predictor as in Table~\ref{tab:validation_protocols}.

\section{Reading the Numbers Across Tables}
\label{sec:app_reading}

Some models appear with two different numbers in the paper.
Each case has one of four causes: a different metric, a different set of validation videos, a different set of labels, or a separately trained run.
Table~\ref{tab:app_reading} lists them.

\paragraph{The error budget (Table~\ref{tab:error_budget}).}
The RGB + 3D tracklets model was trained twice with the same inputs.
The main run is the one reported in Table~\ref{tab:representation_tradeoff}.
Run D is the model trained under condition D (4RC depth and WAFT tracks) in the study of Table~\ref{tab:3d_cross_eval}.
The last two rows of Table~\ref{tab:error_budget}, for flow tracks lifted with estimated depth, take both numbers from run D: a synthetic $z_E$ error of 0.444\,dex and a real $z_E$ error of 1.188\,dex.
With the main run, these two rows would read 0.441 and 1.249\,dex.
The third row of Table~\ref{tab:error_budget}, RGB with exact 3D tracks of the visible surface, differs from the teacher above it in more than one way: it has a different network and it adds RGB.
The $+0.189$\,dex of that row is therefore the cost of replacing the teacher and its exact deformation features by a video predictor on exact tracks, not the cost of a single factor.

\paragraph{Training runs and selection.}
Every model of the main paper is trained once, with one seed; App.~\ref{sec:app_seeds} adds two more seeds.
All direct predictors and students are reported after the fifth epoch, the teacher is selected on synthetic validation, and no checkpoint or hyperparameter is selected on real videos.
We chose the input of \model{}, 2D tracklets, after comparing the inputs on \model{}-Real.
Renders of the real scenes, which need no material labels, also rank lifted tracklets far behind 2D tracklets and select \model{} (Table~\ref{tab:validation_protocols}).

\paragraph{What the evidence covers.}
The real data are 16 objects and 80 drops in one capture setup, with one camera and a gripper release, plus 40 recordings of one object under changed conditions.
The result that we consider established within this setup is the reversal between 2D and lifted tracklets: lifted tracklets are more accurate than 2D tracklets in simulation and less accurate on real drops. For stiffness, the reversal holds under every metric, depth source, and training condition, and at every epoch. Lifted tracklets also lose on the rendered replays with exact labels. For yield stress, the reversal holds for RGB + 2D tracklets against RGB + 3D tracklets and for the two students (App.~\ref{sec:app_reference_free}), with the exceptions listed in App.~\ref{sec:app_epoch5_results}.
The advantage of the teacher--student training over the direct predictor on 2D tracklets is smaller than this reversal.
It is not significant over objects: its object-level interval includes zero.
\model{} does not have the lowest total replay loss (Table~\ref{tab:representation_tradeoff}). PhysFlow started from its predictions ends slightly closer to the references, 0.668 against 0.677\,dex, but it takes about 250 times longer per video (49\,min against 12\,s), and its deformation replay loss and $\mathrm{Err}_{\mathrm{ref}}$ are higher (17.2\% against 12.3\%, and 0.115 against 0.107; Table~\ref{tab:external_comparisons}).

\section{More Long-Range Trackers}
\label{sec:app_more_trackers}
Fig.~\ref{fig:app_tracking} uses one long-range tracker, CoTracker3 in its online mode.
Here we run ten more trackers and tracker modes on the same 16 frames, $C-3$ to $C+12$, with the same query points at $C-3$ as in that figure. We compare them with our short tracklets.
Each tracker is called through its released code and weights with its default settings.
A track is valid in a frame when two conditions hold. First, the tracker's own visibility output marks the point as visible; we use the visibility threshold of the tracker's released code. Second, the point lies inside the object mask.

\begin{figure*}[t]
\centering
\includegraphics[width=\textwidth]{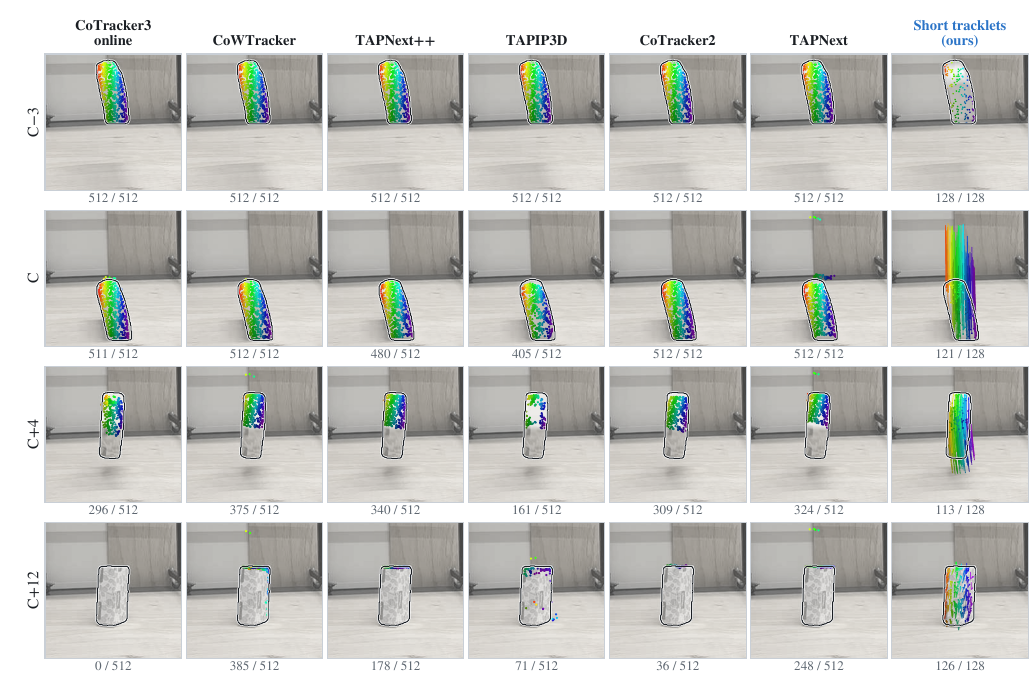}
\caption{\textbf{A failure case of long-range trackers.} Napkin pack, drop 1, at $C-3$ (queries), $C$, $C+4$, and $C+12$ (rows); one column per tracker. Each dot is a tracked query point that the tracker itself marks visible. Each query keeps one color in all frames; the color is given by the position of the query on the object at $C-3$. The line is the object mask. The numbers give visible points / queries; unlike Table~\ref{tab:app_trackers}, they also count visible points outside the object mask. Last column: our short tracklets, drawn as line segments; each frame shows the tracklets of the five-frame window that contains it. Table~\ref{tab:app_trackers} gives the counts over all 80 drops.}
\label{fig:app_trackers_fail}
\end{figure*}

\paragraph{Consistency over the real drops.}
A material predictor needs tracks on the object in every drop.
We therefore measure coverage: the share of the object mask that lies within 8 pixels of a valid track, on the $224\times224$ crop.
Table~\ref{tab:app_trackers} reports coverage at the last frame of the window, $C+12$, for all 80 drops. If a tracker returns fewer than three valid queries for a drop, the drop has no usable tracks and counts as a coverage of zero.
Several recent trackers cover more than half of the object in most drops but fail on a subset of drops.
The best of them, CoWTracker, covers less than half of the object mask at $C+12$ in 10 of the 80 drops. Each of the other long-range trackers does so in 16 to 67 of the 80 drops.
Our tracklets restart on the visible surface at $C-3$, $C+1$, $C+5$, and $C+8$ (window frames 0, 4, 8, and 11). At $C+12$ they cover at least 66\% of the object mask in every drop.
Fig.~\ref{fig:app_trackers_fail} shows one of the failures.
When the napkin pack rotates, the face that was visible at $C-3$ turns away: the trackers lose their points or collapse them onto an edge.
Our tracklets carry no visibility estimate, so we count a tracklet point as valid when it lies inside the object mask. For the long-range trackers we also require that the tracker marks the point as visible. If we count the long-range tracks by position alone, as we count our tracklets, CoWTracker and TAPNext++ each cover less than half of the object at $C+12$ in 2 drops.

\paragraph{Relation to the main paper.}
Sec.~\ref{sec:representations} states that the long-range trackers we tested lose most tracks during impact (Fig.~\ref{fig:app_tracking}).
Table~\ref{tab:app_trackers} gives the full picture for eleven long-range trackers and tracker modes.
Each of them covers less than half of the object at $C+12$ in some of the 80 drops: in 10 drops for CoWTracker, and in 16 to 67 drops for the others.
Our short tracklets do so in none of the 80 drops.
A material predictor needs tracks on the object in every drop, so we use the short tracklets.

\begin{table}[h]
\centering\small
\caption{\textbf{Long-range trackers and short tracklets on the 80 real drops.} Tracks: number of the 80 drops with usable tracks (at least three valid queries). Coverage: share of the object mask within 8 pixels of a valid track. Last two columns: number of the 80 drops with coverage below 0.5 at $C+12$, and number with coverage below 0.5 in the worst frame, the frame between $C+1$ and $C+12$ with the lowest coverage.}
\label{tab:app_trackers}
\begin{adjustbox}{max width=\linewidth}
\begin{tabular}{@{}lccc@{}}\toprule
 & & \multicolumn{2}{c}{Drops with coverage below 0.5} \\\cmidrule(l){3-4}
Tracker & Tracks & at $C+12$ & worst frame \\\midrule
Short tracklets (ours) & 80 & 0 & 10 \\
CoWTracker & 80 & 10 & 13 \\
TAPNext++ & 79 & 18 & 26 \\
TAPIP3D (also given depth) & 75 & 16 & 21 \\
CoTracker2 & 79 & 24 & 31 \\
TAPNext & 79 & 33 & 39 \\
AllTracker & 79 & 49 & 51 \\
CoTracker3, offline & 79 & 47 & 48 \\
CoTracker3, online (Fig.~\ref{fig:app_tracking}) & 77 & 57 & 59 \\
LocoTrack & 79 & 64 & 65 \\
BootsTAPIR & 67 & 60 & 63 \\
TAPIR & 77 & 67 & 69 \\
\bottomrule\end{tabular}
\end{adjustbox}
\end{table}

\section{Poisson's Ratio in the Reference Fit}
\label{sec:app_nu}
Table~\ref{tab:fixed_reference} lists the Poisson's ratios that are fixed in the reference fit.
They were fixed before the tensile data were processed.
For some objects they therefore differ from the apparent values of App.~\ref{sec:app_real_measure}.
We keep the values of Table~\ref{tab:fixed_reference}, for two reasons.

First, the apparent values are not reliable.
App.~\ref{sec:app_real_measure} reads them from the change of one specimen width between two photographs, at 0\% and at about 5\% strain.
The width changes by only a few pixels: for the first box specimen, by 4.3 pixels of a width of 679 pixels.
One pixel of annotation error therefore changes the apparent Poisson's ratio by a quarter of its value.
For the silicone pad the apparent value is 0.58, above the limit of 0.5 for an isotropic solid.

\begin{figure}[h]
\centering
\includegraphics[width=\linewidth]{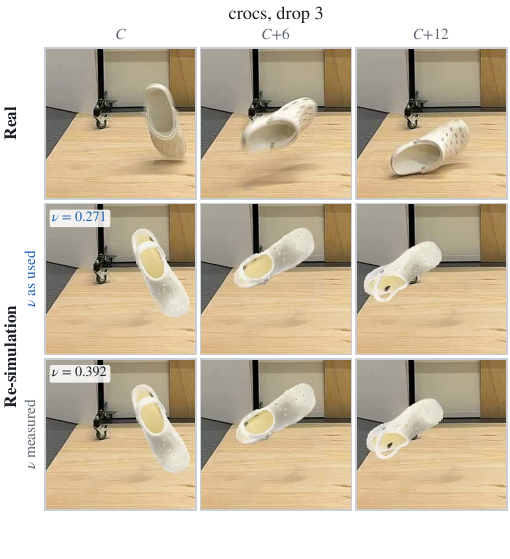}
\caption{\textbf{Re-simulation with a different Poisson's ratio.} Drop 3 of the crocs at the collision frame $C$ and 6 and 12 frames later. Top: the real video. Middle: the reference simulation, with $\nu$ as used in the fit. Bottom: the same simulation with the apparent value of $\nu$.}
\label{fig:app_nu}
\end{figure}

Second, Poisson's ratio has little effect on the simulated drop.
Fig.~\ref{fig:app_nu} shows this for the crocs.
We simulate its five drops again with the reference material and the reference initial state, and change only $\nu$, from the value used in the fit (0.271) to the apparent value (0.392).
After the rigid motion between the two simulations is removed, their corresponding particles are 0.62\,mm apart on average, and the frames cannot be told apart.
One pixel of the $224\times224$ crop given to the predictors covers about 3\,mm on this drop, so the difference is about a fifth of a pixel.
For the box and the hanger, the same test gives 0.03 and 0.20\,mm.
This agrees with the four-head models of App.~\ref{sec:app_controls} (Fig.~\ref{fig:app_four_heads}): the student, which reads the video, does not learn $\nu$, while the teacher, which reads the exact 3D deformation, learns part of it.

\section{Training Seeds}
\label{sec:app_seeds}
The main paper trains every model once.
Here we train nine of the eleven models of Table~\ref{tab:representation_tradeoff} two more times, each time with a different random seed. The two models we do not retrain are 3D tracklets only and the variant with depth augmentation.
Only the seed changes. The code, the data, the schedule, and the number of GPUs are those of the original run.
The seed decides the initial weights of the adapter and the heads, and the order of the training samples.
We checked the procedure with the original seed: after the first training step, all weights agree with the original run to within $2\times10^{-6}$.
Both students are trained again with the same teacher. The teacher itself is not trained again.

Table~\ref{tab:app_seeds} gives two errors for each model and each seed.
The first is the $z_E$ error on synthetic validation.
The second is the parameter distance on the 80 real drops.
Both are in dex, and lower is better.
``Mean $\pm$ std'' is the mean and the sample standard deviation over the three seeds.

\paragraph{How much the seed matters.}
The synthetic error changes little with the seed. For every model, its three values differ by at most 0.013\,dex.
The parameter distance on real drops changes more. Its standard deviation over the three seeds is between 0.01 and 0.06\,dex, depending on the model.

\paragraph{Lifted tracklets fail with every seed.}
The main paper finds that tracklets lifted to 3D are the most accurate input in simulation and the least accurate on real drops.
This holds for each of the three seeds.
Take the two students, which differ only in their tracklets.
On synthetic validation, the student with 3D tracklets has the lower error with every seed.
On real drops, its parameter distance is higher than that of \model{} by 0.487, 0.368, and 0.497\,dex with the three seeds.
The smallest of these gaps is six times the largest standard deviation over seeds in Table~\ref{tab:app_seeds}.
The two direct predictors behave the same way: RGB + 3D tracklets is worse than RGB + 2D tracklets by 0.465, 0.543, and 0.529\,dex.
With every seed, the model with 3D tracklets is worse than the same model with 2D tracklets on 12 of the 16 objects, and the object-level 95\% interval of the difference (computed as in App.~\ref{sec:app_rev_bootstrap}) lies above zero.

\paragraph{\model{} and the other models.}
\model{} has the lowest parameter distance of these nine models with each of the three seeds.
The second-best model is RGB + 2D tracklets + cues with every seed. It is behind \model{} by 0.062, 0.015, and 0.032\,dex.
These margins are small.
We therefore average the distance of each object over the three seeds and repeat the object-level comparison of App.~\ref{sec:app_rev_bootstrap}.
\model{} is then closer to the references than RGB + 2D tracklets + cues, RGB + 2D tracklets, and RGB, each on 12 of the 16 objects.
The mean difference in parameter distance, \model{} minus the other model, is $-0.036$\,dex (95\% interval $[-0.103, 0.035]$) against RGB + 2D tracklets + cues, $-0.054$\,dex ($[-0.112, 0.010]$) against RGB + 2D tracklets, and $-0.072$\,dex ($[-0.150, 0.007]$) against RGB.
All three intervals include zero.
With the original seed alone, the interval against RGB + 2D tracklets also includes zero, and the interval against RGB, $[-0.206, -0.029]$, does not.

\paragraph{Adding 2D tracklets to RGB.}
Averaged over the three seeds, RGB + 2D tracklets is closer to the references than RGB, 0.764 against 0.783\,dex; with the original seed alone, it was 0.745 against 0.797.
The difference over the three seeds, $-0.018$\,dex, is within the variation between seeds: its object-level 95\% interval, $[-0.053, 0.017]$, includes zero.
\model{}, which also reads 2D tracklets, is the best of the nine models with every seed.

\paragraph{Ranking by synthetic validation.}
Table~\ref{tab:validation_protocols} reports Kendall's $\tau=-0.71$ between the synthetic error and the real parameter distance of the eight predictors that share the RGB branch.
A negative $\tau$ means that synthetic validation orders the predictors the wrong way.
With the two new seeds, $\tau$ is $-0.57$ and $-0.64$.
Computed on the three-seed means, it is $-0.71$.

\paragraph{Direct simulation difference.}
$\mathrm{Err}_{\mathrm{ref}}$ (App.~\ref{sec:app_err_ref}) simulates each drop twice from the same initial state, once with the predicted material and once with the reference material, and compares the two simulations.
The last column of Table~\ref{tab:app_seeds} gives its mean and standard deviation over the three seeds.
For \model{} it is 0.109 $\pm$ 0.001, and for the student with 3D tracklets it is 0.146 $\pm$ 0.007.
For every model in the table, the three values differ by at most 0.013.

\begin{table}[h]
\centering\small
\caption{\textbf{Three training seeds.} Columns 2--5: $z_E$ error on synthetic validation (dex). Columns 6--9: parameter distance on the 80 real drops (dex). Both are measured after the fifth epoch, for the original seed (1701) and for the two new seeds. Last column: direct simulation difference $\mathrm{Err}_{\mathrm{ref}}$ on the 80 drops. ``Mean $\pm$ std'' is the mean and the sample standard deviation over the three seeds.}
\label{tab:app_seeds}
\begin{adjustbox}{max width=\linewidth}
\begin{tabular}{@{}lccccccccc@{}}\toprule
 & \multicolumn{4}{c}{Synthetic $z_E$ $\downarrow$} & \multicolumn{4}{c}{Parameter distance $\downarrow$} & $\mathrm{Err}_{\mathrm{ref}}\downarrow$ \\\cmidrule(lr){2-5}\cmidrule(lr){6-9}\cmidrule(l){10-10}
Input & 1701 & seed 2 & seed 3 & mean $\pm$ std & 1701 & seed 2 & seed 3 & mean $\pm$ std & mean $\pm$ std \\\midrule
\model{} & 0.481 & 0.475 & 0.476 & 0.477 $\pm$ 0.003 & 0.677 & 0.735 & 0.719 & 0.710 $\pm$ 0.030 & 0.109 $\pm$ 0.001 \\
Student, 3D tracklets & 0.443 & 0.443 & 0.444 & 0.443 $\pm$ 0.001 & 1.164 & 1.103 & 1.215 & 1.161 $\pm$ 0.056 & 0.146 $\pm$ 0.007 \\
RGB + 2D tracklets & 0.482 & 0.476 & 0.476 & 0.478 $\pm$ 0.003 & 0.745 & 0.784 & 0.765 & 0.764 $\pm$ 0.019 & 0.117 $\pm$ 0.003 \\
RGB + 3D tracklets & 0.444 & 0.445 & 0.451 & 0.447 $\pm$ 0.004 & 1.210 & 1.327 & 1.294 & 1.277 $\pm$ 0.061 & 0.156 $\pm$ 0.005 \\
RGB & 0.478 & 0.475 & 0.476 & 0.476 $\pm$ 0.002 & 0.797 & 0.777 & 0.774 & 0.783 $\pm$ 0.012 & 0.119 $\pm$ 0.001 \\
RGB + 2D tracklets + cues & 0.487 & 0.479 & 0.480 & 0.482 $\pm$ 0.004 & 0.739 & 0.751 & 0.751 & 0.747 $\pm$ 0.007 & 0.114 $\pm$ 0.001 \\
RGB + dense flow & 0.472 & 0.463 & 0.465 & 0.466 $\pm$ 0.005 & 0.833 & 0.822 & 0.850 & 0.835 $\pm$ 0.014 & 0.124 $\pm$ 0.002 \\
RGB + depth & 0.477 & 0.467 & 0.471 & 0.471 $\pm$ 0.005 & 0.789 & 0.765 & 0.759 & 0.771 $\pm$ 0.016 & 0.117 $\pm$ 0.004 \\
2D tracklets only & 0.764 & 0.769 & 0.756 & 0.763 $\pm$ 0.007 & 1.080 & 1.028 & 1.054 & 1.054 $\pm$ 0.026 & 0.145 $\pm$ 0.005 \\
\bottomrule\end{tabular}
\end{adjustbox}
\end{table}

\paragraph{Replay losses.}
Table~\ref{tab:app_seeds_replay} reports the replay losses of Table~\ref{tab:representation_tradeoff} for the three seeds of five models: \model{}, the student with 3D tracklets, RGB + 2D tracklets, RGB + 3D tracklets, and RGB.

\begin{table}[h]
\centering\small
\caption{\textbf{Replay with three training seeds.} Relative increase of the deformation (columns 2--5) and total (columns 6--9) replay losses over the reference on the 80 real drops, for the original seed (1701) and the two new seeds. ``Mean $\pm$ std'' is the mean and the sample standard deviation over the three seeds.}
\label{tab:app_seeds_replay}
\begin{adjustbox}{max width=\linewidth}
\begin{tabular}{@{}lcccccccc@{}}\toprule
 & \multicolumn{4}{c}{$\Delta_{\mathrm{deform}}^{\mathrm{replay}}$ $\downarrow$} & \multicolumn{4}{c}{$\Delta_{\mathrm{total}}^{\mathrm{replay}}$ $\downarrow$} \\\cmidrule(lr){2-5}\cmidrule(l){6-9}
Input & 1701 & seed 2 & seed 3 & mean $\pm$ std & 1701 & seed 2 & seed 3 & mean $\pm$ std \\\midrule
\model{} & 12.3\% & 6.6\% & 13.4\% & 10.8 $\pm$ 3.7\% & 10.1\% & 11.6\% & 11.9\% & 11.2 $\pm$ 0.9\% \\
Student, 3D tracklets & 28.5\% & 23.5\% & 29.1\% & 27.0 $\pm$ 3.1\% & 17.0\% & 15.8\% & 18.0\% & 16.9 $\pm$ 1.1\% \\
RGB + 2D tracklets & 13.1\% & 15.0\% & 10.1\% & 12.7 $\pm$ 2.5\% & 4.9\% & 10.0\% & 9.3\% & 8.1 $\pm$ 2.8\% \\
RGB + 3D tracklets & 31.2\% & 34.3\% & 28.8\% & 31.4 $\pm$ 2.7\% & 14.4\% & 21.1\% & 20.5\% & 18.6 $\pm$ 3.7\% \\
RGB & 17.9\% & 14.5\% & 24.0\% & 18.8 $\pm$ 4.8\% & 6.2\% & 9.9\% & 10.2\% & 8.8 $\pm$ 2.3\% \\
\bottomrule\end{tabular}
\end{adjustbox}
\end{table}

\section{Additional Controls}
\label{sec:app_controls}
This section describes two controls.

\begin{figure*}[t]
\centering
\includegraphics[width=\textwidth]{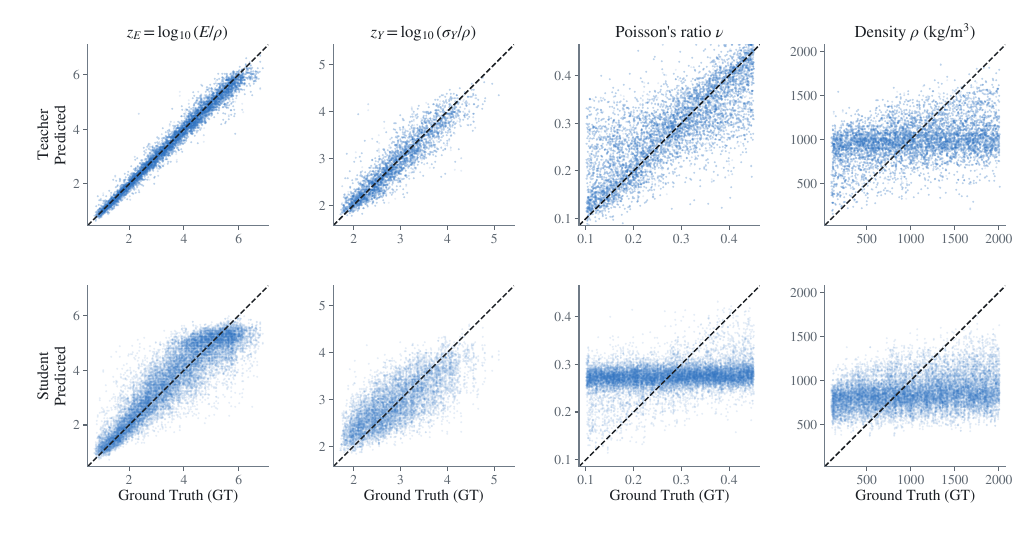}
\caption{\textbf{Teacher and student with four heads on synthetic validation.} Predicted value against ground truth (GT) for the two ratios, Poisson's ratio, and the density. Top: the teacher, which reads exact 3D deformation. Bottom: the student, which reads the video. Each dot is one video; the dashed line is perfect prediction.}
\label{fig:app_four_heads}
\end{figure*}

\paragraph{Replay with a changed reference stiffness.}
We test how sharply the drops determine the reference stiffness.
We replay each drop with the reference material and change only its stiffness, by a factor of $10^{-1}$, $10^{-0.5}$, $10^{+0.5}$, or $10^{+1}$.
The yield stress, the density, and Poisson's ratio are those of the reference.
The initial state is searched in the same way as for a predicted material (App.~\ref{sec:app_behavior}).
A stiffness outside the range of our training data and reference search ($10^4$ to $10^9$\,Pa) is not simulated.
Table~\ref{tab:app_ref_offset} reports the same replay measures as Table~\ref{tab:external_comparisons}.

The replay losses rise with the size of the change, on both sides of the reference.
A change of 0.5\,dex raises the total loss by 8.6--10.6\% and the deformation loss by 13.5--31.4\%.
A change of 1\,dex raises them by 19.3--24.8\% and 32.6--53.2\%.
For comparison, the materials predicted by \model{} raise them by 10.1\% and 12.3\% on all 80 drops (Table~\ref{tab:external_comparisons}).
On the same drops as each row, they raise them by 12.4\% and 9.5\% (the 55 drops of the two softer rows), 10.3\% and 13.6\% (the 65 drops of the row $\times10^{+0.5}$), and 11.0\% and 14.3\% (the 60 drops of the row $\times10^{+1}$).

\begin{table}[h]
\centering\small
\caption{\textbf{Replay with a changed reference stiffness.} All numbers are relative to the reference material; the replay measures are those of Table~\ref{tab:external_comparisons}. Drops: number of drops simulated. A changed stiffness outside the range of our training data and reference search ($10^4$ to $10^9$\,Pa) is not simulated. This removes five objects from the rows $\times10^{-1}$ and $\times10^{-0.5}$ (glove, Kitty, neck pillow, pad, and pillow), three objects from the row $\times10^{+0.5}$ (bowl, box, and hanger), and four from the row $\times10^{+1}$ (the same three and the bottle).}
\label{tab:app_ref_offset}
\begin{adjustbox}{max width=\linewidth}
\begin{tabular}{@{}lcccc@{}}\toprule
Material & $\Delta_{\mathrm{total}}^{\mathrm{replay}}\downarrow$ & $\Delta_{\mathrm{deform}}^{\mathrm{replay}}\downarrow$ & $\mathrm{Err}_{\mathrm{ref}}\downarrow$ & Drops \\\midrule
Reference stiffness $\times10^{-1}$ & 24.8\% & 32.6\% & 0.135 & 55 \\
Reference stiffness $\times10^{-0.5}$ & 10.6\% & 13.5\% & 0.072 & 55 \\
Reference stiffness $\times10^{+0.5}$ & 8.6\% & 31.4\% & 0.092 & 65 \\
Reference stiffness $\times10^{+1}$ & 19.3\% & 53.2\% & 0.145 & 60 \\
\bottomrule\end{tabular}
\end{adjustbox}
\end{table}

\paragraph{Density and Poisson's ratio.}
Our models predict the two ratios $E/\rho$ and $\sigma_Y/\rho$, because a drop does not determine density and Poisson's ratio (App.~\ref{sec:app_identifiability}).
We test this in simulation.
We train the teacher and the student again with four heads instead of two.
The two additional heads predict Poisson's ratio $\nu$ and the density $\rho$ (trained on $\log_{10}\rho$; we report the error in kg/m$^3$, because the density is sampled uniformly in $\rho$).
All other settings are those of \model{}.
Table~\ref{tab:app_four_heads} compares the error on synthetic validation with two baselines.
The first is a constant prediction, the mean of the training set.
The second shows how much the ratios alone reveal about the density.
In \model{}-Sim, $E$ ranges from 10\,kPa to 1\,GPa and $\rho$ from 100 to 2000\,kg/m$^3$, and the two are sampled independently.
The smallest possible ratio, $E/\rho = 10^4/2000 = 5$\,m$^2$/s$^2$, occurs only when $E$ is at its minimum and $\rho$ at its maximum.
A model that knows only $E/\rho$ can therefore guess the density of such a video without sensing it.
Likewise, the largest ratios occur only when $\rho$ is near 100\,kg/m$^3$, while in the middle of the range any density is possible.
The baseline ``From ratios'' measures this effect: for each pair of true $z_E$ and $z_Y$, it predicts the mean density (or Poisson's ratio) of the training videos with these ratios.
It uses no information about the density itself.
A model that beats the constant prediction but not this baseline shows no sign of sensing the density.

The student, which reads the video, learns neither quantity (Fig.~\ref{fig:app_four_heads}, bottom row).
Its error for Poisson's ratio is 0.083, against 0.088 for the constant prediction.
Its error for the density is 463\,kg/m$^3$, against 469\,kg/m$^3$ for the constant prediction and 381\,kg/m$^3$ for the true ratios alone.
The teacher, which reads the exact 3D deformation of the whole surface, does not learn the density either (434\,kg/m$^3$).
It does read part of Poisson's ratio from the exact deformation (error 0.048, against 0.088 for both baselines).
Poisson's ratio shows in how much the object expands sideways when it is compressed at impact.
The teacher sees this lateral and volumetric change in 3D, while in a video it is very small.
App.~\ref{sec:app_nu} gives an example on a real drop: changing $\nu$ of the crocs from 0.271 to 0.392 moves the particles by 0.62\,mm on average, while one pixel of the $224\times224$ crop given to the predictors covers about 3\,mm.
The change is about a fifth of a pixel.
Sec.~\ref{sec:physical_model} states that a drop does not determine Poisson's ratio. This holds for what the video shows: exact 3D deformation, which only the teacher sees in simulation, carries some information about it.
The two additional heads do not change the errors of $z_E$ and $z_Y$: they are 0.164 and 0.196\,dex for the teacher, against 0.161 and 0.192 in Table~\ref{tab:teacher_student}, and 0.482 and 0.357\,dex for the student, against 0.481 and 0.356.

\begin{table}[h]
\centering\small
\caption{\textbf{Teacher and student with four heads.} Mean absolute error on synthetic validation. Constant: error of always predicting the training mean. From ratios: error of predicting the mean value of the training videos with the same true $z_E$ and $z_Y$. The teacher is evaluated on the 14,293 videos with usable surface geometry. Dashes: not applicable, because this baseline is given the true $z_E$ and $z_Y$.}
\label{tab:app_four_heads}
\begin{adjustbox}{max width=\linewidth}
\begin{tabular}{@{}lcccc@{}}\toprule
Quantity & Constant & From ratios & Teacher $\downarrow$ & Student $\downarrow$ \\\midrule
$z_E=\log_{10}(E/\rho)$ (dex) & 1.252 & -- & 0.164 & 0.482 \\
$z_Y=\log_{10}(\sigma_Y/\rho)$ (dex) & 0.554 & -- & 0.196 & 0.357 \\
Poisson's ratio $\nu$ & 0.088 & 0.088 & 0.048 & 0.083 \\
Density $\rho$ (kg/m$^3$) & 469 & 381 & 434 & 463 \\
\bottomrule\end{tabular}
\end{adjustbox}
\end{table}

\end{document}